\documentclass[letterpaper, 10 pt, conference]{article}  

\usepackage[letterpaper,bindingoffset=0.2in,%
            left=1in,right=1in,top=1in,bottom=1in,%
            footskip=.25in]{geometry}
\usepackage{times}
\usepackage{epsfig}
\usepackage{graphicx}
\usepackage{amsmath}
\usepackage{amssymb}
\usepackage{mathrsfs}
\usepackage[title]{appendix}
\usepackage{placeins} 
\usepackage{dblfloatfix} 
\usepackage{subcaption} 
\usepackage{enumitem} 
\usepackage{subcaption}
\usepackage{booktabs}
\usepackage[nice]{nicefrac}
\usepackage{multirow}
\usepackage{arydshln}
\usepackage{forest}
\usepackage{authblk}

\usepackage{titlesec}

\titleformat{\section}
{\normalfont\Large\bfseries}{\thesection}{1em}{}
\titleformat{\subsection}
{\normalfont\large\bfseries}{\thesubsection}{1em}{}
\titleformat{\subsubsection}
{\normalfont\normalsize\bfseries}{\thesubsubsection}{1em}{}
\titleformat{\paragraph}[runin]
{\normalfont\normalsize\bfseries}{\theparagraph}{1em}{}
\titleformat{\subparagraph}[runin]
{\normalfont\normalsize\bfseries}{\thesubparagraph}{1em}{}

\newcommand{\etal}{\textit{et al}. }
\newcommand{\ie}{\textit{i}.\textit{e}. }
\newcommand{\eg}{\textit{e}.\textit{g}. }
\newcommand{\etc}{\textit{etc}.}

\newcommand{\specialcell}[2][l]{%
  \begin{tabular}[#1]{@{}l@{}}#2\end{tabular}}

\graphicspath{{images/}}

\usepackage[pagebackref=true,breaklinks=true,letterpaper=true,colorlinks,bookmarks=true]{hyperref}

\date{}

\begin{document}

\title{\LARGE \bf Behavioral Research and Practical Models of\\ Drivers' Attention}
\author{Iuliia Kotseruba and John K. Tsotsos\\
\small Department of Electrical Engineering and Computer Science \\
\small York University, Toronto, ON, Canada \\
\tt\small {yulia\_k,tsotsos}@eecs.yorku.ca
}

\maketitle

\begin{abstract}
Driving is a routine activity for many, but it is far from simple. When on the road, drivers must concurrently deal with multiple tasks. Besides keeping the vehicle in the lane, they must observe other road users, anticipate their actions, react to sudden hazards, and deal with various distractions inside and outside the vehicle. Failure to notice and respond to the surrounding objects and events can potentially cause accidents. 

The ongoing improvements of the road infrastructure and vehicle mechanical design have made driving safer overall. Nevertheless, the problem of driver inattention has remained one of the primary causes of accidents. Therefore, understanding where the drivers look and why they do so can help eliminate sources of distractions and identify optimal attention allocation patterns for safer driving. The outcomes of research on driver attention have implications for a range of practical applications such as revising existing policies, improving driver education, enhancing road infrastructure and in-vehicle infotainment systems (IVIS), as well as designing systems for driver monitoring (DMS), advanced driver assistance (ADAS) and highly-automated driving (HAD).

This report covers the literature on how drivers' visual attention distribution changes depending on various factors, internal to the driver, \eg their driving skill, age, and physical state, and external ones, \eg the presence of outside distractions and automation.  Various aspects of attention allocation during driving have been explored across multiple disciplines, including psychology, human factors, human-computer interaction, intelligent transportation, and computer vision, each with specific perspectives and goals, and different explanations for the observed phenomena. Here, we link cross-disciplinary theoretical and behavioral research on driver's attention to practical solutions. Furthermore, limitations are discussed and potential directions for future research in this fast-moving field are presented. Statistical data and conclusions shown here are based on over 175 behavioral studies, nearly 100 practical papers, 20 datasets, and over 70 surveys published in the last decade. A curated list of papers used for this report is available at \url{https://github.com/ykotseruba/attention_and_driving}.

\end{abstract}

%
%

\section{Introduction}
Driver inattention has been recognized as a major cause of accidents since at least the late 1930s \cite{1938_AJP_Gibson} and, sadly, continues to be so. To this day, much of the research on drivers' attention is motivated by safety. Early studies on drivers' behavior addressed attention only indirectly via observations of vehicle control. Some of the factors considered were engagement in secondary tasks \cite{1936_AJP_Ryan}, lighting \cite{1960_AJO_Richards} and environmental conditions \cite{1958_QJEP_Bursill}, fatigue \cite{1956_PRSM_Davenport}, and the effect of alcohol consumption \cite{1940_JAMA_Newman}, all of which are still relevant. 

\begin{figure}[t!]
\centering
\begin{subfigure}{0.59\linewidth}
\centering
  \includegraphics[height=1.25in]{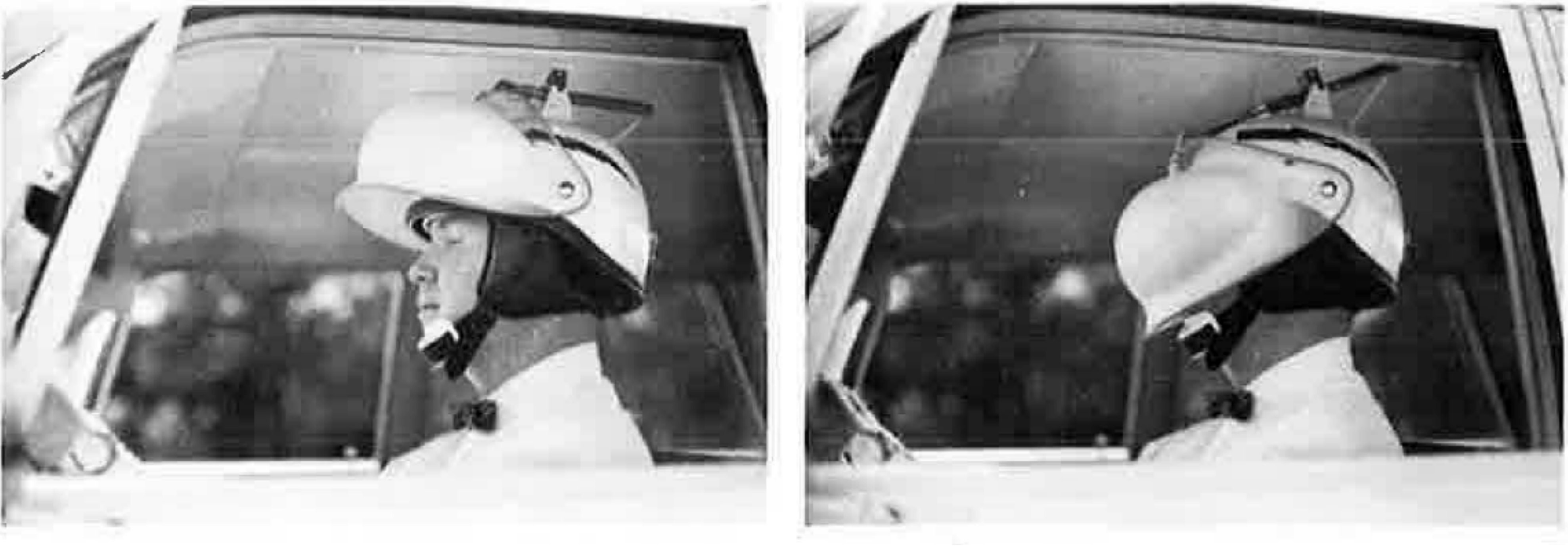}
  \caption{Helmet}
\end{subfigure}
~
\begin{subfigure}{0.39\linewidth}
\centering 
 \includegraphics[height=1.25in]{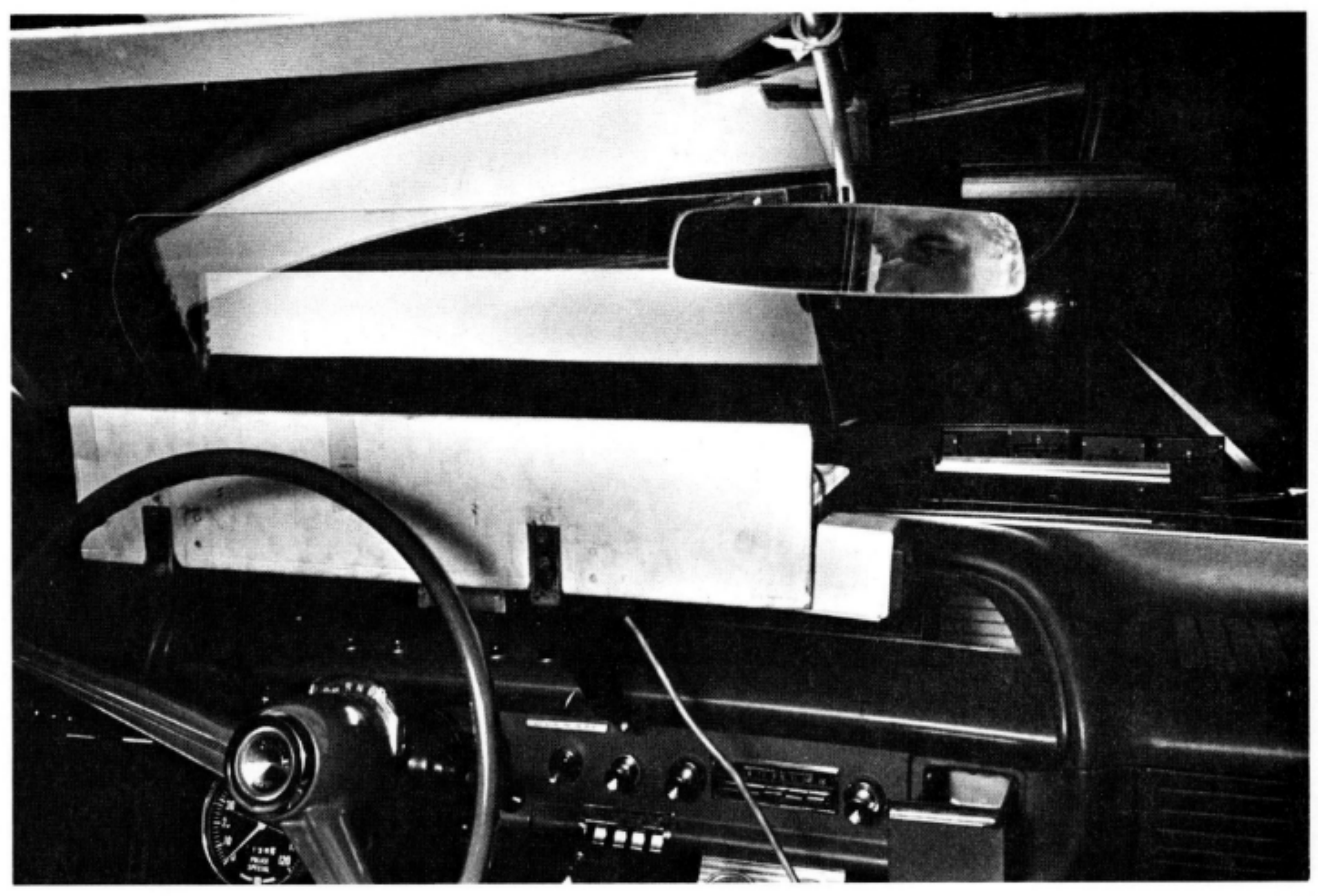}
 \caption{Veiling luminance device}
\end{subfigure}
\caption[Devices used for studying drivers' attention]{Devices used for studying drivers' attention. Source: a) \cite{1967_TR_Senders}, b) \cite{1970_TechRep_Rockwell}.}
\label{fig:attention_devices}
\end{figure}

Driving is primarily a visuomanual task, \ie drivers' motor actions are guided by vision, with some information obtained via auditory or somatosensory modalities. Thus the main questions of interest are \textit{where}, \textit{when}, and \textit{why} drivers look in different environments and situations. A natural first step towards the answers is gathering objective and precise recordings of drivers' gaze. Since eye-tracking equipment was not easily accessible until the 1970s when some of the first commercial offerings started to appear, other methods of observing drivers' visual behaviors were explored. These included an aperture device that reduced the driver's peripheral vision \cite{1966_HumanFactors_Gordon}, a helmet with a translucent shield that could be lowered to control the driver's visual input \cite{1967_TR_Senders}, and wooden planks in front of the windshield to artificially limit the view of the road \cite{1970_TechRep_Rockwell} (Figure \ref{fig:attention_devices}).

\begin{figure}[!t]
\centering
\begin{subfigure}{0.4\linewidth}
\centering
  \includegraphics[height=1.25in]{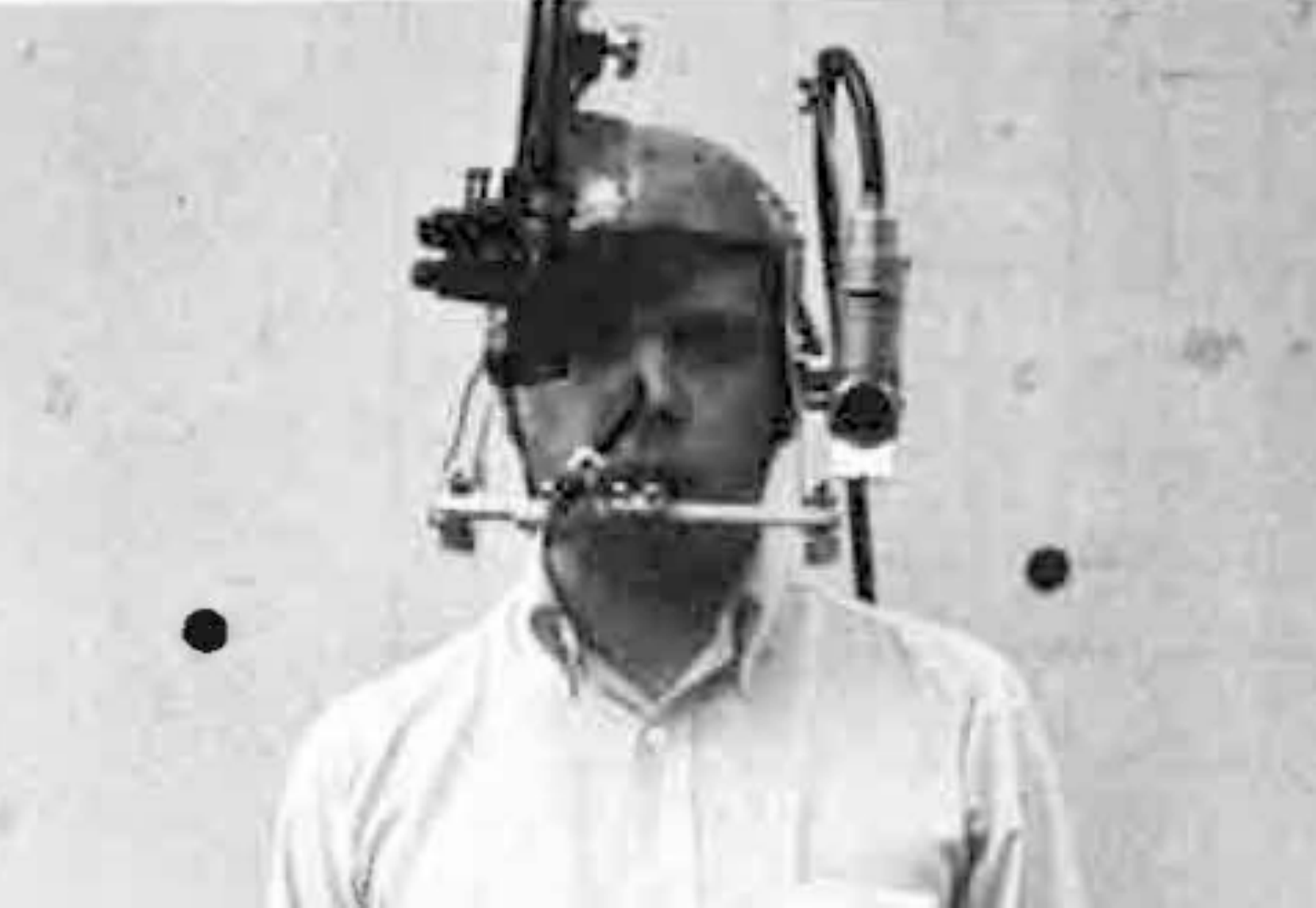}
  \caption{}
\end{subfigure}
~
\begin{subfigure}{0.4\linewidth}
\centering 
 \includegraphics[height=1.25in]{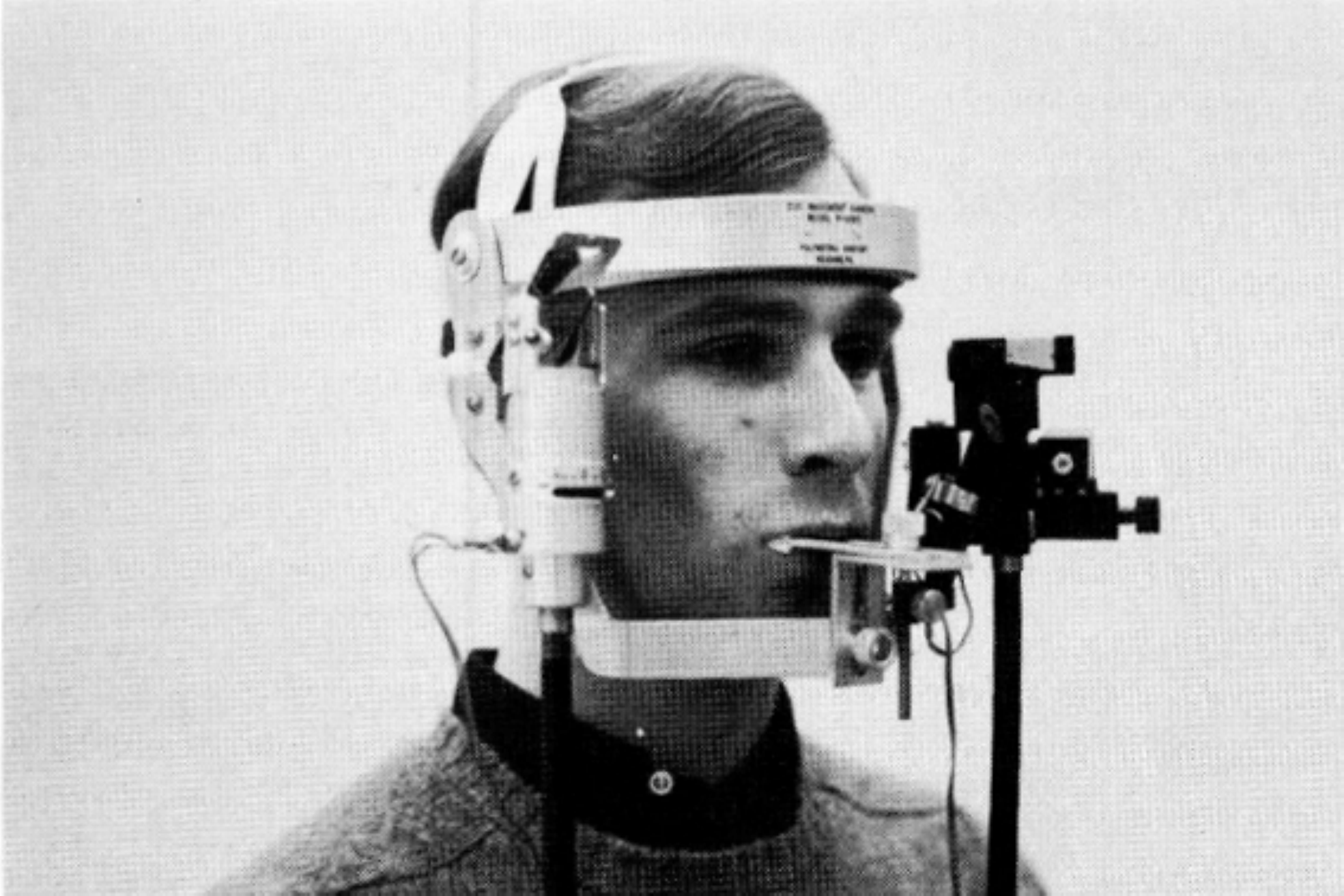}
 \caption{}
\end{subfigure}
\caption[Early mobile eye-trackers]{Mobile eye trackers used in 1970. Source: a) \cite{1969_TechRep_Kaluger}, b) \cite{1970_TechRep_Rockwell}.}
 \label{fig:early_eye_tracker}
\end{figure}

Early models of mobile eye-trackers (shown in Figure \ref{fig:early_eye_tracker}), despite being unwieldy, intrusive, and lacking in precision, afforded valuable insights into drivers' gaze behavior. Modern eye-tracking technology offers even more detailed and high-resolution recordings of eye movements. Although gaze does not provide complete information about drivers' perception and decision-making, eye-tracking data has proved useful in both theoretical and applied research on drivers' attention. Continued efforts are dedicated to discovering common patterns of drivers' visual behavior, their causes, and possible links to road accidents. This knowledge can inform regulation and help improve driver training. For example, studies indicate that bright colors and motion of the digital roadside billboards distract the drivers; therefore policies prescribe how many billboards can be installed along the roads and how often the images on them should transition. Driver training is another problem. Worldwide, new drivers are over-represented in crash statistics, partly because they do not possess visual skills for safe driving after finishing driving school, as established in the early work by Zell \cite{1969_TechRep_Zell} and Mourant \& Rockwell \cite{1970_SAE_Mourant}, and confirmed by countless studies since. Research towards improving driver education includes developing methods that accelerate the acquisition of the necessary visual scanning strategies.

Recent technological advances in intelligent transportation, computer vision, and AI promise to reduce accidents by monitoring where the driver is looking to prevent them from engaging in distracting activities, daydreaming, or falling asleep at the wheel. Driver assistive technologies, such as forward collision warning, further improve safety by attracting drivers' attention to the source of imminent danger. An ultimate solution to driving safety is replacing the driver with an autonomous system controlling the vehicle. 

Besides safety considerations, driving provides a useful testbed for studying human visual attention because it is a widespread activity familiar to many people of different ages and backgrounds. A traffic scene is a prime example of a noisy, dynamic, and complex environment that humans can actively navigate and explore by selectively sampling the information. The driving task itself consists of multiple competing sub-tasks, each with different and sometimes conflicting demands. Consequently, driving makes it possible to investigate the control of eye movements to prioritize and accomplish current goals and study the link between vision and motor actions. 

Attention during driving is explored across multiple disciplines that provide different perspectives and explanations of the observed phenomena. This report summarizes the past decade of research on various factors that have been linked to drivers' attention allocation, internal mechanisms of attention, and efforts towards building practical models of attention while driving. The literature review focuses on studies that explicitly measure drivers' attention using eye-tracking data as opposed to studies that rely on indirect measures such as driving performance. The final set of publications reviewed comprises 175 behavioral studies, 100 practical works, and 65 surveys published since 2010 (inclusive).

This report is structured as follows: Section \ref{ch:what_driver_sees} provides the necessary background for why eye movements are needed and the advantages and limitations of using gaze as a proxy of drivers' attention. Section \ref{ch:data_collection} covers various gaze data collection methodologies, and Section \ref{ch:attention_measures} lists measures of attention based on gaze and links them to various phenomena observed in behavioral literature. Section \ref{ch:attention_factors} introduces external and internal factors studied with respect to drivers' attention, establishes connections between them, and highlights historical trends in the literature. Section \ref{ch:behavioral} continues by reviewing the literature associated with each group of factors in detail. Analytical models of drivers' attention are covered in Section \ref{ch:psychological_models}. Practical works, including publicly available datasets, algorithms for predicting drivers' gaze and monitoring drivers' behaviors, as well as applications of attention in self-driving, are reviewed in Section \ref{ch:practical}. Discussion of the results, outstanding questions for future research, and conclusions are presented in Section \ref{ch:discussion_conclusions}.

\section{What can and cannot be revealed about attention by gaze?}
\label{ch:what_driver_sees}

Even though vision does not constitute $90\%$ of the sensory input to driving \cite{1996_Perception_Sivak}, as previously claimed \cite{1988_JAGS_Reuben, 1980_Perception_Hills}, it is generally acknowledged that drivers' decisions are largely based on what they see, and accidents may happen when drivers fail to look at the right place at the right time \cite{2015_TR_Victor}.

Driving is not a single activity but rather a set of concurrent tasks with different and sometimes conflicting visual demands. For example, drivers observe the forward path, sides of the road, and lane markings to keep the vehicle in the center of the ego-lane and within a safe distance of the cars ahead. Besides controlling the vehicle, they must follow the traffic rules and keep track of signs and traffic signals. They might sometimes need to verify the route to the destination by checking the guiding signs or in-vehicle navigation system. Various maneuvers such as changing lanes, turning, and avoiding obstacles require looking at different parts of the scene around the vehicle. Finally, unexpected hazards, \eg a lead vehicle braking or pedestrian crossing the road, may suddenly appear. 

Besides task demands and anomalous events, physical and emotional condition of the drivers affect where and when they look, as well as their driving skills and the presence of distractions. The analysis of statistical relationships between changes in drivers' gaze patterns, their internal states, and external factors comprises the bulk of behavioral research (covered in Section \ref{ch:behavioral}) and is widely used in practical applications (Section \ref{ch:practical}). Even though drivers' gaze has been an indispensable tool in attention research for decades now, eye movements provide an incomplete representation of the drivers' sensory experience and processing. Thus, the nature of internal mechanisms involved in attention allocation, processing of the selected regions, and generation of motor actions remain an area of active research. This section will briefly discuss the biological characteristics and limitations of human vision and visual attention, how they relate to various driving-related tasks, and what insights into drivers' attention can and cannot be gained from the gaze.

\begin{figure}[!t]
\centering
\begin{subfigure}{0.6\linewidth}
\centering
  \includegraphics[height=2.5in]{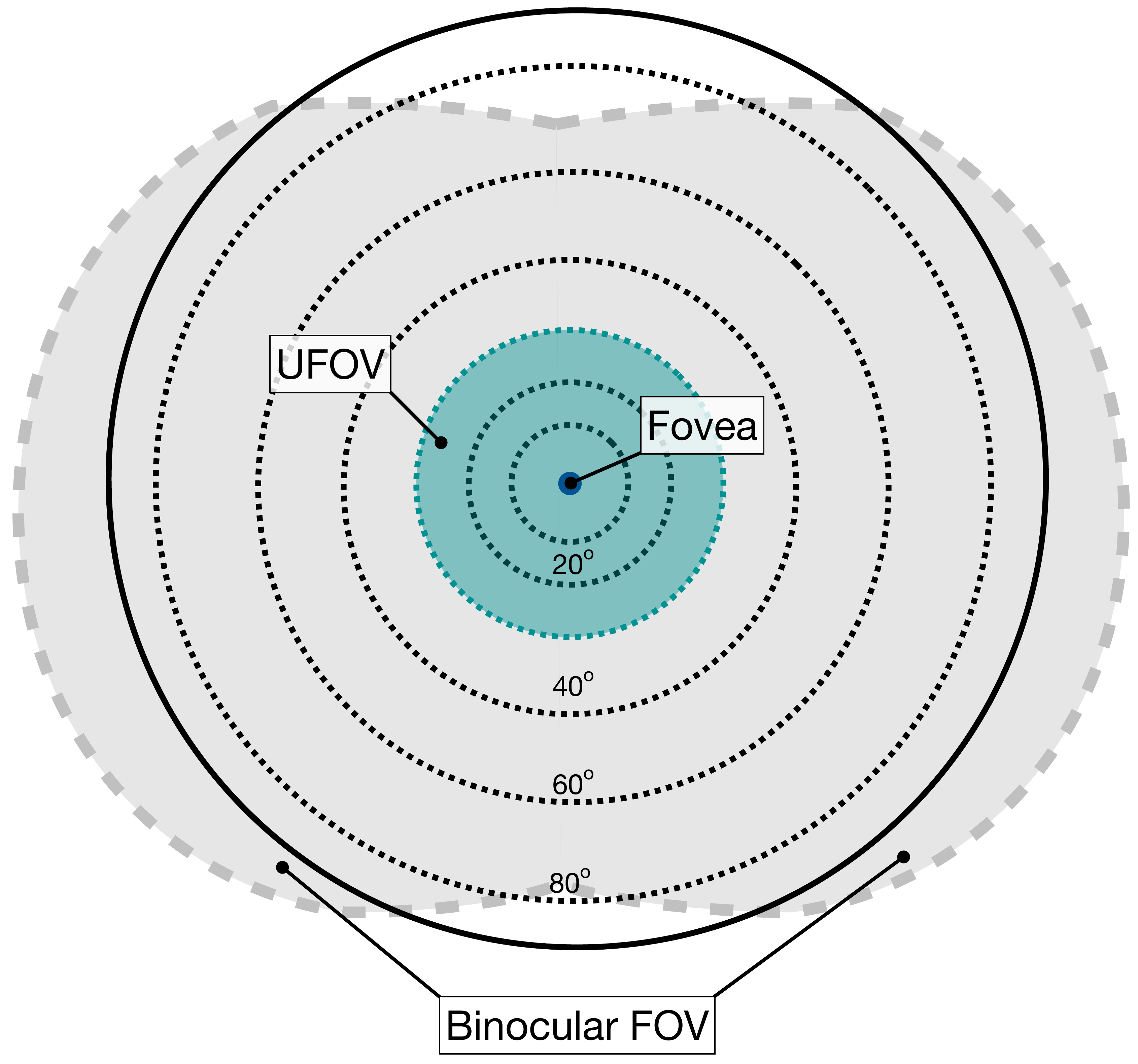}
  \caption{Human binocular visual field}
    \label{fig:visual_field}
\end{subfigure}
~
\begin{subfigure}{0.9\linewidth}
\centering 
 \includegraphics[height=1.5in]{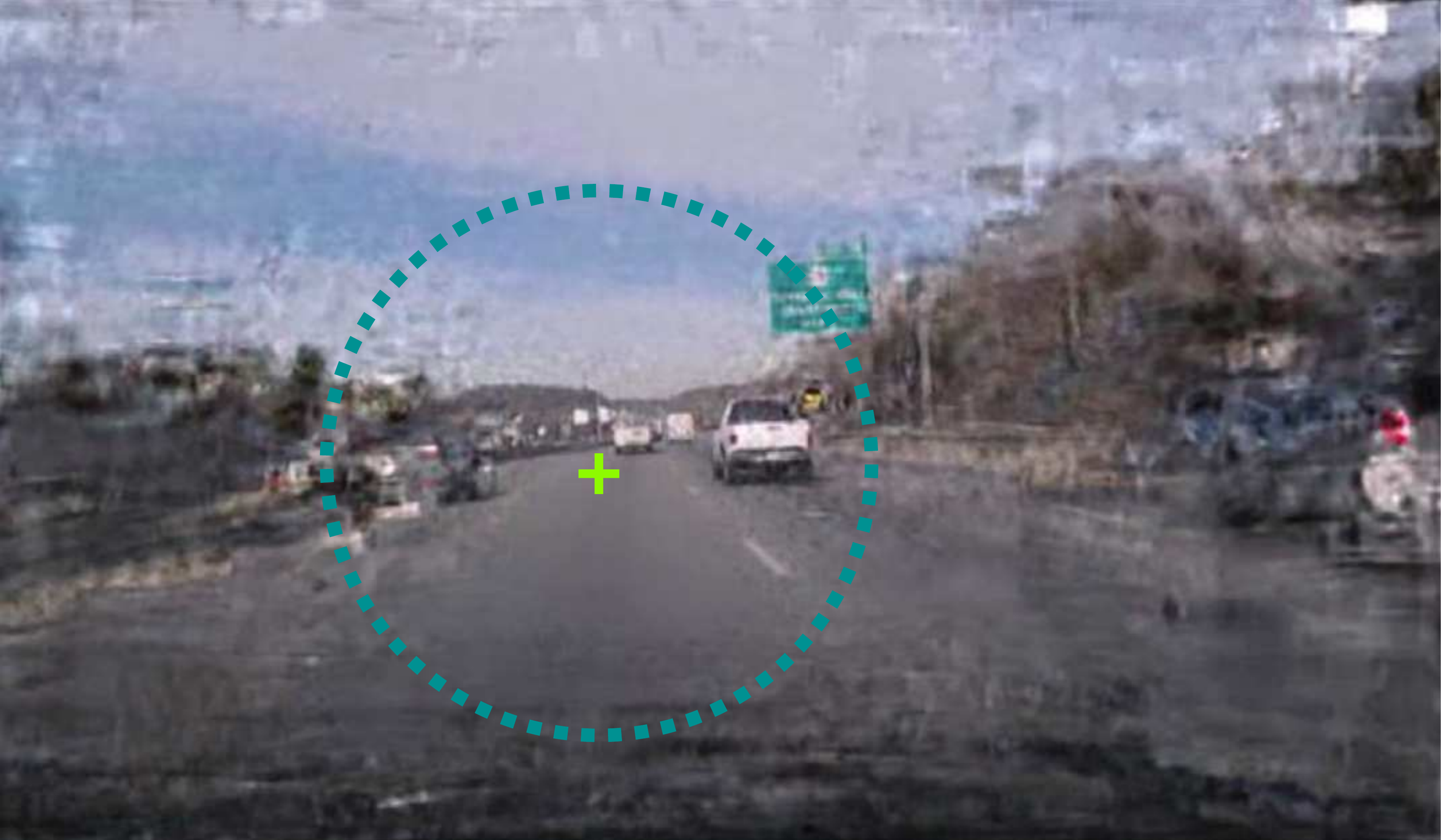}
  \includegraphics[height=1.5in]{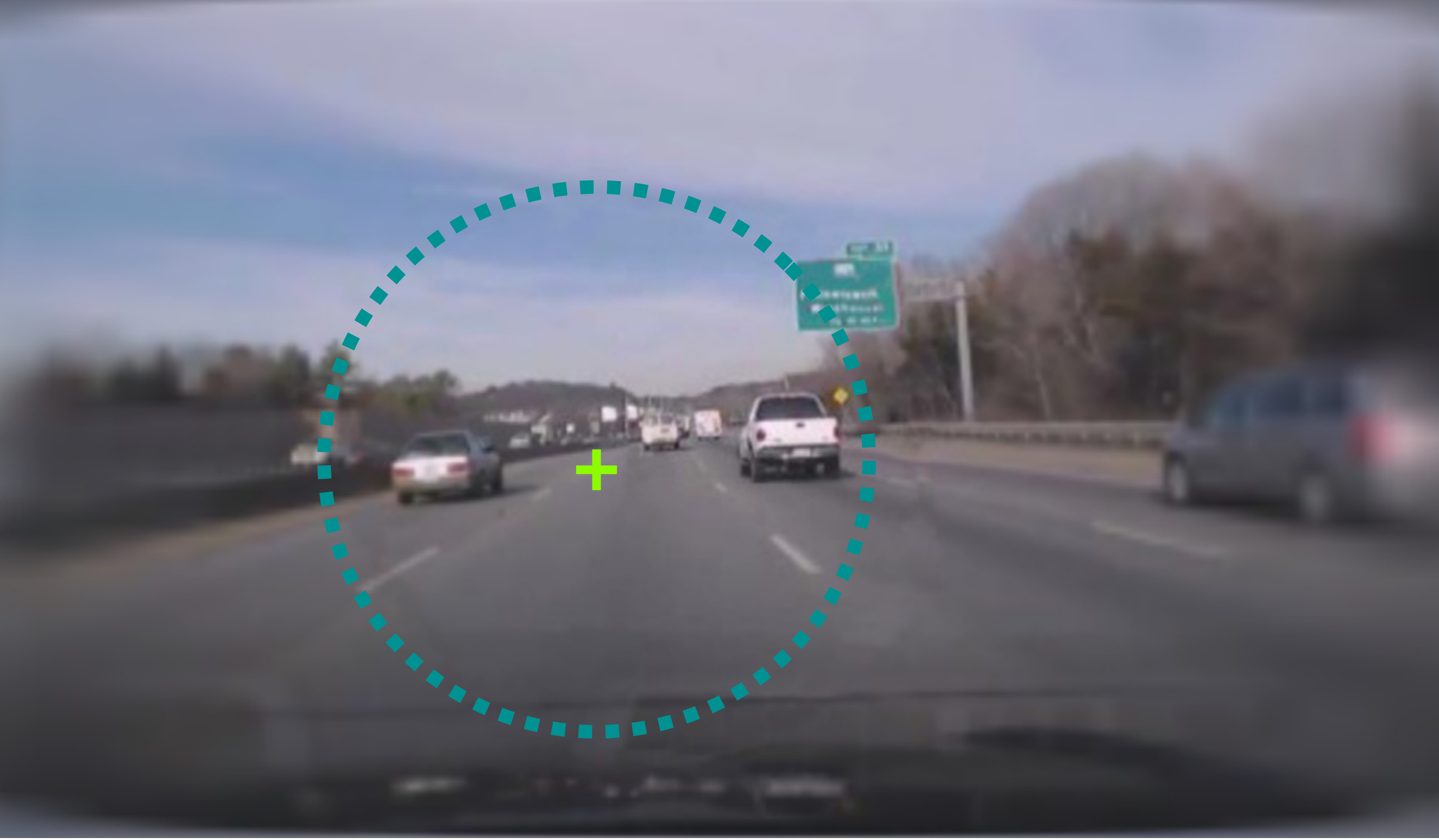}
  \caption{Foveated images of the traffic scene}
  \label{fig:visual_field_foveated}
  \end{subfigure}
  \caption[Visualization of the human visual field and foveated traffic scene]{a) Visualization of the human binocular visual field (gray area), fovea, and useful field of view (UFOV); b) Foveated images of the traffic scene. Right: distortion in the periphery according to the Texture Tiling Model \cite{2012_FPsych_Rosenholtz}. Left: loss of resolution in the periphery according to Geisler \& Perry model \cite{1998_HV_Geisler, 2016_JEMR_Tsotsos}. In both images the green cross shows the point of gaze and the blue circle indicates the UFOV with a diameter of $\approx 30^\circ$. Source: \cite{2017_AppliedErgonomics_Wolfe}.}
\end{figure}

\subsection{Covert, overt and divided attention}
\textbf{Limitations of gaze as a proxy for attention.} Human vision is foveated, \ie visual acuity is highest in the center of the visual field (fovea) and degrades towards the periphery due to the non-uniform distribution of receptors in the retina \cite{1935_AO_Osterberg,1990_JCN_Curcio} and neural tissue dedicated to processing visual stimuli in the cortex (cortical magnification) \cite{2011_Encyclopedia_Cohen}. The visual field is commonly divided into fovea ($2^\circ$ in diameter), parafovea (up to $5^\circ$), together referred to as the central vision, and the periphery that covers the remainder of the visual field. Since only a small portion of the environment in the parafovea (Figure \ref{fig:visual_field}) can be captured with high resolution, eye movements are necessary to bring new targets into the central visual field. Such sampling of visual information via eye movements is referred to as overt attention. Covert attention, or changing the focus of attention without explicit gaze change, plays a role in planning future eye movements. Overall, attention towards various elements of the scene is enabled by a set of mechanisms that perform selection, restriction, and suppression of features, objects or areas \cite{2011_MIT_Tsotsos}, and is influenced by both the demands of the current task and involuntary attraction towards salient regions (as explained later in Section \ref{sec:bu_td_attention}). Even though gaze is strongly correlated with overt spatial attention, it does not provide full information of the driver's sensory experience for reasons discussed below.

\noindent
\textbf{Foveal vision $\neq$ awareness.} Looking at a specific object or area in the scene does not guarantee processing; in other words, the relationship between awareness and attention is not symmetrical. Even if an object or event falls within the foveal region, it may still be missed. It is well-documented that people often fail to notice things that are in plain sight. One example of this is change blindness -- missing a significant and obvious change in the environment. This phenomenon may be caused by incomplete, nonexistent, or overwritten internal visual representations as well as comparison failures \cite{2011_WIR_Jensen}. The established experimental paradigm for studying change blindness is as follows: the scene is briefly shown to a subject, followed by a blank, and then again the same scene is shown, either unchanged or with some objects removed or added \cite{2005_Elsevier_Rensink}. Similar conditions occur naturally during driving due to blinking, saccades, and physical occlusions, however, motion and illumination cues usually mitigate the effects of change blindness \cite{2012_SAE_Young}. When these additional cues are reduced or removed, drivers fail to notice changes to the traffic signs even when explicitly told to look for them \cite{2014_TransRes_Metz, 2016_TransRes_Harms} and miss changes to other objects in the environment \cite{2017_AccidentAnalysis_Beanland}.

\noindent
\textbf{Divided attention.} Drivers are often required to divide their attention between two or more concurrent tasks, which typically affects their performance in some or all tasks. A common theory behind this observation is that attention allocates limited resources between tasks with stronger negative effects when different tasks compete for the same resources \cite{2000_JEP_Recarte}. For example, driving as a visuomanual activity is affected by the tasks that are visual (reading), manual (eating), or visuomanual (texting), although cognitive distractions likewise have adverse effects (discussed later in Sections \ref{sec:inattention_taxonomy} and \ref{sec:secondary_tasks_effects}). 

Inattention blindness, \ie failure to notice an event or object when attention is diverted towards another activity \cite{2011_WIR_Jensen}, is a commonly observed phenomenon caused by divided attention. Some examples are reduced attention to safety-critical objects such as signs due to ``zoning out'', \ie driving without awareness, especially on familiar routes \cite{2013_TrafficPsychology_Charlton}, and failing to see hazards while talking to a passenger \cite{2010_AccidentAnalysis_White}. Shallow processing due to inattention blindness may lead to ``looked but failed to see'' (LBFTS) errors that cause many accidents, particularly violations of right-of-way at T-intersections. In such cases, drivers fail to see the approaching motorcycle, bicycle, or another car and pull out in front of it \cite{2008_TransRes_Crundall, 2010_AccidentAnalysis_White, 2011_VR_Underwood, 2018_HumanFactors_Robbins}.

\noindent
\textbf{The role of peripheral vision.} The area outside of the fovea, referred to as periphery, comprises more than $95\%$ of the visual field that spans over $200^\circ$ horizontally and $125^\circ$ vertically \cite{2020_iPerception_Strasburger} (see Figure \ref{fig:visual_field}). Although acuity decreases with eccentricity, peripheral vision is not merely a blurry version of foveal vision. For instance, perception of contrast \cite{1987_JOSA_Legge}, color \cite{2009_JoV_Hansen}, motion \cite{1992_PP_Warren, 2012_VR_Traschulz}, size, shape \cite{2016_iPerception_Baldwin} and category \cite{2016_JoV_Boucart} of objects is different in foveal and peripheral vision. Some of these differences (\eg acuity \cite{1991_JOSA_Banks}) stem from a higher concentration of receptors in the fovea and cortical magnification and can be minimized by scaling the stimuli so that they appear larger in the periphery. However, some differences, \eg motion perception or symmetry detection, are not scalable (see \cite{2011_JoV_Strasburger} for a comprehensive review). Crowding, the inability to recognize objects in cluttered scenes, is another well-known phenomenon often associated with peripheral vision \cite{2016_AnnRevVisScience_Rosenholtz} (although it occurs in fovea as well \cite{2020_iPerception_Strasburger}). More specifically, when several objects appear close together in the periphery, it may be difficult to tell them apart, even though crowding does not affect object detection \cite{2011_Cell_Whitney}. For example, a driver can detect with peripheral vision a group of pedestrians waiting to cross at the curb and some of their collective properties (\eg orientation), but will not be able to identify them. In general, the effects of crowding, besides eccentricity, are mediated by a range of factors, \eg spacing of objects, similarity of appearance, motion \cite{2011_Cell_Whitney, 2020_iPerception_Strasburger}. 

Most of the research on peripheral vision is conducted using artificial stimuli (\eg gratings and characters) presented at eccentricities not exceeding $20^\circ$, whereas investigations of real-world scene perception are rare and fragmented \cite{2017_JoV_Loschky}. Therefore less is known about peripheral perception for everyday tasks. In driving, for instance, relevant cues often appear in the periphery, away from the center of the road, where both crowding and loss of resolution are apparent (see Figure \ref{fig:visual_field_foveated}). Nevertheless, most research on attention and driving focuses on foveal vision, while peripheral vision has been historically relegated to detecting global motion cues, such as optical flow \cite{1972_HumanFactors_Mourant}. This property is useful for lane maintenance; in fact, experienced drivers can control the vehicle relying on peripheral vision alone, as demonstrated by Summala \etal \cite{1996_HumanFactors_Summala}. Besides lateral control \cite{1999_Perception_Crundall,2002_ACP_Crundall}, peripheral vision plays a role in hazard detection \cite{2010_AccidentAnalysis_Shahar, 2019_APP_Wolfe, 2020_JEP_Wolfe} and recognition of road signs \cite{2018_Ergonomics_Costa}.

Even though some driving tasks can be done with peripheral vision, it is not safe to do so. The forced-peripheral conditions elicited slow responses to road hazards such as suddenly braking lead vehicles \cite{1998_AccidentAnalysis_Summala}. Likewise, reaction times and miss rates for target detection increased with eccentricity \cite{2003_JEP_Recarte, 2019_APP_Wolfe}. Limitations of peripheral vision may contribute to LBFTS errors and reduce sign identification ability. For instance, at intersections, motorcycles and vehicles located the same distance away have different spatial frequencies and present different sizes of the images on the retina. Vehicles appear as large blocks of moving color, whereas motorcycles are small and easily blend into the surrounding clutter with high-frequency textures such as road markings, signposts, and trees \cite{2008_TransRes_Crundall}. For similar reasons, signs with singular shapes (\eg cross sign) can be identified even far in the periphery, however, uniformly shaped signs with pictograms or text are more easily confused with one another and other elements of the road scene \cite{2018_Ergonomics_Costa}.

\subsection{What can be captured with one glance?}

Useful (or functional) field of view (UFOV) is a theoretical construct widely used in the transportation literature (see reviews in \cite{2010_VisionResearch_Owsley,2016_SJOVS_Thorslund,2017_AppliedErgonomics_Wolfe}. UFOV, first proposed by Sanders in 1970 \cite{1970_Ergonomics_Sanders}, describes a visual field from which information can be extracted in a single fixation. The diameter of UFOV is  $30^\circ$ around the fovea as shown in Figure \ref{fig:visual_field}. Based on this measure, Ball \etal \cite{1991_HumanFactors_Ball} developed a test (also called UFOV) to map out the functional field of view of older drivers who may be at risk for an accident due to vision impairments. UFOV test examines processing speed, divided and selective attention using foveal identification and peripheral localization tasks. These tasks are performed separately and together, with and without distractors, while the presentation duration and eccentricity of stimuli within UFOV are manipulated. These conditions simulate peripheral detection and orientation tasks that drivers perform while focusing on the forward roadway.

The inability to perceive safety-critical information due to reduced UFOV in older drivers has been linked to increased risk of crashes in multiple independent studies \cite{1998_JAMA_Owsley, 2005_OVS_Clay, 2006_JAGS_Ball, 2010_AccidentAnalysis_Belanger, 2015_AccidentAnalysis_Cuenen}. Besides age, the reduction of UFOV can be caused by changes in cognitive workload \cite{1988_JOpnSocAm_Ball, 2003_TransRes_Underwood, 2010_TRR_Bian} and fatigue \cite{2004_VisionResearch_Roge}.

Despite its widespread use, UFOV has limitations. The implication of UFOV that areas outside of it are minimally processed or not processed at all have been questioned lately \cite{2017_AppliedErgonomics_Wolfe}. ``Visual tunneling'', \ie narrowing of the UFOV during an increased workload, implies a reduction of sensitivity in the periphery of the visual field \cite{1999_TechRep_Martens}. More recent literature, however, suggests that the performance reduction is not localized in the periphery but covers the entire retina and the apparent tunneling is a result of interference of the two concurrent tasks \cite{2012_SAE_Young}. The original definition of UFOV covers only a small portion of the visual field, however peripheral areas beyond UFOV are also useful. As shown in Figure \ref{fig:visual_field_foveated}, cars and edges of the road remain distinguishable at high eccentricities. As a result, later studies considered extending UFOV to $70^\circ$ in diameter \cite{2009_Gerontechnology_Itoh}. Lastly, UFOV does not take into account depth. There is evidence suggesting that declining detection performance is a function of target eccentricity as well as distance. In other words, attention is optimal at some 3D point in the scene and declines as targets are farther away from that position in 3D \cite{1990_Perception_Psychophysics_Andersen, 2011_AccidentAnalysis_Andersen}.


\subsection{What determines where we look?}
\label{sec:bu_td_attention}
What prompts changes in gaze locations is a major topic of research in attention. Mechanisms of attentional control are commonly divided into the \textit{bottom-up} and \textit{top-down} \cite{1998_Attention_Yantis}. The former is viewed as a process guided primarily by the properties of the scene and depends on the saliency of objects which attract gaze, unlike featureless areas \cite{1980_CognitivePsychology_Treisman, 1998_PAMI_Itti}. In contrast, top-down attention is driven by the demands of the current task \cite{1967_EyeMovements_Yarbus}. In other words, even salient stimuli may fail to attract attention if they are irrelevant to the task. Some researchers argue that the bottom-up/top-down dichotomy does not describe a full range of selection phenomena. \textit{Selection history} \cite{2013_TrendsCognSci_Awh} or \textit{memory-based attention} \cite{2011_JoV_Tatler, 2012_Cell_Hutchinson}, \ie attentional control based on past experience and rewards, is proposed as an additional category of control mechanisms that are neither fully stimulus- nor task-driven. Although this category is not explicitly considered in driving literature, multiple studies indicate that where and when drivers look depends on their experience \cite{1998_Perception_Chapman, 2010_AccidentAnalysis_Konstantopoulos} and familiarity with the route \cite{2007_TransRes_Martens, 2008_HumanFactors_Borowsky}.

Both bottom-up and top-down control are involved in driving, but their relative contributions or how they are integrated remain open research problems (see Sections \ref{ch:psychological_models} and \ref{ch:practical} for existing theories and implementations). Experimental evidence suggests that top-down attention plays a large role in driving. For instance, drivers' visual strategies change when the priorities of driving tasks are manipulated. For instance,  drivers monitor their speedometer more often when maintaining speed is prioritized over vehicle following) \cite{2012_JoV_Sullivan}. The fact that the horizontal spread of search is not affected during nighttime driving similarly suggests that it is not dictated by the bottom-up stimuli whose influence would be reduced in the dark \cite{2010_AccidentAnalysis_Konstantopoulos}. Furthermore, change blindness depends on the relevance of objects to the task, \eg changes involving trees and buildings were missed more often than changes involving vehicles and signs \cite{2017_AccidentAnalysis_Beanland}. 

Bottom-up strategies for driving are investigated as well. Experiments show that it is possible to make stop-and-go judgments after viewing scenes with high-saliency regions masked or scenes where only high-saliency regions are visible (saliency is determined by a bottom-up algorithm \cite{2001_NatRevNeuroscience_Itti}). This is not surprising as many important objects in the scene, such as signs, traffic lights, and vehicle signals, are salient by design. At the same time, because bottom-up cues are not optimal more information from larger non-salient areas is required for better decision-making which reinforces the role of experience in locating critical information within the scene \cite{2014_ACP_Mccarley}. Low bottom-up saliency contributes to LBFTS errors involving motorcyclists and bicyclists at the intersections \cite{2008_TransRes_Crundall, 2011_VR_Underwood}. Here too, top-down strategies may mitigate some of these effects as drivers who are also experienced motorcyclists make fewer mistakes \cite{2008_TransRes_Crundall}.



\subsection{Summary}
The neurophysiological properties of vision in the central visual field differ from those in the periphery. Since sharp vision is available only within a small area in the fovea, the observed scenes are sequentially processed via eye movements towards different areas and objects. It is often presumed that during driving, most of the information is processed in the fovea; therefore, gaze can tell where the drivers looked at any given moment of time but cannot serve as a proxy of their attention. Another common assumption is that drivers' eye movements are guided primarily by the demands of the task. Thus gaze can, to some extent, shed light on their decision-making.

On the other hand, gaze has a number of limitations:
\begin{itemize}
\item{Gaze} alone does not reveal whether visual processing was successful. Ample experimental evidence shows that in many instances foveal vision does not guarantee awareness, \eg due to attention being diverted elsewhere. 
\item{Gaze} does not reveal anything about covert changes in attention and provides very limited insight into processing in the periphery which comprises over $95\%$ of the visual field. As recent studies indicate, peripheral vision, despite its limitations, plays a larger role in driving than initially thought. 
\item{Eye} movements are prompted by a complex interplay between involuntary attraction to salient areas and task-driven attention, but the relative contributions of the bottom-up and top-down factors are unknown. Besides theoretical interest, understanding internal mechanisms that result in the observed gaze distribution has practical implications, \eg with regards to the effects of the driving experience or environment on gaze allocation.
\end{itemize}

\section{How to collect drivers' gaze data?}
\label{ch:data_collection}

Besides the limitations of gaze as a proxy for drivers' attention that stem from the biological characteristics of human vision, its utility may be further affected by the data collection procedures. Driving has unique characteristics, such as the active nature of the task, dynamic environment and observer, 3D visual field, and, particularly, the risk of harming oneself or others. Unlike other visual tasks commonly studied via eye movement analysis, such as reading, visual search, and saliency, driving requires a data collection methodology that can accommodate its properties. 

\subsection{Conditions for gaze data collection} 
When collecting gaze behavior data, there are three important considerations:  \textit{active vehicle control}, \textit{physical and visual realism} of the in-vehicle and external environment, and \textit{gaze recording equipment}. Driving is a visuomotor task, meaning that there is a strong link between motor actions and gaze. Therefore, allocation of attention should be studied by analyzing fixation patterns and other gaze-related statistics coupled with motor actions. Ideally, such data should be recorded in a naturalistic setting in an actual vehicle while the person is actively engaged in the task, except when it is not safe or legal (\eg for studying hazard perception or drowsy driving). Driving simulators are often used as an alternative to real vehicles but differ significantly in the level of physical and visual realism they provide. Finally, equipment for recording gaze should have high resolution and precision without limiting the driver physically. 

\begin{figure}[t!]
\centering
%

\includegraphics[width=1\linewidth]{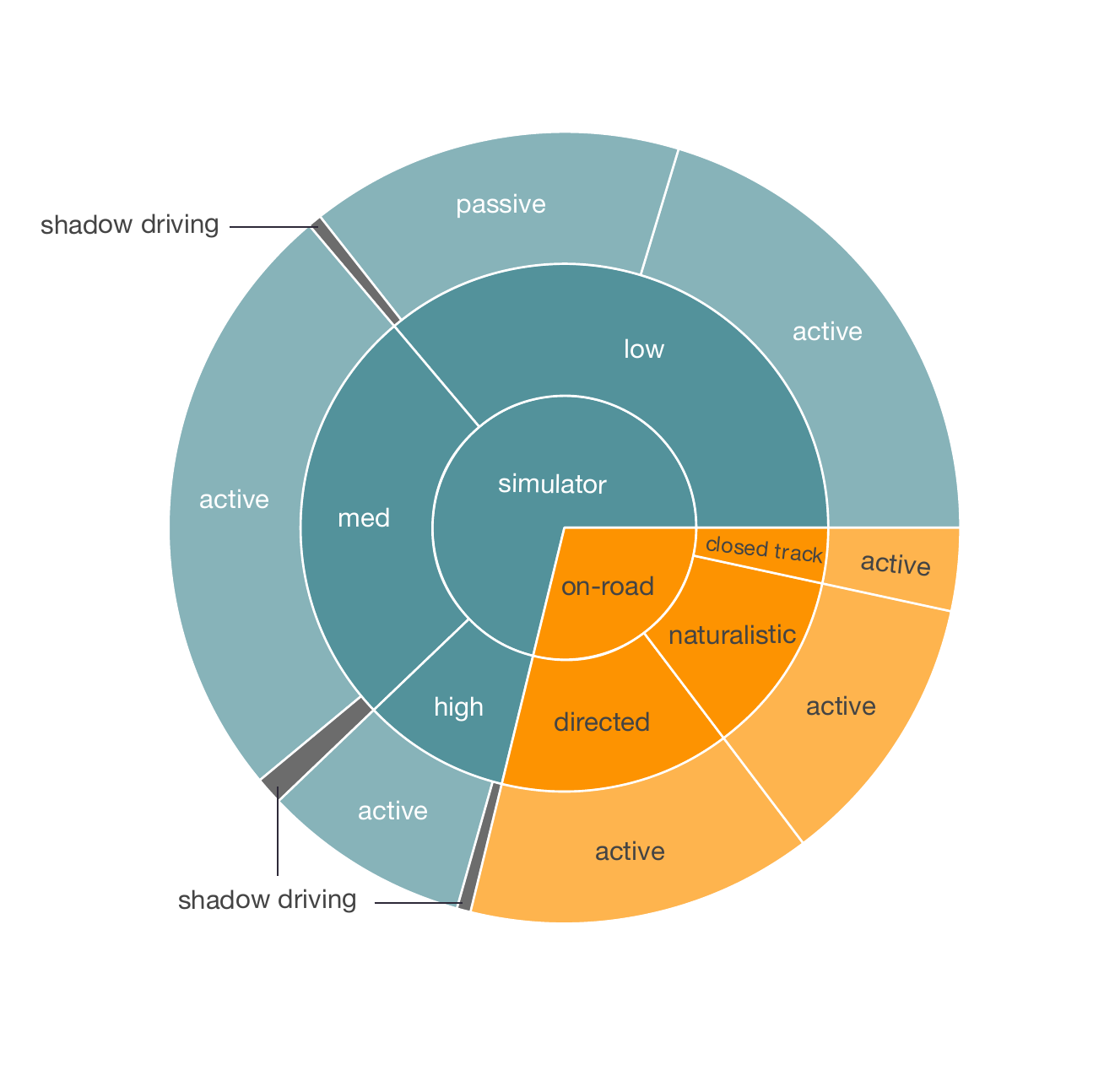}
  \caption[Data collection methods used in behavioral studies]{Data collection methods used in behavioral studies. The innermost circle represents the types of environments: driving simulator or real vehicle. The inner ring shows fidelity/realism of the environment and the outermost ring specifies whether the control of the vehicle was active (the experiment subject drove the car) or passive (the experiment subject watched a prerecorded driving video or vehicle control was automated). ``Shadow driving'' indicates that subjects used vehicle controls to mimic the prerecorded driving videos. The areas of the individual segments are proportional to the number of papers that used the corresponding data collection method.}
   \label{fig:data_collection_sunburst}
\end{figure}

The diagram in Figure \ref{fig:data_collection_sunburst} shows the relative proportions of data collection methods in the behavioral literature. It can be seen that only one-third of the studies are conducted on-road, and the rest in driving simulators. Nearly $20\%$ of all experiments are done without active control of the vehicle, instead, subjects passively view the prerecorded driving videos on the monitor. ''Shadow driving`` is a recently introduced technique where subjects are asked to use vehicle controls to match driver's actions in the recorded videos. Below we discuss these experimental setups, their advantages, and limitations.

\subsubsection{On-road studies}
All on-road studies are conducted in instrumented commercial vehicles and usually involve active vehicle control (except for studies involving highly-automated vehicles), but within this group, there are gradations in the level of realism they provide. We further subdivide on-road studies from the least to most restrictive into \textit{naturalistic}, \textit{directed}, and \textit{closed track}.

\begin{itemize}
\itemsep0em

\item{\textit{Naturalistic}} studies, \ie when participants use instrumented vehicles for their daily activities, comprise about $40\%$ of all on-road studies. While this method offers a look into real driving patterns, it is not without downsides: 1) Because driving tasks are not specified in advance, drivers' behaviors are highly variable. Hence, a large volume of data should be gathered to study rare or specific scenarios. This, in turn, leads to other issues such as processing, storing, and analyzing large volumes of data. 2) There may be geographic and population biases as the studies are conducted with a limited population of drivers, and within a certain geographic area. 3) Specifically for studies of driver attention, the data may be of insufficient quality. It is costly to instrument the vehicle with eye-tracking equipment, which requires maintenance and may suffer from data loss. As a result, nearly all large-scale naturalistic studies resort to annotating drivers' gaze from driver-facing cameras installed inside the vehicle. These issues are further discussed in Sections \ref{sec:eye_tracking_data_loss} and \ref{sec:manual_gaze_coding}.


\item{\textit{Directed}} on-road studies are also conducted on public roads but allow for more control over the experimental conditions. The drivers are given a route and may be accompanied by the researcher during the experiment to deal with technical issues as they arise. Fixed route and/or task reduces the variability of the driver's behavior, thus removing the need for large-scale data collection. However, if the vehicle used in the experiment is not owned by the participant there is a concern that lack of familiarity with the vehicle may affect their behavior \cite{2018_TransRes_Young, 2019_SafetyScience_Kuo} and that few minutes of practice in the new vehicle before the start of the experiment are not sufficient \cite{2019_AppliedErgonomics_Costa, 2018_TransRes_Kidd}. Furthermore, to reduce risk, on-road experiments are often conducted during off-peak hours \cite{2013_TrafficInjuryPrevention_Dukic, 2014_TransRes_Costa, 2018_TransRes_Young} which may bias the results. Despite these restrictions, it is still difficult to replicate conditions exactly across participants as there is inevitable variability in the environmental conditions and vehicle control. Thus ecological validity of the on-road studies imposes a loss of experimental control and data interpretation challenges \cite{2013_JoV_Lappi}.

\item{\textit{Closed track}} studies are conducted in specially built environments or isolated parking lots. Such environments have been used for experiments that would be otherwise risky to conduct on public roads (\eg hazard response during automated driving \cite{2018_HumanFactors_Victor} or parking \cite{2012_SAE_Kim}). At the same time, reduced risk and lack of interaction with normal traffic may reduce participants' sense of danger and bias their behavior.
\end{itemize}

\subsubsection{Driving simulators}
\label{sec:driving_simulators}
Driving simulators offer advantages unattainable in on-road conditions: an identical route for all subjects, the ability to trigger events at known times and to limit vehicle control variability. Besides replicating the same conditions across many subjects, some simulation experiments are more scalable and cost-efficient (for the history and extensive review of driving simulators see \cite{2011_Handbook_Caird}). Despite these desirable properties, the validity of the simulation studies, \ie whether the conclusions translate to on-road conditions, is not guaranteed and should be verified. Absolute validity, \ie matching results from the simulated and on-road studies, is preferred. In most cases, relative validity is acceptable, meaning that the trends, patterns, and effects found in the simulation study are similar to those observed on-road but differ in absolute numerical terms \cite{2011_Handbook_Caird}.

\begin{figure}[t!]
\centering
%
\includegraphics[width=0.8\linewidth]{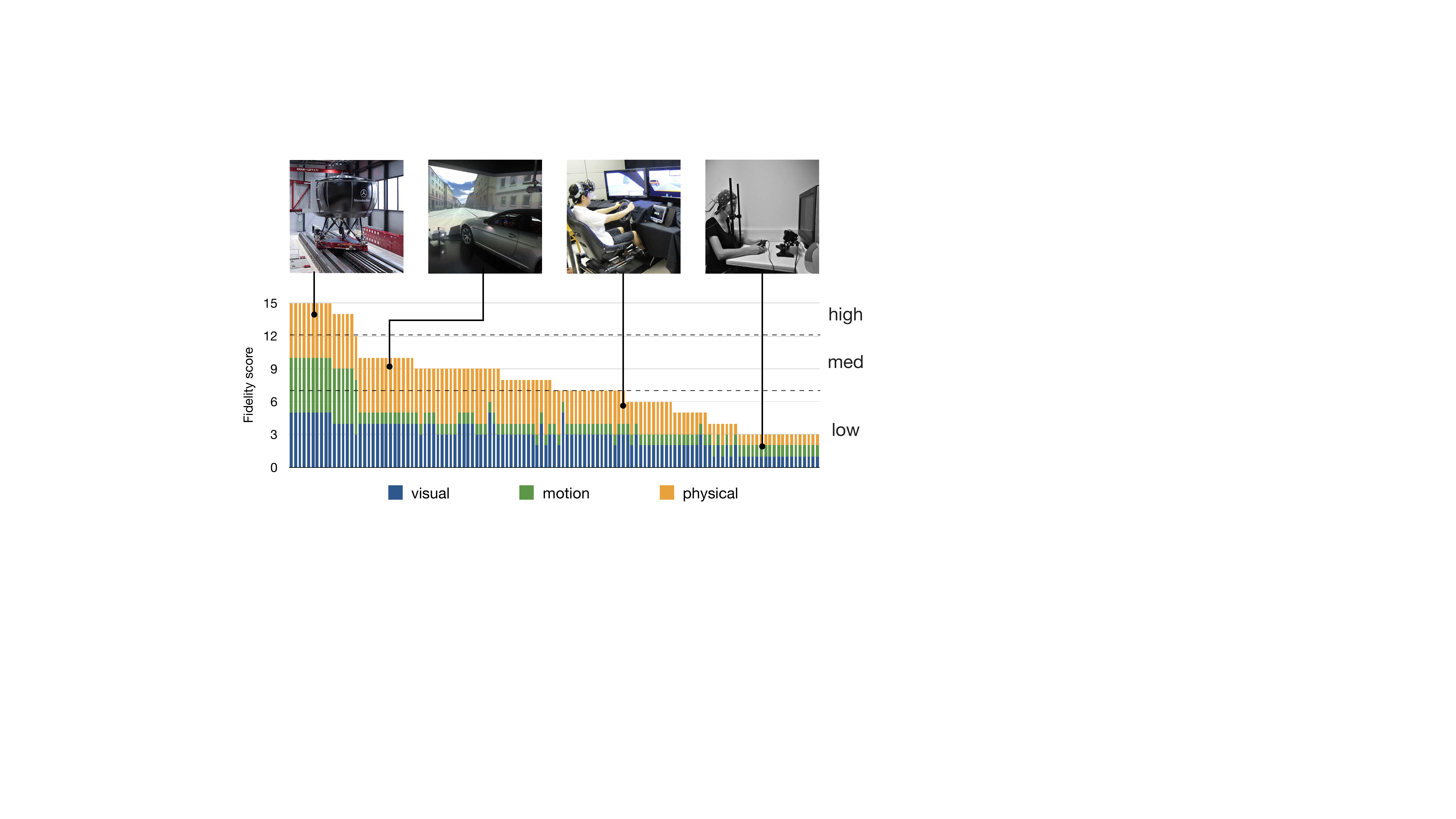}
\caption[Distribution of driving simulator fidelity scores and illustrations of various simulators]{A bar plot showing the distribution of simulator fidelity scores. Each stacked bar corresponds to a single study. Scores for visual, motion and physical fidelity are shown in blue, green, and yellow colors respectively. Based on the total score, simulators are divided into three groups: high ($12-15$), medium ($7-11$) and low ($1-6$). Examples of simulators above the bar plot (left to right): 1) high-fidelity simulator with the dome providing $360^\circ$ FOV, full cab inside the dome with a movable base, 2) medium-fidelity simulator with a full cab and no moving base with 3 projectors providing $135^\circ$ FOV; 3) low-fidelity PC-based simulation with arcade seat and 4 monitors with FOV $<180^\circ$; 4) low-fidelity simulation with a single monitor and a joystick. Sources: 1) \cite{2019_TMC_Fan}, 2) \cite{2019_TransRes_Lu}, 3) \cite{2012_TR_Kaber}, 4) \cite{2013_TransportRes_Savage}.}
   \label{fig:simulator_score_distribution}
\end{figure}

\noindent
\textbf{Simulator fidelity and validity}. The simulator fidelity defines how faithfully it can reproduce the environment inside and outside the vehicle along multiple dimensions such as visual appearance, motion, and sound. A large variety of commercial and custom-made simulators range from a PC-based simulation showing recorded driving footage on a monitor to state-of-the-art setups with $360^\circ$ field of view and an actual vehicle placed on a moving base (see Figure \ref{fig:simulator_score_distribution}). In the literature, simulators are broadly divided by fidelity level into \textit{low}, \textit{medium}, and \textit{high}, but there are no fixed criteria for how these levels are defined, to the best of our knowledge. As a result, the same simulator model may be assigned different fidelity types in different studies. For consistent fidelity assessment, we adopt the scoring system proposed in \cite{2019_SafetyScience_Wynne} where the visual, motion, and physical fidelity scores determine the fidelity of the simulator (Table \ref{tab:fidelity_scoring}). Based on the total score, simulators are divided into three groups: high ($12-15$), medium ($7-11$), and low ($1-6$). Figure \ref{fig:simulator_score_distribution} shows the distribution of the scores for simulators used in the $125$ behavioral studies reviewed.

\begin{table}[t!]
  \centering
   \resizebox{\linewidth}{!}{   
    \begin{tabular}{l|l|l|l|l|l}
    \specialcell{Score\\Fidelity type} & \multicolumn{1}{c|}{1} & \multicolumn{1}{c|}{2} & \multicolumn{1}{c|}{3} & \multicolumn{1}{c|}{4} & \multicolumn{1}{c}{5} \\
    \midrule
    Visual & PC monitor & \specialcell[l]{Projector/PC monitor\\$> 25''$ diag}  & \specialcell{$< 180^\circ$ FOV\\(PC monitors)} & \specialcell{$180-270^\circ$ FOV\\(PC monitors)} & \specialcell{$> 270^\circ$ FOV\\(projector)} \\ \hdashline
    Motion & No motion base & -     & \specialcell{$< 6 DoF$ or \\partial plate} & -     & \specialcell{Motion-base or\\full motion plate} \\ \hdashline
    Physical & Keyboard/joystick & \specialcell{PC with\\steering wheel} & \specialcell{Arcade seat with\\steering wheel} & \specialcell{Vehicular controls,\\ no or incomplete cab} & \specialcell{Full vehicular cab\\and controls} \\
    \end{tabular}%
  }%
  \caption[Driving simulator fidelity scoring table]{Simulator fidelity scoring table with criteria for evaluating fidelity of visual input, motion, and physical vehicle controls. Adapted from \cite{2019_SafetyScience_Wynne}.}
  \label{tab:fidelity_scoring}%
\end{table}%

Contrary to intuition, simulator's fidelity is not directly associated with validity and depends on what measures are being compared, as shown in the recent meta-review of validation studies by Wynne \etal \cite{2019_SafetyScience_Wynne}. Although the review focuses primarily on validating driving performance measures such as steering, speed variation, and lane maintenance, the authors noted that only non-valid results are reported for gaze measures regardless of the simulator sophistication. In our review, only a handful of results were validated, mostly across different types of simulators. For example, Kim \etal \cite{2019_HumanFactorsErgonomics_Kim} examined the effect of vehicle control on attention distribution and Mangalore \etal \cite{2019_TRR_Mangalore} showed that hazard perception using a virtual reality (VR) headset produces the same results as an advanced driving simulator. The study by Robbins \etal \cite{2019_AppliedErgonomics_Robbins} is the only one where the results obtained in an on-road experiment are compared to those from a high-fidelity driving simulator. They conclude that the absolute validity of the simulator is reached for medium- to high-demand situations. 
%

\noindent
\textbf{Visual fidelity.} Scores proposed in \cite{2019_SafetyScience_Wynne} are defined by the field of view, which has been shown to affect validity \cite{2014_TR_Alberti}. Recently, several studies proposed to use virtual reality (VR) headsets as a tool for studying driver attention. VR headsets produce $360^\circ$ immersive environment, thus eliminating the cost of projector arrays and the need to install a full vehicle cab. Besides, many models may be fitted with eye-tracking equipment. The validity of VR headsets for a variety of tasks is not yet established, as, to date, only one study validated a VR headset in a driver hazard anticipation experiment \cite{2019_TRR_Mangalore}.

One other aspect of visual fidelity, the visual realism of the simulation, is less studied. Despite significant improvements computer graphics quality over the past years, most driving simulators still look undoubtedly artificial. But does photo-realism matter at all for driving studies? So far, the studies on this are scarce, and evidence is inconclusive. For instance, Ciceri \etal \cite{2014_TransRes_Ciceri} found that experienced drivers had distinctly different gaze patterns when watching driving footage vs video game recordings. In a more recent study, Kim \etal \cite{2019_HumanFactorsErgonomics_Kim} showed that gaze behavior was the same when watching real or computer-generated videos, however, only hazard perception scenarios were considered. A validation study by Robbins \etal \cite{2019_AppliedErgonomics_Robbins} established that visual behavior at real and simulated intersections was similar as long as the task demands were at least moderate. Differences in gaze measurements were small when the driving maneuver was easy, such as going straight, and more pronounced when turning right (the experiment was conducted in the UK). 

The amount of visual detail provided by the simulated environment matters too, as shown in the study by van Leeuwen \etal \cite{2015_Ergonomics_vanLeeuwen}. During the experiment, the route remained the same for all participants while the simulated graphics was manipulated. Depending on the condition, the full environment was shown, with or without roadside objects, or only lane markings and the road center were visible. Significant differences were observed in the steering activity, lane-keeping, and speed between the highest and the lowest fidelity conditions. 

\noindent
\textbf{Physical and motion fidelity.} Driving simulators vary significantly in the appearance and authenticity of controls, from a full vehicle cab to a steering wheel mounted on the office desk to a computer keyboard. Although visuomotor coordination during driving is well-studied and strong links have been established between driving performance and visual attention measures (as will be discussed later in Section \ref{sec:vehicle_control}), there are not many investigations into changes in attention distribution without active vehicle control. For instance, it was found that drivers who controlled the vehicle scanned the environment less and looked closer to the vehicle. Their reaction time to hazards was $1-1.5$s longer than that of the subjects who were passively watching prerecorded videos \cite{2015_VC_Mackenzie}. When negotiating the curves, the drivers who were actively steering made fewer look-ahead fixations and focused on the areas that were important for visuomotor coordination \cite{2012_PLOS_Mars}. 

In some studies, to compensate for the lack of vehicle control in low-fidelity simulators, the subjects are asked to ``shadow drive'', \ie rotate the steering wheel and use pedals to match what they see in the prerecorded driving videos \cite{2020_TransRes_Li, 2014_TransRes_Ciceri}. Alternatively, drivers can use the wheel to control a digital crosshair superimposed on top of the driving video to simulate vehicle control \cite{2019_HumanFactors_Kim, 2019_HumanFactorsErgonomics_Kim}. According to these studies, the mimicking task creates a more immersive experience and increases cognitive workload in simulation, matching real driving.  Kim \etal \cite{2019_HumanFactorsErgonomics_Kim} show that ``shadow driving'' leads to the same visual attention allocation as active vehicle control.

We were not able to find research that validated eye-tracking data across intermediate levels of physical validity or measured the effect of vehicle motion on gaze patterns.

\subsubsection{Effects of laboratory setting and experiment design}

\begin{figure}[t!]
\centering
\begin{subfigure}[c]{1\textwidth}
	\begin{subfigure}[c]{0.6\textwidth}
	\centering
 	\includegraphics[width=1\textwidth]{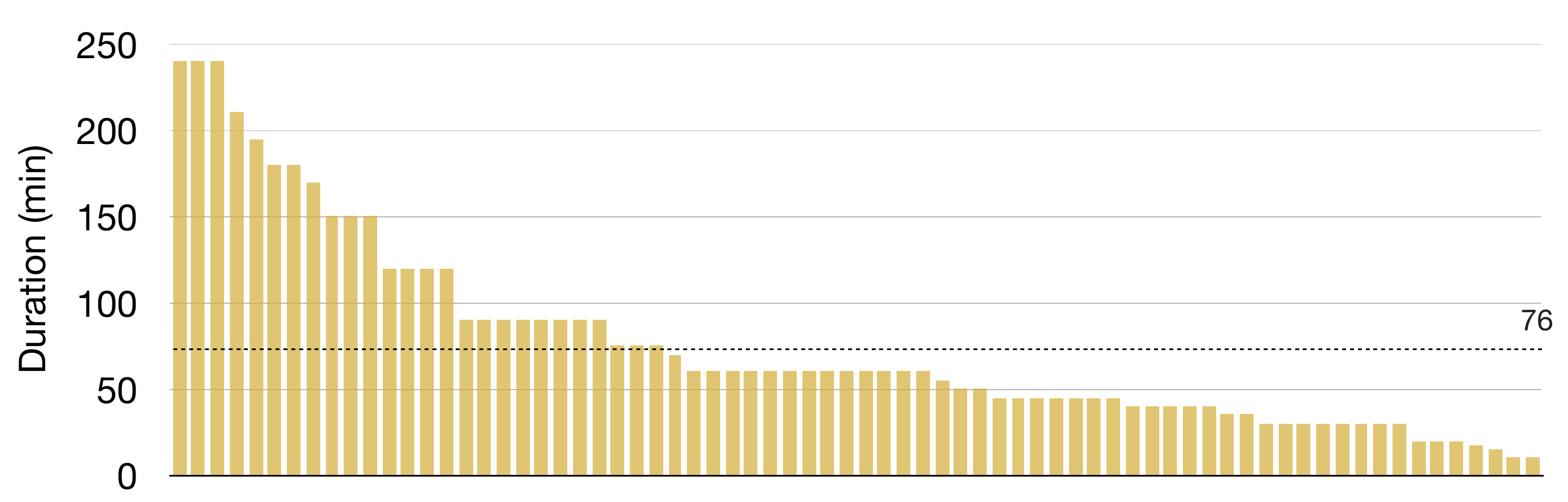}
  	\caption{Session durations}
  	\label{fig:session_durations}
	\end{subfigure}
\hspace{-0.5em}
	\begin{subfigure}[c]{0.4\textwidth}
	\centering
	 \includegraphics[width=1\textwidth]{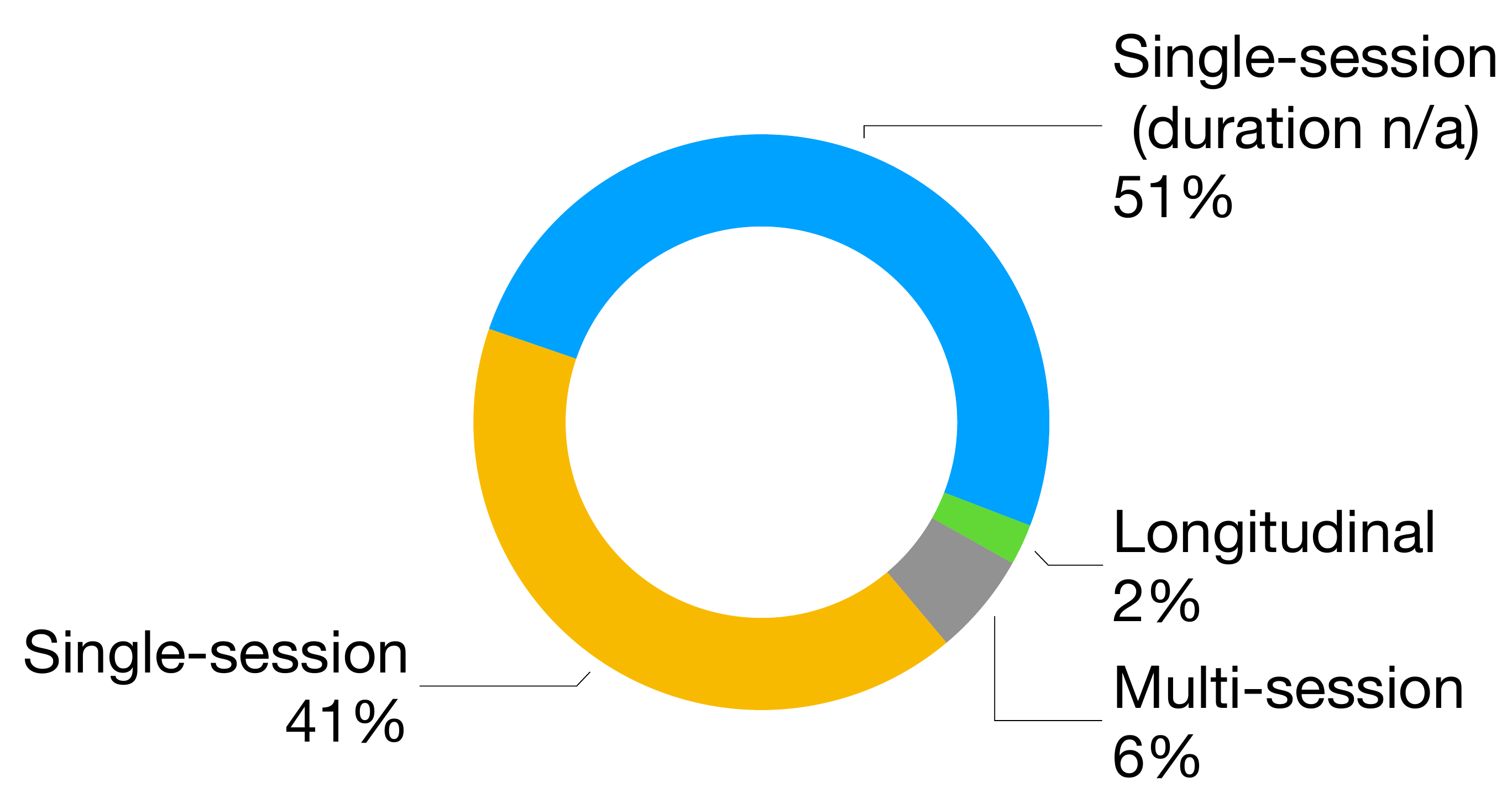}
	  \caption{Sessions}
	  \label{fig:study_types}	
	\end{subfigure}
\end{subfigure}

\begin{subfigure}[c]{1\textwidth}
	\begin{subfigure}[c]{0.6\textwidth}
	\centering
	 \includegraphics[width=1\textwidth]{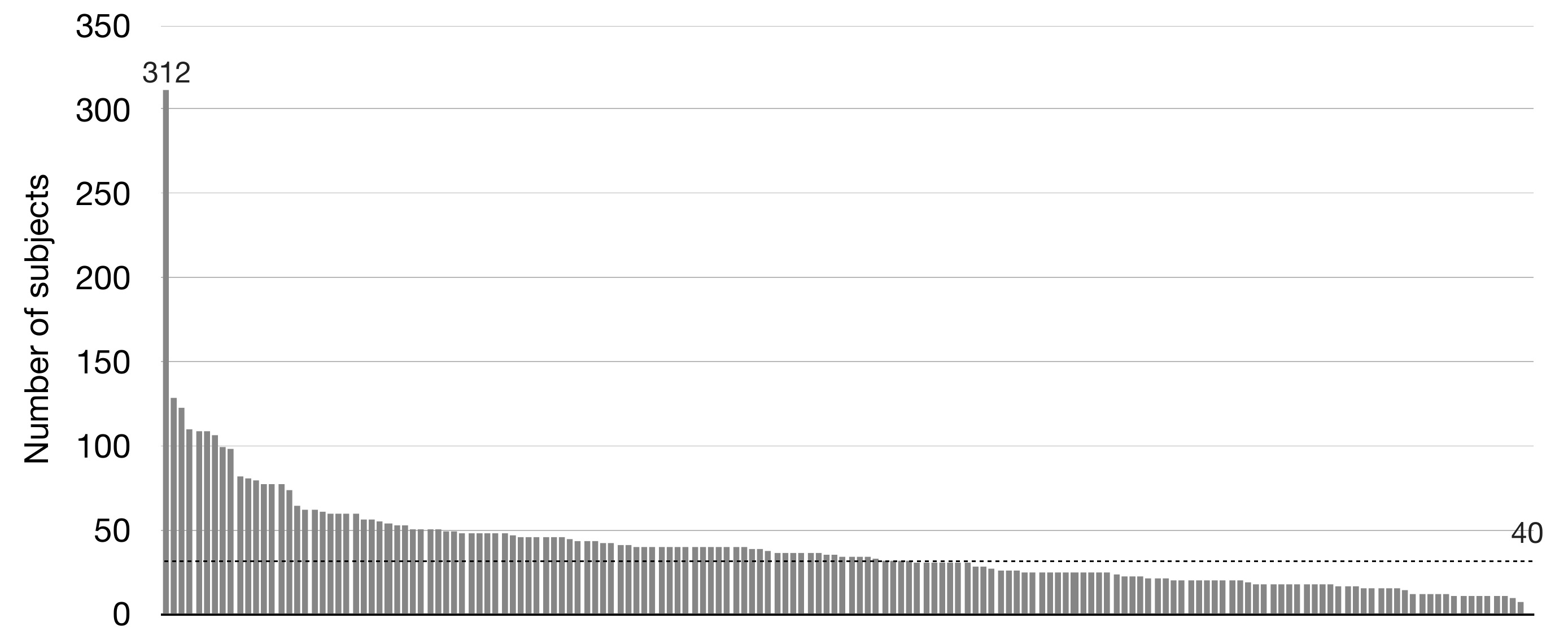}
	  \caption{Number of participants}
	  \label{fig:number_subjects}
	\end{subfigure}
\hspace{-0.6em}
	\begin{subfigure}[c]{0.4\textwidth}
	\centering
	 \includegraphics[width=1\textwidth]{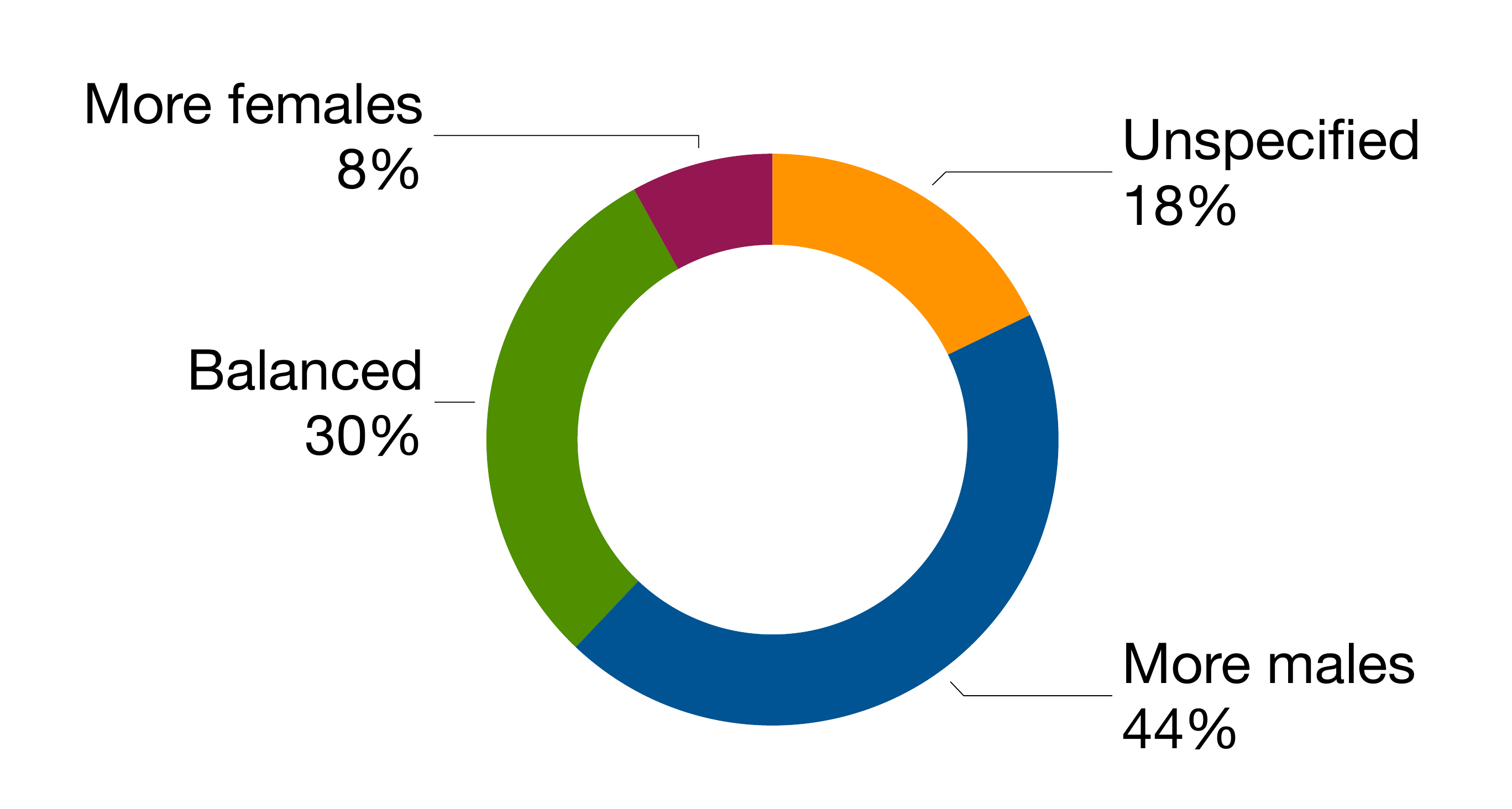}
	 \caption{Gender balance}
	 \label{fig:gender_balance}
	\end{subfigure}
\end{subfigure}
\caption[Distributions of experiment session durations, types of sessions, number of subjects per study and gender balance across studies]{a) Bar plot of single-session study durations, one bar shown per study. The dashed line indicates the average session duration. b) Pie chart showing proportions of studies that are conducted within one session, during multiple sessions, and longitudinal studies which take course over weeks or months. 51\% of the studies do not specify duration. c) Bar plot of the number of subjects, one bar shown per study. The dashed line indicates the average number of participants. 10 studies that did not specify the number of subjects are not shown. d) Pie chart showing participant gender balance across studies. Balanced studies are ones where the difference between the numbers of female and male participants is $\leq 5\%$ of the total number of participants.}
\end{figure}

Besides the environmental realism, the experimental procedure in the laboratory setting itself affects the transfer of findings to the on-road driving conditions. Below we discuss issues that include the short duration of experiments, a small number of subjects, high exposure to warning signals, and low workload and arousal levels of participants (see also \cite{2014_THMS_Ho} for a more detailed analysis of these and other factors). 

\vspace{0.5em}
\noindent
\textbf{Experiment duration}. As shown in Figure \ref{fig:session_durations}, half of the studies take just over 1 hour that may also include the instruction, post-experiment surveys or interviews, breaks between trials, \etc, so the actual time spent driving is likely shorter. It was observed in the 100-Car Naturalistic Driving Study that during the first hour, participants were more attentive and involved in fewer incidents \cite{2006_TechRep_Dingus}. Thus, 5-10 minutes given to participants to familiarize themselves with the vehicle or simulator before experimental drives might not be sufficient \cite{2016_TransRes_Fitzpatrick}. Only $6\%$ of the studies involve multiple sessions spread across different days, \eg study on the effect of hazard perception training for young drivers \cite{2011_DrivSymposium_Taylor} and the experiment on the effect of route familiarity conducted over 28 sessions \cite{2018_TransRes_Young}. Longitudinal studies where information is collected over several weeks \cite{2019_SafetyScience_Kuo} or months \cite{2014_JAH_SimonsMorton} comprise only $2\%$ of all studies we reviewed. 

\vspace{0.5em}
\noindent
\textbf{Prevalence effects.} In simulated studies, the frequency of events does not correspond to their real-world distribution \cite{2005_Nature_Wolfe}. For example, during a single hazard perception session, drivers may be exposed to tens or even hundreds of instances of events that are exceedingly rare in real driving environments. In addition to that, drivers are often told to anticipate hazards or are primed to expect them based on the preceding trials. Thus, there is a concern that such conditions may artificially inflate their performance \cite{2020_JEP_Wolfe}.

\vspace{0.5em}
\noindent
\textbf{Lower workload/arousal} is an issue, particularly for studies that do not involve any vehicle control and require the subjects to passively view prerecorded videos. Since the participants' actions do not affect the behaviors of other road users, there is no risk involved. Therefore, participants may perform better on visual search tasks, \eg pedestrian detection \cite{2019_TransRes_Chen}. When active control is required, simulator studies usually provide optimal driving conditions, with reduced traffic and good weather \cite{2014_HumanFactorsErgonomics_Schieber, 2018_NatSciReports_Shiferaw}. Lower workload may also result from the visually impoverished simulated environment, even in high-fidelity simulators \cite{2019_AppliedErgonomics_Robbins}.  

\vspace{0.5em}
\noindent
\textbf{Number and diversity of participants.} Most of the studies involve a small number of participants. As shown in Figure \ref{fig:number_subjects}, the largest study involved $312$ subjects, while the smallest study used only $1$, with $40$ subjects on average. The majority of the studies use more males and only $30\%$ of the studies are balanced across gender. Demographics of the participants are often limited to university students or car company employees who are more likely to be exposed to and have a positive attitude towards new technologies, affecting conclusions of studies involving automation \cite{2020_Information_Feierle}.

\subsection{Recording drivers' gaze} 

\subsubsection{Eye trackers} 


Unlike the early days, where custom-built setups were common, most experimental work was done using commercially available eye-trackers with different capabilities and costs in the past decade. Below we will discuss properties of eye-trackers relevant to the studies in the automotive domain: \textit{type of the eye-tracker}, \textit{sampling rate} (measured in Hz), \textit{accuracy}, and \textit{precision}.

\begin{itemize}
\itemsep0em
\item{\textit{Type of the eye-tracker.}} Eye-trackers are subdivided into static (remote and tower-mounted) and head-mounted. Naturally, eye-trackers that restrict head movement offer a better quality of data. For instance, tower-based eye-trackers have high precision and support sampling rates up to 2000 Hz \cite{2017_Eye_Link_Specs} but require the use of a chin rest. Modern head-mounted eye-trackers allow for normal head movement and support reliable head tracking and sampling rates up to 120 Hz. Remote eye-trackers are the least intrusive ones since the image of the driver's face and eyes is analyzed via camera(s) installed on the vehicle dashboard. According to \cite{2011_OUP_Holmqvist}, tower-mounted systems have the highest accuracy and precision, followed by head-mounted and remote eye-trackers. 

\begin{figure}[t!]
\centering
 	\includegraphics[width=0.7\textwidth]{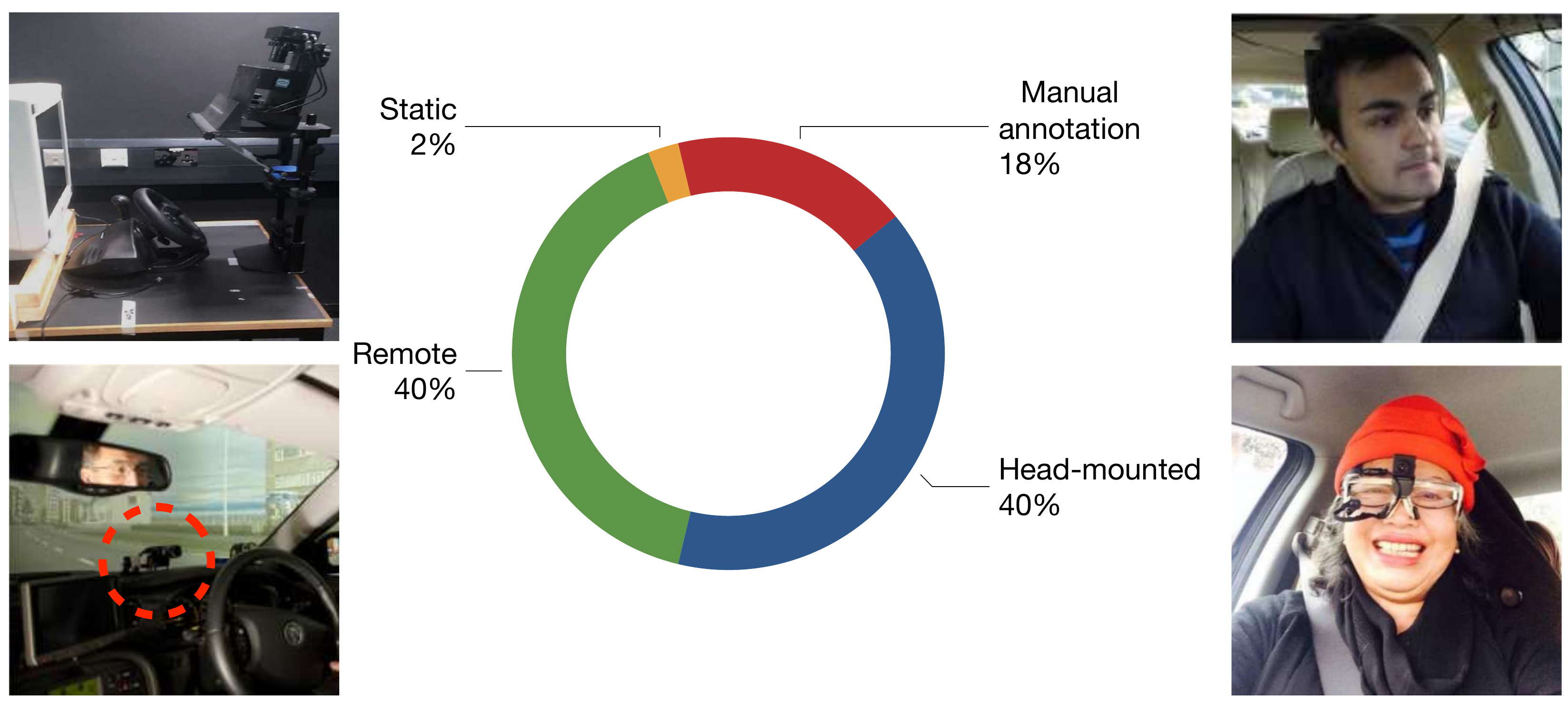}
  	\caption[Types of eye-trackers used in the studies]{Pie chart showing the distribution of different eye-trackers used in the studies. Illustrative examples of each eye-tracker type are shown next to labels. Sources: static \cite{2015_VC_Mackenzie}, manual annotation \cite{2018_TIV_Vora}, head-mounted  \cite{2018_AccidentAnalysis_Sun} and remote \cite{2013_TR_Jamson}.}
  	\label{fig:eye_tracking_types}
\end{figure}

As shown in Figure \ref{fig:eye_tracking_types}, only a few ($2\%$) studies use static tower-mounted eye-trackers. Such experiments require passively watching driving videos since most driving tasks cannot be performed without some head movement. Less invasive head-mounted and remote trackers are used equally often (each in $\approx 40\%$ of studies), and the remaining $18\%$ of works rely on manual coding of gaze from videos (as will be discussed in Section \ref{sec:manual_gaze_coding}).

Despite the growing number of commercially available eye-trackers, they are rarely compared head-to-head on the same task. One such study by Funke \etal \cite{2016_HumanFactorsErgonomics_Funke} evaluated multiple remote eye-trackers and reported on the calibration success rate, accuracy, and precision metrics. The authors concluded that calibration was severely affected when subjects wore eyeglasses and that the quality of eye-tracking data degraded with eccentricity (measured as the distance from the center of the screen). There are no comparative studies of eye-trackers relevant to the driving task to the best of our knowledge. Meta-studies such as \cite{2017_IEEEAccess_Kar} aggregate the results from multiple sources but the exact numerical comparisons are impossible due to variations in the reported metrics and task conditions.

\item{\textit{Sampling rate}} of 30-60 Hz is common as this is also a typical resolution of video cameras. What rate is sufficient or optimal depends on what is being measured. For instance, one of the primary metrics used in many studies of drivers' attention is fixation duration  (see Section \ref{sec:position_measures}). When sampling at $50$ Hz, there will be a $20$ ms window between samples, meaning that the fixation duration may be over- or under-estimated at most by this amount. At $250$ Hz this window shrinks to $2$ ms. When measuring fixation duration, the low sampling frequency may be mitigated by simply collecting more data, but for other common measures such as saccadic velocity and acceleration, a higher sampling frequency is a must \cite{2011_OUP_Holmqvist}. In the studies we reviewed, there appears to be no relationship between sampling rate and amount of data gathered (in terms of the number of participants and/or the duration of experiments), however, studies involving measurements of saccades do tend to use higher sampling rates ($>60$ Hz).

\item{\textit{Accuracy and precision}} measure the quality of the eye-tracking data. The accuracy of the eye-tracker is defined as the average difference between the actual position of the stimulus and the measured gaze position. \textit{Precision} reflects how reliably it can reproduce repeated gaze measurements.

Most commercial eye-trackers provide accuracy and precision for various conditions and gaze angles. Typical accuracy in optimal conditions is within $0.25-0.5^\circ$ of visual angle for static and head-mounted eye trackers, and up to $1^\circ$ for remote eye-trackers. Precision is rarely reported; where available it ranges between $0.01-0.05^\circ$ root mean square (RMS) error. Both precision and accuracy degrade significantly in low illumination conditions and for gaze angles beyond $15^\circ$ 

Most in-lab experimental procedures include drift checks before each trial and recalibration of the equipment during breaks to ensure good recording quality. In on-road studies, calibration is typically done once before each driving session. A considerable length of some on-road experiments (\eg 25 min \cite{2018_TransRes_Young}, 40 min \cite{2013_TrafficInjuryPrevention_Dukic, 2016_JTSS_Grippenkoven}) and suboptimal on-road conditions with changing illumination and frequent large gaze angles \cite{2014_AccidentAnalysis_Lehtonen} make it more challenging to maintain data recording quality. Only a few on-road studies incorporate stops for recalibration \cite{2015_PONE_Itkonen, 2016_PLOS_Cheng} or trigger recalibration automatically \cite{2013_TITS_Ahlstrom}.
\end{itemize}

\subsubsection{Manual coding of gaze}
\label{sec:manual_gaze_coding}

\begin{figure}[t!]
	\centering
	 \includegraphics[width=0.6\textwidth]{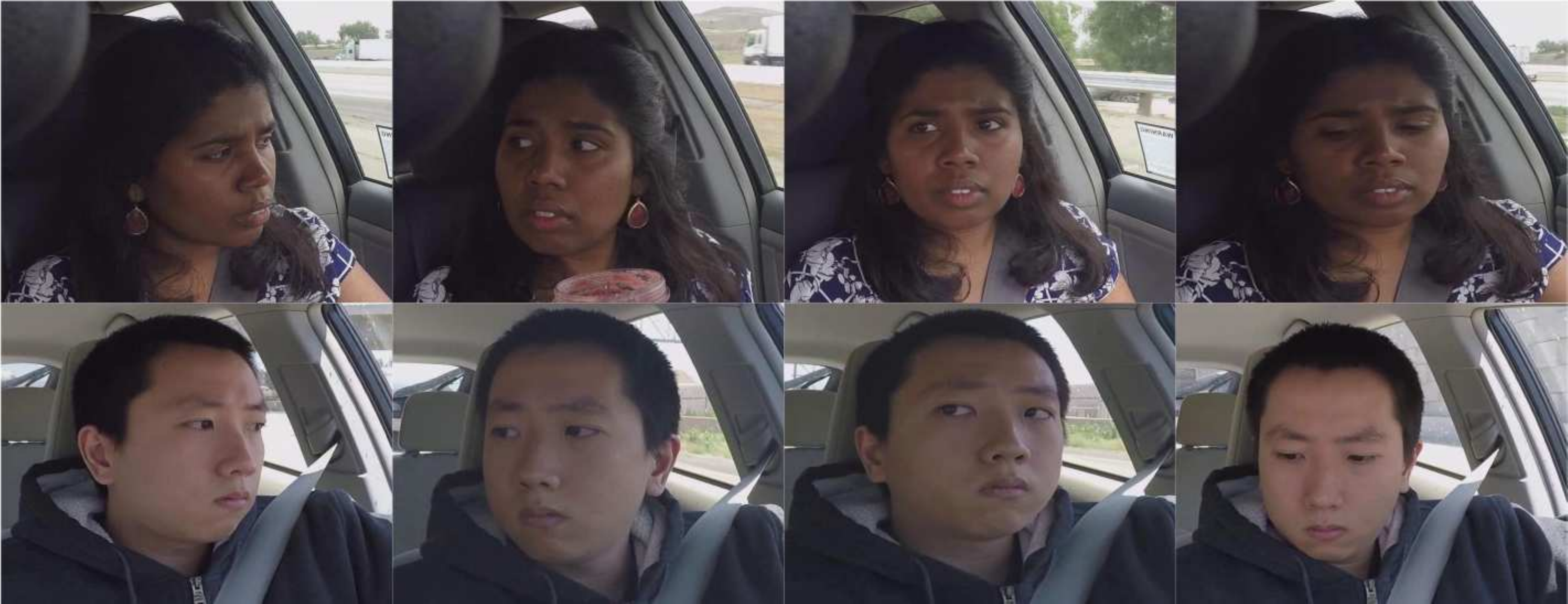}
	  \caption[Example images from driver-facing cameras]{Images from driver-facing cameras. Drivers are looking at (from left to right): left mirror, right mirror, rearview mirror and radio. Source: \cite{2016_ITSC_Vasli}.}
	  \label{fig:manual_gaze_coding}
\end{figure}

Manual coding of gaze is the process of annotating each frame of the video from the driver-facing camera with a text label specifying in what approximate direction the driver is looking at the moment, \eg rearview mirror, straight ahead, or eyes closed (see definitions of areas of interest in Section \ref{sec:AOI}). It is more common in naturalistic driving studies due to the low cost, low maintenance, and non-intrusiveness of the cameras. At the same time, human annotation of the data is a labor-intensive process, often requiring multiple trained annotators to reduce subjectivity \cite{2016_TransRes_Munoz, 2015_TR_Victor, 2014_HFES_Liang}. Due to subtle differences in drivers' appearances depending on what area they focus (\eg moving eyes down to switch between the road ahead and speedometer) and large individual differences between drivers (as shown in Figure \ref{fig:manual_gaze_coding}), consistent annotations are difficult to obtain, thus typical inter-rater reliability reported in the papers is around 90\% \cite{2006_TR_Klauer,2014_TransRes_Costa}.

Another aspect is the sampling rate of the driver-facing cameras which can be as low as 10 Hz \cite{2018_TITS_Morando, 2017_AccidentAnalysis_Seppelt, 2014_TransRes_Tivesten, 2014_HFES_Liang, 2013_AccidentAnalysis_Peng, 2011_AccidentAnalysis_Owens} and rarely exceeds 30 Hz (the most common frame rate for commercial cameras). Thus, manual annotation of fixations may miss many fine details. For instance, eye movements occur at a high rate of once every 0.2 to 0.5 seconds \cite{2012_SAE_Kim}, and at 10 Hz many short fixations are missed, increasing bias towards more prolonged glances. It is also difficult to establish a gaze direction precisely from the video, hence gaze is coded with respect to coarsely defined areas of interest \cite{2014_TransRes_Tivesten} (see Section \ref{sec:AOI}).

\subsubsection{Data loss} 
\label{sec:eye_tracking_data_loss}

\begin{figure}[t!]
\centering
\begin{subfigure}{0.29\linewidth}
\centering
 \includegraphics[height=1in]{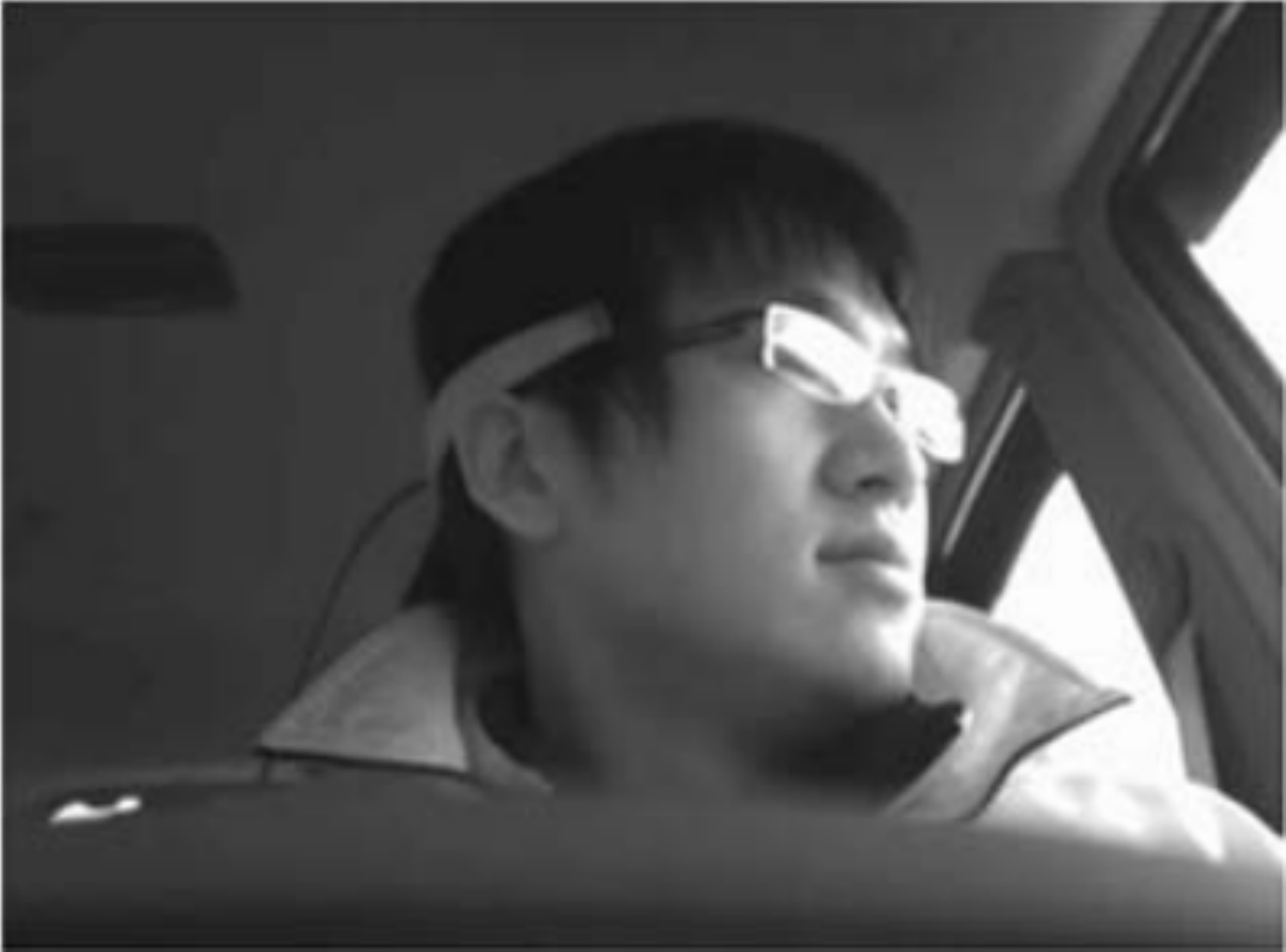}
  \caption{Reflection from eyewear}
\end{subfigure}
\begin{subfigure}{0.29\linewidth}
\centering

 \includegraphics[height=1in]{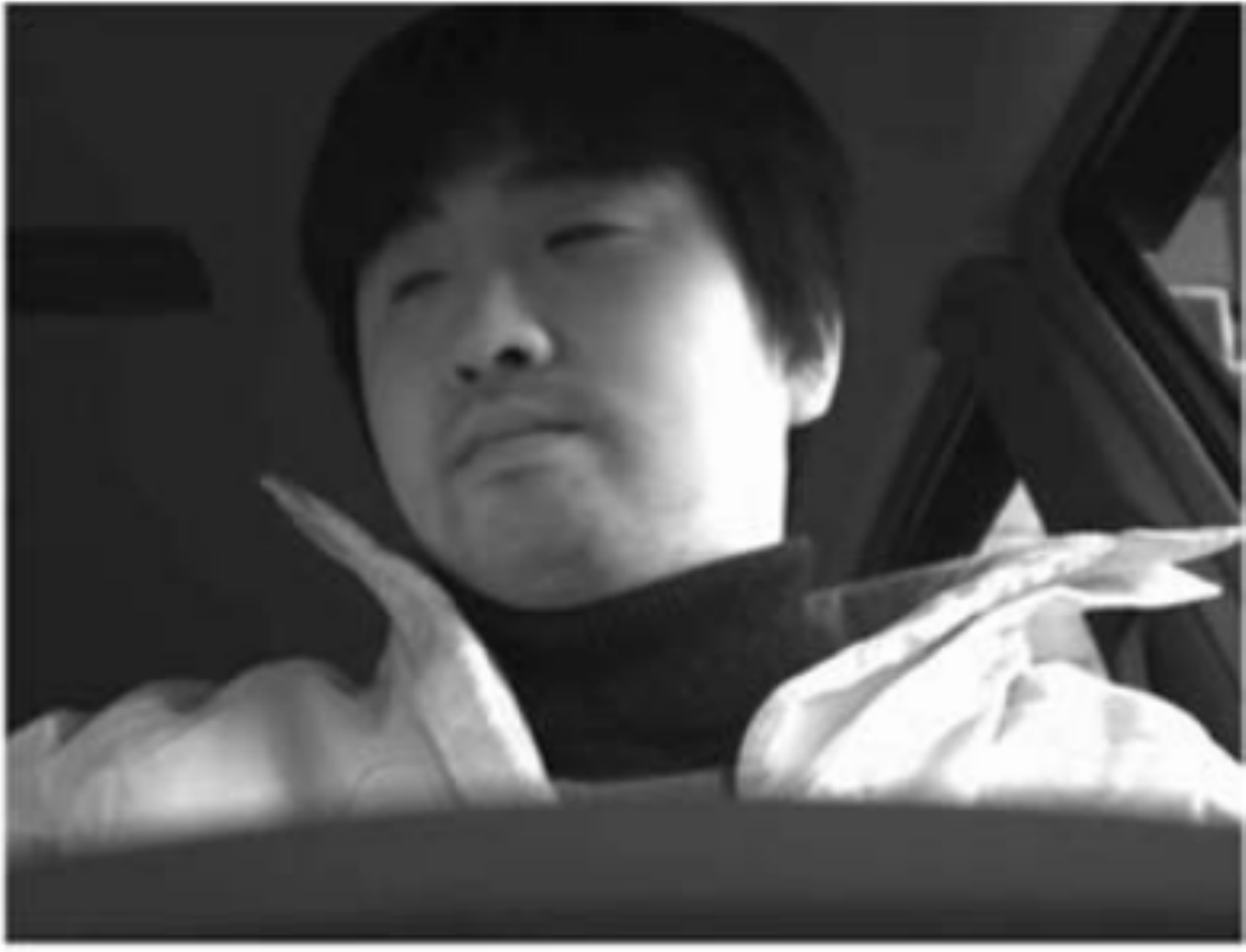}
  \caption{Blink}
\end{subfigure}
\begin{subfigure}{0.29\linewidth}
\centering

 \includegraphics[height=1in]{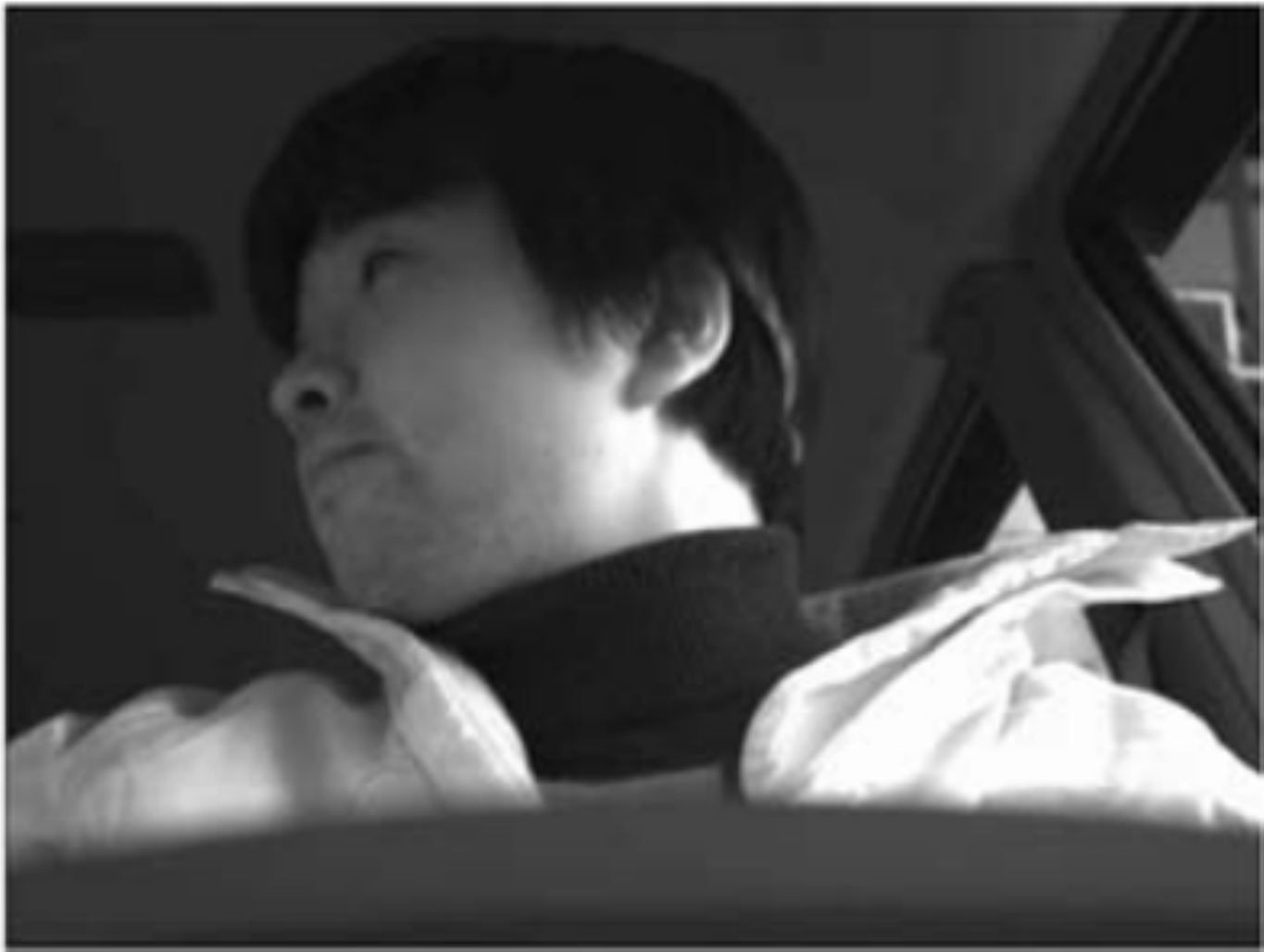}
  \caption{Large head angle}
\end{subfigure}
\caption[Samples from driver-facing camera where driver's gaze could not be estimated]{Examples of the common issues leading to data loss. Source: \cite{2011_TITS_Lee}.}
   \label{fig:data_loss_examples}
\end{figure}

More than $12\%$ of the reviewed behavioral studies report issues with gaze data collection in both laboratory and naturalistic settings. The reasons include the inability to calibrate the eye-tracker \cite{2014_TR_Alberti, 2016_TR_Stavrinos} (especially at far eccentricities \cite{2018_HumanFactors_Robbins}), large tracking errors up to $3^\circ$ \cite{2015_PONE_Itkonen}, and significant data loss from $10\%$ up to $50\%$ \cite{2010_HumanFactorsErgonomics_Reimer,2016_AccidentAnalysis_Kountouriotis,2014_TR_Merat,2013_TrafficInjuryPrevention_Dukic,2013_Ergonomics_Lehtonen,2019_TransRes_Clark}. Changes in the illumination and presence of eye-wear also affect studies that rely on manual annotation of gaze from videos (see Figure \ref{fig:data_loss_examples}). For example, in \cite{2018_TransRes_Kidd} it is reported that over $25\%$ of recorded data was not suitable for coding due to glare, and in \cite{2014_TransRes_Tivesten, 2013_AccidentAnalysis_Peng, 2010_TR_Klauer} the videos with subjects wearing sunglasses were not used.

Data loss is an issue not only for research, since it effectively reduces the participant pool, but also for practical applications that depend on eye-tracking data, such as driver monitoring systems. Sun glare, the presence of facial hair, glasses, and baseball caps, as well as physical occlusions (\eg driver's hand), can result in inaccurate or missing data \cite{2013_AccidentAnalysis_Peng, 2014_TransRes_vanLeeuwen, 2016_AutoUI_Hurtado}. For example, in a field test of the distraction algorithm \cite{2013_TITS_Ahlstrom}, data was of sufficient quality only for $77\%$ of the total distance driven by the participants.

\subsection{Summary}
The fidelity of the environment and the type of eye-tracking equipment determine how the drivers' gaze is recorded. Realistic surroundings and vehicle controls offered by on-road studies often come at the cost of losing control over subjects' behaviors and other aspects of the experiment. On the other hand, simulation studies allow the same scenario to be replicated for participants, which is useful for studying specific behaviors. Still, validation is needed to ensure that the results translate to on-road conditions. 

The choice of equipment for recording gaze matters. Eye-trackers, remote or head-mounted, can precisely capture where the driver is looking and allow for free head movement, however most models are expensive and require maintenance, limiting their use in large-scale naturalistic driving studies. As an alternative, approximate attended locations can be derived from images of drivers' faces captured by the driver-facing camera via labor-intensive manual annotation. 

Significant data loss may result from eye-tracker failures, illumination changes, the presence of eye-wear, and occlusions. The design of the experiments should account for such events by carefully selecting suitable equipment for the given conditions and increasing the number of participants.

\section{Measuring drivers' attention}
\label{ch:attention_measures}

There are four types of movements that human eyes can make: 
\begin{enumerate}
\itemsep0em
\item{\textit{Saccades}} or fast eye movements to change gaze from one location to another;
\item{\textit{Stabilizing movements}} stabilize gaze against head and body motion (vestibulo-ocular reflex) and retinal slip (optokinetic reflex); 
\item{\textit{Smooth pursuit}} movements align fovea with targets that move at speeds not exceeding $15^\circ$s$^{-1}$ (saccades are used at higher velocities); 
\item{\textit{Vergence}} movements help focus on objects at different distances \cite{2011_OUP_Liversedge}. 
\end{enumerate}

Research on driving and attention focuses predominantly on the episodes when the gaze is stationary, referred to as \textit{fixations}, and transitions between them or \textit{saccades}. Other types of eye movements are rarely considered. Although vergence may reveal information about the 3D location of gaze \cite{2011_OUP_Liversedge} or drowsy state of the driver \cite{1994_TechRep_Wierwille}, detecting this type of eye movements is difficult in the driving environment due to changing lighting conditions \cite{2019_VisionResearch_Hooge}. Likewise, microsaccades are not studied in the driving domain due to challenges in recording them outside of laboratory conditions, as well as lack of understanding of their underlying neurological mechanisms and function in spatial attention \cite{2016_VisionResearch_Poletti}.

Raw gaze data recorded via eye-trackers is a set of timestamps and corresponding 2D coordinates. Although it is possible to analyze raw data, further processing is necessary to extract fixations and saccades and alleviate noise issues. Glance is a coarser unit, usually obtained by manual coding of gaze from videos ,but can also be derived from eye-tracking data. Gaze coordinates, fixations, and glances are often assigned to predefined areas of interest (AOI) inside or outside the vehicle. Multiple measures derived from these data points can be used to examine various aspects of the driver's attention allocation, \eg how long and how frequently the driver was looking at specific areas. Below we discuss data preprocessing and standard attention measures in detail.

\subsection{Preprocessing raw eye-tracking data} 
\label{sec:data_preprocessing}
\subsubsection{Fixation definitions across studies}


Fixation is defined as a brief period of time lasting from a fraction of a second and up to several seconds when eyes are held steady at a single location. In reality, eyes rarely remain completely still due to small eye movements such as drifts, tremors and microsaccades (also called fixational saccades) \cite{2011_OUP_Liversedge}. Even though research on microsaccades has seen a steady rise, they are not addressed in the driving literature to the best of our knowledge. 

Fixation extraction from eye-tracking data is done based on several parameters: \textit{fixation duration threshold}, \textit{gaze velocity}, and \textit{dispersion} (the spread of gaze coordinates that are aggregated into a single fixation). Since microsaccades cannot be distinguished from saccades based on physical parameters (amplitude or velocity) \cite{2008_JoV_Otero-Millan}, care should be taken when determining fixation and saccade detection settings. Only some of the behavioral and virtually none of the practical studies specify what fixation duration cut-off was used and even fewer mention velocity or dispersion thresholds set for fixation detection algorithms. Values for these thresholds vary significantly from study to study, from as low as $50$ ms \cite{2012_JoV_Sullivan} and up to $333$ ms \cite{2013_TR_Romoser}. In some cases, the authors use the default settings provided by the eye-tracking software, \eg $60$ ms (Tobii \cite{2019_AppliedErgonomics_Robbins}), $75$ ms (Fovio \cite{2019_TransRes_Hashash}), or $80$ ms  (BeGaze SMI software  \cite{2013_TrafficInjuryPrevention_Dukic, 2014_JEMR_Lemonnier, 2015_TR_Lemonnier, 2015_TR_Eyraud, 2019_DrugAlcoDependence_Shiferaw}). The most common setting is $100$ ms ($3$ frames in a video recorded at $30$ fps \cite{2010_AccidentAnalysis_White, 2013_ACP_Garrison} or 1 frame at $10$ fps \cite{2017_JSR_Wang}), however proper justification for this choice is rarely given. Holmquist \etal \cite{2011_OUP_Holmqvist}, often cited in this regard, use $100$ ms threshold as an example but never claim that it is an appropriate choice for all purposes. 

Since driving is a dynamic task with continuous motion in the scene, lower fixation durations are used in some studies. For instance, Costa \etal \cite{2019_AppliedErgonomics_Costa} suggest using a low threshold of $66$ ms arguing that in dynamic visual scenes short fixations and saccades account for a significant portion of all fixations ($5.9\%$) and almost as many as $100$ ms fixations which comprise $7.9\%$ of the total.

Some studies use a higher threshold of $200$ ms \cite{2018_TR_Vlakveld,2014_TransRes_vanLeeuwen}. The justification for this choice is the work by Velichkovsky \etal \cite{2002_TransRes_Velichkovsky}. The authors found that fixations below $90$ ms result from large saccades, are followed by very small saccades, and can be interpreted as stops on the way of correcting the eye position. Fixations of $90-140$ ms produce large saccades of over $4^\circ$, beyond the parafoveal region of the retina, \ie aim at blobs not individualized objects - the case for preattentive processing. Longer fixations $140-200$ ms are related to focal processing and initiate saccades within the parafoveal region where objects are easily seen and continuously attended.

Besides duration, a wide range of dispersion thresholds are used in the literature: $0.5^\circ$ \cite{2019_AppliedErgonomics_Robbins}, $1^\circ$ \cite{2016_JEMR_Sun, 2011_TR_Metz, 2013_AccidentAnalysis_Borowsky, 2010_AccidentAnalysis_Borowsky,2016_JARMAC_Wood}, $1.6^\circ$ \cite{ 2015_OPO_Lee,  2016_OptometryVisionScience_Lee}, $2^\circ$ \cite{2014_TransRes_vanLeeuwen, 2015_BMCGeriatrics_Urwyler}, $3^\circ$ \cite{2018_TR_Vlakveld}, $4.6^\circ$ \cite{2015_TR_Lemonnier, 2015_TR_Eyraud} and $5^\circ$ \cite{2019_TransRes_Louw}. In the vision literature, $1^\circ$ amplitude threshold is considered as a practical upper threshold for involuntary fixational microsaccades, therefore high dispersion threshold may overestimate fixation duration and exclude voluntary saccades \cite{2011_OUP_Liversedge}. Velocity thresholds are very rarely specified and also vary significantly from $35$ deg/s \cite{2012_JoV_Sullivan, 2014_TR_Alberti} to $100$ deg/s \cite{2018_AppliedErgonomics_Zahabi}. 

Another difficulty in properly identifying fixations is the driving environment itself, where everything is in relative motion. Smooth pursuit eye movements (when gaze follows a moving target) are common in such settings, but many fixation detection algorithms included with eye-tracking software do not identify them and provide inaccurate fixation data. One could use the unprocessed raw data, however, it may be noisy due to vehicle vibration in on-road conditions \cite{2015_PONE_Itkonen}. To ensure correct fixation detection, some authors modify the experiment to reduce such eye movements \cite{2016_JEMR_Sun}, and others resort to manual inspection of eye-tracking data to select appropriate settings \cite{2014_AccidentAnalysis_Lehtonen,2013_TR_Romoser} or devise custom algorithms \cite{2015_PONE_Itkonen}.

\textbf{Why are these settings important?} Holmquist \etal \cite{2011_OUP_Holmqvist} provide ample evidence that dispersion, velocity, and duration settings used for fixation detection algorithms can significantly change the distribution of the data,  affecting averages and variances (which are used to compute most measures) and variance-based tests commonly found in the literature. Consequently, under-specification of these settings makes the results difficult to reproduce and renders studies uncomparable.  Especially in cases when the effects reported for fixation measures are small, fixation detection thresholds may affect the conclusions of the study, therefore proper justification of the chosen settings and analysis of their effect should ideally be provided.

\subsubsection{Glance definitions across studies}
Glance is a coarser unit of analysis derived either from the raw gaze data or from detected fixations and saccades. Unlike fixations, which refer to a short time when the eyes remain stationary, glance refers to gaze maintained within some area of interest that is typically larger than the foveal region and requires more than one fixation and saccade to view. ISO 15007 standard \cite{2020_ISO} used in many studies \cite{2013_AppliedErgonomics_Zhang, 2013_TR_NHTSA, 2014_TransRes_Birrell, 2014_TR_Reimer, 2016_AccidentAnalysis_Belyusar, 2014_TransRes_Tivesten, 2015_TR_Victor, 2019_JEMR_Ojstersek} defines glance as starting from the moment the gaze moved toward an area of interest to the moment it moved away, \ie it includes the transition time to the area in addition to all fixations and saccades within the area. Alternatively, in some studies, a glance is defined as a sum of consecutive fixations \cite{2018_TransRes_Kraft} or gaze data points \cite{2017_AppliedErgonomics_Ahlstrom} excluding the saccades, which under-estimates glance duration.

\subsubsection{Areas of interest} 
\label{sec:AOI}
As mentioned above, fixation or raw gaze data are often aggregated by regions or areas of interest (AOI). There are many alternative terms in the literature for AOI: region of interest (ROI) \cite{2012_AccidentAnalysis_Borowsky}, important areas \cite{2012_AccidentAnalysis_Werneke}, visual scanning areas \cite{2017_JSR_Wang}, driver interest regions \cite{2016_PLOS_Cheng}, gaze zones \cite{2017_IV_Martin, 2016_TITS_Lundgren, 2014_IV_Tawari}, gaze regions \cite{2016_IET_Fridman}, glance areas \cite{2014_ITSC_Pech}, and fixation areas \cite{2012_TITS_Jimenez}. The meaning of all these terms is the same -- predefined areas in the scene, which can be either \textit{static} or \textit{dynamic}. Below we consider some of the most common AOIs in the literature.

\noindent
\textbf{Static AOIs} are usually defined with respect to the driver's frame of reference. Dividing gaze into \textit{on-road} and \textit{off-road} is the most basic way of measuring inattention and is by far the most popular. There are differences across studies in how the forward roadway area is defined precisely (and whether it is defined at all). Depending on the study, it may refer to the entire windshield \cite{2017_AdvErgonomics_Feldhutter, 2019_TransRes_Hashash} or only part of it in front of the driver \cite{2017_AccidentAnalysis_Seppelt, 2014_ITSC_Tawari, 2014_ITSC_Pech}. A common approach is to define a circular area around the \textit{road center} represented by the mode of driver's gaze location computed over a certain time interval or with a fixed sliding window. The radius for the circle can be $6^\circ$ \cite{2018_TIV_Wang, 2013_TR_Jamson}, $8^\circ$ \cite{2018_TITS_Morando, 2018_HumanFactors_Victor, 2018_AccidentAnalysis_Kircher, 2014_SafetyScience_Young} or $10^\circ$ \cite{2012_TITS_Jimenez}. In one study, an ellipse is used instead \cite{2014_TR_Merat}.

\textit{Off-road gaze} may be defined simply as outside the forward roadway or as gaze towards specific areas, \eg instrument cluster \cite{2020_Information_Feierle, 2019_IJHCI_Navarro}, entertainment system \cite{2012_TR_Kaber, 2013_PUC_Kujala, 2015_HumanFactors_Liang, 2015_AccidentAnalysis_Peng, 2019_TransRes_Miller}, any interior vehicle objects/areas \cite{2011_AccidentAnalysis_Owens} or a secondary task \cite{2013_TransRes_Benedetto, 2014_TR_Reimer, 2015_IET_Yang, 2017_CHB_Stenberger}.  However, such coarse definitions of AOIs may conflate driving and non-driving related glances. For instance, this split does not consider looking at the side and rear-view mirrors necessary for safe driving, and does not imply inattention, but only a few studies correct for this bias \cite{2014_TransRes_Tivesten}.

\begin{figure}[t!]
\centering
 \includegraphics[width=0.8\textwidth]{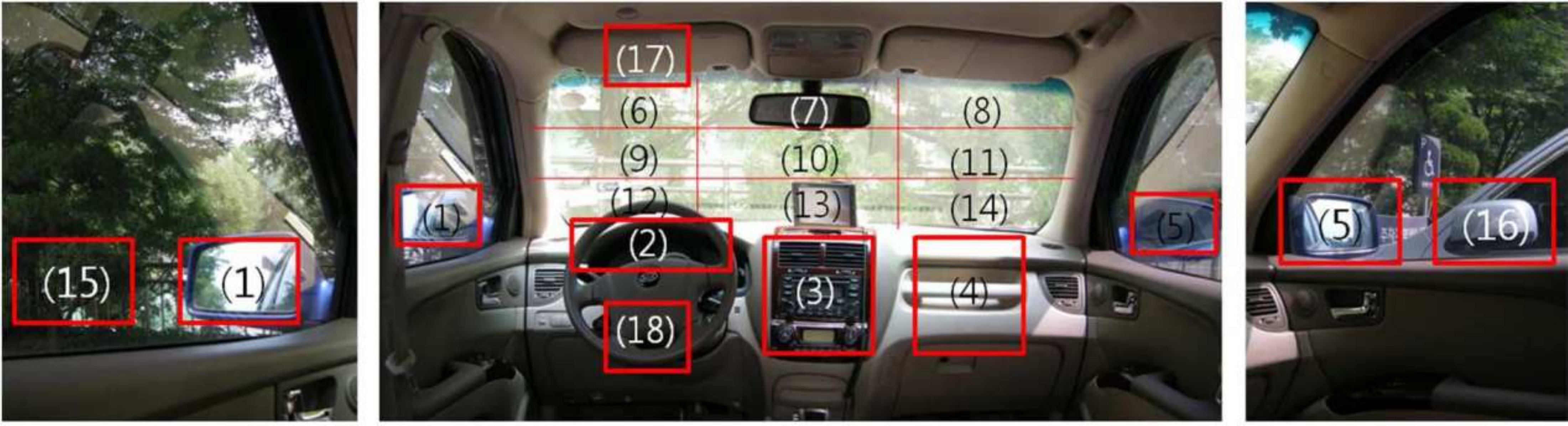}
  \caption[Illustration of the static areas of interest defined for the vehicle interior]{Vehicle interior divided into 18 areas of interest. Source: \cite{2011_TITS_Lee}}
   \label{fig:static_AOI_examples}
\end{figure}

As a result, many behavioral studies and practical works use more fine-grained \textit{AOIs inside the vehicle} for detailed driver behavior analysis and modeling. The number of AOIs and their spatial extent varies significantly depending study's purposes from 3 zones \cite{2013_TR_Wortelen} to as many as 18 \cite{2015_TITS_Vicente, 2016_AccidentAnalysis_Belyusar} (as shown in Figure \ref{fig:static_AOI_examples}). For example, rear-view and side mirrors may be treated as a single area \cite{2014_TransRes_Birrell}, considered separately \cite{2017_AppErgonomics_Lu}, or combined with the corresponding window \cite{2018_TIV_Martin}. Additionally, looking over the shoulder \cite{2016_AccidentAnalysis_Kidd}, towards front seat passengers \cite{2016_AccidentAnalysis_Morando}, glove compartment \cite{2015_TITS_Vicente}, or other objects inside the vehicle \cite{2017_CHI_Fridman} may be considered separate AOIs.

\begin{figure}[t!]
\centering
\begin{subfigure}{0.49\linewidth}
\centering
 \includegraphics[width=0.6\textwidth]{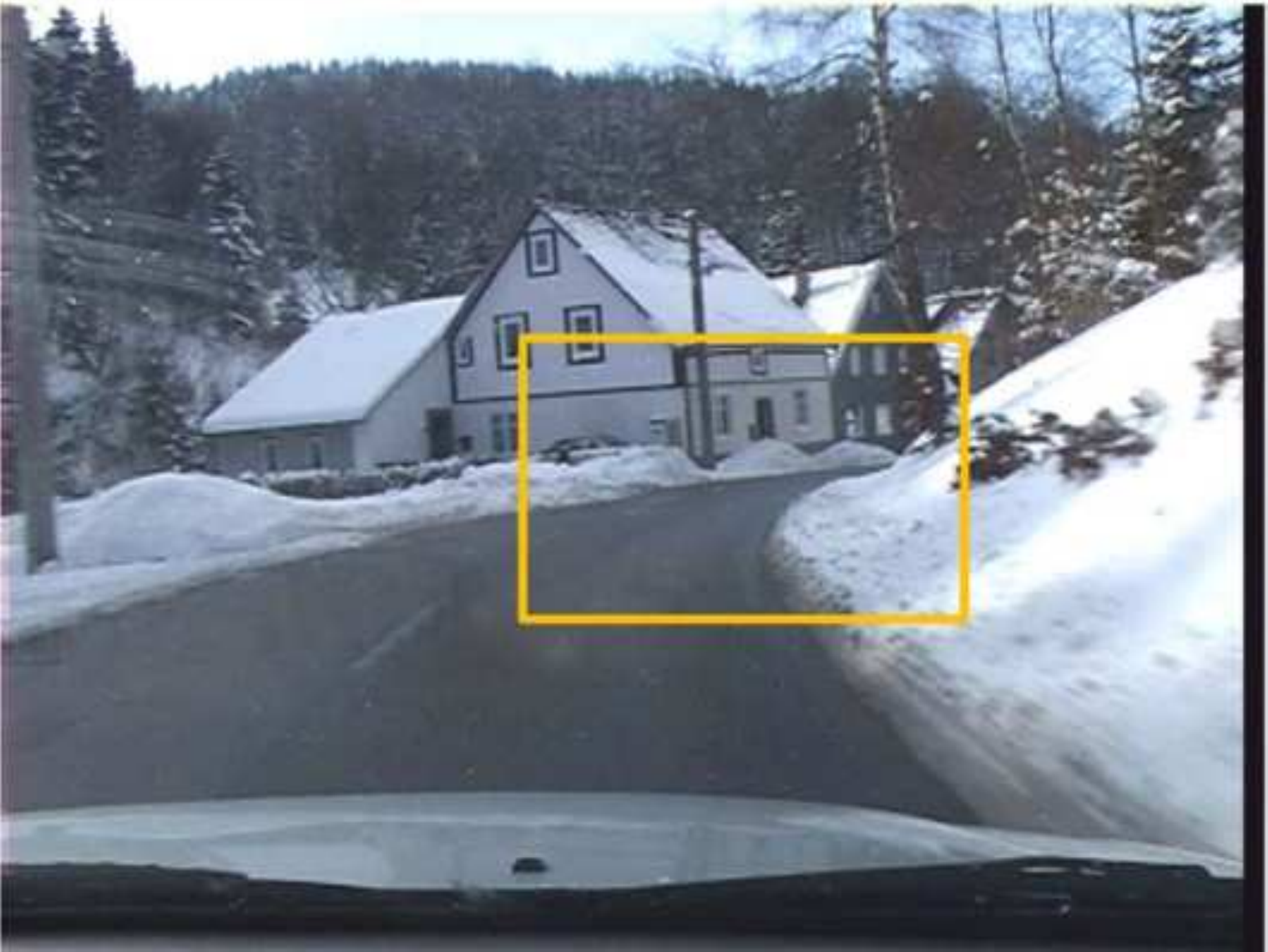}
  \caption{Latent hazard}
  \label{fig:latent_hazard}
\end{subfigure}
\begin{subfigure}{0.49\linewidth}
\centering
 \includegraphics[width=0.6\textwidth]{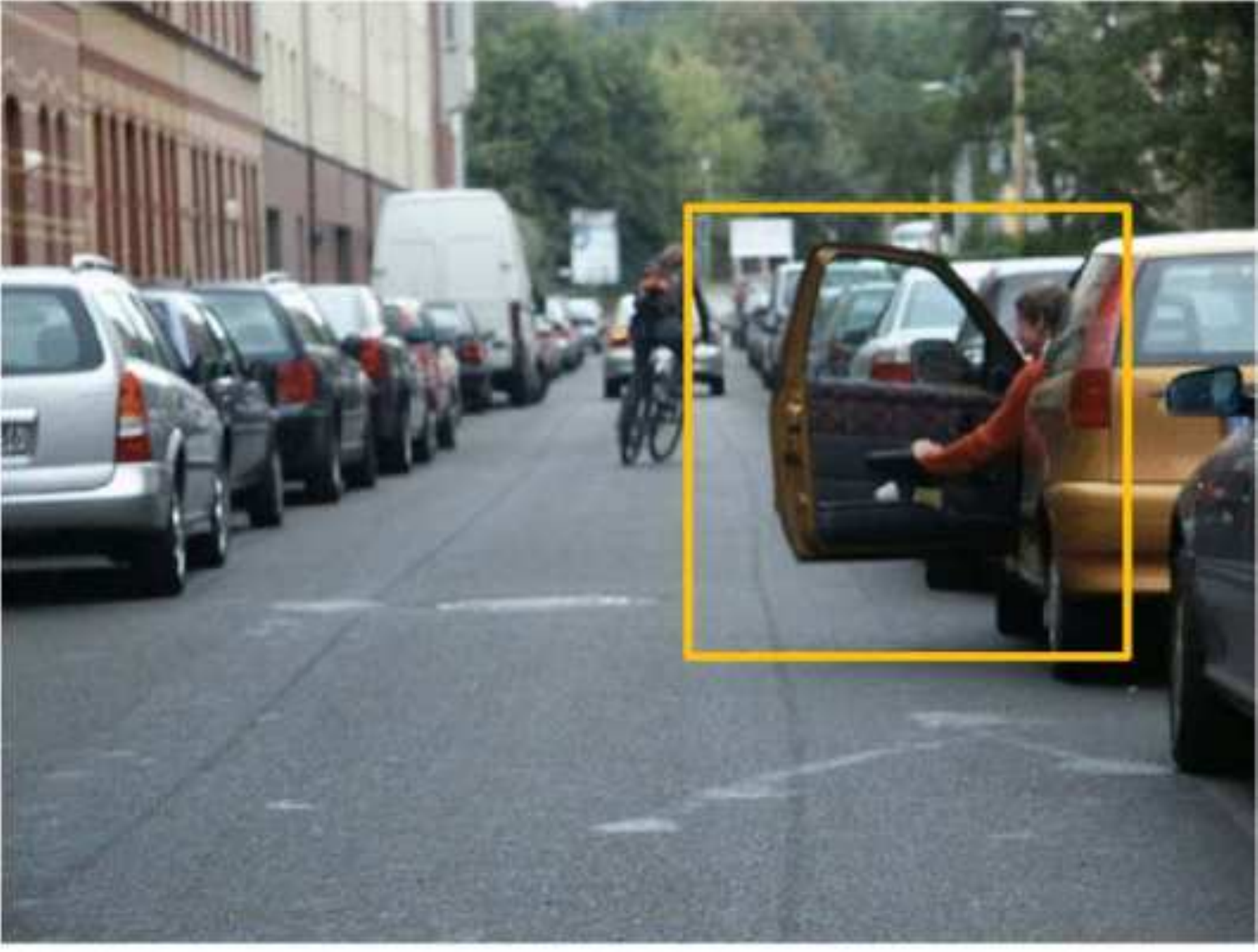}
  \caption{Materialized hazard}
  \label{fig:materialized_hazard}
\end{subfigure}
\caption[Examples of dynamic areas of interest for latent and materialized hazards]{Examples of dynamic AOIs. Source: \cite{2016_JoV_Huestegge}.}
   \label{fig:dynamic_AOI_examples}
\end{figure}

\noindent
\textbf{Dynamic AOIs} are primarily used for objects outside the vehicle. These areas are not fixed and may need to be updated per frame due to the motion of the object and/or ego-motion of the vehicle. Dynamic AOIs are predominantly used in studies involving hazard perception (Section \ref{sec:primary_driving_task}) and external distractions (Section \ref{sec:billboards_signs}). For example, to evaluate drivers' hazard perception skills, dynamic AOIs are defined around certain categories of road users, \eg pedestrians \cite{2012_AccidentAnalysis_Borowsky, 2015_OPO_Lee} or approaching vehicles \cite{2010_Perception_vanLoon, 2011_VR_Underwood} (see Figure \ref{fig:materialized_hazard}). If a hazardous situation has not materialized or involves multiple objects, more abstract hazard zones are defined as bounding boxes around the relevant area in the scene \cite{2019_TRR_Mangalore, 2012_AccidentAnalysis_Werneke, 2016_JoV_Huestegge, 2017_DrivingAssessmentConference_Yamani} (Figure \ref{fig:latent_hazard}). The assumption is that drivers will not anticipate a hazard in time or will fail to identify it at all if they do not look at the predefined region. However, since such hazard zones are tailor-made for each particular situation, they carry a degree of subjectivity. 

Dynamic AOIs are also useful for studies of external distractors where typical targets are roadside objects, \eg billboards and signs \cite{2016_TR_Stavrinos, 2016_JAR_Wilson, 2013_ACP_Garrison, 2018_AppliedErgonomics_Costa, 2016_AutoUI_Hurtado}. Given that many objects appear very small in the driving scene while still being recognizable (\eg pedestrians take up as little as $1\%$ of the total area \cite{2019_TransRes_Chen}), dynamic AOIs (usually bounding boxes or, rarely, ellipses \cite{2015_OPO_Lee}) are enlarged to allow for parafoveal vision and eye-tracking imprecision \cite{2019_TransRes_Chen, 2013_IV_Bar, 2015_TrafficInjuryPrevention_Borowsky, 2012_AccidentAnalysis_Borowsky}. Furthermore, some works automatically segment the driving scenes into multiple categories (roadside areas, horizon, sky, buildings, \etc) \cite{2019_ITSC_Fang, 2017_SNPD_Shinohara, 2017_AccidentAnalysis_Beanland, 2018_CVPR_Ramanishka, 2018_PAMI_Palazzi} to study more general patterns of driver's attention.

\subsection{Eye-tracking measures}
\label{sec:eye_tracking_measures}
\begin{figure}[t!]
\centering
  \includegraphics[width=1\linewidth]{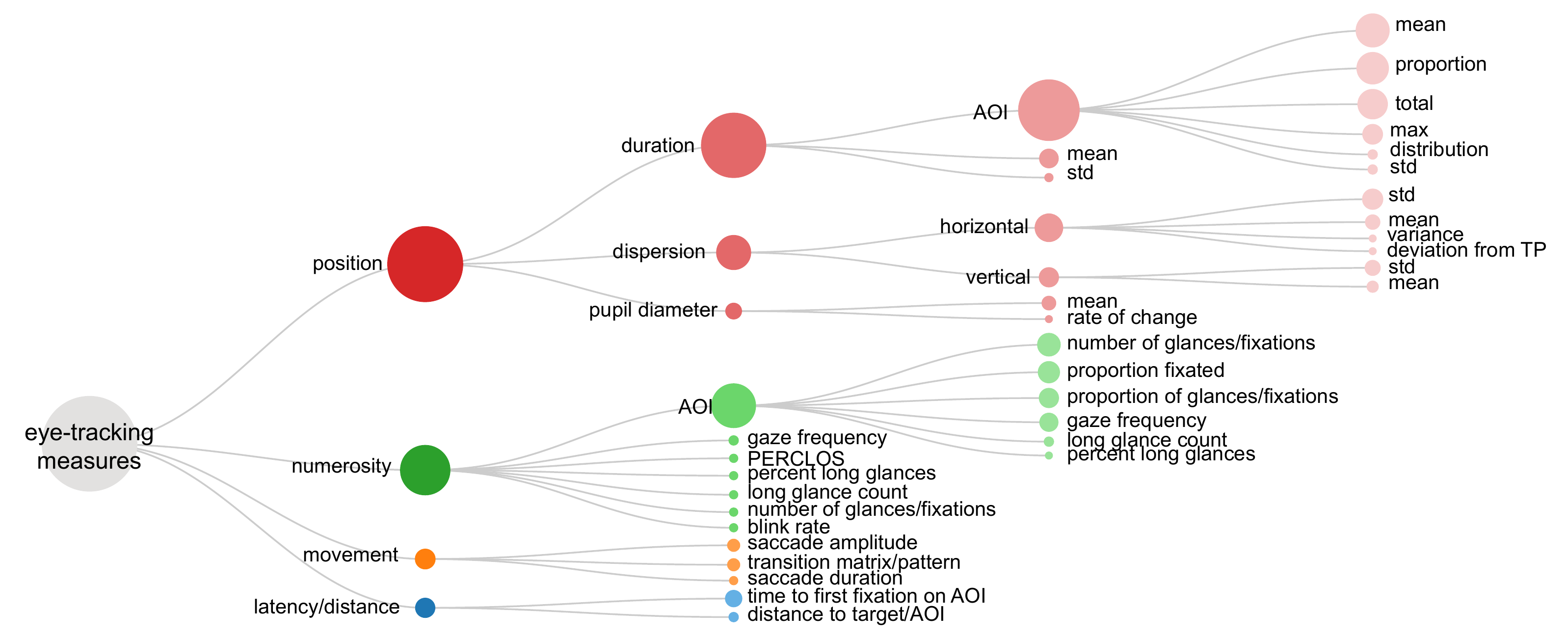}
  \caption[A dendrogram of eye tracking measures]{Eye-tracking measures used in the reviewed behavioral studies. Size of the node reflects the number of papers that used this measure. Only measures used in at least 3 studies are shown.}
   \label{fig:gaze_measures}
\end{figure}

Driver gaze data, \eg fixation coordinates, saccades (eye movements between the fixations), pupil size, and state of the eye, are further processed and aggregated for statistical analysis and modeling. Below, we will list the most common measures and their purposes and discuss any conflicting definitions and specifics of the application of these measures.

We identified nearly a hundred different eye-tracking measures used in the literature and organized them into four major categories following the taxonomy proposed in \cite{2011_OUP_Holmqvist}: \textit{position}, \textit{numerosity}, \textit{latency/distance}, and \textit{movement}. \textit{Position} category refers to positional data and is further subdivided into \textit{duration} and \textit{dispersion} of gaze, \textit{blink duration}, \textit{saccade duration}, and \textit{pupil diameter}. \textit{Numerosity} measures include frequencies, counts, ratios, percentages, and distributions of various events, primarily fixations, fixation durations, and saccades within the entire scene and specific areas of interest. \textit{Latency/distance} measures the timing of events and distances to targets in the scene and within the visual field. Finally, \textit{movement} measures serve to analyze eye movement speed, amplitude, and transition patterns between various areas of interest. The taxonomy is illustrated in Figure \ref{fig:gaze_measures}. For clarity, we will focus only on the common measures in each category, \ie those used in at least 3 papers. Note also that we use \textit{gaze} as a generic term that may refer to raw gaze data, fixations, or glances depending on what unit of analysis was used in each particular study (see Section \ref{sec:data_preprocessing} for definitions).

\subsubsection{Position measures}
\label{sec:position_measures}
\textit{Gaze duration} (mean and standard deviation) has been linked to depth \cite{2019_TransRes_Hashash, 2019_TransRes_Chen} and speed of processing \cite{2016_JEMR_Sun, 2012_AccidentAnalysis_Crundall}, task demand (shorter gaze durations for more complex maneuvers \cite{2019_AppliedErgonomics_Robbins}), visibility conditions (longer fixations during low visibility compared to normal daytime driving  \cite{2010_AccidentAnalysis_Konstantopoulos}), and relevance (drivers look longer when hazards are present \cite{2013_ACP_Garrison, 2012_AccidentAnalysis_Crundall, 2019_TransRes_Chen}). Shorter fixation durations combined with increased saccade amplitudes and frequencies suggest increased visual exploration in the presence of potential hazards \cite{2010_AccidentAnalysis_Konstantopoulos, 2016_PLOS_Yan}. Changes in these measures are associated with various stages of decision-making processes at intersections \cite{2014_JEMR_Lemonnier}.


\vspace{0.5em}
\noindent
\textit{Mean, maximum, and total duration of gaze on AOI}, depending on the type of AOI, can be used to assess the driver's attention or inattention. For instance, total and mean fixation duration on pedestrians convey the level of attention and strategy for dividing attention between pedestrians and other driving-related elements of the scene \cite{2012_AccidentAnalysis_Borowsky}. A more commonly used measure is the duration of off-road gaze (total eyes-off-road time or TEORT \cite{2015_TR_Victor}). Particularly, total time looking away from the road and single long off-road glances are associated with increased crash risk \cite{2014_JAH_SimonsMorton}. As a result, multiple studies report correlations between off-road glance statistics and secondary task type and completion time, \eg sending and receiving messages on a cell phone \cite{2013_PUC_Kujala, 2011_AccidentAnalysis_Owens, 2014_ACM_Reimer, 2015_AccidentAnalysis_Peng}, and increased off-road glances in the presence of roadside billboards and signs \cite{2013_TrafficInjuryPrevention_Dukic, 2015_AppliedErgonomics_Kaber}.

\vspace{0.5em}
\noindent
\textit{Percentage of time gazing at AOI} is calculated as the ratio of time looking at the AOI to the total time (\eg while completing a task or being exposed to stimuli of interest). Depending on what basic unit of measurement is used, raw data, fixation or glance, or how the measure is computed, the duration of the saccades may or may not be included. Sometimes not only the time gazing at an AOI but also the size of the AOI is taken into account \cite{2015_TR_Eyraud}. Other terms used for this measure are attention ratio \cite{2018_AHAT_Feldhutter, 2020_Information_Feierle}, the proportion of gaze \cite{2018_HumanFactors_Robbins}, and visual time sharing \cite{2013_TrafficInjuryPrevention_Dukic}.

This measure can help understand how drivers allocate their gaze during different tasks. For example, when using parking assist, drivers look more at the camera rather than rely on mirrors \cite{2012_SAE_Kim, 2016_AccidentAnalysis_Kidd, 2018_TransRes_Kidd}. If a lead vehicle is present, drivers increase dwell time to this AOI, especially during rear-end collision avoidance \cite{2018_TRR_Li}. When making a right turn at an intersection, the percentage of gaze spent looking to the left or right changes depending on the phase (approaching, waiting, or accelerating), traffic density, and the presence of pedestrians \cite{2014_CogTechWork_Werneke}.

\vspace{0.5em}
\noindent
\textit{Percent road center (PRC)}, first proposed in \cite{2005_TransRes_Victor} and commonly used in the literature \cite{2013_TR_Jamson, 2014_TR_Merat, 2018_TITS_Morando, 2019_TransRes_Hashash, 2014_SafetyScience_Young}, specifically measures what proportion of time the driver spends glancing at the road AOI (see definitions of the forward roadway and road center in Section \ref{sec:AOI}). Other terms in the literature for this measure are percent eyes on-road \cite{2015_TransRes_Bargman} or on-path \cite{2015_DDI_Tivesten, 2016_AccidentAnalysis_Morando} and monitoring ratio (in automation applications) \cite{2016_HumanFactors_Hergeth}.

Percent of time spent looking on the road is a sensitive indicator of inattention. For instance, the percentage of on-road gaze decreases when the driver is operating a cell phone \cite{2014_SafetyScience_Young, 2019_TransRes_Hashash} and also depends on the task complexity (length of the entered message) and traffic density (which has the opposite effect) \cite{2015_AccidentAnalysis_Peng}. Drivers tend to look more off-road when exposed to billboards \cite{2016_TR_Stavrinos, 2016_AccidentAnalysis_Belyusar}.

PRC is also indicative of driving context, \eg dwell times off-road, on near and far road ahead differed depending on the road type (urban, suburban, or rural) \cite{2018_TransRes_Young}. PRC is higher during maneuvers and when the lead or oncoming vehicles are present \cite{2014_TransRes_Tivesten}, or when driving at night \cite{2015_BMCGeriatrics_Urwyler}.

\vspace{0.5em}
\noindent
\textit{Gaze duration/location distribution} represented by a histogram or density plot helps show how various events and conditions affect both temporal and spatial allocation of gaze. For instance, in \cite{2015_TransRes_Bargman} distributions of off-road glance durations are shifted due to the effect of various visual-manual tasks compared to baseline driving. In \cite{2014_TransRes_Birrell} distributions of glance durations to an in-vehicle infotainment system (IVIS) are used to evaluate how well new interface reduces long glances away from the road. Data from naturalistic studies indicate that the frequency of long off-road glances increased half a minute before the crash or near-crash events \cite{2014_HFES_Liang}. Visualizations of the proportion of glances towards different areas inside and outside the vehicle further help understand the temporal allocation of attention before frontal collision warning events \cite{2015_DDI_Tivesten} or when adaptive cruise control was engaged \cite{2016_AccidentAnalysis_Morando}.

\vspace{0.5em}
\noindent
\textit{Horizontal} and \textit{vertical spread} is measured as mean, standard deviation, and variance of gaze position.  Gaze spread indicates the breadth of visual exploration of the scene and is associated primarily with driving experience and workload changes. For instance, expert drivers tend to have a larger horizontal spread of gaze than novices \cite{2010_AccidentAnalysis_Konstantopoulos}. They also adjust their gaze to the available field of view to observe the entire scene, improving their response to hazards \cite{2014_TR_Alberti}. Professional taxi drivers were shown to have narrower vertical dispersion than experienced and novice drivers, presumably because they allocate their attention more efficiently \cite{2013_AccidentAnalysis_Borowsky}. A different study did not find a significant relationship between experience and vertical spread, presumably because it was confounded by the experiment design, requiring the drivers to attend to the speedometer and rear-view mirror \cite{2010_AccidentAnalysis_Konstantopoulos}.

With regards to workload, several studies measured the horizontal spread of fixations during secondary tasks. Cognitive tasks \cite{2013_TransportRes_Savage} and emotional conversations \cite{2011_TR_Briggs} were shown to decrease the horizontal gaze dispersion, whereas some visual tasks increased the horizontal and vertical gaze dispersion \cite{2016_AccidentAnalysis_Kountouriotis}. Similarly, when subjects passively viewed videos, their horizontal and vertical gaze was spread more widely compared to when they were actively controlling the vehicle, likely because of the increased demand on vision and attention by the driving task itself. Furthermore, passive viewers monitor the road less and can focus on the scenery instead \cite{2015_VC_Mackenzie}.

\vspace{0.5em}
\noindent
\textit{Mean pupil diameter.} First demonstrated by Kahneman and Beatty in the 1960s \cite{1966_Science_Kahneman,1967_PerceptionPsychopysica_Kahneman}, pupil size changes depending on the relative mental effort needed to perform a task - larger pupil diameter is associated with higher task demand. Since then, this measure has been confirmed as a sensitive indicator of mental workload in many contexts, including driving. In driving studies, mean pupil diameter has been used to measure increasing workload due to secondary tasks \cite{2015_TransRes_Niezgoda, 2014_SafetyScience_Lemercier, 2018_TRR_Li} and driving task duration \cite{2017_TrafficInjuryPrevention_Wang}. Pupil diameter has also been associated with differences in the working memory capacity  \cite{2016_JARMAC_Wood} and age \cite{2017_TrafficInjuryPrevention_Wang} (older adults have smaller pupil sizes). 

Due to the significant individual variations in pupil size, some studies instead of mean absolute diameter measure the relative rate of change of the pupil diameter with respect to individual baselines measured before the experiment \cite{2017_TrafficInjuryPrevention_Wang} or across the entire experimental run \cite{2015_TransRes_Niezgoda}.

Pupil diameter changes depending on the available illumination due to pupillary light reflex \cite{1958_ArchOpht_Lowenstein} making the use of this measure problematic in on-road studies where lighting is difficult to control.  Therefore, with few exceptions \cite{2017_TrafficInjuryPrevention_Wang}, studies using this measure are conducted in laboratory conditions. Even then, care should be taken not to confound the results. For example, in the study on the effect of visibility on attention, Konstantopoulos \etal \cite{2010_AccidentAnalysis_Konstantopoulos} reported that pupil diameter was significantly larger in simulated night driving conditions, and results were affected more by the lighting conditions than the mental workload. Likewise, pupil dilation changes could not be solely attributed to hazard levels in the study by Lu \etal \cite{2020_TR_Lu}.

\subsubsection{Numerosity measures}

\vspace{0.5em}
\noindent
\textit{Gaze frequency} is measured as the number of fixations\slash glances made anywhere in the scene  \cite{2016_JEMR_Sun, 2016_OptometryVisionScience_Lee} or, more commonly, towards a particular AOI  (\eg side windows \cite{2018_AccidentAnalysis_Kircher}, cell phone \cite{2014_SafetyScience_Young}, off-road \cite{2018_HumanFactors_Victor} or on-road \cite{2019_TransRes_Hashash}) per standardized unit of time (second \cite{2016_OptometryVisionScience_Lee, 2018_TRR_Li}, minute \cite{2018_AccidentAnalysis_Kircher, 2018_NatSciReports_Shiferaw, 2019_TransRes_Hashash, 2019_SafetyScience_Kuo, 2010_TransEng_Bonmez} or hour \cite{2018_HumanFactors_Victor}). Gaze frequency is associated with the efficiency of information intake, which for the driving case means fewer fixations towards task-specific areas \cite{2015_TR_Eyraud}.

Sometimes, gaze frequency refers to the proportion of fixations\slash glances to an AOI out of the total number of fixation/glances during the predefined time interval \cite{2015_AppliedErgonomics_Kaber, 2018_AppliedErgonomics_Zahabi, 2019_TRR_Pankok, 2015_AppliedErgonomics_Kaber, 2018_AppliedErgonomics_Zahabi, 2017_AppliedErgonomics_Zahabi} (see \textit{Percentage of glances/fixations} below).

Frequency of gaze indicates the rate at which the driver samples the environment, which may change depending on circumstances or the drivers' characteristics. For example, drivers check side mirrors more frequently when merging onto a highway \cite{2018_AccidentAnalysis_Kircher}, look around when prompted by navigation instructions \cite{2016_OptometryVisionScience_Lee} or when feeling drowsy \cite{2019_SafetyScience_Kuo} and fixate less on the road when distracted by secondary tasks \cite{2012_TR_Kaber, 2019_TransRes_Hashash}.

\vspace{0.5em}
\noindent
\textit{Number of glances/fixations} overall or on a specific AOI over an arbitrary time interval (\eg while performing a maneuver \cite{2016_JEMR_Sun} or while engaging in a secondary task \cite{2014_SafetyScience_Young}) is closely related to gaze frequency, however, care should be taken in comparing the results.

The number of fixations or glances to the scene is affected by driving conditions (fewer fixations during more demanding night driving \cite{2010_AccidentAnalysis_Konstantopoulos}), presence of hazards (high-level hazards are inspected with fewer fixations \cite{2010_TR_Huestegge}), vehicle control (during automation road near the vehicle was visited less since there was no need to maintain lateral position of the vehicle \cite{2019_IJHCI_Navarro}) and distractions (\eg gaze frequency is higher when undistracted \cite{2011_TR_Briggs}). Concerning the latter, the number of glances is sensitive to different types of visual-manual secondary tasks \cite{2011_AccidentAnalysis_Owens, 2014_TransRes_Tivesten, 2014_TR_Reimer}.

The number of glances towards areas inside and outside the vehicle helps analyze scanning activities and frequency of visual search \cite{2016_PLOS_Yan} during regular driving and when performing maneuvers \cite{2019_JSR_Li}. For instance, drivers who made more glances towards the forward roadway and conflict vehicles at intersections were more likely to avoid collisions \cite{2016_PLOS_Yan}. This is supported by naturalistic driving data which shows that in the moments preceding crashes drivers made more glances away from the road \cite{2017_AccidentAnalysis_Seppelt}. The frequency of sampling is also affected by perceived risk (\eg approaching motorcycle vs car) and, to some extent, the saliency of the hazard \cite{2011_VR_Underwood}.

\vspace{0.5em}
\noindent
\textit{Proportion (percentage) of glances\slash fixations} measures the ratio of glances\slash fixations to the target AOI and glances\slash fixations to all regions over some time period. This measure is primarily used to understand how attention is spread across multiple areas depending on various factors such as the task, maneuver, or driving conditions. For instance, when drivers are asked to maintain a constant speed, they predictably make more glances towards the speedometer \cite{2012_JoV_Sullivan}. The use of park assist reduces the percentage of glances over the shoulder \cite{2016_AccidentAnalysis_Kidd}, forward roadway, and side windows \cite{2018_TransRes_Kidd}. Perceived risk also has an effect, \eg the proportion of fixations towards the target vehicle was greater when cars and motorcycles were approaching from a far distance compared to bicycles but greater when bicycles approached from a medium or near distance compared to cars and motorcycles \cite{2018_HumanFactors_Robbins}.

\vspace{0.5em}
\noindent
\textit{Proportion of AOIs fixated.} In studies that use this measure, if the fixation falls within the area of interest, it is counted as detection. Duration is not explicitly taken into account, only hit or miss is considered (see common values of thresholds for fixation duration in Section \ref{sec:data_preprocessing}). 

This measure helps understand how drivers pay attention to different objects, events, or categories of road users. For example, drivers look at 25\% of traffic signs even though they are relevant to driving \cite{2014_TransRes_Costa, 2016_TR_Topolsek}). When it comes to hazards, drivers notice actualized ones more often than potential, although, with more driving experience, latent hazard detection skills improve \cite{2012_AccidentAnalysis_Crundall_1, 2019_TRR_Mangalore, 2017_DrivingAssessmentConference_Yamani}. High-level hazards (in terms of braking affordance and safety relevance) tend to attract more gaze than low-level hazards \cite{2016_JoV_Huestegge}. The environment also plays a role in what is considered hazardous by drivers \cite{2017_AccidentAnalysis_Beanland}. 

It would be wrong to assume that fixation on an AOI equals processing, therefore detection accuracy is combined with drivers' reactions (\eg brake response, maneuver, keypress, \etc) \cite{2010_AccidentAnalysis_White, 2012_AccidentAnalysis_Crundall, 2016_JTSS_Grippenkoven}. There are 4 possible situations: a) failed to see and react, b) did not fixate but reacted, c) looked at the AOI but did not react, and d) looked and reacted. Scenario d) is the ideal situation, scenario a) suggests inattention, scenario b) can help identify targets that were detected using extrafoveal vision \cite{2016_JoV_Huestegge}, whereas c) reveals ``looked-but-failed-to-see'' errors \cite{2010_AccidentAnalysis_White}.

\vspace{0.5em}
\noindent
\textit{Long glance off-road count/percentage.} Long glances off-road are defined as instances where the driver continuously looks away from the forward roadway. Based on the data from the 100-Car naturalistic study \cite{2006_TR_Klauer}, single long glances away from the road for more than 2 seconds significantly increase the risk of a crash. As a result, National Highway Traffic Safety Administration (NHTSA) guidelines for in-vehicle electronic devices \cite{2013_TR_NHTSA} mandate that devices that require the driver to divert their gaze away from the road for more than 2s (and for the cumulative duration of off-road glances over 12s to complete the task) are inappropriate and encourage the use of voice interfaces to mitigate this issue. However, analysis of the more recent naturalistic driving dataset gathered as part of the second Strategic Highway Research Program (SHRP2) \cite{2016_TR_Hankey} shows that in the specific case of the rear-end crashes, shorter glances led to most of the incidents \cite{2015_TR_Victor}. Nevertheless, the 2s threshold continues to be widely used in the literature \cite{2013_AccidentAnalysis_Peng, 2014_TR_Reimer, 2016_AccidentAnalysis_Belyusar, 2017_CHB_Stenberger, 2016_AutoUI_Smith, 2019_TransRes_Miller}, although shorter durations of 1.6s \cite{2013_PUC_Kujala} and longer glances of 3s \cite{2012_TRR_Divekar}, 4s, and 8s \cite{2018_HumanFactors_Victor} are also considered.

Links were established between the increased number of long off-road glances and secondary tasks such as the use of in-vehicle navigation and radio tuning \cite{2014_ACM_Reimer}, receiving or sending messages via cell phone \cite{2015_AccidentAnalysis_Peng}, external distractions \cite{2016_AccidentAnalysis_Belyusar, 2019_AppliedErgonomics_Costa}, as well as risk-taking behavior \cite{2015_AccidentAnalysis_Peng} and exposure to partially and conditionally automated driving \cite{2018_HumanFactors_Victor, 2019_TransRes_Miller}. At the same time, since studies did not report any accidents, safety implications cannot be established. With regards to vehicle control, the effect of long off-road glances is inconclusive. For instance, for lane-keeping ability statistical significance shown is either marginal \cite{2013_AccidentAnalysis_Peng} or none \cite{2012_TRR_Divekar, 2016_AutoUI_Smith}. 

\vspace{0.5em}
\noindent
\textit{Blink rate (frequency)}. Blink rate is measured as the number of blinks per second \cite{2020_TransRes_Li, 2018_TRR_Li, 2016_AccidentAnalysis_Wang} or minute \cite{2015_TransRes_Niezgoda}. For some applications, it may be measured before or after certain events \cite{2020_TransRes_Li}. Blink duration and rate are somewhat controversial. Both measures are affected by changes in workload, but there are conflicting reports as to \textit{how} they are affected. For instance, Niezgoda \etal \cite{2015_TransRes_Niezgoda} and Savage \etal \cite{2013_TransportRes_Savage} report increased blink rates for the subjects engaged in auditory cognitive tasks while driving. The opposite trend is shown in the study by Li \etal \cite{2018_TRR_Li} where drivers were talking or operating cell phones during collision avoidance, whereas no change in blink frequency was found in \cite{2020_TransRes_Li} for drivers watching videos before and after taking over control from automation. Savage \etal \cite{2013_TransportRes_Savage} suggest that the nature of the secondary task and its visual attention demands cause the discrepancy in the trends; however, more investigation is necessary to establish this. Despite being somewhat uncertain, blink measures are used in practice for distracted driving detection \cite{2015_TITS_Liu, 2015_TITS_Li, 2016_TITS_Li}. Increased blinking rate is also associated with drowsiness \cite{2018_NatSciReports_Shiferaw} and has been used for fatigue detection in practical studies \cite{2011_OptEng_Jo, 2013_AdvMechEng_Jin, 2019_IEEEAccess_Zhang, 2019_IEEEAccess_Deng}.

\vspace{0.5em}
\noindent
\textit{PERCLOS (percentage of eye closure)}, defined as the proportion of time during which the eyes are $80-100\%$ closed over $1$ minute interval, was introduced in \cite{1994_TechRep_Wierwille} as the most robust single measure of alertness. Since then, the superiority of PERCLOS over other measures and its use as the only metric for alertness has been disputed \cite{2011_DrivSymposium_Trutschel}; however, it still appears in behavioral and practical studies, often supplemented by other measures.

To compute PERCLOS, the threshold for eye closure and time window for computing the measure must be decided. The original threshold of $80\%$ is the most common setting, although other values have been used, \eg $70\%$ in \cite{2010_IV_Friedrichs} and $75\%$ used as default in the popular eye-tracking system faceLab \cite{2015_TITS_Liu}. The majority of the studies do not use the originally proposed window of $1$ min and instead compute PERCLOS over a range of durations: from seconds ($10$s \cite{2011_OptEng_Jo}, $30$s \cite{2016_IV_Schmidt}) to minutes ($3$ min \cite{2015_TITS_Liu}, $6$ min \cite{2010_JCTA_Zhang}, $20$ min \cite{2019_SafetyScience_Kuo}, $30$ min \cite{2016_TraffInjuryPrevention_Jackson}). 

For practical applications, it is necessary to determine what values of PERCLOS correspond to drowsiness. Some studies establish hard thresholds experimentally \cite{2019_SafetyScience_Kuo, 2016_IV_Schmidt, 2011_OptEng_Jo, 2010_JCTA_Zhang}. Others, to mitigate the effect of individual variations, use the beginning of the driving sequence to compute a baseline \cite{2013_IJVT_Sigari} or manually annotate videos into drowsy/non-drowsy and divide the measured PERCLOS values accordingly \cite{2013_AdvMechEng_Jin}.

PERCLOS is also more reliable when computed over a long time interval \cite{1998_FHWA_Dinges} which delays detection of the drowsiness and makes the measure unsuitable for detection of microsleeps that last few seconds or when the subjects fall asleep with their eyes open \cite{2010_IV_Friedrichs}.

\subsubsection{Latency/distance measures}
\vspace{0.5em}
\noindent
\textit{Time to the first fixation on an AOI} is measured from the time the target first appeared in the scene in absolute units (ms) \cite{2016_JARMAC_Wood} or in relative terms to allow comparisons between different time windows \cite{2012_AccidentAnalysis_Crundall_1}. This measure is an indicator of how fast drivers focus on certain types of objects or hazardous events. For instance, drivers fixate faster on actualized hazards rather than precursors \cite{2012_AccidentAnalysis_Crundall_1}, notice pedestrians intending to cross earlier \cite{2019_TransRes_Chen}, and look at the approaching cars sooner than motorcycles \cite{2012_AccidentAnalysis_Crundall}. These results suggest that the severity of the hazard may play a role in faster reaction times. Although one study has shown that experienced drivers spot hazards faster than novices \cite{2012_AccidentAnalysis_Crundall_1}, other studies investigating hazards did not find effects of driving experience \cite{2010_TR_Huestegge, 2019_TransRes_Chen}. As expected, distractions caused by a secondary task led to slowed reactions to hazards \cite{2016_JARMAC_Wood}. In contrast, subjects who passively watched videos without the need to control the vehicle could detect hazards faster due to decreased workload \cite{2015_VC_Mackenzie}.

In studies involving automation, \textit{eyes on road} is frequently measured to assess how quickly a driver can regain control after automation failure. The time it took to first fixate on the road after the take over request is positively correlated with the number of hazards detected \cite{2018_TR_Vlakveld}. Driving assistance can further improve drivers' reactions via visual \cite{2012_ACM_Pomarjanschi} or auditory warnings \cite{2019_TransRes_Lu}. 

\vspace{0.5em}
\noindent
\textit{Distance to target} can refer to the angular distance to target within the field of view (eccentricity) or distance between the target and ego-vehicle in the scene. Few studies looked at the effect of eccentricity on hazard perception with somewhat contradictory conclusions. One reported that large angular distance between the first fixation on the target and previous gaze location could affect driver awareness of the hazard \cite{2019_HumanFactorsErgonomics_Kim}, however, according to a different study, for high-level hazards, this may not be true as they were found equally quickly across a wide range of eccentricities \cite{2016_JoV_Huestegge}. The effect of physical distance away from the ego-vehicle has been considered only for signs. Results show a linear dependence between the distance when the sign is first fixated and size of the sign, speed of the ego-vehicle, and text length  \cite{2019_AppliedErgonomics_Costa, 2018_AppliedErgonomics_Costa}. 

\subsubsection{Movement measures}

\vspace{0.5em}
\noindent
\textit{Saccade amplitude} is the distance (in degrees of visual angle) traveled from its onset to offset. Decreased saccade amplitudes have been associated with high cognitive and perceptual load caused by following navigation instructions \cite{2016_OptometryVisionScience_Lee}, dealing with difficult traffic scenarios with small braking affordance \cite{2010_TR_Huestegge}, and reduced visual conditions (\eg artificially blurred scenes) \cite{2015_OPO_Lee}. In one study, saccade amplitudes increased when visual exploration was required (when giving way at the intersection), therefore this measure may be correlated with different stages of decision making \cite{2014_JEMR_Lemonnier}.

Saccade amplitudes change with age -- older drivers tend to make smaller saccades, presumably due to reduced accuracy of saccades whose endpoints do not land as precisely on the targets as do saccades of younger adults  \cite{2015_OPO_Lee, 2016_OptometryVisionScience_Lee}. Driving experience has been shown to have a small effect on amplitude. For instance, experienced drivers tend to scan the scene more thoroughly, thus making shorter mean saccades \cite{2010_TR_Huestegge}. Sleep deprivation and duration of driving cause saccade amplitudes to increase, in other words, spatial distances between fixations increase, leading to more dispersed gaze \cite{2018_NatSciReports_Shiferaw}.


\vspace{0.5em}
\noindent
\textit{Saccade duration} is measured in ms and is correlated with saccade amplitude and velocity. The duration of saccades has been linked to increased workload due to secondary tasks or challenging driving conditions. For example, in the study by Li \etal  \cite{2018_TRR_Li} the smallest saccade durations were observed for hand-held, followed by hands-free and no phone conditions. The authors suggest that reducing the saccadic eye movements and blink inhibition are the most efficient strategies for minimizing missed visual information while using a cell phone. Cheng \etal \cite{2016_PLOS_Cheng} measured drivers' scanning durations relative to glance durations when entering and exiting the highway and found that scanning decreased during these phases compared to regular driving on the highway. Throughout the more demanding merging phase, saccade duration was affected by traffic density.

\vspace{0.5em}
\noindent
\textit{Transition matrix/pattern.} Transition probabilities between multiple AOIs help understand drivers' visual exploration strategies and are used in practical studies modeling driver's inattention (see Section \ref{sec:driver_monitoring}). Transition probability between areas A and B is expressed as the number of gaze transitions between these areas normalized by the sum of all transition events over the observation period. Transition matrices have been used to visualize changes in attention allocation to different areas due to secondary task involvement \cite{2017_JSR_Wang}, the use of in-vehicle navigation systems \cite{2014_TransRes_Birrell}, or the presence of digital billboards next to the road \cite{2019_JTEPBS_Zhang}. Visual exploration patterns indicated by transitions can help infer internal decision processes, \eg at intersections while performing different maneuvers \cite{2013_SafetyResearch_Scott, 2015_TR_Lemonnier, 2019_JSR_Li}.

\subsection{Vehicle control}
\label{sec:vehicle_control}

\subsubsection{Vehicle control measures}

\begin{figure}[t!]
\centering
  \includegraphics[width=0.7\linewidth]{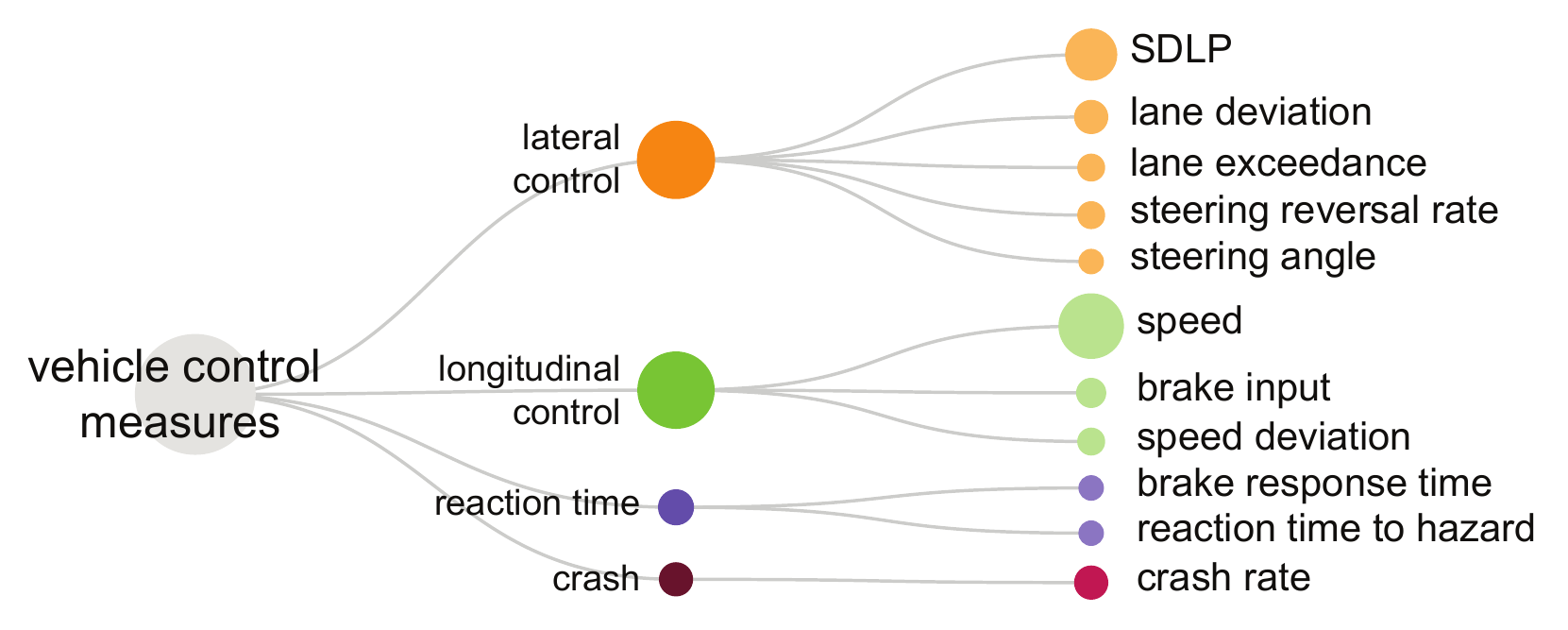}
  \caption[A dendrogram of driving performance measures]{Commonly used driving performance measures. Size of the node reflects the number of papers that used the corresponding measure. Only measures used in at least 3 papers are shown.}
   \label{fig:driving_performance_measures}
\end{figure}

More than half of the behavioral studies link attention to driving performance using a set of measures shown in Figure \ref{fig:driving_performance_measures}. Here we rely on the taxonomy proposed in \cite{2011_Handbook_Caird}.

\vspace{0.5em}
\noindent
\textit{Longitudinal control} refers to changes in ego-vehicle speed and distance to the vehicle ahead.
	\begin{itemize}
	\item{\textit{Ego-vehicle speed}} is primarily a measure reflecting the driving style, typically computed as an average across the entire ride or specific portions of it for more fine-grained analysis. Many papers report speed reduction as a compensatory strategy in high workload conditions such as reduced visibility \cite{2014_TransRes_vanLeeuwen}, performing secondary tasks \cite{2010_HumanFactorsErgonomics_Reimer, 2014_TR_Reimer} while navigating curves \cite{2016_AccidentAnalysis_Kountouriotis}, or due to age-related deficiencies \cite{2012_HumanFactors_Reimer, 2017_AppliedErgonomics_Zahabi, 2018_HumanFactorsErgonomics_Zahabi}. Speed reduction has also been associated with hazard anticipation, \eg scanning at level crossings \cite{2016_JTSS_Grippenkoven} or slowing down to notice potentially hazardous objects sooner \cite{2012_AccidentAnalysis_Crundall_1, 2014_TR_Alberti}.

	\item{\textit{Speed deviation}} is measured with respect to a pre-defined target or posted speed limit and has been linked to changes in cognitive workload, \eg due to a non-driving related activity (NDRT) \cite{2011_TR_Briggs, 2014_SafetyScience_Young, 2017_AppliedErgonomics_Zahabi} combined with complex driving conditions such as construction zones \cite{2015_AppliedErgonomics_Kaber}.

	\item{\textit{Brake and accelerator input}}	are indicators of safe and smooth driving. Braking in response to hazard may indicate that the driver has not examined the scene fully \cite{2019_SAP_Bozkir} or was distracted by external objects \cite{2016_AccidentAnalysis_Belyusar}. Hard braking and accelerating operations are reduced if the driver is anticipating the changes in the environment (\eg given in-vehicle traffic information \cite{2015_IET_Yang} or when given cues during take-over from automation \cite{2014_HumanFactorsErognomics_Lorenz, 2018_ITSC_Yang}). Fewer acceleration events may indicate added cognitive demand \cite{2012_HumanFactors_Reimer}. Older drivers braked and accelerated harder after take-over from automation \cite{2017_AccidentAnalysis_Clark}.
	
	\item{\textit{Headway}} is measured in terms of distance \cite{2012_TR_Kaber}, time \cite{2016_AccidentAnalysis_Morando, 2015_DDI_Tivesten} or time-to-collision (TTC) \cite{2015_TransRes_Bargman, 2013_TR_Jamson} to the lead vehicle. When the safety margin is low, a faster and stronger response is required from the driver to avoid crashing into the lead vehicle \cite{2012_TR_Kaber}. In some studies, inverse TTC ($TTC^{-1}$) is used to operationalize the change rate of lead-vehicle looming and has been linked to crash risk in rear-end collisions where braking was used as an evasive maneuver \cite{2015_TR_Victor, 2015_TransRes_Bargman}. When the vehicle ahead is far away (large headway), drivers are more likely to engage in secondary activities \cite{2015_TR_Victor}. Drivers already engaged in secondary visual tasks tend to compensate for distraction by increased headway \cite{2012_TR_Kaber}. At the same time, a higher cognitive load leads to the variation of headway distance \cite{2016_CHI_Lee} or time \cite{2016_AutoUI_Smith}. There is evidence that engaging automation (\eg adaptive cruise control) improves headway compared to manual control \cite{2013_TR_Jamson, 2015_DDI_Tivesten, 2018_TITS_Morando}.	
	\end{itemize}

\vspace{0.5em}
\noindent
\textit{Lateral control} is typically measured in terms of the lateral position of the vehicle within the ego-lane or by examining the frequency and magnitude of steering wheel rotations.
	\begin{itemize}
	\itemsep0em
	\item{\textit{Lane deviation}} is a common measure for evaluating lateral control. It is typically measured as mean (MLP) or standard deviation of lane position (SDLP) with respect to road centerline (lane changes are not taken into account). There is conflicting evidence regarding the effects of secondary tasks on lateral control and their interpretation. Several studies show that off-road glances due to non-driving-related activities \cite{2013_AccidentAnalysis_Peng, 2012_HumanFactors_Lee, 2016_CHI_Lee, 2016_AutoUI_Smith, 2016_AccidentAnalysis_Zeeb} or external distractions \cite{2019_JTEPBS_Zhang} lead to increased SDLP. However, there are experiments where SDLP reduced or did not change under increased visual \cite{2014_TR_Reimer, 2014_HumanFactorsErgonomics_Schieber} and cognitive load \cite{2010_HumanFactors_Cooper,2014_SafetyScience_Lemercier, 2014_SafetyScience_Young, 2016_AccidentAnalysis_Kountouriotis, 2014_HumanFactors_He}, which points to inhibition of top-down interferences with the automatized task such as lane-keeping \cite{2010_HumanFactors_Cooper, 2013_ACP_Garrison}. Alternatively, reduced SDLP may not necessarily mean improvement of lateral control. When combined with reduced lateral scanning, it may indicate rigid steering \cite{2016_AccidentAnalysis_Kountouriotis}. Other findings suggest that lateral control is more demanding than longitudinal, and therefore high cognitive load makes the drivers focus more on the lateral control diverting resources from other tasks \cite{2014_HumanFactors_He}. Some studies also point out that lateral control depends on the secondary task complexity and how it is presented to the driver (\eg when using head-up display, there was no difference from baseline conditions even for medium load) \cite{2014_HumanFactorsErgonomics_Schieber}. Professional drivers, \eg police officers making regular patrols, retain their lateral control even when performing demanding tasks \cite{2018_AppliedErgonomics_Zahabi}.

	\item{\textit{Lane exceedance}} (\textit{departure} or \textit{excursion}) is defined as the proportion of time outside of ego-lane or the number of times the vehicle left the ego-lane (it is not always specified whether the vehicle left the lane partially or fully). Lane excursions are extremely rare under normal driving conditions but may be caused by distractions (\eg visual secondary task \cite{2010_CHI_Jensen, 2019_TransRes_Louw}), especially for inexperienced drivers due to their poor use of peripheral vision for lateral control \cite{2012_TRR_Divekar}. Age-related deficiencies in lane-keeping have also been observed \cite{2015_BMCGeriatrics_Urwyler}.

	\item{\textit{Steering reversal rate}} measured per time or distance indicates the number of steering wheel direction changes. Increased steering adjustments indicate lane-keeping errors and attempts to correct them. Increased steering reversal rate has been observed under high cognitive load \cite{2014_HumanFactors_He, 2012_HumanFactors_Reimer, 2011_AccidentAnalysis_Owens} and is associated with age-related factors \cite{2011_AccidentAnalysis_Owens, 2012_HumanFactors_Reimer}.

	\item{\textit{Steering angle}}, continuously measured, correlates with horizontal eye movements (see also discussion in Section \ref{sec:visuo_motor_coordination}). Correspondingly, distractions of different types lead to changes in correlations, \cite{2013_TITS_Yekhshatyan} as does automation of lateral control \cite{2018_TIV_Wang}.
	\end{itemize}

\vspace{0.5em}
\noindent
\textit{Reaction time} measures assess how quickly the driver can react to hazardous events by applying brakes or steering.  Specifically, \textit{brake response time} is the time from the hazard appearance to brake onset. As with other vehicle control measures, it takes the drivers longer to react if they are distracted (\eg texting \cite{2019_TransRes_Hashash}) or sleep-deprived \cite{2018_AppliedErgonomics_Zahabi}. With early warning provided by driver assistance systems or better visibility of the scene, the response time \cite{2019_TransRes_Lu, 2018_ITSC_Yang}, distance \cite{2019_SAP_Bozkir}, and TTC \cite{2016_AutoUI_Borojeni} to the target  may improve.

\vspace{0.5em}
\noindent\textit{Crash rate} and \textit{crash risk} are direct measures of driving safety. Because of the rarity and severity of such events, this measure is reported infrequently, especially in on-road studies. When using data from large-scale naturalistic datasets, near-crash events, where drivers avoided an imminent accident by performing a rapid evasive maneuver, are used together with crash events to increase statistical significance \cite{2015_TransRes_Bargman, 2018_AccidentAnalysis_Lee}. 

Crash risk has been directly linked to insufficient visual information intake due to long off-road glances \cite{2014_JAH_SimonsMorton, 2015_TransRes_Bargman, 2018_AccidentAnalysis_Lee}, sleep deprivation \cite{2016_TraffInjuryPrevention_Jackson}, and limited visibility of the scene  \cite{2016_PLOS_Yan, 2014_TR_Alberti}.

\subsubsection{Visuo-motor coordination}
\label{sec:visuo_motor_coordination}

\begin{figure}[t!]
\centering
  \includegraphics[width=0.8\linewidth]{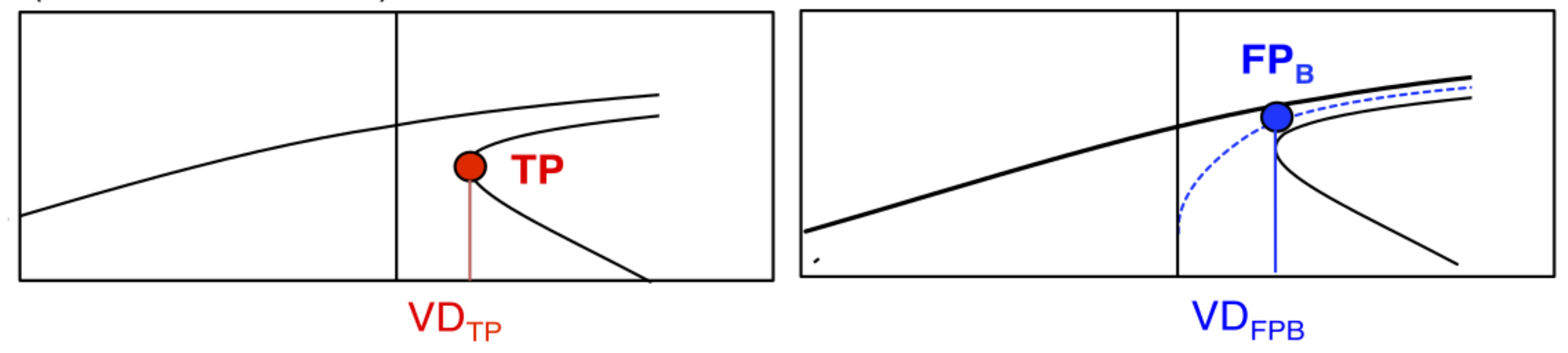}
  \caption[Illustration of steering points for curve negotiation]{Illustration of two steering points: tangent point (TP) and future path (FP$_B$) point beyond the tangent point. VD refers to the egocentric visual direction in each case. Source: \cite{2014_JoV_Lappi}}
   \label{fig:TP_vs_FP}
\end{figure}

Road geometry, \ie whether it is straight or curved, gives rise to different visual strategies. Driving on curved roads is perhaps one of the most studied visually guided tasks with $25$ years of research accumulated to date. One of the long-standing debates is where drivers look in order to steer the vehicles on curved roads (for a detailed review see \cite{2014_JoV_Lappi}). Some of the first to address this problem were Land and Lee in their seminal work \cite{1994_Nature_Land} where they proposed the tangent point (TP), \ie point where the inside road edge reverses its direction, as a possible candidate. They also established the relationship between the TP, the curvature of the road, and steering input. As an alternative, a future path (FP) point for guiding steering was proposed by Wann and Swapp \cite{2000_Nature_Wann}. A schematic illustration of both points is shown in Figure \ref{fig:TP_vs_FP}. Despite many new alternatives proposed, TP remains the default theory for visually guided steering on curved roads. Several recent publications have questioned this status. Due to gaze recording imprecision and spatial proximity of TP and FP, some studies were not conclusive \cite{2012_PLOS_Mars, 2013_JoV_Lappi}, especially when analyzing entering the curve since the area spanned by the curve in the field of view is too small \cite{2015_PONE_Itkonen}. Some researchers suggest that the dichotomy between TP and FP may be false, and perhaps the drivers rely on multiple reference points as they navigate the curve \cite{2013_JoV_Lappi}.

\subsection{Summary}
Data processing is required to detect various types of eye movements from raw gaze coordinates and aggregate statistics for further analysis and visualization. Fixations, glances, and saccades are most commonly used since they specify the locations where the driver held their gaze and transitions made in between. Other types of eye movements are rarely considered due to a lack of algorithms and equipment with sufficient resolution for detecting them. However, movements such as smooth pursuit and vergence can provide information about the objects being tracked and the depth of the fixated objects.

Fixations are the basic unit for the majority of eye-tracking measures. However, definitions of fixations found in the literature are inconsistent and often under-specified. The choices for various thresholds that define how fixations are detected can affect statistical properties of the aggregated data and, as a result, may lead to incorrect conclusions and contradictions across studies. Therefore, precise definitions and justifications of the chosen parameters are necessary to ensure consistent and reproducible results.

Many measures have been developed to analyze different aspects of attention such as duration of gaze on a target, frequency of visits to certain areas, reaction times to events, and state of the eyes indicative of the drivers' condition. For the most part, attention measures are well defined, but because there are several ways of assessing the same property of driver's attention, there is a staggering number of measures, many of which are used in only a handful of studies. This further complicates aggregation and comparisons of the results across studies.

Given the strong association between visual attention and motor actions during driving, attention and vehicle control measures are often analyzed together. A large body of research is dedicated to understanding visually guided tasks such as curve negotiation. Besides visuo-motor coordination, a combination of eye-tracking and vehicle data is useful for studying hazard perception, divided attention, and estimating safety risks.

\section{What affects drivers' attention?}
\label{ch:attention_factors}
\section{External and internal factors}
We analyzed a comprehensive set of papers published since 2010 (inclusive) where driver's attention was explicitly measured (using measure(s) described in Section \ref{ch:attention_measures}) while factors such as state of the driver and/or properties of the scene were manipulated. The following discussion will be organized around two groups of factors that have been studied: internal and external (see Table \ref{tab:factors}). 

\noindent
\textbf{Internal factors} include driver demographic characteristics (age and gender), the primary driving task (\eg highway driving, hazard perception, vehicle following, \etc), involvement in secondary non-driving related activities such as eating/drinking or operating a cell phone, and associated internal driver's state (\eg distracted or drowsy). The driving experience is another important factor in this group which includes duration, frequency of driving, experience with other vehicle types (\eg motorcycle or bicycle), and geographical location where the driving experience was obtained.

\noindent
\textbf{External factors} include properties of the environment (lighting and visibility conditions), road type (highway, urban, rural), and geometry (curved or straight), as well as roadside objects that may inform or distract the driver (signs and billboards). Automation is included in this group as well. Here, automation is used as a catch-all term that encompasses any system that can take over the control of the vehicle, assist during certain tasks (\eg parking), monitor the driver's state, and provide guidance and/or warnings to the driver. As these features are becoming standard in most commercial vehicles, it is increasingly important to consider the effect the automation may have on driver's attention.


\begin{figure}[t!]
\centering
\begin{subfigure}{0.85\linewidth}
 \includegraphics[width=1\linewidth]{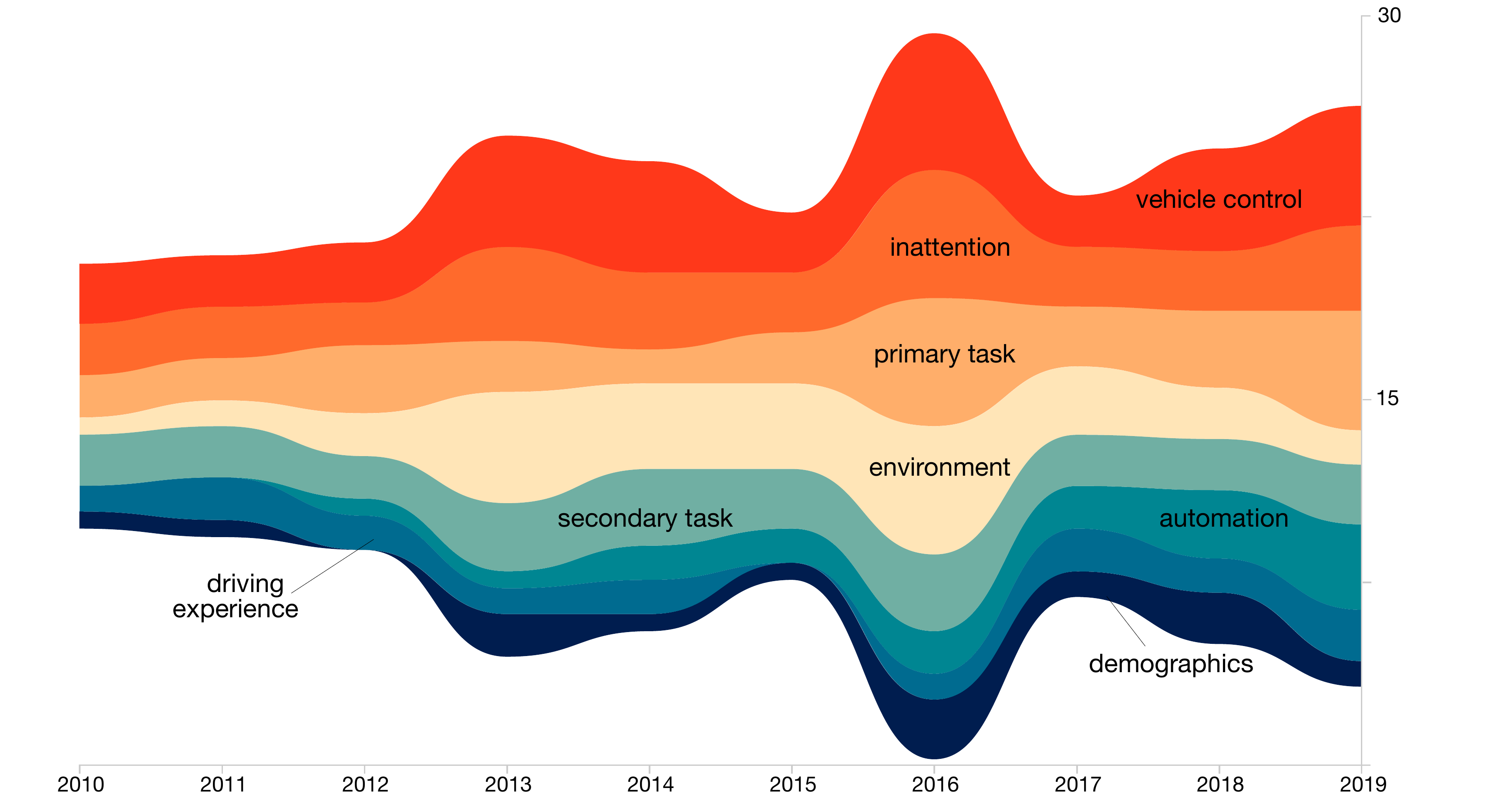}
  \caption{Behavioral studies}
\end{subfigure}
\begin{subfigure}{0.85\linewidth}
  \includegraphics[width=1\linewidth]{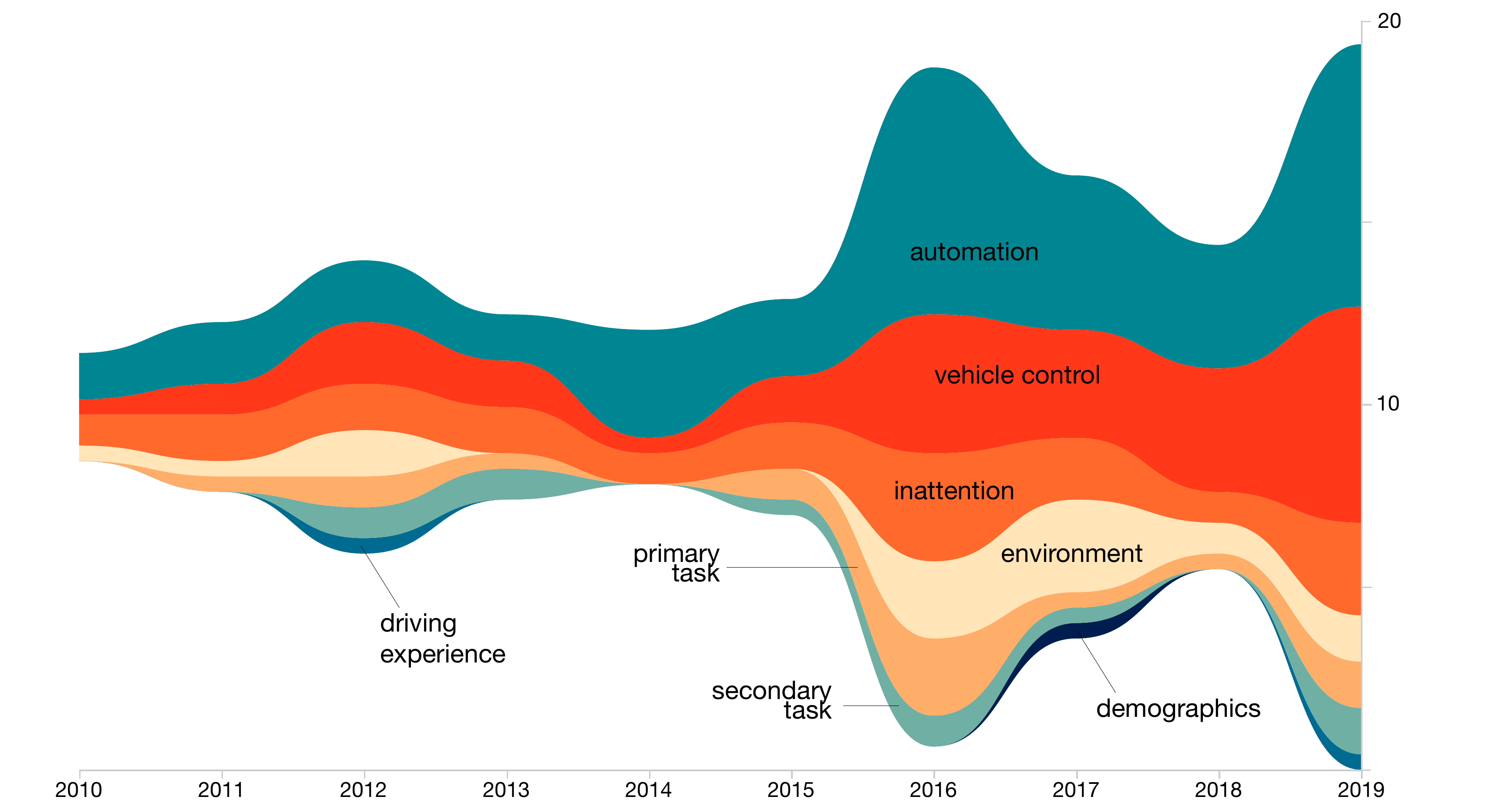}
\caption{Practical works}
\end{subfigure}
\caption[Streamgraphs of temporal distribution of topics in behavioral and practical studies]{Streamgraphs showing the distribution of papers on various factors in behavioral and practical works from 2010 to 2019 inclusive. Vertical axis shows the number of papers. Individual rivulets correspond to different factor and sorted by the total number of papers that considered this factor in descending order from top to bottom of the graph.}
   \label{fig:topics_streamgraph}
\end{figure}

Finally, since driving is a visuomotor task, vehicle control is considered explicit factor. Multiple studies have shown that where we look is highly dependent on the task and, in many cases, correlates with motor activity. This is also true for driving and has implications for conducting the studies and analyzing their results as well as for the design of practical systems.

Figure \ref{fig:topics_streamgraph} shows temporal trends in research on various factors in behavioral studies and practical implementations. It shows the overall growth of interest towards various aspects of attention related to driving, particularly in practical applications, with a peak around 2016. In behavioral studies, various factors were more or less uniformly covered. Note that automation emerged as a factor since 2011 as lane assist, park assist, emergency braking, and other similar technologies became commonly available in commercial vehicles. Even though all vehicle control solutions available on the market are level 2 according to SAE classification, there has been a lot of interest in understanding driver behavior and attention allocation when using highly automated driving that requires little to none human supervision.

Practical works focus heavily on several factors such as automation, vehicle control, and monitoring driver inattention. Recently, properties of the environment and attention towards other road users gained more prominence.

\begin{table}[t!]
  \centering
\resizebox{\textwidth}{!}{%
\begin{tabular}{l|l|l|l|l|ll}
\multicolumn{5}{c|}{Internal factors}                                                                                                                                                      & \multicolumn{2}{c}{External factors}                       \\ \hline
\begin{tabular}[c]{@{}l@{}}Primary \\ driving task\end{tabular} & \begin{tabular}[c]{@{}l@{}}Secondary \\ non-driving task\end{tabular} & Driving experience  & Demographics & Inattention & \multicolumn{1}{l|}{Environment}    & Automation           \\ \hline
hazard anticipation/detection                                   & visual                                                                & automation          & age          & distraction & \multicolumn{1}{l|}{billboard/sign} & lateral control      \\
hazard response                                                 & manual                                                                & duration            & gender       & drowsiness  & \multicolumn{1}{l|}{intersection}   & longitudinal control \\
parking                                                         & cognitive                                                             & location            &              &             & \multicolumn{1}{l|}{road geometry}  & driver monitoring    \\
taking over control                                             &                                                                       & training            &              &             & \multicolumn{1}{l|}{road type}      & warning/guidance     \\
vehicle following                                               &                                                                       & other vehicle types &              &             & \multicolumn{1}{l|}{traffic}        & explainability       \\
                                                                &                                                                       & racing              &              &             & \multicolumn{1}{l|}{visibility}     & failure              \\
                                                                &                                                                       & gaming              &              &             & \multicolumn{1}{l|}{vegetation}     & take over request   
\end{tabular}%
}
  \caption{External and internal factors and their sub-factors used in this report.}
  \label{tab:factors}%
\end{table}%



\subsection{Understanding driver inattention}
\label{sec:inattention_taxonomy}

Previous sections discussed what the driver attends to and why, however, it is driver inattention that poses a safety concern. In the literature, concepts related to driver inattention, such as distraction and driver's state, are often conflated. As will be explained later, here, distraction is a subtype of inattention, and the driver's state is characterized in terms of their physiological and emotional condition that may lead to inattention (Section \ref{sec:driver_state}).

\begin{figure}[t!]
\centering
\resizebox{0.7\textwidth}{!}{
\begin{forest}
[DRIVER INATTENTION, draw
	[Restricted\\attention, l=19mm, align=center, draw]
	[Misprioritized\\attention, l=19mm, align=center, draw]
	[Neglected\\attention, l=19mm, align=center, draw]
	[Cursory\\attention, l=19mm, align=center, draw]
	[Diverted\\attention, l=19mm, align=center, draw
		[Non-driving-related, draw]
		[Driving-related, draw]
	]
]
\end{forest}
}
  \caption[Driver inattention taxonomy by Regan \etal.]{Driver inattention taxonomy by Regan \etal \cite{2011_AccidentAnalysis_Regan}}
  \vspace{-1.5em}
  \label{fig:inattention_taxonomy_regan}
\end{figure}
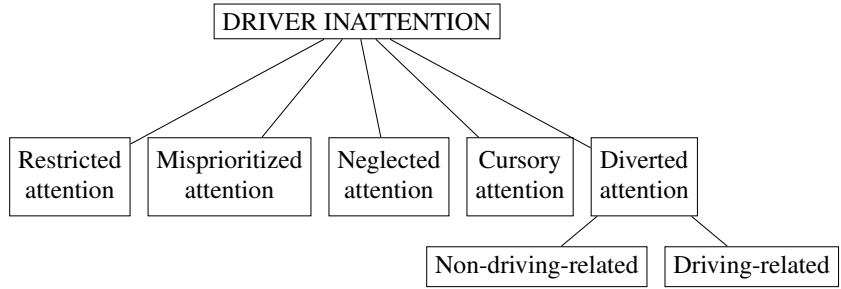

This report adopts the definition of inattention proposed by Regan \etal \cite{2011_AccidentAnalysis_Regan}: ``insufficient, or no attention, to activities critical for safe driving''. According to the taxonomy by the same authors, distraction (or diverted driver attention) together with restricted, misprioritized, neglected, and cursory attention, are considered subtypes of inattention (shown in Figure \ref{fig:inattention_taxonomy_regan}). Distraction is further subdivided into non-driving-related and driving-related (\eg noticing flashing fuel warning light and continuously thinking about finding the nearest gas station). Within this framework, internal and external factors (listed in Section \ref{ch:attention_factors}) give rise to various processes causing inattention or increasing alertness \cite{2014_AAAM_Regan}. Other definitions and taxonomies of inattention derived from in-depth analyses of crash reports also exist (see \cite{2013_TechReport_Engstrom} for an extensive review). 


In what follows, we will primarily focus on two types of inattention, restricted and diverted, that comprise the bulk of behavioral and practical research on this topic. Restricted attention is brought by functional limitations, such as microsleeps, blinks, and saccades, that prevent the driver from properly observing the scene and are caused by changes in the driver's state (see Section \ref{sec:driver_state}). Diverted attention due to non-driving-related activities is another widely studied category that includes distractions due to various secondary tasks (Section \ref{sec:secondary_task}) and the presence of external objects (Section \ref{sec:billboards_signs}). These two types of inattention have been investigated theoretically and modeled in practice (see recent reviews of research on drowsiness and fatigue \cite{2020_IJITSR_Doudou, 2018_TITS_Sikander, 2016_TheorIssErgonSci_Lenne}, and distraction \cite{2019_JEMR_Ojstersek, 2019_TITS_Khatib, 2018_TIES_Parnell, 2017_IET_Cunningham, 2016_SigProcMag_Aghaei}). Other types of inattention are difficult to find in the literature. For example, there are no studies involving misprioritized attention and only a few works on cursory and neglected attention, primarily at intersections (see Section \ref{sec:intersections}).

\subsection{Summary}
Besides biological characteristics and limitations of human vision discussed in the previous section, other factors affect drivers' attention allocation.  Some are internal to the driver, such as their state, driving skill, demographics, and driving or non-driving related tasks they are engaged in. External influences include the effect of the environment itself, \eg roadside objects, traffic density, weather, and visibility conditions, as well as automation of some of the vehicle control functions. Research trends over the past decade show that, overall, behavioral studies covered these factors fairly evenly and considered many of them in combination, whereas practical works focused primarily on implementing automation for vehicle control and driver monitoring.

At least 5 different types of inattention have been identified, however some of them, such as misprioritized or neglected attention, can be identified only in hindsight after an accident has already occurred. Therefore, the focus in the literature is on the restricted and diverted attention, caused by drowsiness and non-driving task distractions, respectively.

\section{Behavioral studies of attention}
\label{ch:behavioral}
\subsection{Primary driving task}
\label{sec:primary_driving_task}
In most experiments, subjects are asked to follow the rules of the road, maintain the vehicle in the center of the chosen lane, and not exceed the posted speed limit. In some highly automated scenarios, no input from the driver is required at all \cite{2020_Information_Feierle}. The most common driving tasks examined in multiple studies are discussed below: \textit{identifying/responding to hazards}, \textit{vehicle following}, and \textit{taking over control} (from automation). Other tasks are considered only in a handful of studies, such as maintaining speed \cite{2012_JoV_Sullivan}, racing \cite{2017_PONE_VanLeeuwen}, and parking \cite{2012_SAE_Kim, 2018_TransRes_Kidd, 2016_AccidentAnalysis_Kidd}.

\vspace{0.5em}
\noindent
\textbf{Vehicle following} is a common task performed by drivers in traffic. It has been established that drivers use the lead vehicle (LV) as a point of information, \eg they make fewer gazes at the speedometer and instead look at LV and near road region before LV to adjust speed and maintain the vehicle within lane limits \cite{2016_AccidentAnalysis_Kountouriotis, 2018_TITS_Morando, 2019_IJHCI_Navarro}. However, the presence of LV also increases rear-end collision risk especially if the driver is involved in other non-driving-related activities. Naturalistic studies indicate that even socially acceptable and mildly distracting tasks such as radio tuning significantly increase crash risk \cite{2015_TransRes_Bargman, 2017_TransRes_Louw}. Some distracted drivers appear to be aware of the danger and tend to increase time headway to LV \cite{2012_TR_Kaber} and monitor the LV even when driving is automated \cite{2018_TITS_Morando}. According to counterfactual simulations, further increasing time spent looking on the road, reduction of long glances away from the road and time-compressing glance distribution help in reducing the odds of collision \cite{2015_TransRes_Bargman}. When rear-end collision is imminent, drivers tend to fixate longer on the LV during avoidance maneuvers \cite{2018_TRR_Li}. Drivers are also cued by looming and brake lights of the vehicle ahead as well as by ADAS such as forward collision warning (FCW) which helps to direct their attention to the LV \cite{2016_AccidentAnalysis_Morando, 2015_DDI_Tivesten}, however shortly after the critical period drivers tend to look away from the road towards the rear-view mirror and side windows to assess the situation \cite{2015_DDI_Tivesten}. 

Lead vehicles can also be used to direct the drivers \cite{2013_TR_Romoser, 2012_TRR_Divekar, 2018_HumanFactors_Victor} and control the timing of the experiment \cite{2013_TR_Romoser}, \eg in the study by Victor \etal \cite{2018_HumanFactors_Victor} the lead vehicle obscured impeding conflict object before making an evasive maneuver that revealed the object to the participants driving behind. Given that drivers tend to focus on the lead vehicle more when it is present, few studies used it to examine the extent of spatial attention in 2D  \cite{2016_HumanFactors_Gaspar} or 3D \cite{2011_AccidentAnalysis_Andersen} by projecting stimuli at different eccentricities with respect to the LV.
 
\vspace{0.5em}
\noindent
\textbf{Hazard anticipation, detection, and response}. Hazards are defined as objects and situations that can cause harm to the driver, their vehicle, other road users or their property. Identifying and avoiding hazards is an important driving skill and is part of driving tests in some countries such as the UK and Australia. For obvious reasons, virtually all experiments involving hazards are conducted in simulations. Few studies conducted in real vehicles used static objects \cite{2018_HumanFactors_Victor} or moving props \cite{2012_SAE_Kim}. Large-scale naturalistic driving studies have also been used to analyze driver behavior before crashes and near-crashes \cite{2017_AccidentAnalysis_Seppelt, 2018_AccidentAnalysis_Lee, 2016_AccidentAnalysis_Morando, 2014_HFES_Liang, 2010_TR_Klauer}, however, they do not offer high-resolution gaze data and rarely involve vulnerable road users. 

Hazardous scenarios presented in simulation experiments differ by the number of instances, \textit{hazard instigator} (vehicles, VRUs, other objects), and \textit{type} (actualized or potential). Some of the most common \textit{hazard instigators} are pedestrians or cyclists \cite{2019_TransRes_Lu, 2019_SAP_Bozkir, 2012_ACM_Pomarjanschi, 2010_AccidentAnalysis_White, 2019_HumanFactorsErgonomics_Kim, 2019_TransRes_Chen, 2012_AccidentAnalysis_Borowsky, 2010_AccidentAnalysis_Borowsky}, stationary obstacles in the ego-lane \cite{2018_ITSC_Yang, 2017_AdvErgonomics_Feldhutter, 2014_HumanFactorsErognomics_Lorenz, 2019_AccidentAnalysis_Stahl, 2011_TR_Metz, 2015_VC_Mackenzie}, lead vehicle braking \cite{2018_ITSC_Yang, 2019_AccidentAnalysis_Stahl, 2019_AccidentAnalysis_Vogelpohl}, and parked vehicles pulling out \cite{2011_TR_Metz, 2013_AppliedErgonomics_Zhang}.

\textit{Hazard type} indicates whether the hazard requires an immediate response (referred to as critical \cite{2019_HumanFactorsErgonomics_Beanland} or materialized \cite{2013_AccidentAnalysis_Borowsky}) or is only potentially dangerous (also called credible \cite{2019_HumanFactorsErgonomics_Beanland} or latent \cite{2016_PLOS_Yamani}). Examples of materialized hazards are stalled vehicles in the ego-lane or pedestrians in the path of the vehicle. Latent hazards may involve conflict vehicles in an intersection with limited visibility or a pedestrian stepping from behind a parked car. Types of hazards can be distinguished by braking affordance \cite{2010_TR_Huestegge, 2016_JoV_Huestegge}, the type of processing involved, and hazard eccentricity \cite{2015_TRR_Samuel} or type of precursor \cite{2012_AccidentAnalysis_Crundall_1}.

The response to hazard is measured from the hazard onset and until the driver reacts (by pressing a key on the keyboard or making a maneuver). Gaze data can help determine when the driver first fixated the hazard since its onset and for how long, which allows analyzing hazard perception time and processing depth \cite{2012_AccidentAnalysis_Crundall}. Additionally, for latent hazards, launch and target zones may be defined in the scene, \ie areas where the subject should look to notice the cues for a potential hazard (launch) and areas where the hazard may become visible (target) \cite{2015_TRR_Samuel, 2016_PLOS_Yamani, 2019_TRR_Mangalore}.

Overall, it is difficult to generalize findings across studies involving hazards because of widely different scenarios presented to subjects and other related factors that are being examined at the same time. For example, some secondary tasks have been shown to impact response to hazards negatively (Section \ref{sec:secondary_tasks_effects}). Hazard detection skill depends on the amount of exposure to various on-road situations, and even though novice and expert drivers do not differ in what they consider hazardous \cite{2010_TR_Huestegge}, it is more difficult for inexperienced drivers to identify and react to latent hazards \cite{2013_AccidentAnalysis_Borowsky, 2012_AccidentAnalysis_Crundall_1} (also see Section \ref{sec:differences_between_novice_experienced}). There is evidence that the safety impact of the hazard may also affect how fast it is identified \cite{2016_JoV_Huestegge, 2017_AccidentAnalysis_Beanland}, whereas saliency and eccentricity \cite{2015_TRR_Samuel, 2016_JoV_Huestegge} have little effect.

\vspace{0.5em}
\noindent
\textbf{Taking over manual control.} Drivers may also be required to supervise partially or fully-automated vehicles. Discussion of effects of automation, returning to manual control, and proposed assistive functions is provided in Section \ref{sec:automation}.

\subsection{Driving experience}
\label{sec:driving_experience}


\subsubsection{Definitions of driving experience}

The vast majority of the papers we reviewed explicitly state that recruited subjects had a prior driving experience, except for one experiment that specifically looked for non-drivers \cite{2014_TransRes_Ciceri}, and in about $5\%$ of studies (all conducted in a driving simulator) it was not clear whether the subjects had driving experience or not.

\begin{figure}[t!]
\centering
  \includegraphics[width=0.8\linewidth]{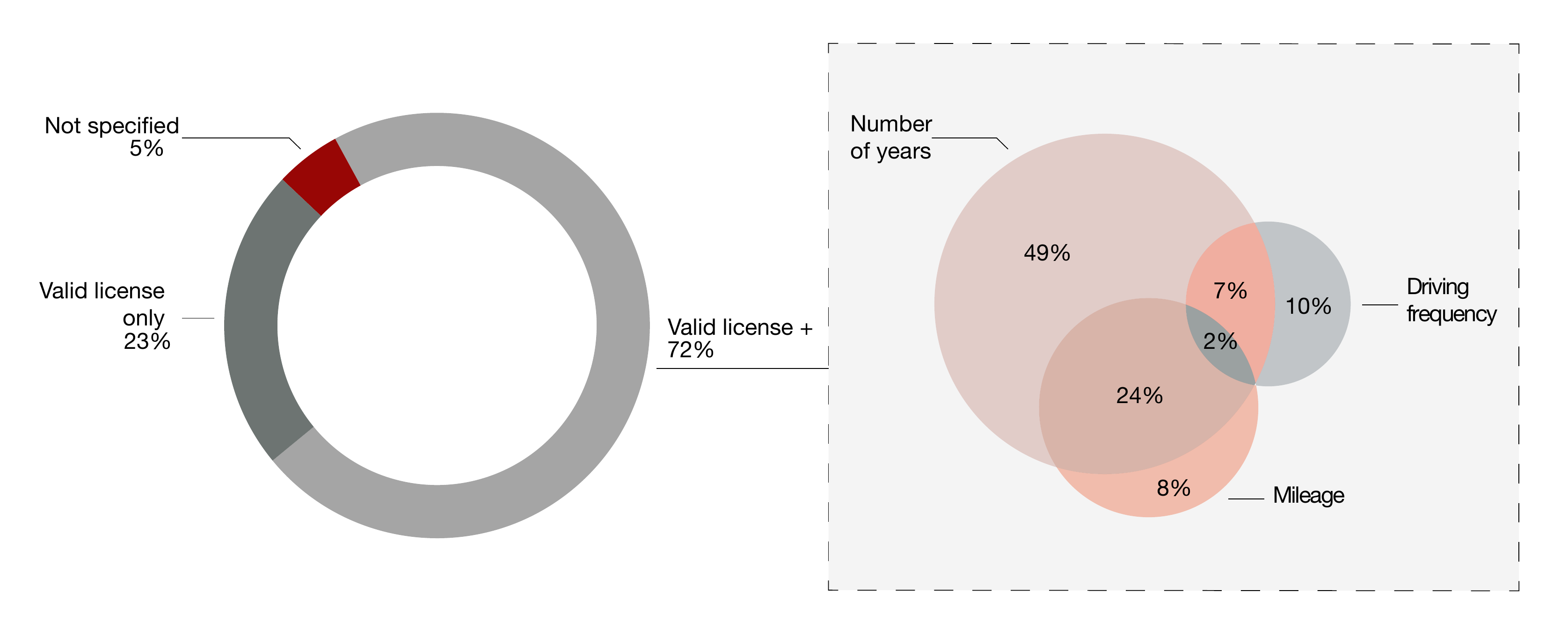}
  \caption[Definitions of driving experience]{Diagrams showing how driving experience is specified in the studies. Pie chart on the left shows the proportion of studies that required only the valid driver's license or provided more information. Venn diagram on the right further elaborates what proportion of studies provided additional information about driving experience of subjects in terms of years, frequency or miles driven or combinations of the three.}
   \label{fig:driving_experience_def}
\end{figure}

The way driving experience was specified differed significantly from study to study (see Figure \ref{fig:driving_experience_def}). In nearly $\nicefrac{1}{4}$ of papers, only having a valid license was required from the participants. The remaining studies used additional quantitative criteria, most often just the number of years driving. However, licensure duration does not reflect the qualitative differences between drivers who passed their driving test several years ago but only drive occasionally and those who make daily long commutes or are professional taxi drivers. The frequency of driving and annual mileage, are provided only for one-third of all studies.

Qualitative descriptors, such as \textit{novice}, \textit{experienced}, and \textit{professional}, are also not consistently defined in the literature. The most common cut-off for distinguishing between novices and experienced drivers is $2$ years \cite{2018_TRR_Li, 2017_SNPD_Shinohara, 2017_AutoUI_Shinohara, 2016_JoV_Huestegge, 2010_TR_Huestegge} although the justification for this choice is rarely provided. In one of the few studies that looked at driving skill development, Olsen \etal \cite{2007_TransRes_Olsen} showed that after $6$ months of independent driving novice drivers improve their mirror and window scanning but still struggle in more complex driving situations or when engaged in secondary tasks. As a result, it is more common to compare groups of drivers with larger gaps in terms of driving experience. An extreme example is a study comparing beginner drivers with an average of $24$ hours of solo driving experience and driving instructors with more than $9$ years of teaching experience \cite{2010_AccidentAnalysis_Konstantopoulos}. In general, novices with $1$ year or less of driving experience are compared to experienced drivers with $4-15$ years driving or to professional taxi drivers \cite{2019_HumanFactorsErgonomics_Beanland, 2014_TR_Alberti, 2013_SafetyResearch_Scott, 2020_TR_Zheng, 2019_TransRes_Chen, 2018_TRR_Li, 2013_AccidentAnalysis_Borowsky}.

Other less studied aspects of experience, besides licensure duration and frequency of driving, include the location (\eg country where the driver gained their experience), experience with other vehicle types (\eg bicycle \cite{2018_HumanFactors_Robbins} or motorcycle \cite{2012_AccidentAnalysis_Crundall}), mode of driving (\eg racing experience \cite{2017_PONE_VanLeeuwen}), and previous exposure to partially automated driving \cite{2020_Information_Feierle}.


\subsubsection{Differences between novices and experienced drivers}
\label{sec:differences_between_novice_experienced}
The fact that novice drivers are over-represented in crash statistics may be attributed to limitations in their visual search strategies compared to more experienced drivers. Early studies pointing to this date back more than $40$ years \cite{1972_HumanFactors_Mourant}. Differences between the groups were examined in terms of fixation duration \cite{2010_TR_Huestegge, 2010_AccidentAnalysis_Konstantopoulos}, saccade amplitude \cite{2010_TR_Huestegge}, the number of fixations \cite{2010_AccidentAnalysis_Konstantopoulos, 2013_AccidentAnalysis_Borowsky}, vertical \cite{2013_AccidentAnalysis_Borowsky, 2014_AccidentAnalysis_Lehtonen} and horizontal \cite{2014_TR_Alberti, 2010_AccidentAnalysis_Konstantopoulos} spread of search. However, based on the recent meta-analysis by Robbins \etal \cite{2019_AccidentAnalysis_Robbins} only the horizontal spread of search remains statistically significant when all studies are pooled together. As mentioned in Section \ref{sec:position_measures}, reduced horizontal dispersion of gaze of novice drivers indicates that they insufficiently anticipate and scan the environment around them, increasing the chances of missing potential hazards.

\vspace{0.5em}
\noindent
\textbf{Hazard perception and anticipation.} Novices and experienced drivers generally agree about what situations are considered hazardous \cite{2010_TR_Huestegge,2010_AccidentAnalysis_Borowsky} but experienced drivers detect more hazardous events overall and are better anticipate them. In contrast, novice drivers fixate more on hazards rather than their precursors. Experienced drivers find relevant cues (\eg a pedestrian obscured by a parking vehicle) early and monitor them continuously to take proactive measures \cite{2019_AccidentAnalysis_Stahl, 2012_AccidentAnalysis_Crundall_1, 2010_AccidentAnalysis_Borowsky, 2013_AccidentAnalysis_Borowsky}. The difference in hazard anticipation ability can be quite significant. One study reports that experienced drivers anticipated hazards in 90\% of the cases compared to 62\% for novices \cite{2019_TRR_Mangalore}. 

The saliency of the hazard has been shown to have no effect on how fast the hazard was detected \cite{2010_TR_Huestegge,2016_JoV_Huestegge,2012_TRR_Divekar}, nor did the presence of people \cite{2016_JoV_Huestegge}, but the severity of the hazard mattered. Braking affordance affected how rapidly the hazard was detected for both novice and experienced drivers as more imminent hazards consistently attracted more gazes \cite{2016_JoV_Huestegge}. Crundall \etal \cite{2012_AccidentAnalysis_Crundall_1} further analyzed the ability of drivers with different amounts of experience to respond to 3 types of hazards: behavior prediction, environmental prediction and dividing, and focusing of attention. In this study, learners missed more behavior precursors but detected the same number of behavior hazards, albeit slower than other groups. At the same time, learners missed environmental hazards despite fixating as often as other groups of drivers on the precursors. 

There are somewhat conflicting explanations in the literature as to what causes these differences. Some attribute it to the poor ability of novice drivers to allocate their attention efficiently (as evidenced by larger vertical dispersion of gaze) and looking at irrelevant locations \cite{2013_AccidentAnalysis_Borowsky}. Others claim that novice drivers know where to look but are slower at processing the hazards and initiating appropriate responses since their time to fixate on the hazard is similar to that of experienced drivers \cite{2010_TR_Huestegge}. Although previous studies indicate that experienced drivers better utilize peripheral vision to detect targets early without fixating on them, there is evidence that driving experience mainly helps to decrease hazard processing in foveal vision \cite{2016_JoV_Huestegge}.

\vspace{0.5em}
\noindent
\textbf{Distractions.} Distractions such as roadside billboards and secondary tasks cause both novice and experienced drivers to take their gaze off-road for extended periods of time \cite{2012_TRR_Divekar, 2011_AppliedErgonomics_Edquist, 2011_CTW_Nabatilan}. However, novice drivers are more affected by it, detecting fewer hazards  \cite{2010_AccidentAnalysis_Konstantopoulos}, showing slowed reaction times \cite{2011_AppliedErgonomics_Edquist}, and making more lane departures \cite{2012_TRR_Divekar}.

\subsubsection{Experience and familiarity of the environment}
A large proportion of drivers rarely drive in a different country and use their vehicles predominantly for commuting within a limited area (\eg to go shopping or to work) \cite{2011_Statistics_Canada, 2017_Household_Travel_Survey}, presumably driving the same routes multiple times. However, little is known about how the familiarity of the route affects attention. There is some evidence which agrees with the intuition that familiarity encourages inattention, \ie drivers tend to pay less attention to critical elements of the scene if they are familiar with the route \cite{2013_TrafficPsychology_Charlton, 2016_AccidentAnalysis_Burdett}. For example, one study reports a significant decrease in attention to the road ahead and traffic signs in terms of the proportion of dwell time and increase of dwell time on non-relevant areas \cite{2018_TransRes_Young}. It should be noted that these conclusions were derived from trips made by a single subject, an expert driving instructor, and did not involve any hazard events. A more recent study using a larger and more diverse pool of participants in a simulation experiment revealed that familiarity may lead to inattention to hazards on some familiar roads but ultimately depends on the context. Specifically, the authors found that effects from road type on changes in attention to hazards were larger than those of familiarity regardless of driving experience \cite{2019_HumanFactorsErgonomics_Beanland}. A similar conclusion was reached in a cross-cultural study involving drivers from the UK and Malaysia \cite{2013_TR_Lim}. It was found that hazard perception skills were transferable overall. Although experienced UK drivers were more sensitive to the familiar hazards in familiar environments, there was no difference in visual patterns depending on the clip origin, suggesting that visual strategy is moderated more by the immediate driving environment rather than familiarity.

Few studies looked at cross-cultural effects. For example, drivers from the US had greater mean and total fixation durations than Japanese participants \cite{2017_AutoUI_Shinohara} and Malaysian drivers' fixation durations were shorter compared to UK drivers when viewing Malaysian traffic scenes \cite{2013_TR_Lim}. But definite conclusions cannot be reached due to different levels of visual complexity in videos from different countries. 

\subsubsection{Experience with other vehicle types}
Motorcyclists and bicyclists represent a small fraction of all road users but are involved in a disproportionate number of accidents, particularly at intersections. Questions of interest are why do drivers so often fail to give way and does dual experience (either as a motor- or pedal cyclist) lead to different visual patterns. There is evidence that indeed dual drivers are more cautious towards vulnerable road users (VRUs) \cite{2011_VR_Underwood, 2012_AccidentAnalysis_Crundall} and spend more time looking at them in terms of the number of fixations \cite{2011_VR_Underwood} and fixation duration \cite{2012_AccidentAnalysis_Crundall}. Since it takes all groups of drivers the same amount of time to notice conflicting VRUs \cite{2012_AccidentAnalysis_Crundall}, this suggests processing rather than a perceptual issue. However, these results were obtained in a low-fidelity simulator where no vehicle control was necessary, and a more recent study conducted in a high-fidelity simulator did not find significant effects of dual experience \cite{2018_HumanFactors_Robbins}. 

\subsubsection{Improving scanning habits of novice drivers} 
Research towards developing better driver training techniques suggests that it is possible to improve the visual scanning patterns of novice drivers. While simply viewing recorded experts' eye movements was not sufficient \cite{2017_DrivingAssessmentConference_Yamani}, training programs that combine simulation training with hazardous situations and instruction sessions explaining the appropriate gaze directions appear to have an effect \cite{2011_TransRes_Vkalveld, 2017_DrivingAssessmentConference_Yamani}. In a study conducted by Vlakveld \etal \cite{2011_TransRes_Vkalveld}, novice drivers had correct gaze patterns in 47\% of the scenarios similar to those shown during the training and 33\% correct gazes in types of scenarios they have not been previously exposed to. Significant improvements have been achieved in a different study \cite{2011_DrivSymposium_Taylor} which also confirmed that the effects of hazard training remain after 6-8 month delay but still do not reach the performance of the experienced drivers. It is, however, difficult to assess whether such interventions improve the crash rates of the novice drivers.

\subsection{Demographics}
\subsubsection{Age}
\label{sec:driver_age}
Age is strongly correlated with crash rates. In the USA, the risk of accidents is the highest for teen drivers who tend to take the most risks while lacking necessary visual and driving skills. Crash risk gradually decreases for drivers in their 30s, 40s, and 50s, reaches a minimum for drivers in their 60s, only to increase again for drivers in their 70s and beyond \cite{2014_Research_Brief}. One of the reasons is that older drivers, despite having decades of driving experience, begin to see an age-related functional decline affecting their driving performance. Thus understanding the effects of age on driving safety is a pressing concern given that the proportion of elderly drivers has been steadily increasing. In Canada, drivers in the 65+ age group comprised $12\%$ in 2000 \cite{2000_Traffic_Collision_Statistics} and nearly $18\%$ in 2016 \cite{2016_Canadian_Collision_Stats}. Similarly, in 2017, $17\%$ of all licensed drivers in the USA and $22\%$ in Japan were over the age of 65 \cite{2017_Highway_Statistics, 2017_Japan_Driver_Statistics}.

There is no definition of \textit{older} in the literature, although a cut-off of 60 or 65 years has been common \cite{1991_HumanFactors_Waller, 1993_GeriatricMedicine_Retchin, 2020_JAppGeront_Ratnapradipa}. Some authors point out that the information processing decline is minimal in the late 60s \cite{2018_AccidentAnalysis_Sun, 2016_PLOS_Yamani}; however, the majority of the studies we reviewed focus on healthy older drivers in their 60s and 70s, and some even include drivers 50+ years old \cite{2013_AccidentAnalysis_Peng, 2013_AccidentAnalysis_Dozza, 2019_TransRes_Miller}, whereas drivers above the age of 80 are rarely considered (\eg \cite{2013_TR_Romoser, 2019_TransRes_Clark}). This may explain some of the inconsistencies and small statistical effects of age reported in the studies.

Effects of aging manifest themselves more in situations with complex road geometry and traffic, especially when compounded by other distractions. For instance, slowed visual processing resulted in longer fixations of older drivers at intersections \cite{2016_JEMR_Sun}. Older drivers also made fewer fixations on important objects at intersections and roundabouts which was related to deficits in selective attention capacity \cite{2016_JEMR_Sun, 2018_AccidentAnalysis_Sun, 2013_TR_Romoser, 2013_SafetyResearch_Scott} or to compromised ability to multitask resulting in prioritization of glances necessary for vehicle control \cite{2016_PLOS_Yamani}. Similarly, when operating a hand-held cellphone while driving, older drivers had more trouble keeping the vehicle in the lane and paying attention to the road than younger groups \cite{2011_AccidentAnalysis_Owens}. Difficulty to adjust search patterns in the presence of distractors was also shown in studies on hazard detection not requiring vehicle control \cite{2016_OptometryVisionScience_Lee}.

Overall, older drivers have more conservative visual attention allocation strategies \cite{2018_HumanFactorsErgonomics_Zahabi} and focus longer on task-relevant objects \cite{2015_BMCGeriatrics_Urwyler} than younger groups. In simulation studies, when exposed to external distractors, such as billboards, older drivers were not as accurate at detecting relevant signs \cite{2018_HumanFactorsErgonomics_Zahabi}, spent less time looking off-road \cite{2011_AppliedErgonomics_Edquist, 2016_TR_Stavrinos}, had lower off-road glance duration and frequency \cite{2018_HumanFactorsErgonomics_Zahabi}, and adjusted their driving by slowing down to read the roadside signs, presumably to compensate for the reduced visual function \cite{2017_AppliedErgonomics_Zahabi, 2018_HumanFactorsErgonomics_Zahabi}. However, in an on-road experiment, older drivers made more long glances at billboards than younger age groups \cite{2016_AccidentAnalysis_Belyusar}.

One of the positive outcomes of automation and ADAS is improved mobility of the elderly, however, the interaction of this group of drivers with new technology is still not well understood. Several recent studies looked into attention allocation changes when taking over from automation after being engaged in a secondary task, but none found significant age-related effects for gaze measures \cite{2017_AccidentAnalysis_Clark, 2019_TransRes_Miller, 2019_TransRes_Clark}, possibly due to the small sample size and inclusion of relatively young drivers in their 50s.

\subsubsection{Gender}
There are significant differences in traffic violations of male and female drivers, especially for people 18-35 years of age. Male drivers are more often involved in accidents, receive more traffic fines and make more insurance claims \cite{2018_Fatality_Facts}, which is typically associated with their propensity towards risk-taking \cite{2011_AccidentAnalysis_Rhodes} and overconfidence in their driving ability \cite{2005_SSJ_Bergdahl}. However, only a few studies looked for gender-related differences in attention allocation, even though most recruited participants of both genders.

No gender-related differences in eye-movement patterns were found in rear-end collision avoidance scenarios in the presence of distractions \cite{2018_TRR_Li}. Dozza \etal \cite{2013_AccidentAnalysis_Dozza} found that response times in near-crash situations were slower for all drivers due to inattention but did not differ significantly by gender. Miller \etal \cite{2019_TransRes_Miller} showed that drivers exposed to partial automation were more likely to engage in secondary tasks and made longer off-road glances even after returning to manual driving, however, no gender differences were revealed.  Likewise, no differences were found in the vehicle control due to inattention with eyes-on-road (\eg talking to passengers) and eyes-off-road (\eg looking inside the vehicle) \cite{2013_AccidentAnalysis_Peng}. However, drivers, especially females, made more looked-but-failed-to-see (LBTFS) errors when they were involved in a conversation with research confederate. This resulted in a higher number of incidents with pedestrians compared to solo trips \cite{2010_AccidentAnalysis_White}.  

In accord with crash statistics, female drivers showed better abilities to obtain important information when approaching intersections; they looked more often and focused longer on the conflict vehicles and were involved in fewer collisions in a simulation study \cite{2016_PLOS_Yan}. In a study by Reimer \etal  \cite{2014_TR_Reimer}, young male drivers spent less time looking at the road and were more prone to make continuous long glances ($>2$s) towards a cell phone while driving than females, although subjects of both genders indicated that phone use is distracting and rated themselves as safe drivers.

Overall, the studies on gender differences are too sparse to make any definite conclusions about gender differences in attention allocation.

\subsection{Driver's state}
\label{sec:driver_state}
Driver internal state can be characterized in terms of physiological (\eg drowsy, drunk) and emotional factors (\eg stressed). However, there are more studies on the associated changes in driving performance and comparatively few papers relating these factors directly to visual attention.

\vspace{0.5em}
\noindent
\textbf{Drunk driving} is illegal in almost every country in the world because it increases collision risk due to the adverse influence of alcohol on neurocognitive mechanisms. In particular, alcohol-induced effects on visual attention include a reduction in the top-down regulation of gaze control leading to a more dispersed and random distribution of gaze \cite{2019_DrugAlcoDependence_Shiferaw}, deficiencies in visuomotor coordination, longer reaction times and increased distractibility (see the review by Shiferaw \etal \cite{2014_CurrDrivAbuseRev_Shiferaw}).

\vspace{0.5em}
\noindent
\textbf{Drowsiness} after sleep deprivation or fatigue caused by the monotony of driving has been associated with many adverse effects on driving performance \cite{2005_AAP_Philip}, but its impact on attention is less studied. Part of the difficulty is that unlike alcohol intoxication, drowsiness is more difficult to induce and detect. Self-reported drowsiness (\eg measured using Stanford Sleepiness Scale \cite{1973_Psychophysiology_Hoddes} or Karolinska Sleepiness Scale \cite{1999_IJN_Akerstedt}) often used in studies does not always correlate with the objective physiological measures.

\begin{itemize}
\itemsep0em
\item{\textit{Sleep deprivation.}} Even one night without sleep has been shown to change attention allocation significantly. A naturalistic study of shift workers showed that $62\%$ of trips were drowsy according to the PERCLOS metric \cite{2019_SafetyScience_Kuo}. In lab conditions, PERCLOS of sleep-deprived drivers during the $30$ min drive was more than double that of the control group \cite{2016_TraffInjuryPrevention_Jackson}. Drowsy drivers also spend more time with their eyes closed due to the reduced rate and duration of blinks and fewer fixations. At the same time, larger saccade amplitudes and gaze entropy measures indicate a more dispersed distribution of gaze and random patterns of visual scanning. Together these effects point to impairments of top-down modulation of visual attention \cite{2018_NatSciReports_Shiferaw}. Besides losing visual information due to long blinks, drowsy drivers look off-road more often and for extended periods of time, possibly seeking novel stimuli to offset drowsiness \cite{2019_SafetyScience_Kuo}.

\item{The \textit{monotony}} of driving has been studied primarily in the context of automated driving since these effects require a much longer time to manifest themselves during manual driving \cite{2014_THMS_Ho}. For instance, when driving manually, drivers reached a self-reported medium level of fatigue after $45-50$ min, whereas when automation was engaged, half of the subjects reached that level after only $20$ min \cite{2019_AccidentAnalysis_Vogelpohl}. Task-induced fatigue is worse when combined with sleep-deprivation \cite{2018_NatSciReports_Shiferaw, 2019_AccidentAnalysis_Vogelpohl}. Reduced arousal levels compared to manual driving conditions were shown to induce drowsiness as indicated by increased PERCLOS \cite{2013_TR_Jamson} and micro-sleeps \cite{2016_IV_Schmidt}. Drowsiness may also be a result of complacency, as extreme drowsiness was more prevalent in cases where subjects were not given information about the system limitations and were not provided with any warning informing them of possible failures \cite{2018_HumanFactors_Victor}. An obvious consequence is that fatigued drivers are slower to react when the automated system fails and may lack sufficient situation awareness to take appropriate actions \cite{2019_AccidentAnalysis_Vogelpohl}. The gamification of automated driving has been proposed to mitigate boredom \cite{2017_CHB_Stenberger}, but more studies are needed to understand how to reduce the effects of monotony and improve drivers' engagement (see also Section \ref{sec:automation} on effects of automation).
\end{itemize}

\vspace{0.5em}
\noindent
\textbf{Emotional state} and its effects on drivers' attention are not well-known, partly because of ethical issues in the design of such experiments. Jones \etal \cite{2014_AccidentPrevention_Jones} found that both positive and negative valence resulted in changes in the subjective perception of risk when looking at still frames of hazards overlaid with affective imagery. The authors reported shorter fixation durations for modified images of hazards suggesting an inhibitory influence of the emotional state on hazard perception. Although they did not find reduced spread of visual search (also referred to as vision tunneling) associated with emotions reported in the earlier study \cite{1994_TheHeartsEye_DerryBerry}, it could be because of the highly artificial nature of the task. In a simulation study, Briggs \etal \cite{2011_TR_Briggs} simulated emotional involvement by selecting drivers with arachnophobia and measuring changes in their eye measurements during conversations about spiders. Their findings indicate inefficient visual scanning patterns typical of cognitive and visual tunneling resulting in decreased visual awareness and vehicle control. As a compensatory strategy, many of the phobic participants decreased the speed of the vehicle.

\subsection{Secondary task}
\label{sec:secondary_task}
In addition to following the rules of the road, drivers are expected to be fully aware of their surroundings to cooperate with other road users and respond to unforeseen events. However, surveys (USA \cite{2018_Traffic_Safety}), crash reports (EU \cite{2015_Distraction_Study}), and data from naturalistic studies \cite{2015_TR_Victor} indicate that a significant proportion of drivers engage in secondary tasks in practice. Secondary non-driving related tasks (NDRT) are defined as any activity that is not related to controlling the vehicle or improving forward, indirect or rearward visibility \cite{2013_TR_Angell}. Some of the most common activities include operating a cell phone, using in-car navigation and entertainment systems, talking to the passengers, eating or drinking, smoking, \etc \cite{2010_TR_Klauer}.

Given the road crash statistics associated with distracted driving, particular activities such as using cell phones while driving have been heavily researched. A large two-part study by Collet \etal \cite{2010_Ergonomics_Collet1, 2010_Ergonomics_Collet} was dedicated to the effects of phoning on driving behavior. The authors note that epidemiological studies show that frequent phone use in the vehicle leads to an increase in the risk of a car crash due to slowing of reaction time by $15-40\%$ and impaired detection ability, which, in turn, delays brake reaction by $0.5$s \cite{2010_Ergonomics_Collet1}. Phone use, whether hand-held or hands-free, leads to adverse effects on both foveal and peripheral vision \cite{2010_Ergonomics_Collet}. More recent surveys also note increased gaze concentration implying less peripheral awareness and detection sensitivity, more frequent and longer off-road glances, and slowed information processing ability \cite{2016_TR_OviedoTrespalacios, 2018_HumanFactors_Caird}. Systematic reviews of the literature show substantial driving performance decrease across many measures caused by the secondary task involvement \cite{2012_SafetyScience_Young, 2013_AJPH_Ferdinand, 2019_TITS_Khatib}. Cognitive distractions not caused by technology, \eg daydreaming, stress, are another common cause of accidents \cite{2019_TrafficInjuryPrevention_Wundersitz} but are less studied due to difficulties in detecting such states.

\subsubsection{Taxonomy of the secondary tasks}

Although most statistics are reported by grouping the activities by type, a demand-based categorization provides a better understanding of what cognitive functions are affected by the task. In terms of \textit{modality} tasks are typically a combination of one or more of the following \cite{2020_EU_Road_Safety}:
\begin{itemize}
	\itemsep0em
	\item{Visual} - require averting gaze off the road (\eg looking at the in-vehicle navigation system);
	\item{Cognitive} - require thinking (\eg talking to the passenger, solving puzzles or memorizing information);
	\item{Manual} - require to take one or both hands off the wheel (\eg smoking, drinking).
\end{itemize}

Other dimensions have also been considered, for example, \cite{2011_IET_Spiessl} proposed interruptibility (good or bad), degree of interaction (passive or active), and information coding (verbal or spatial) in addition to modality dimension. 


\subsubsection{Effects of secondary tasks}
\label{sec:secondary_tasks_effects}

\vspace{0.5em}
\noindent
\textbf{Visual tasks}, as mentioned before, require the driver to look away from the road to read messages on the roadside signs or their phones.
	\begin{itemize}
	
	\item{\textit{Visual search}} is a common perceptual task where the subject is asked to find the target object among other objects in the environment and verbally report the result. Drivers occasionally have to search for gas station logos on the highway exit signs or to check the distance to their destination on the guiding signs \cite{2013_AppliedErgonomics_Zhang, 2015_AppliedErgonomics_Kaber, 2019_TRR_Pankok}. Reading such signs is generally not difficult for drivers as studies report detection rates in the $96-98\%$ accuracy range \cite{2019_TRR_Pankok, 2015_AppliedErgonomics_Kaber}. Particularly, guide signs that contain a few words within a relatively small area are easily accessible and not as visually demanding as vendor signs \cite{2015_AppliedErgonomics_Kaber, 2019_TRR_Pankok}. Artificial visual search tasks such as searching for target characters in random text strings \cite{2016_AutoUI_Smith} or grids \cite{2015_TRR_Samuel} allow for a larger range and precise control of task complexity. Sometimes, two approaches are combined, \eg searching for characters in roadside signs \cite{2012_TRR_Divekar}.
	
	In all studies, attention on the road ahead was affected by the visual search task as evidenced by longer and more frequent glances off-road \cite{2016_AutoUI_Smith, 2013_AppliedErgonomics_Zhang, 2019_TRR_Pankok, 2018_HumanFactorsErgonomics_Zahabi, 2017_AppliedErgonomics_Zahabi}. Although some minor effects of attention allocation changes on driving performance were detected, they are inconsequential for road safety \cite{2013_AppliedErgonomics_Zhang, 2015_AppliedErgonomics_Kaber, 2019_TRR_Pankok}. More importantly, there was significant degradation of hazard perception and anticipation. Even fast-moving pedestrians failed to attract attention in the presence of distractions \cite{2012_TRR_Divekar}. Likewise, drivers were not able to obtain sufficient information for latent hazards given shorter forward roadway glances when distracted \cite{2015_TRR_Samuel, 2012_TRR_Divekar}. However, a reduction in hazard detection was not linked to crash risk directly.

	\item{\textit{Text reading}}. Drivers may sometimes need to read road message signs that serve as a reminder (\eg fasten the seatbelts) or notification about changing traffic conditions. In experiments, subjects were expected to read and report the contents of the messages. As with the visual search task, longer off-road glances were observed \cite{2019_TransRes_Louw, 2016_AutoUI_Hurtado, 2014_HumanFactorsErgonomics_Schieber, 2011_AppliedErgonomics_Edquist} and increased non-linearly as a function of the message length and speed \cite{2014_HumanFactorsErgonomics_Schieber}. Similarly, reading text from the in-vehicle screen or mobile device reduces the amount of time looking at the road \cite{2015_AccidentAnalysis_Peng, 2014_SafetyScience_Young}, but the effect is not as drastic as during text entry.
\end{itemize}

\vspace{0.5em}
\noindent
\textbf{Visual-manual tasks} are associated with more random scanning patterns, a larger number of glances, shorter mean glance durations, and shorter on-road glance durations \cite{2017_JSR_Wang}. Some studies note that visual-manual tasks have a higher impact because they compete with cognitive and attentional resources needed for driving tasks \cite{2007_DistractedDriving_Young}.

\begin{itemize}
\itemsep0em
\item{\textit{Surrogate Reference Task (SuRT)}} is a generic visual-manual secondary task where the user is required to identify a target stimulus within an array of distractors shown on display and select it (via touchscreen or button press) \cite{ISO_14198}. It is frequently used to simulate smartphone usage, especially during highly-automated driving \cite{2018_ITSC_Yang, 2017_AdvErgonomics_Feldhutter, 2019_AutomotiveUI_Walch}. However, when this task is being continuously performed, the effects on gaze compared to baseline condition are not reported \cite{2019_AutomotiveUI_Walch, 2013_TransRes_Benedetto, 2014_HumanFactorsErognomics_Lorenz, 2018_ITSC_Yang}. One study reports more frequent but shorter glances on the road as a result \cite{2017_AdvErgonomics_Feldhutter}. If drivers perform the task on a mobile device, they tend to hold it higher to increase their dwell time on the road and to observe the changes \cite{2018_ITSC_Yang}.

\item{\textit{Text entry}} may be needed to answer text messages or to enter an address into the navigation system. Entering text is more disruptive to driving than reading messages since it leads to longer off-road glances and overall fewer glances on the road \cite{2011_AccidentAnalysis_Owens, 2014_SafetyScience_Young, 2014_TransRes_Tivesten, 2015_AccidentAnalysis_Peng, 2019_TransRes_Hashash}. Some drivers appear to have better compensatory strategies than others resulting in considerable variability of long-glances off-road \cite{2014_SafetyScience_Young}. Typing is more affected by the message length than reading and has been linked to other risky behaviors \cite{2014_SafetyScience_Young}. Errors when entering text further disrupt dual-tasking by increasing task time and total time looking away from the road \cite{2016_CHI_Lee}. Replacing manual entry with a voice interface does not solve the eyes-off-road problem because drivers often continue to monitor the screen for confirmation that their command was understood  \cite{2014_ACM_Reimer}. \textit{Dialing a number on a cell phone}  leads to fewer long glances than texting, but still more than reading \cite{2014_TransRes_Tivesten}. Tactile keyboard may mitigate some of negative effects only for those used to this method of entry \cite{2014_TR_Reimer}.

\item{\textit{Radio tuning}} is a standard task proposed as a benchmark for a reasonable level of distraction during driving \cite{2012_NHTSA_Driver_Distraction}. Usually, it involves three steps -- selecting a radio function, selecting a radio band, and tuning to the station using a knob -- although other variations also exist. As with other secondary tasks, it primarily affects long glances off-road  \cite{2014_ACM_Reimer, 2015_TransRes_Bargman} and total eyes off road time \cite{2018_AccidentAnalysis_Lee}. \textit{Playlist search} is another common secondary task \cite{2010_TransEng_Bonmez, 2012_HumanFactors_Lee, 2013_PUC_Kujala}. Although searching through the short list of about 20 songs is comparable in complexity to a tuning task, a longer list of over 80 items significantly undermined driving performance and required more glances to the device \cite{2012_HumanFactors_Lee}.  
\end{itemize}

\vspace{0.5em}
\noindent
\textbf{Cognitive tasks}. Mind-wandering is difficult to identify but cannot be ruled out in studies, especially ones involving automation where drivers have more spare attention capacity \cite{2020_Information_Feierle}. Thus, artificial cognitive tasks are designed to induce such a state in the drivers or to increase their cognitive load in general.

\begin{itemize}
\item{\textit{Delayed digit recall (N-back task)}} is an artificial task initially proposed for measuring working memory capacity \cite{1958_JEP_Kirchner}. During the experiment, subjects listen to digits presented and repeat each as presented (0-back), repeat previous digit (1-back) or digit two steps before (2-back). As expected, task performance for the 0-back task is very high at 99\% but degrades to 85\% for the 2-back task \cite{2012_HumanFactors_Reimer} (similar results are reported when the subjects were not driving \cite{2018_FrontiersPsychology_Gajewski}). Somewhat contradictory effects of this task are reported. For example, two studies indicate a decrease in horizontal gaze dispersion due to high cognitive demand, suggesting reduced awareness of the periphery \cite{2010_HumanFactorsErgonomics_Reimer, 2012_HumanFactors_Reimer}, but in a more recent experiment, the gaze distribution decreased uniformly across the visual field affecting foveal vision as well \cite{2016_HumanFactors_Gaspar}. Another study found an effect on blink rate but not other gaze measures or driving performance \cite{2015_TransRes_Niezgoda}.

\item{\textit{Conversations}} with passengers or on the handsfree phone despite being common and socially acceptable can nevertheless lead to distraction \cite{2017_AccidentAnalysis_Clark, 2018_TRR_Li}, especially if they affect drivers' emotional state \cite{2011_TR_Briggs}. When talking to someone, drivers fixate less on critical objects \cite{2013_ACP_Garrison} and make more LBFTS errors \cite{2010_AccidentAnalysis_White}.

\end{itemize}

\vspace{0.5em}
\noindent
\textbf{Visual-cognitive tasks} such as \textit{watching videos} are frequently requested and performed during automated driving. They may mitigate some effects of automation fatigue \cite{2013_TR_Jamson} at the cost of significantly reducing drivers' attention towards the forward roadway \cite{2016_AccidentAnalysis_Zeeb, 2013_TR_Jamson, 2020_Information_Feierle, 2020_TransRes_Li}.

There is little work on the combination of all three modalities \cite{2012_TR_Kaber, 2013_TITS_Yekhshatyan, 2016_AutoUI_Borojeni} and only a few studies that compare effects of different secondary tasks \cite{2016_AccidentAnalysis_Zeeb, 2010_TR_Klauer} or same tasks in different mediums \cite{2020_TransRes_Li} and complexity levels \cite{2016_AutoUI_Smith}.

\subsection{Environment}
Drivers are constantly exposed to a rich stream of visual information from the environment. To some of these elements, the driver must pay close attention (\eg traffic signs and construction zones), while others should be ignored (\eg billboards) despite being conspicuous by design. Besides, the properties of the road (curved, straight, intersection) and traffic (dense or sparse) combined with other environmental factors (such as wind and visibility) may contribute to increased cognitive and attentional load. 

\subsubsection{Intersections}
\label{sec:intersections}
Intersections comprise only a small portion of the road network but lead to many accidents with severe consequences. More than $30\%$ of fatal crashes in the USA \cite{2020_FHWA_Intersection_Safety} and nearly a quarter in Europe occur at intersections \cite{2015_EU_Traffic_Safety}. Many of these crashes are caused by the driver not understanding the traffic situation due to lack of knowledge or lack of observation, which could be caused by physical obstruction, inattention, or neglect. 

Overall, driving through intersections is more visually taxing than driving on straight roads \cite{2016_JEMR_Sun}. Visual behavior at the intersections also depends on \textit{geometric structure}, \textit{type} (e.g signalized), \textit{conflicting road user type} (\eg bicycle vs car), \textit{traffic density}, and \textit{driver's experience} and \textit{age} (see Section \ref{sec:driver_age}). More specifically,  the \textit{geometric structure} of an intersection can affect the angle of approach of the conflicting vehicle. For example, a perpendicular angle of approach results in better estimation of TTC than obtuse angles since sideways movement provides easier to interpret optical flow information than the movement towards the observer. As a result, fixation durations for the most challenging conditions (obtuse angle of approach and high ego-vehicle speed) were the highest, suggesting increased visual processing load \cite{2010_Perception_vanLoon}. Reduced field of view at intersections due to buildings or vegetation also forces the drivers to scan the environment more since the conflict vehicle is not easily observable \cite{2016_PLOS_Yan}. The \textit{type} of intersection focuses drivers' attention on different areas of the road. For example, when a stop sign or traffic signal is present, less time is spent observing the intersecting zone and gaze is more focused on the ego-lane or the signal \cite{2014_JEMR_Lemonnier, 2015_TR_Lemonnier, 2019_JSR_Li}, but when the drivers have priority, they continue to scan the intersecting areas in anticipation of possible interactions \cite{2015_TR_Lemonnier}. Drivers also react differently depending on the \textit{conflicting road user type}. Motorcycles and cyclists are smaller than cars and are perceived as less risky \cite{2011_VR_Underwood, 2018_HumanFactors_Robbins}. They are also less salient and not easy to notice, therefore drivers react to them slower and fixate on them more, especially when farther away \cite{2011_VR_Underwood}. When approaching busy intersections, drivers gaze more at the traffic and start looking at pedestrians when waiting for a gap or preparing to leave the intersection \cite{2014_CogTechWork_Werneke}.



\subsubsection{Billboards and traffic signs}
\label{sec:billboards_signs}

Roadside signs provide important information such as speed limit or distance to the next gas station. Signs are not conspicuous, except stop and construction zone signs, requiring deliberate attention. However, drivers often miss signs \cite{2017_AccidentAnalysis_Beanland}, \eg they notice $1$ in $4$ and recall only $7\%$ of the signs according to one study \cite{2014_TransRes_Costa}. A part of the reason is that roadside signs are difficult to identify with peripheral vision while drivers focus on the forward roadway. Experiments show that only signs with very distinct shapes and colors, \eg octagonal and cross signs \cite{2018_Ergonomics_Costa}, are immune to peripheral degradation even when presented at high eccentricity. Signs are also more likely to be overlooked if the driver is distracted \cite{2013_ACP_Garrison}. Other contributing factors are inattentional blindness, high attentional load, or lack of proper visual scanning routines \cite{2014_TransRes_Costa}.

\begin{figure}[t!]
\centering
\begin{subfigure}{0.29\linewidth}
  \includegraphics[width=1\linewidth]{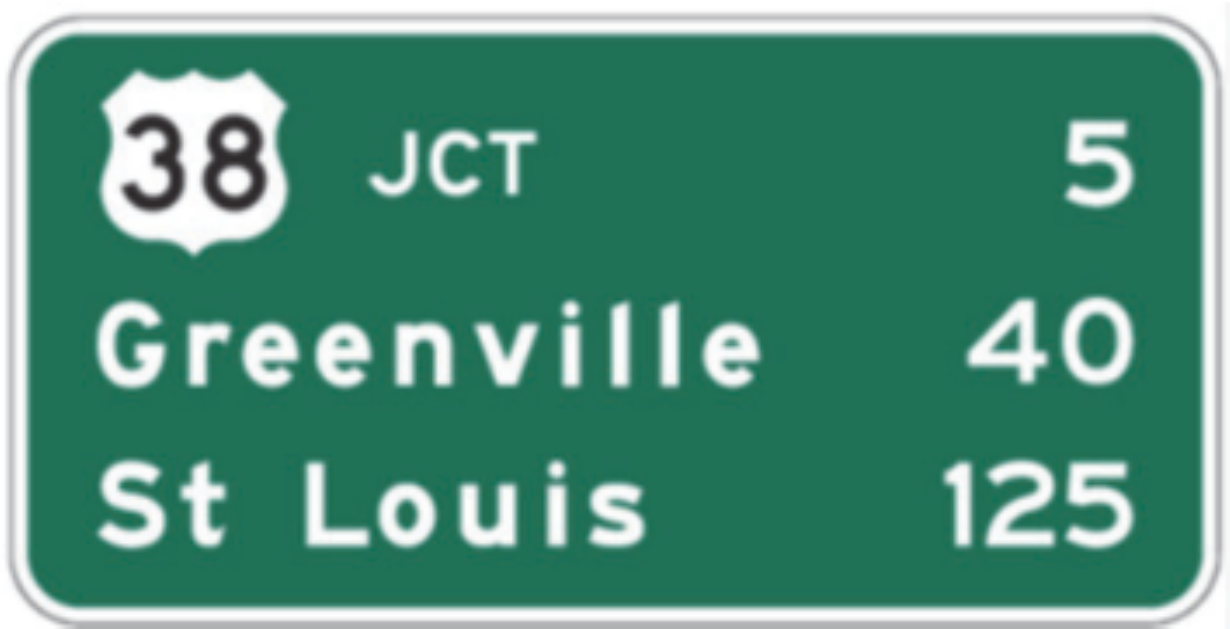}
\caption{Guidance sign}
\label{fig:guidance_sign}
\end{subfigure}
\begin{subfigure}{0.29\linewidth}
  \includegraphics[width=1\linewidth]{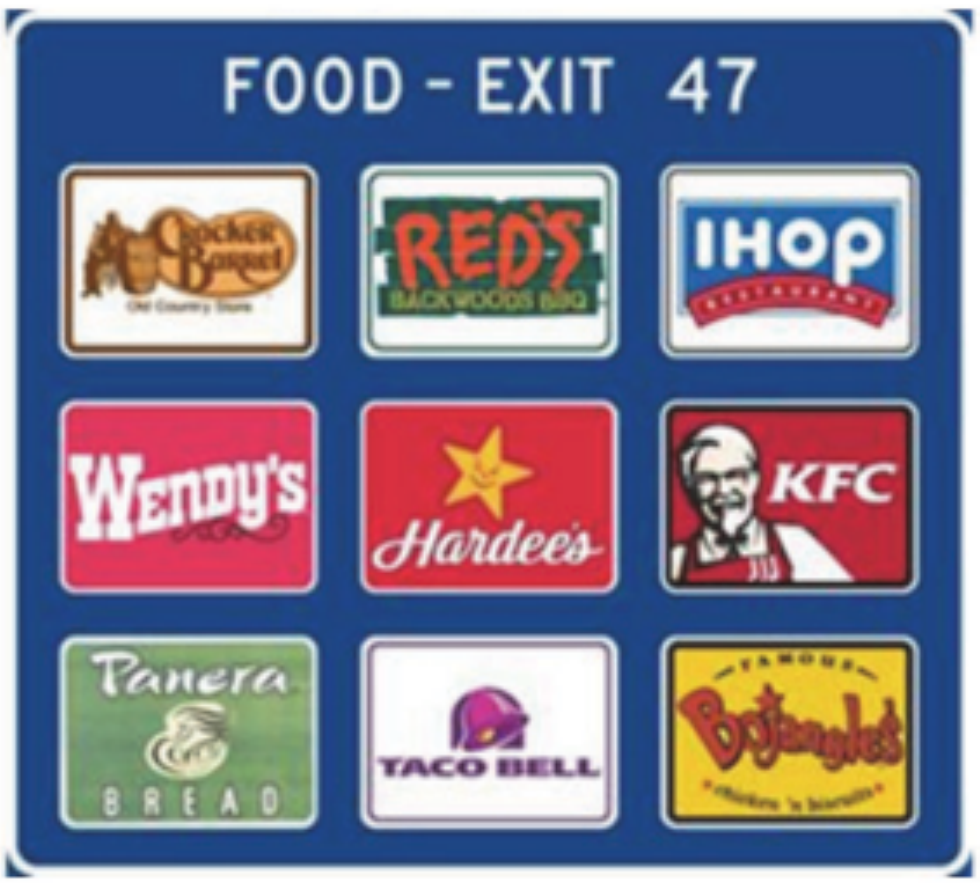}
\caption{9-panel logo sign}
\label{fig:9-panel_logo}
\end{subfigure}
\begin{subfigure}{0.29\linewidth}
 \includegraphics[width=1\linewidth]{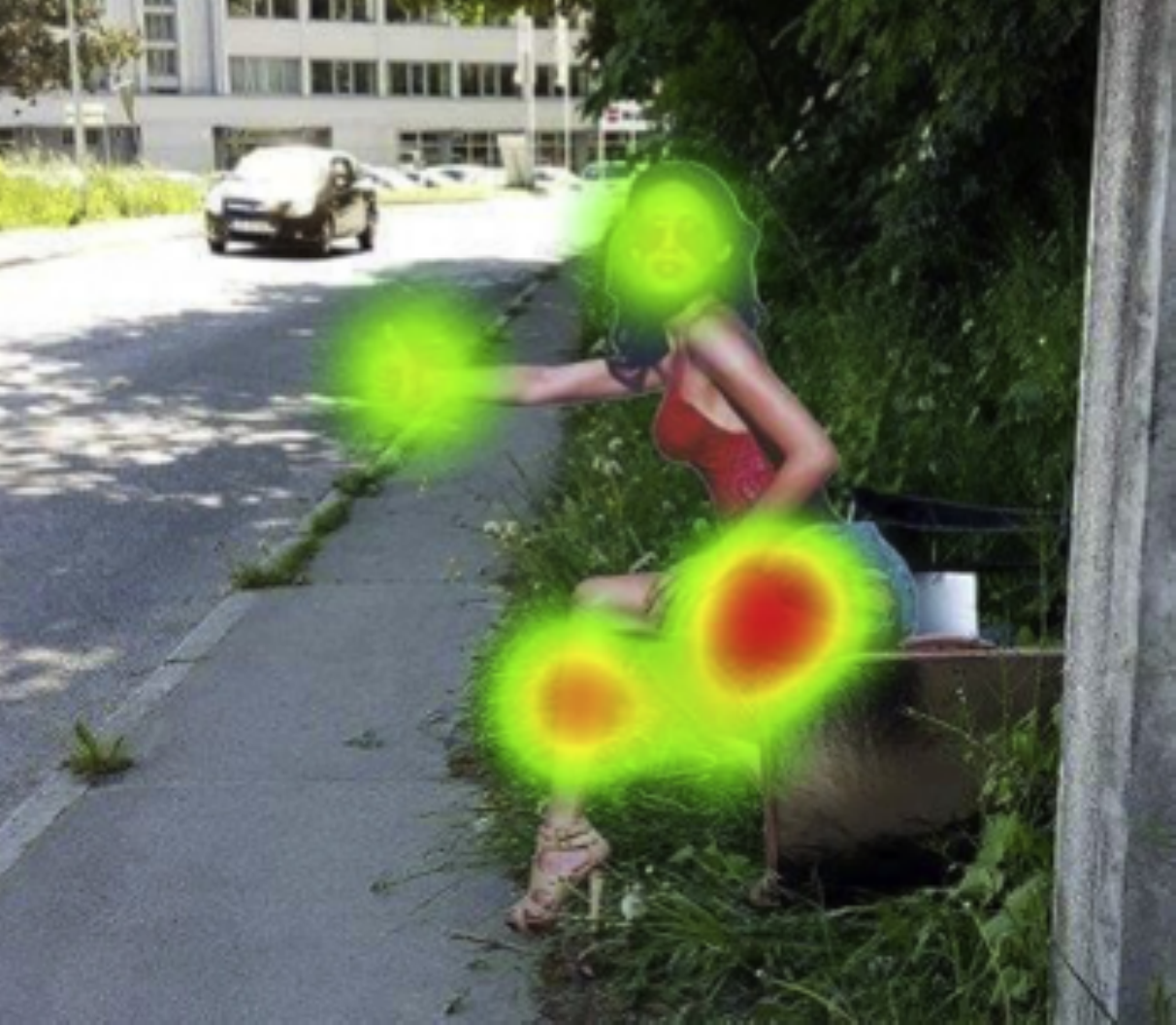}
  \caption{Provocative sign}
  \label{fig:provocative_sign}
\end{subfigure}
\caption[Examples of typical roadside signs and billboards used in the experiments]{Examples of roadside signs: a) guiding sign, b) 9-panel logo sign at the highway exit, c) "Hitchhiker" advertising with overlaid participants gazes. Sources: a) and b) \cite{2019_TRR_Pankok}, c) \cite{2016_TR_Topolsek}.}
   \label{fig:billboards_and_signs}
\end{figure}

Occasionally, drivers need to read the guiding signs to find the distance to the target destination or to check what vendors are available at the next exit (Figures \ref{fig:guidance_sign} and \ref{fig:9-panel_logo}). Simple distance signs with few destinations do not put a significant strain on the drivers and do not affect their vehicle control \cite{2018_HumanFactorsErgonomics_Zahabi, 2019_TRR_Pankok, 2014_HumanFactorsErgonomics_Schieber}. More complex logo signs require more attention and result in longer off-road glances \cite{2019_TRR_Pankok}, although there is conflicting evidence regarding the influence of sign visual complexity on driver attention. One study found no differences between standard 6- and 9-panel signs \cite{2015_AppliedErgonomics_Kaber}, another reported effect of the number of panels and target location on off-road glance duration \cite{2013_AppliedErgonomics_Zhang}, and yet another study showed increased attentional demand but only for unfamiliar targets in the 9-panel layout \cite{2017_AppliedErgonomics_Zahabi}. Some degradation of driving performance was also found. For instance, speed reduction was associated with complex 9-panel signs \cite{2013_AppliedErgonomics_Zhang}, and lane-keeping errors were more frequent when drivers attempted to read long text sequences from signs while maintaining high speed \cite{2014_HumanFactorsErgonomics_Schieber}.

Unlike signs, billboards with product advertising are irrelevant for the driving task but are designed to be salient (see Figure \ref{fig:provocative_sign}). They attract more attention than any other roadside object (especially large billboards, ads with provocative content \cite{2016_TR_Topolsek}, or visually complex ads \cite{2019_AppliedErgonomics_Costa}), resulting in higher fixation rates, dwell times, and proportion of long off-road glances \cite{2013_TrafficInjuryPrevention_Dukic, 2019_AppliedErgonomics_Costa, 2011_AppliedErgonomics_Edquist}. Digital advertising with frequent transitions between content is particularly distracting as drivers are predisposed to react to motion in the periphery \cite{2016_AccidentAnalysis_Belyusar, 2016_TR_Stavrinos, 2013_TrafficInjuryPrevention_Dukic}. Although long off-road glances have been associated with increased accident rates \cite{2015_TR_Victor}, there is no sufficient evidence to establish a direct link between changes in visual behavior due to the billboard presence and risk of crashes, according to the recent meta-studies \cite{2015_TrafficInjuryPrevention_Decker, 2019_TR_OviedoTrespalacios}.


\subsubsection{Traffic density}
In dense traffic or construction zones, the drivers may need to make frequent speed adjustments and have shorter braking affordances. As a result, they pay more attention to the rear-view mirrors \cite{2013_AccidentPrevention_Wong} as maintaining situation awareness becomes more difficult \cite{2017_AppErgonomics_Lu}. Drivers hesitate to start secondary tasks \cite{2013_AccidentPrevention_Wong, 2012_HumanFactors_Lee} or compensate by making shorter gazes away from the forward roadway \cite{2012_HumanFactors_Lee, 2014_TransRes_Tivesten, 2015_AccidentAnalysis_Peng}. When approaching an intersection or merging on a highway, drivers tend to scan more to find a gap and monitor the approaching vehicles \cite{2016_PLOS_Cheng, 2012_AccidentAnalysis_Werneke}.

\subsubsection{Road geometry and visibility}
It has been shown that road geometry (straight or curved) has an effect on gaze allocation (see discussion in Section \ref{sec:vehicle_control}).

Depending on the location, rural or urban, drivers may be more primed for specific events. For example, drivers identify more pedestrians and fixate on them longer in urban residential areas than more sparsely populated suburbs \cite{2012_AccidentAnalysis_Borowsky}. On rural roads, drivers are more sensitive to oncoming vehicles, perhaps because they pose a larger threat (since no median barrier was present) or because few moving objects attracted attention  \cite{2014_TransRes_Tivesten}. Busy urban locations are more visually demanding, leading to more frequent and short fixations with the wide spatial distribution and increased ``looked but failed to see'' errors compared to rural areas \cite{2017_AccidentAnalysis_Beanland}. 

Since most experiments are conducted during daytime and in clear weather, less is known about the impact of lighting and weather on drivers' visual behavior. Due to a combination of safety and scheduling issues, only a handful of in-vehicle experiments are conducted at night \cite{2019_JTEPBS_Zhang, 2018_TITS_Morando, 2017_JSR_Wang, 2013_AccidentAnalysis_Dozza} and none in adverse weather conditions. Even in some naturalistic studies, only a fraction of the data was recorded during bad weather \cite{2014_TransRes_Tivesten, 2018_TransRes_Hammit}. Surprisingly, bad weather and visibility are rarely modeled in a simulation where safety is not a concern.

Existing evidence shows that bad visibility conditions due to weather or time of day do impact drivers' attention. In bad visibility conditions, sampling rates are reduced, and fixations are longer, suggesting increased processing time \cite{2010_AccidentAnalysis_Konstantopoulos}. During rain, drivers tend to focus more on the road \cite{2014_TransRes_Tivesten}. Driving at night affects mean fixation durations \cite{2017_JSR_Wang} and visual exploration. Since the preview distance of the road is shorter and detection is degraded, drivers focus more on the forward path and driving-relevant objects \cite{2015_BMCGeriatrics_Urwyler, 2018_TITS_Morando}.

\subsection{Driving assistance and automation}
\label{sec:automation}
For a long time, safety and driver-assistive technology were available primarily in luxury commercial vehicles. Over the past decade, features such as adaptive cruise control (ACC), emergency braking, lane departure warning (LDW), lane keeping system (LKS), park assist and many others are becoming standard in new mid-price segment vehicles and soon may be a requirement in some countries (\eg in EU \cite{2019_Road_Safety}). Consequently, research on the impact of assistive and automated features on attention allocation has seen a significant increase. Specifically of interest are partially- and conditionally automated systems that correspond to SAE Level 2 (where two or more driving functions are automated, \eg ACC+LKS) and Level 3 (autonomous but requires intervention), although Level 1 systems where only one aspect of driving is automated (\eg steering) are also considered.  Most experiments are conducted in medium- to high-fidelity simulations and only a few on-road using commercially available technology (\eg Level 2 automated driving in Volvo \cite{2018_TITS_Morando} and Tesla vehicles \cite{2019_HumanFactors_Gaspar} or Lincoln Active Park Assist \cite{2018_TransRes_Kidd}).



\subsubsection{General effects of automated driving}
Significant changes in drivers' attention allocation occur as various aspects of driving are automated and as the role of the driver switches from being an operator to a more supervisory role. For example, when steering was automated significant decrease in visuomotor coordination was observed \cite{2018_TIV_Wang}, and drivers stopped looking at the near road section, suggesting that they were not engaged in the lane position adjustment control loop \cite{2019_IJHCI_Navarro}. Drivers were also less likely to observe the environment around them when using park assist and instead focused on the rear-camera view \cite{2018_TransRes_Kidd}. Multiple studies report that when level 1-3 automation is enabled, drivers divert their gaze from the road and are more prone to engaging in non-driving activities \cite{2017_AdvErgonomics_Feldhutter, 2017_AccidentAnalysis_Clark, 2020_TransRes_Li, 2018_TITS_Morando, 2016_AccidentAnalysis_Morando, 2018_AHAT_Feldhutter, 2016_AccidentAnalysis_Morando, 2019_HumanFactors_Gaspar, 2018_TransRes_Kraft}. Decreased attention may also depend on the situation or trust in automation since drivers were less distracted when traffic or visibility conditions degraded \cite{2013_TR_Jamson, 2017_TransRes_Louw, 2019_HumanFactors_Gaspar}. At the same time, a recent meta-study showed that when drivers are motivated or instructed to pay attention to the traffic, they have better situation awareness because they are not engaged in manual driving \cite{2014_TransRes_Winter}.

\subsubsection{Returning to manual control}


Until reliable driverless technology becomes more widely available, conditionally-automated vehicles will rely on human intervention in the case of software failure. However, studies suggest that when exposed to automation, it takes drivers longer to react to events following limitations or failures of automated driving, especially if they are distracted \cite{2017_AdvErgonomics_Feldhutter} or fatigued \cite{2019_AccidentAnalysis_Vogelpohl}, and they might deactivate automation without being fully aware of the situation. As a result, research on this topic considers how and when to issue a take-over request (TOR) and how to restore driver's situation awareness to ensure a smooth and safe transition to manual driving.

Quiet automation failure is the worst-case scenario as drivers may not realize that they need to intervene, especially if distracted, or may do so late. Unsurprisingly, in such conditions, only a small proportion of drivers avoids undesirable events. For example, in one study, $75\%$ of participants could not prevent lane departure when automation switched off without warning \cite{2018_AHAT_Feldhutter}. Another experiment showed that even drivers that were observing the road made lane excursions \cite{2019_TransRes_Louw}. In a closed track study by Victor \etal \cite{2018_HumanFactors_Victor}, $38\%$ of participants did nothing when the vehicle drifted to an adjacent lane and over $30\%$ could not avoid crashes with stationary objects despite being briefed about system limitations prior to the ride and while keeping their eyes on the road and hands on the wheel. Similarly, in the experiment conducted by Lu \etal \cite{2019_TransRes_Lu}, issuing a warning to watch for hazards instead of an explicit TOR was not sufficient. As a result, many subjects failed to avoid collision with pedestrians. Although more systematic research is needed to establish links between attention patterns, automation failures, and criticality of the events, simply maintaining attention on the road is insufficient to prompt timely and adequate response from the drivers.

Explicit take-over requests help indicate failure and elicit a better reaction by forcing the driver to focus on the front scenario and mirrors \cite{2020_TransRes_Li}, especially if issued in multiple modalities. Audio warnings are more typical, but combinations of acoustic and visual signals \cite{2019_TransRes_Lu} coupled with seat vibration \cite{2019_TransRes_Clark} have also been proposed. Drivers also need time to observe the scene and think of a course of action. The exact time required depends on the complexity and severity of the situation, and longer is generally better \cite{2017_AccidentAnalysis_Clark}. For instance, drivers detected $49\%$ of hazards when given $6$s to take over compared to only $29\%$ with $3$s \cite{2018_TR_Vlakveld}. An early warning to start monitoring the scene \cite{2019_TransRes_Lu} and audio guidance towards relevant areas of interest \cite{2019_TransRes_Clark} were shown to be effective to improve situation awareness. Visual guides, \eg LED lights installed under the windshield, can be used to direct attention. They are more effective when they indicate the path to avoid the hazard \cite{2016_AutoUI_Borojeni} rather than point at the obstacle itself \cite{2018_ITSC_Yang}.

\subsubsection{Guidance and warning}
\label{sec:automation_guidance_warning}

Although various safety features are common in vehicles, most research focuses on their safety implications, not on how they affect drivers' gaze allocation. The results of the field test of the driver distraction detection algorithm showed no significant differences in gaze distributions in drives with and without the monitoring function enabled. Some positive changes were observed such as increased glancing towards the forward roadway and fewer long off-road glances \cite{2013_TITS_Ahlstrom}. Data from the EuroFOT indicates that forward collision warning (FCW) system warnings prompted the drivers to redirect their glances at medium and high eccentricity locations, presumably to restore their situation awareness \cite{2016_AccidentAnalysis_Morando}.

\begin{figure}[t!]
\centering
  \includegraphics[width=0.7\linewidth]{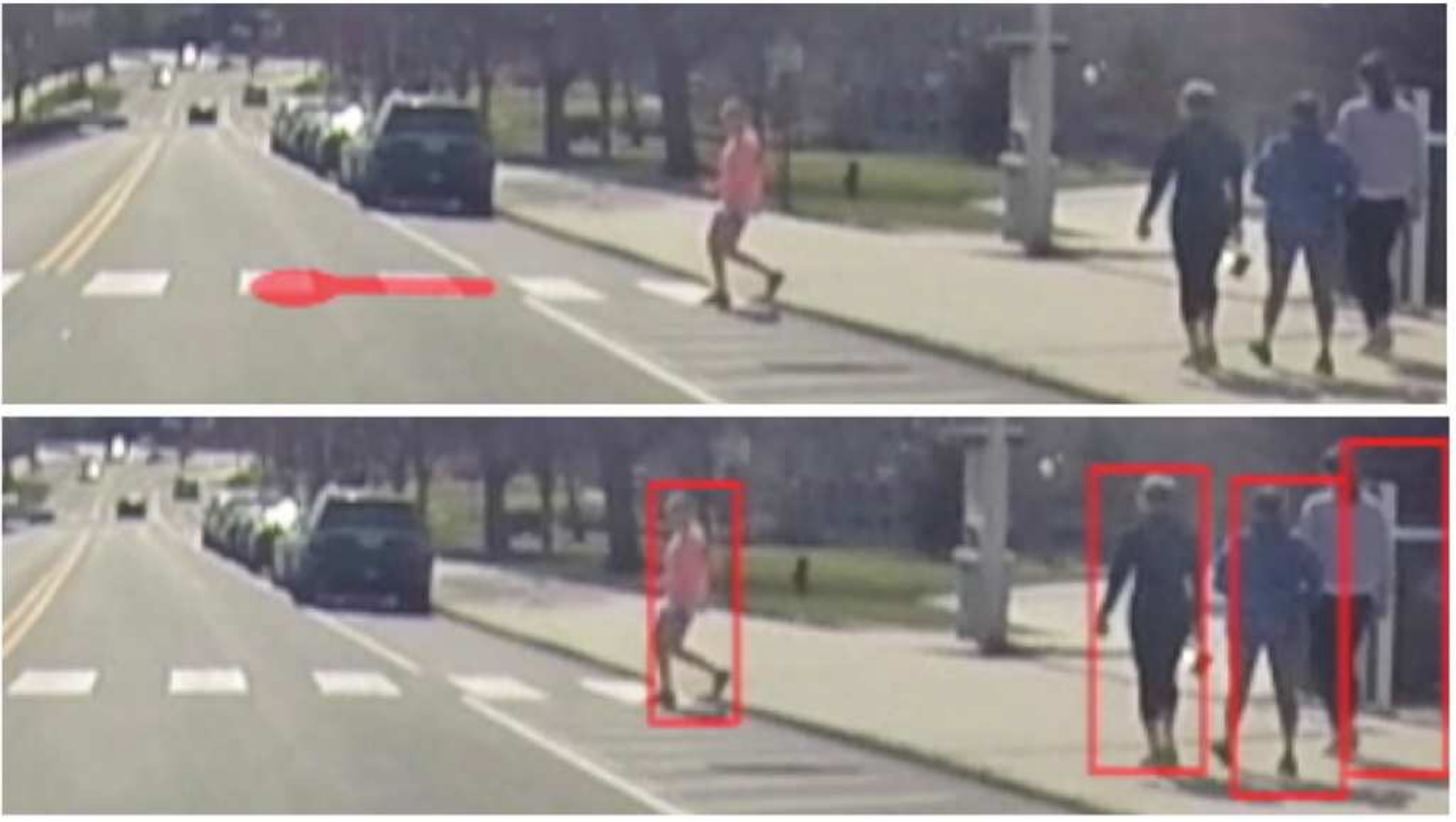}
\label{fig:cue_examples}
\caption[Augmented reality visualizations used to guide driver's gaze]{How to attract driver's attention to pedestrian? ''Virtual shadow`` (top) is conspicuous and, unlike bounding boxes (bottom), does not pull all attention away from other elements of the scene. Source: \cite{2019_HumanFactors_Kim}.}
   \label{fig:virtual_shadow}
\end{figure}

A relatively new trend is guiding drivers' gaze towards hazards or important elements of the traffic scenes. Several studies showed that highlighting hazards and important objects helped drivers notice them sooner and brake smoother in response compared to the control group \cite{2019_SAP_Bozkir, 2012_ACM_Pomarjanschi}. Similarly, highlighting maneuver-relevant cues leads to a more optimal allocation of attention: the number of fixations to relevant areas increased while the proportion of dwell time on salient but unrelated zones reduced \cite{2015_TR_Eyraud}. 

How this guidance is realized also matters. Providing too many cues may be detrimental, \eg highlighting traffic signs and line markings when only one is needed for the maneuver \cite{2015_TR_Eyraud} or highlighting all visible pedestrians \cite{2019_HumanFactors_Kim}. Cues themselves should not be too conspicuous. For instance, bounding boxes may attract attention to the pedestrians but may obscure their posture and fine movements or distract the driver from other driving-related objects. A more subtle cue such as virtual shadow \cite{2019_HumanFactors_Kim} (Figure \ref{fig:virtual_shadow}) or thin lines converging on the pedestrian \cite{2012_ACM_Pomarjanschi} may be preferable. Although these techniques are promising, they so far have been tested only in simulation, and many technical challenges remain before such guidance is possible in real vehicles. For instance, superimposing the visual cues on the scene requires augmented reality (AR) windshields or glasses. Ideally, visual warnings should also be aware of where the driver is looking, however, even with precise eye-tracking, it may be difficult to reliably establish which object the driver is paying attention to (summarized in Section \ref{sec:driver_awareness}).

\subsection{Summary}

More than 30 internal and external factors, and their combinations are considered in the behavioral literature. Two categories, driving experience and inattention due to secondary task engagement, are particularly important and have been extensively studied. 

Beginner drivers comprise a small portion of the overall driver population but are over-represented in crash statistics. Decades of research identified multiple differences in different aspects of visual behaviors of novice and experienced drivers. Perhaps the most prominent one is insufficient scanning of the environment by beginner drivers that leads to failure to notice hazards. It is not fully understood yet what causes these deficiencies in visual behavior, how much experience is needed to acquire proper visual skills, and whether this process can be accelerated via targeted training.

Driver inattention is another well-studied topic. According to multiple sources, a significant proportion of drivers engage in secondary non-driving-related tasks, which cause them to avert their gaze from the road. Visual and visual-manual tasks such as receiving and entering text messages are considered the most dangerous since they interfere with the cognitive abilities needed for driving. Cognitive distractions, \eg mind-wandering, are no less dangerous but are much more difficult to induce and identify. The interaction between different types of tasks performed simultaneously or tasks involving visual, cognitive, and manual modalities are rarely studied.

The effect of automation on attention is being actively researched given the wide-spread availability of various driver assistance systems. Automating lateral and longitudinal control of the vehicle leads to loss of visuomotor coordination and reduced attention to the front scenario. Highly-automated driving also makes the drivers prone to engaging in secondary non-driving activities. Thus, one of the most studied topics is how to alert the driver and help them regain situation awareness when automation fails. 

\section{Analytical models of attention}
\label{ch:psychological_models}

Previous sections discussed behavioral studies that focus on investigating correlations between changes in drivers' gaze allocation when various related factors are manipulated. In contrast, only relatively few works attempt to formulate a more general framework that can explain phenomena observed in the experiments and even make predictions. As such, analytical models are different from the practical models of drivers' gaze (presented in Section \ref{sec:practical_driver_attention}) that are to a lesser extent based on hypotheses and implicitly learn patterns of attention distribution from large datasets. This section will focus on two approaches to modeling the human gaze: human performance modeling and psychologically-grounded models.

\subsection{Human performance modeling}
\subsubsection{Background and formulation of SEEV model}

SEEV, which stands for Salience, Effort, Expectancy, and Value, is a well-known human performance model that identifies the role of bottom-up and top-down factors on where and when people look in dynamic environments \cite{2001_TechReport_Wickens}. Based on these factors, the model predicts the percentage of dwell time spent within each AOI. Initially proposed for modeling pilot behavior, the model has also been extensively tested in the driving domain \cite{2006_JEP_Horrey, 2012_AccidentAnalysis_Werneke, 2013_TransRes_Benedetto, 2013_TR_Wortelen, 2014_CogTechWork_Werneke, 2015_TR_Lemonnier}. 

SEEV provides a descriptive and prescriptive model of attending to AOIs that serve different purposes. The probability of scanning an area of interest $A$ is provided by the following \textit{descriptive} model of scanning \cite{2013_TransRes_Benedetto}:

\begin{equation*}
P(\mathrm{A}) = s\mathrm{S} - ef\mathrm{EF} + (ex\mathrm{EX})(v\mathrm{V}),
\end{equation*}
where salience ($\mathrm{S}$) is determined by the physical properties of the events, effort ($\mathrm{EF}$) depends on the physical distance between the previously fixated and current AOI and/or demands of other concurrent tasks, expectancy ($\mathrm{EX}$) is expressed in terms of bandwidth or event rate, and value ($\mathrm{V}$) is the reward associated with processing information in an AOI or the cost of failing to attend. Coefficients $s, ef, ex$, and $v$ represent the relative influence of these factors. 

According to this equation, to be effective, visual attention should be primarily guided by top-down factors, while bottom-up factors should be minimized. Irrelevant salient objects may act as distractors while, at the same time, relevant elements that require too much effort to observe will likely be overlooked. Understanding these parameters of human attention may serve as guidance for the design of better interfaces where important information should be highlighted (\eg flashing indicators in the instrument panel) and require little effort to access (\eg placing a camera for parking in the rear-view mirror).

A \textit{prescriptive} (or optimal) model of attending to an AOI is expressed in terms of tasks that this AOI is involved in and their relevance. Its formulation is given below:

\begin{equation*}
P(\mathrm{A}_j) = \sum\limits_{t=1}^n [(b\mathrm{B}_t)(r\mathrm{R}_t)\mathrm{V}_t - \mathrm{EF}_t],
\end{equation*}

where $\mathrm{B}$ is the bandwidth of the informational events for task $t$, $\mathrm{R}$ is the relevance of the AOI for the task, value $\mathrm{V}$ defines the priority of the task and $\mathrm{EF}$ is the effort associated with accessing the AOI. The prescriptive model, unlike the descriptive one, does not contain non-optimal bottom-up saliency influences and effort factors and instead focuses on bandwidth and priority of the subtasks.

\subsubsection{Applications}
The prescriptive model was verified on human experimental data and explained changes in gaze patterns of drivers performing dual tasks (driving and reading numbers from the screen), while the priority and difficulty of the tasks was manipulated. Specifically, Expectancy (information change) and Value (task priority) were two dominating factors, \eg dwell time on the road increased as bandwidth for the in-vehicle task increased. In addition, AOIs associated with higher priority tasks were attended preferentially, sometimes at a cost to other AOIs \cite{2006_JEP_Horrey, 2013_TR_Wortelen}. Dwell times predicted by the SEEV model had a high linear correlation ($r^2>0.9)$ with the observed data. SEEV has also been applied to tasks involving interaction with other road users.  Lemonnier \etal \cite{2015_TR_Lemonnier} used SEEV to analyze gaze behavior when approaching intersections which involved two tasks: driving and interacting with other drivers. Intersecting road Expectancy and Value were manipulated by changing traffic density and the right of way (stop, yield, and priority) respectively. The model showed that these two top-down factors explained human experimental data (with average error of $6\%$) and reflected the differences in visual exploration behavior in various situations. In a different study, the authors looked at the effect of Expectancy and Value during phases of passing the T-intersection (approaching, waiting, and accelerating) with varying traffic density and vulnerable road users (VRUs) present. These results show distinct attention allocation strategies for each phase, \eg in the approach phase, traffic density from the left is dominant, however in the waiting phase its importance reduces, while the presence of VRUs becomes important and gains dominance in the acceleration phase \cite{2012_AccidentAnalysis_Werneke, 2014_CogTechWork_Werneke}. At the same time, the SEEV model was not able to replicate the phenomenon of spatial attention asymmetry in an experiment involving a standard lane change test (LCT), where drivers read information from two identical road signs located on either side of the road. In the experiment, human subjects showed a clear preference towards left signs, but according to SEEV, right signs should be attended more frequently since their Saliency was the same and Effort (head and eye movements) was lower for the right sign \cite{2013_TransRes_Benedetto}.

Two notable limitations of SEEV are not implementing a visual processing pipeline and not addressing dynamic switching of gaze. Instead it calculates cumulative gaze towards AOIs for each subtask. This is helpful for interface design but not sufficient for understanding dynamic processes during driving. An extension of SEEV, called Adaptive Information Expectancy (AIE), takes dynamics into account \cite{2013_TR_Wortelen}. Another issue with SEEV is parameter setting. In the original paper, model parameters were defined a priori based on the assumptions and hypotheses regarding the task priority, AOI relevance, and task bandwidths. In the driving domain which is less structured compared to aviation, for which the model was initally developed, these values proved to be difficult to establish \cite{2015_TR_Lemonnier}.

\subsection{Models based on reward and uncertainty}

\begin{figure}[t!]
\centering
 \includegraphics[width=0.7\linewidth]{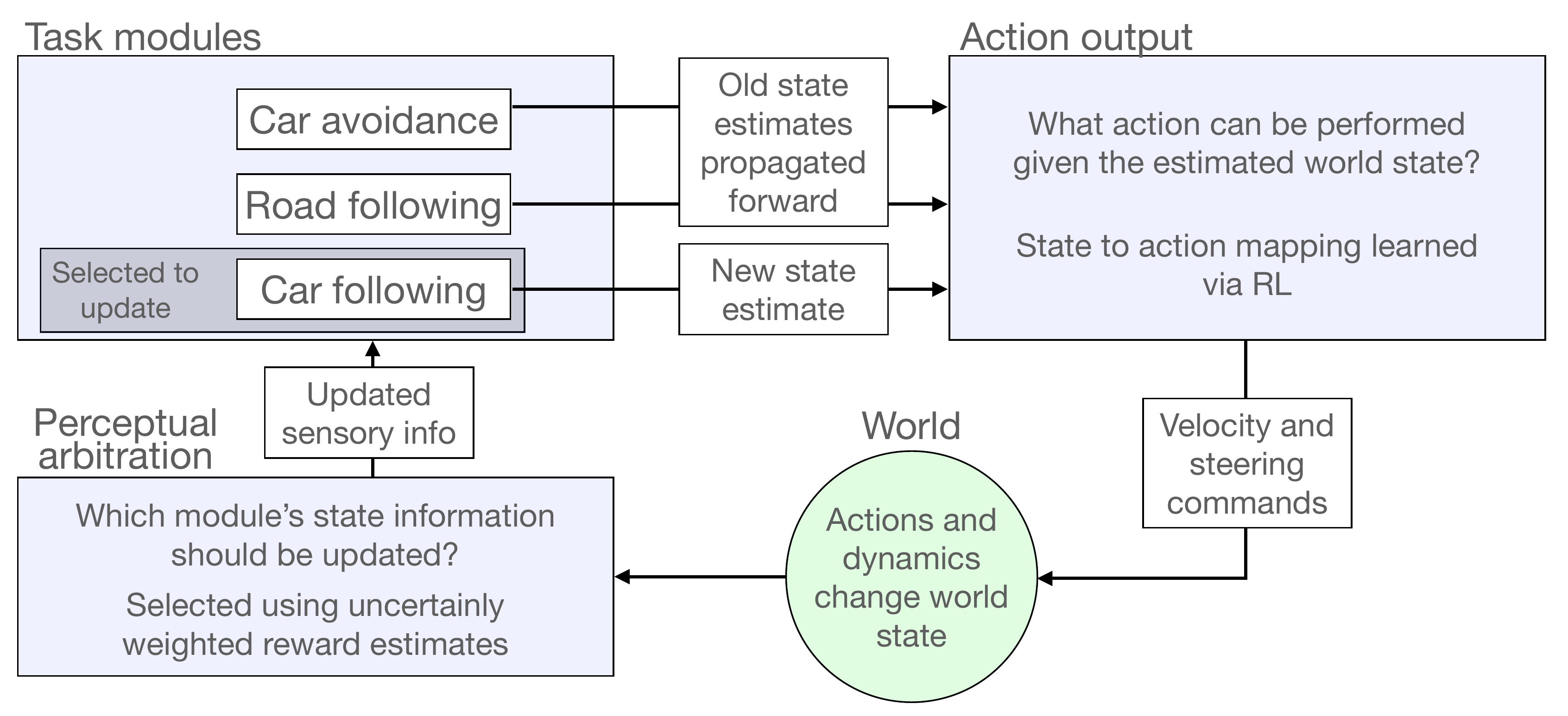}
  \caption[A diagram of the model for task-based attention control proposed by Hayhoe \& Ballard]{A diagram of the model for task-based control of eye movements. Multiple task modules compete for the gaze to improve their state estimates. When the car following is chosen, the system executes visual operations to detect the vehicle ahead and estimate its location. Based on the state, the next action is selected based on reinforcement learning. After the effects of action take place, a new cycle begins. Source: \cite{2014_CB_Hayhoe}}
  \label{fig:reward_uncertainty_model}
\end{figure}

\subsubsection{Background and formulation}

A different route is taken in psychological models of task-based attention allocation proposed by Ballard, Hayhoe, and colleagues in multiple publications \cite{2005_TrendsCogSci_Ballard, 2009_VisualCognition_Ballard, 2014_CB_Hayhoe}. A general diagram of the approach is shown in Figure \ref{fig:reward_uncertainty_model}. Driving and potentially any visually guided behavior (\eg walking \cite{2003_NIPS_Sprague, 2007_ACM_Sprague}, making a sandwich \cite{2003_JoV_Hayhoe}) in this framework is represented as a collection of concurrent tasks, each of which has well-defined visual demand, internal state, and uncertainty. Depending on the level of abstraction, gaze is being used differently. At the behavior level, vision computes state information for the current goal (\eg detecting obstacles in the path for the vehicle control task), at the arbitration level, multiple tasks require different locations of gaze, and, finally, context defines the set of appropriate behaviors \cite{2005_TrendsCogSci_Ballard}. Thus, eye movements depend on the ongoing goals and are driven by reward and uncertainty. Based on these criteria, the scheduler activates an appropriate behavior, which then executes a series of visual operations (or visual routines \cite{1984_Cognition_Ullman}) that deploy gaze to relevant areas. Once new perceptual information is received, the model state is updated and action selection can be done. Current context and internal state define what behaviors are relevant and one is chosen for execution.

A similar idea is behind the \textit{threaded theory of cognition}, which also uses a modular structure to represent tasks and incorporates dynamic scheduling for sensory-motor arbitration. Several dual-task driving scenarios were represented within the well-known ACT-R cognitive architecture \cite{2004_PsychReview_Anderson, 2008_PsychReview_Salvucci}, however they did not model gaze.

\subsubsection{Applications}

Early implementations by Salgian \etal \cite{1998_ICCV_Salgian, 1998_Springer_Salgian} modeled three common behaviors: understanding stop signs and red light, and following the lead vehicle. Each behavior had visual routines formulated as finite state machines to guide stop sign, traffic lights, and looming detection respectively. A scheduler alternated between each behavior, each running under 100 ms to avoid missing critical information.

Johnson \etal \cite{2013_RSTB_Johnson} proposed a more complex solution. They modeled temporal gaze data from the experiment where task priority and uncertainty were manipulated. Specifically, subjects were asked to follow the lead vehicle while prioritizing either maintaining constant speed or constant headway distance. Besides task priority, uncertainty was changed by adding noise to the gas pedal signal. The experiment showed that both reward and uncertainty were significant determinants of how gaze was deployed. As expected, gaze patterns changed when task priorities were switched. In turn, under noisy conditions drivers made more frequent and longer glances at the speedometer when speed maintenance was prioritized. The computational model incorporated three tasks that ran concurrently: following the lead vehicle, maintaining speed, and maintaining lane, with state variables expressed as the headway distance, vehicle speed, and deviation from the lane center respectively.

The uncertainty at time $t$ is expressed as the difference between current error estimate  $\sigma$ and priority $\rho$ for task $i$ as follows:
\begin{equation*}
 \zeta^{(i)}(t)=\sigma^{(i)}-\rho^{(i)}.
\end{equation*}
To ensure that high-priority tasks do not get selected all the time, the model incorporates a soft barrier model which defines the probability of selecting a module using a Boltzmann distribution over each of the priority weighted module uncertainties:

\begin{equation*}
 P(\phi(t))=i|\zeta^{(1)}(t),\dots,\zeta^{(N)}(t)=\frac{\exp{\zeta^{(i)}(t)}}{Z},
\end{equation*}

\noindent
where $\phi(t)$ is a global variable for the index of the next module to be updated implemented as a softmax decision mechanism and $Z$ normalizes the discrete distribution. If module $i$ is above the threshold, \ie when $\sigma^{(i)}(t)>\rho^{(i)}$, then it is likely going to be selected.

The distributions of gaze obtained from this model replicated human data and captured sensitivity to priority and noise as well as the low influence of noise in low-priority tasks and vice versa. Measured KL divergence between ground truth and model predictions was smaller compared to the center gaze baseline, models based on bottom-up saliency and task priority. The limitation of this model is that it does not implement visual processing and relies on data provided by the simulation with added noise.

\subsection{Other theoretical proposals}
Several recent theories of gaze allocation yet to be implemented in practice are worth mentioning as these proposals address aspects of attention allocation during driving beyond what is being considered in the existing models and much of the experimental work. 

Lappi formulated seven qualitative laws describing gaze behavior in the wild summarizing experimental findings \cite{2016_NBR_Lappi} and verified their validity using eye-tracking data from an expert driver \cite{2017_FPsych_Lappi}. The following behaviors are described: 1) repeatable gaze patterns, 2) gaze on task-relevant AOIs, 3) interpretable roles of fixations, 4) fixating on targets ``just in time'', 5) intermittent sampling, 6) memory used to (re)orient in 3D and 7) eye/head/body/locomotor coordination. Existing models can demonstrate only some of these phenomena, such as task-relevant gaze and stereotypical patterns during curve negotiation. Majority of these models are reactive, \ie predict gaze based on the required action, despite evidence that gaze usually precedes action.

Information Acquisition Theory proposed by Wolfe \& Rosenholtz \cite{2020_HumanFactors_Wolfe} instead of focusing on what area or object the driver is looking, consider the information contained in the entire visual field and thus emphasize the role of peripheral vision in providing the information to guide eye movements. As discussed in Section \ref{ch:what_driver_sees}, peripheral vision is sufficient to perform some driving tasks but is not optimal. Nevertheless, incomplete preattentive representation of the scene in the periphery complements and guides information acquisition in the foveal regions. This formulation puts more emphasis on the experience as a way to improve preattentive representation and shape anticipation skills. The proposed theory also argues for the necessity of current task information to understand the allocation of gaze and suggests that large-scale datasets associating actions, scenes, and gaze may provide a solution.

Kircher \& Ahlstrom consider driver (in)attention following a systems approach and formulate a concept of Minimum Required Attention (MiRA) \cite{2017_HumanFactors_Kircher}. According to MiRA, the driver is attentive if their information intake is sufficient for the demands on the system. Thus, sampling defines attention, not processing, interpretation, or communication. For example, within this framework, looked-but-failed-to-see errors would not be classified as caused by inattention since the information was sampled but not processed correctly. Meeting the minimum required attention does not automatically imply safety either. The authors argue that it is meaningless to consider a single agent responsible for the safety of the system influenced by environmental factors, elements of infrastructure, and other road users. Operationalizing MiRA for various stereotypical situations and maneuvers would be necessary first based on the existing empirical data for various conditions (\eg traffic density, relative speed, visibility). Information intake may be visual (guided by top-down and bottom-up influences) as well as multisensory. Once it is achieved, evaluating driver behavior against MiRA will allow classifying them as attentive or inattentive without the necessity of hindsight (which is currently the case).

\subsection{Summary}
Analytical models of drivers' attention can potentially explain the observed eye movements beyond correlation analysis offered by the behavioral studies. Besides their explanatory function, such models can make predictions for unknown conditions or generate data for vehicle interface design. In general, analytical models are hard to develop and validate against human data. For example, a widely used SEEV model of attention provides a simple formula for optimal gaze distribution across defined AOIs. On the downside, it does not implement visual processing, produces only aggregate statistics and has many parameters that must be tuned manually, which is particularly challenging in an unstructured driving domain. 

A different approach by Ballard, Hayhoe, and colleagues is based on reward and uncertainty and provides a framework for understanding how visual requirements of different driving-related tasks interact to produce dynamic gaze patterns. To date, several prototypes showed a good fit to human data for a combination of two-three driving tasks, such as following the lead vehicle, maintaining speed, and maintaining lane. Further research is necessary to implement more realistic visual processing and to extend this framework to a full range of activities during driving. 

Several recent theoretical proposals were put forward that emphasize the role of peripheral vision and a more holistic approach to drivers' information intake but are have not been implemented and validated against human data.

\section{Practical solutions}
\label{ch:practical}
Behavioral studies provide a wealth of evidence that attention allocation can be affected by multiple external and internal factors and is tightly linked to vehicle control. As a result, practical solutions are developed under the assumption that changes in gaze patterns and driving performance can be detected and associated with changes in the environment, driver's behavior, and state. Knowing where the driver is looking and what they are doing or about to do can, in turn, be used for various driver monitoring applications, such as inattention and drowsiness detection. More advanced scenarios include also understanding what the driver is or is not aware of and issuing active warnings to direct their gaze. High-quality behavioral data collected in diverse conditions from a large population of drivers is crucial for developing algorithms, particularly those that rely on supervised machine learning.

\subsection{Datasets}
Most publicly available datasets for studying drivers' attention were released in the past 5 years, thus nearly $90\%$ of behavioral and more than two-thirds of practical studies we reviewed were conducted using unpublished data.

Table \ref{tab:datasets} shows detailed properties of the datasets published in the past 10 years or used in studies that we reviewed. We divide the datasets into 3 groups: \textit{Eye-tracking datasets}, \textit{Other datasets}, and \textit{Large-scale naturalistic driving studies (NDS) and field operational tests (FOT)}. In view of the discussion in Section \ref{ch:data_collection} we will focus on how data was collected.

\begin{table}[t!bp]
  \centering
  \resizebox{1\textwidth}{!}{
    \begin{tabular}{r|llllllllllll}
          & Dataset &  \multicolumn{1}{l}{Year} & \multicolumn{1}{l}{Type} & Data  & \multicolumn{1}{l}{Annotations} & \specialcell{Annotation/\\recording\\conditions} & \multicolumn{1}{l}{\specialcell{Vehicle\\control}} & \multicolumn{1}{l}{\rotatebox{60}{\# subjects}} & \rotatebox{60}{\# frames} & \rotatebox{60}{\# videos} \\ \hline
    \multicolumn{1}{r|}{\multirow{9}[0]{*}{\rotatebox{90}{Eye-tracking datasets}}} & \href{https://github.com/taodeng/CDNN-traffic-saliency}{Deng \etal 2020} \cite{2020_TITS_Deng} & 2020  & \multicolumn{1}{l}{normal driving} & V$_S$, ET &   -    & simulation: low & \multicolumn{1}{l}{-} & 28    & 77K$^\ast$      & \multicolumn{1}{l}{16} \\
          & \href{https://github.com/JWFangit/LOTVS-DADA}{DADA-2020} \cite{2019_ITSC_Fang} & 2019  & \multicolumn{1}{l}{hazard perception} & V$_S$, ET & \multicolumn{1}{l}{BB, TL} & simulation: low & \multicolumn{1}{l}{-} & 20    & 658K  & 2000 \\
          & \href{https://github.com/taodeng/traffic-eye-tracking-dataset}{Deng \etal 2018}  \cite{2018_TITS_Deng} & 2018 & \multicolumn{1}{l}{normal driving} & I$_S$, ET &   -    & simulation: low & \multicolumn{1}{l}{-} & 20    & \multicolumn{1}{l}{100} & - \\
          & \href{http://imagelab.ing.unimore.it/dreyeve}{DR(eye)VE} \cite{2018_PAMI_Palazzi, 2016_CVPRW_Alletto} & 2018 & \multicolumn{1}{l}{normal driving} & V$_S$, ET, PD, EV &    -   & on-road: directed & \multicolumn{1}{l}{+} & 8     & 555K  & \multicolumn{1}{l}{74} \\
          & \href{https://bdd-data.berkeley.edu/}{BDD-A} \cite{2018_ACCV_Xia} & 2018 & \multicolumn{1}{l}{hazard perception} & V$_S$, ET &   -    & simulation: low & \multicolumn{1}{l}{-} & 45    & 378K$^\ast$  & \multicolumn{1}{l}{1232} \\
          & \href{https://osf.io/c42cn/}{Taamneh \etal} \cite{2017_NatSciData_Taamneh} & 2017 & \multicolumn{1}{l}{secondary tasks} & V$_S$, ET, PS &   -    & simulation: med & \multicolumn{1}{l}{+} & 68    & -     & \multicolumn{1}{l}{456} \\
          & \href{https://github.com/taodeng/Top-down-based-traffic-driving-saliency-model}{Deng \etal 2016} \cite{2016_TITS_Deng} & 2016  & \multicolumn{1}{l}{normal driving} & I$_S$, ET &     -  & simulation: low & \multicolumn{1}{l}{-} & 40    & \multicolumn{1}{l}{100} &  -\\
          & \href{http://ilab.usc.edu/borji/Resources.html}{USC Video Games} \cite{2011_BMVC_Borji} & 2011 & \multicolumn{1}{l}{normal driving} & V$_S$, ET &   -    & simulation: low & \multicolumn{1}{l}{+} & 10    & 192K  &  108\\
          & \href{https://cvssp.org/data/diplecs/}{DIPLECS Sweden} \cite{2010_ACCV_Pugeault} & 2010 & \multicolumn{1}{l}{normal driving} & V$_S$, ET, EV &  TL     & on-road: naturalistic & \multicolumn{1}{l}{+} & 1    & 159K  &  1\\ \hdashline
    \multicolumn{1}{c|}{\multirow{8}[0]{*}{\rotatebox{90}{Other datasets}}} & \href{https://usa.honda-ri.com/HAD}{HAD}  \cite{2019_CVPR_Kim} & 2019  & \multicolumn{1}{l}{normal driving} & V$_S$, EV & \multicolumn{1}{l}{TL} & simulation: high & \multicolumn{1}{l}{-} & \multicolumn{1}{l}{-} & 3.4M$^\ast$  & \multicolumn{1}{l}{5675} \\
          & \href{https://github.com/JinkyuKimUCB/BDD-X-dataset}{BDD-X} \cite{2018_ECCV_Kim} & 2018 & \multicolumn{1}{l}{normal driving} & V$_S$, EV & \multicolumn{1}{l}{TL} & simulation: low & \multicolumn{1}{l}{-} & \multicolumn{1}{l}{-} & 8.4M  & \multicolumn{1}{l}{6984} \\
          & \href{https://usa.honda-ri.com/HDD}{HDD}   \cite{2018_CVPR_Ramanishka} & 2018 & \multicolumn{1}{l}{normal driving} & V$_S$, EV & \multicolumn{1}{l}{BB, TL} & simulation: low & \multicolumn{1}{l}{-} & \multicolumn{1}{l}{n/a} & 1.2M$^\ast$  & \multicolumn{1}{l}{137} \\
          & \href{http://cv.cs.nthu.edu.tw/php/callforpaper/datasets/DDD/}{DDD}  \cite{2017_ACCV_Weng} & 2017  & \multicolumn{1}{l}{driver monitoring} & V$^{IR}_D$ & \multicolumn{1}{l}{TL}  & simulation: low & \multicolumn{1}{l}{+} & 36    & 486K$^\ast$  & \multicolumn{1}{l}{360} \\
          & \href{https://aliensunmin.github.io/project/dashcam/}{Chan \etal} \cite{2016_ACCV_Chan} & 2016 & \multicolumn{1}{l}{hazard perception} & V$_S$ & \multicolumn{1}{l}{BB, TL} & simulation: low & \multicolumn{1}{l}{-} & \multicolumn{1}{l}{n/a} & -     & \multicolumn{1}{l}{678} \\
          & \href{https://github.com/asheshjain399/ICCV2015\_Brain4Cars}{Brain4Cars} \cite{2015_ICCV_Jain} & 2016 & \multicolumn{1}{l}{driver monitoring} & V$_{S,D}$, EV & \multicolumn{1}{l}{BB, TL} & on-road: naturalistic & \multicolumn{1}{l}{+} & 10    & 2M    &  - \\
          & \href{https://cvssp.org/data/diplecs/}{DIPLECS Surrey} \cite{2015_TranVehTech_Pugeault} & 2015 & \multicolumn{1}{l}{normal driving} & V$_S$, EV &   -  & on-road: naturalistic & \multicolumn{1}{l}{+} & 1    & 54K$^\ast$   &  1\\
          & \href{https://ieee-dataport.org/open-access/yawdd-yawning-detection-dataset}{YawDD} \cite{2014_ACM_Abtahi} & 2014 & \multicolumn{1}{l}{driver monitoring} & V$_D$ & \multicolumn{1}{l}{BB, TL} & on-road: directed & \multicolumn{1}{l}{-} & 107   & \multicolumn{1}{l}{-} & \multicolumn{1}{l}{342} \\
		  & \href{http://www.cvlibs.net/datasets/kitti/}{KITTI} \cite{2013_CVPR_Geiger, Geiger2012CVPR} & 2012 & \multicolumn{1}{l}{normal driving} & V$_S$, EV & \multicolumn{1}{l}{BB, SM, TL} & on-road: directed & \multicolumn{1}{l}{+} & -   & \multicolumn{1}{l}{-} & \multicolumn{1}{l}{-} \\ \hdashline
    \multicolumn{1}{r|}{\multirow{4}[0]{*}{\rotatebox{90}{NDS \& FOT}}} & EOR-FOT \cite{2013_TechRep_Karlsson} & 2013  & \multicolumn{1}{l}{normal driving} & V$_{S,D}$, ET, EV &   -    & on-road: naturalistic &   +    & 19    & -     & - \\
          & \href{https://www.eurofot-ip.eu}{euroFOT} \cite{2012_TechRep_Kessler} & 2012  &    \multicolumn{1}{l}{normal driving}    & V$_{S,D}$, EV & - & on-road: naturalistic &   +    & 1200  & -     & - \\
          & \href{https://www.vtti.vt.edu/facilities/data-center.html}{100-Car NDS} \cite{2005_TechRep_Neale} & 2005  &  \multicolumn{1}{l}{normal driving} & V$_{S,D}$, EV &   -    & on-road: naturalistic &   +   & 249   & -     & - \\
          & RDCW FOT \cite{2006_TechRep_LeBlanc} & 2006  &    \multicolumn{1}{l}{normal driving}    &  V$_{S,D}$, EV &   -   & on-road: naturalistic &   +    & 78    & -     & - \\
    \end{tabular}%
} 
    \caption[Datasets for studying attention and driving]{Datasets for studying attention and driving. Click on the dataset name to go to the corresponding project page (if available). The following abbreviations are used for data types: V - video ($_S$ - traffic scene, $_D$ from the driver-facing camera, $^{IR}$ - infrared camera), ET - eye-tracking, PD - pupil dilation, EV - ego-vehicle information and PS - physiological signals. Abbreviations for annotations include: BB - bounding boxes, TL - text labels, 3DP - 3D pose and SM - semantic maps. For some datasets we provide estimated frame counts (marked with asterisk $^\ast$) based on the sampling rate of the camera and total or average length of videos in the dataset. Annotations/recording conditions column indicates how attention-related data was collected, \eg for the majority of the datasets video data was recorded in-vehicle and later annotated with gaze or manually in-lab.}
 \label{tab:datasets}%
\end{table}%

\subsubsection{Eye-tracking datasets}
Datasets with eye-tracking information are the most relevant for studying the attention allocation of drivers. Within this group, most datasets contain normal driving scenes, two datasets, DADA-2000 \cite{2019_ITSC_Fang} and Berkeley DeepDrive Attention (BDD-A) \cite{2018_ACCV_Xia}, focus on hazardous scenarios, and one dataset by Taamneh \etal \cite{2017_NatSciData_Taamneh} captures drivers' data when engaged in secondary tasks. Hazard perception datasets consist of multiple short fragments starting seconds before the accidents, or anomalous events occur and ending shortly after. Datasets focusing on normal driving conditions contain longer uninterrupted video recordings.

Eye-tracking data in DR(eye)VE \cite{2018_PAMI_Palazzi, 2016_CVPRW_Alletto}, UCF Video Games \cite{2011_BMVC_Borji},  Taamneh et al. \cite{2017_NatSciData_Taamneh} and DIPLECS Sweden \cite{2010_ACCV_Pugeault} was recorded while the subjects controlled the vehicle. DR(eye)VE and DIPLECS were captured on-road while UCF Video Games and Taamneh et al. data were recorded as the participants played driving video games in a driving simulator. Given that it is virtually impossible to replicate the same route for different drivers in naturalistic conditions, videos in on-road datasets are accompanied by a single person's gaze recording. The dataset by Taamneh et al. \cite{2017_NatSciData_Taamneh} contains recordings of 68 subjects, but in other datasets the number of subjects is relatively small: 1 driver in DIPLECS, 8 in DR(eye)VE, and 10 in the UCF Video Games. Such lack of diversity and potential individual biases are undesirable for practical applications.

The remaining datasets in this group use existing driving footage to record eye-tracking data in the lab. This offers several advantages as data from multiple human subjects can be collected for each video segment, improving scalability and reducing bias. Besides, data acquired in a lab setting is of higher quality since there are no vibrations caused by the vehicle, drivers' movements can be restricted, and higher-resolution equipment can be used. On the other hand, low-fidelity simulators, where most of such studies are conducted, introduce other biases caused by lack of vehicle control, lowered risk, and overexposure to rare situations, \eg in hazard perception experiments (see also Section \ref{sec:driving_simulators}). Xia \etal \cite{2018_ACCV_Xia} show that subjects who viewed footage from the DR(eye)VE dataset detected significantly more driving-related objects (vehicles, pedestrians, and motorcycles) than the drivers whose gaze was recorded originally. The authors also validated in-lab data by showing videos attenuated with saliency maps generated from in-lab and in-vehicle to human subjects asking them to choose the one they prefer. 71\% preferred in-lab saliency maps, however, it remains unclear whether patterns generated by the algorithm result in safer driving as no vehicle control was involved.

\subsubsection{Other datasets}

Datasets in this group do not contain eye-tracking data and, instead offer annotated scene videos or videos from driver-facing cameras from which glance data may be inferred. Textual annotations (TL in Table \ref{tab:datasets}) are very common. For example, Honda Research Institute Advice Dataset (HAD) \cite{2019_CVPR_Kim}, Berkeley DeepDrive eXplanation (BDD-X) \cite{2018_ECCV_Kim} and Honda Research Institute Driving Dataset (HDD) \cite{2018_CVPR_Ramanishka} datasets provide textual annotations for prerecorded driving footage to describe and explain drivers' actions and attention allocation in terms of objects or events visible in the scene. HDD, in addition to textual annotations, contains bounding boxes for objects that caused the driver to stop or deviate from the path. This data is useful for understanding when and where the driver should look when performing specific actions during normal driving scenarios. For hazard anticipation, Dashcam Accident Dataset (DAD) \cite{2016_ACCV_Chan}, as the name suggests, provides an extensive collection of dashcam videos of accidents annotated with accident type labels, accident temporal window, and bounding boxes around relevant objects. Since human gaze often does not land precisely on the objects of interest, object annotations with importance scores provided by human annotators may be easier to generate and use in practice, although they may not necessarily reflect the true gaze distribution during driving. 

Yawning Detection Dataset (YawDD) \cite{2014_ACM_Abtahi}, Driver Drowsiness Detection (DDD) \cite{2017_ACCV_Weng}, and Brain4Cars \cite{2015_ICCV_Jain} focus instead on monitoring the driver's actions and whether they pay attention to driving. YawDD is the first publicly available dataset featuring drowsy drivers. Besides yawning, subjects who participated in the study were recorded with their mouth closed and while singing and talking. Overall, the dataset captures a diverse set of participants (different skin colors, age groups, wearing glasses and scarves) and illumination conditions, with the only drawbacks being scripted actions and recording in a stationary vehicle are the drawbacks. Similarly, DDD is a scripted dataset where in addition to acting normal and drowsy (yawning, nodding off), subjects were recorded laughing, talking, and looking to the sides. The dataset was recorded in a low-fidelity simulator while the subjects were playing a driving video game. Brain4Cars provides in-vehicle recordings of the participants along with an outside-facing camera and vehicle information. The purpose of the dataset is to observe and anticipate drivers' actions before they occur based on where the driver looks and outside view.

\subsubsection{Naturalistic driving studies (NDS) and field operational tests (FOT)}
This group includes large-scale naturalistic driving studies (NDS) and field operational tests (FOT). The purpose of NDS is to study the natural behaviors of drivers over an extended period of time and link behaviors to crash statistics. FOTs are conducted to test the adoption and use of driver assistance and driving monitoring technologies to improve their design.

What distinguishes these datasets from others is the scale and duration of data collection. One of the earliest and well-known studies, 100-Car NDS, was conducted in 2002-2004. During this time, approximately 1 year of data was recorded from each of the 100 participating vehicles with 241 primary and secondary drivers. In total, 2 million vehicle miles were recorded with 43 thousand hours of driving data. This dataset contains 82 crashes, 761 near-crashes, and 8295 incidents. The largest NDS to date, SHRP2, was conducted from 2010 to 2013 in the USA in 6 states. This study involved over 3000 drivers who generated 50 million miles of travel, equal to over 1 million hours of naturalistic data. Over 372 crashes of various severity levels were identified in the dataset, including over 100 police-reportable crashes. Another 200 potential crash events were not manually verified. Naturalistic driving studies normally do not provide eye-tracking data but contain coarse glance annotations towards different AOIs for studying attention allocation. 

The amount of data generated by NDSs is so immense that it takes years to process. For example, only a fraction of 2 petabytes of data generated by the SHRP2 project is manually analyzed and annotated. Besides the sheer volume of data, restricted access due to privacy concerns contributes to slow processing. As of now, only portions of data may be accessed upon approval of research proposals submitted to the organization that conducted the study. As a result, only s few practical \cite{2012_HumanFactors_Liang, 2019_arXiv_Baee} and behavioral works \cite{2014_HumanFactors_He, 2013_AccidentPrevention_Wong, 2013_AccidentAnalysis_Dozza, 2010_TR_Klauer, 2015_TransRes_Bargman} use it.

Field operational tests, as mentioned above, pursue different goals. Only one study, Eyes-On-Road (EOR-FOT) \cite{2013_TechRep_Karlsson} conducted by Autoliv and Volvo, tested a prototype system for driver visual attention measurement and monitoring. The test involved 10 drivers who used instrumented vehicles for several months in 2014-2015. This study provided insights into changes in drivers' glance patterns when lateral and longitudinal assistance was engaged \cite{2018_TITS_Morando}. Naturalistic datasets, except EOR-FOT, make available only the coarse glance data manually coded from driver-facing cameras.

Other large-scale studies conducted in Europe and the USA, such as euroFOT \cite{2012_TechRep_Kessler}, Road Departure Crash Warning (RDCW) FOT \cite{2006_TechRep_LeBlanc}, and Integrated Vehicle-Based Safety Systems (IVBSS) FOT, focused on intelligent vehicles equipped with various driver assistance systems (\eg forward collision warning (FCW), adaptive cruise control (ACC), speed regulation system (SRS), lane departure warning (LDW)). The primary goals of these FOTs included testing the systems to reduce false positive alarm rates and to increase drivers' acceptance of these systems. Similarly, data generated by these studies is difficult to access, therefore most publications are by the organizations which conducted the study and their collaborators \cite{2017_JSR_Wang, 2016_AccidentAnalysis_Morando, 2015_DDI_Tivesten, 2014_TransRes_Tivesten, 2013_AccidentAnalysis_Peng}. 

\subsection{Predicting drivers' gaze}
\subsubsection{Predicting gaze in the traffic scene}
\label{sec:practical_driver_attention}

\begin{table}[t!]
  \centering
\resizebox{1\textwidth}{!}{
    \begin{tabular}{llllllll}
    Reference & Input type & \specialcell{Observation\\length} & Output & Dataset & Metrics \\
    \midrule
    \href{https://sites.google.com/eng.ucsd.edu/sage-net}{SAGE-Net} \cite{2020_CVPR_Pal} & RGB & 16 frames & saliency map & \specialcell{DR(eye)VE, JAAD \cite{2017_ICCVW_Rasouli}, BDD-A} & \specialcell{KLDiv, CC, F1, MAE} \\
    Deng \etal 2020 \cite{2020_TITS_Deng} & RGB   & 1 frame & saliency map & Deng \etal 2020 & \specialcell{AUC, NSS, IG, CC,\\SIM, EMD, KLDiv} \\
    Tavakoli \etal 2019 \cite{2019_WACV_Tavakoli}  &  RGB   & 1 frame & saliency map & USC Video Games & AUC, NSS \\
    Ning \etal 2019 \cite{2019_ITSC_Ning} & RGB, OF & 1 frame & saliency map & DR(eye)VE & CC, KLDiv, IG \\
    Palazzi \etal 2018 \cite{2018_PAMI_Palazzi} & RGB, OF, SM & 16 frames & saliency map & DR(eye)VE & CC, KLDiv, IG \\
    Tawari \etal 2018 \cite{2018_ITSC_Tawari} & RGB   & 2 frames & saliency map & custom: on-road & PR curve, CC, mAP \\
    Xia \etal 2018 \cite{2018_ACCV_Xia} &  RGB   & 6 frames & saliency map & BDD-A & KLDiv, CC \\
    Tawari \etal 2017 \cite{2017_IV_Tawari} & RGB   & 1 frame & saliency map & DR(eye)VE & CC \\
    \href{https://github.com/francescosolera/dreyeving}{Palazzi \etal 2017} \cite{2017_IV_Palazzi}  & RGB   & 16 frames & saliency map & DR(eye)VE & CC, KLDiv \\ \hdashline
    Deng \etal 2018 \cite{2018_TITS_Deng} & RGB   & 1 frame & saliency map & Deng \etal 2018 & AUC, ROC, NSS \\
    Ohn-Bar \etal 2017 \cite{2017_PR_Ohn-Bar} & RGB, BB, EV & 2-3 s & BB + score & KITTI & PR curve, mAP \\
    \href{https://github.com/taodeng/Top-down-based-traffic-driving-saliency-model}{Deng \etal 2016} \cite{2016_TITS_Deng}  & RGB   & 1 frame & saliency map & custom: on-road & AUC, ROC, NSS \\
    Borji \etal 2014 \cite{2014_TransSysManCybernetics_Borji} & RGB   & 1 frame & saliency map & USC Video Games & AUC, NSS \\
    \href{http://ilab.usc.edu/borji/Resources.html}{Borji \etal 2012} \cite{2012_CVPR_Borji}  & RGB, PG, EV & 1 frame & saliency map & USC Video Games & NSS, AUC \\
    Borji \etal 2011 \cite{2011_BMVC_Borji} & RGB   & 1 frame & saliency map & USC Video Games & NSS, AUC \\
    \end{tabular}%
}
\caption[Models for predicting driver gaze distribution in the traffic scene]{A summary of properties of the recently proposed models for predicting driver gaze distribution in traffic scenes. Dashed line separates deep-learning models (top rows) and classical vision models (bottom rows) that are based on other machine learning techniques and heuristics. The following abbreviations are used for input and output types: RGB - 3-channel image, OF - optical flow, SM - semantic map, BB - bounding box, EV - ego-vehicle information, PG - previous gaze location.}
  \label{tab:driver_attention_models}%
\end{table}%

Models in this group predict the spatial distribution of the driver's gaze for a given image of the traffic scene. Table \ref{tab:driver_attention_models} lists models published in the past 10 years. These models learn associations between videos (or images) of recorded driving scenes and human gaze data to produce a single-channel saliency map for a given image where higher pixel values correspond to areas that the driver is more likely to look at. The only exception is the model in \cite{2017_PR_Ohn-Bar} which is trained on object annotations with corresponding importance scores assigned by human annotators.

\noindent
\textbf{Heuristic and machine learning approaches.} Drivers often fixate on the vanishing point on straight segments (referred to as ``far road'' in behavioral studies) since it offers an optimum position for viewing the road ahead \cite{1994_Nature_Land, 2014_JEMR_Lemonnier}. This is reflected in the average gaze maps calculated from human gaze data \cite{2016_TITS_Deng, 2018_PAMI_Palazzi}. Based on this finding, Deng \etal \cite{2016_TITS_Deng, 2018_TITS_Deng}  proposed combining low-level features (color, intensity, and orientation), bottom-up saliency maps, and high-level features (vanishing point and center bias). Weights for different components can be set empirically (20/80 split for bottom-up and top-down respectively) \cite{2016_TITS_Deng} or learned via random forest \cite{2018_TITS_Deng}. Similarly, Tavakoli \etal \cite{2019_WACV_Tavakoli} show that task-specific factors such as vanishing point contribute to the resulting distribution of attention more than bottom-up saliency and motion given by the optical flow. Relative weights between bottom-up and top-down factors are computed via regression. The probabilistic Bayesian models proposed by Borji \etal \cite{2011_BMVC_Borji, 2012_CVPR_Borji, 2014_TransSysManCybernetics_Borji} work under the assumption that the selection of the next object to attend is guided primarily by the task-based factors such as properties of the current objects in the scene, the location of the previous fixation, and motor actions. 

\noindent
\textbf{Deep-learning models} use a number of feed-forward and recurrent architectures to learn distributions of human gaze from a single frame or a stack of frames. Many of these approaches have been used in bottom-up saliency prediction, salient object detection \cite{2019_arXiv_Wang}, and semantic segmentation \cite{2019_AIR_Liu}. For example, Deng \etal \cite{2020_TITS_Deng} use a convolution-deconvolution neural network similar to U-Net \cite{2015_ICMIC_Ronneberger} to predict gaze maps and Tawari \etal \cite{2017_IV_Tawari} apply a fully-convolutional CNN (FCN-8s) with skip connections to model driver's attention from a single frame. A more complex model from the same authors computes visual features for a given frame using VGG-16 \cite{2014_arXiv_Simonyan}, concatenates features from select convolutional layers, and passes them through 2 LSTM networks with a dropout layer between them. The weighted MSE loss function addresses the imbalance in training data by reweighting salient and non-salient pixels to reduce bias \cite{2018_ITSC_Tawari}. Xia \etal \cite{2018_ACCV_Xia} use AlexNet \cite{2017_ACM_Krizhevsky} to extract features, upsample them, and feed them into 3 fully-connected (FC) layers followed by a ConvLSTM network. They also propose a human weighted sampling strategy to mitigate the prevalence of common driving scenarios (\eg vehicle following) and to focus on more interesting or hazardous scenarios. Palazzi \etal \cite{2017_IV_Palazzi} propose a two-stream network where coarse branch operates on the original frames and fine branch uses center-cropped images of the scenes. Both branches use a pretrained C3D \cite{2015_ICCV_Tran} network to encode a stack of 16 frames. Encoding produced by the fine branch is upsampled to the original size, combined with the last frame of the input, and is passed through a number of 3D convolutional and pooling layers to refine the prediction. Only the output of the fine branch is used at the test time. An extension of this model \cite{2018_PAMI_Palazzi}, in addition to RGB, uses stacks of optical flow and semantic segmentation maps which are processed in parallel, as described above, and then saliency maps obtained from each feature are summed. Pal \etal \cite{2020_CVPR_Pal} propose a saliency prediction framework SAGE-Net that incorporates an arbitrary saliency model, depth estimation and pedestrian intent prediction if the ego-vehicle speed is below some predefined threshold. The predicted saliency map is element-wise multiplied with the estimated depths and summed with the result of the pedestrian branch. The authors also propose to augment gaze ground truth with object masks for common types of road users and infrastructure computed with Mask R-CNN \cite{2017_CVPR_He}. The object-level map is superimposed with the recordings of drivers' gaze and used for training the model.

\noindent
\textbf{Evaluation and remaining issues.} Unlike some classical models, none of the deep-learning approaches explicitly represent the task as the only input given to the algorithm is visual. The assumption is that gaze changes due to maneuvers and the relative importance of the objects in the scene can be inferred from the raw gaze data and apparent ego-motion of the scene. Although DR(eye)VE provides ego-vehicle information that can be used to represent drivers' actions, most of the data corresponds to driving on straight segments of the road and sparse data available for maneuvers is insufficient for training \cite{2017_IV_Tawari}.

Intended practical applications of the proposed models are not always clearly articulated. If the goal is to mimic drivers' attention, then internal and external factors should be acknowledged and their effects modeled. If the goal is to produce optimal gaze allocation patterns, then filtering out episodes related to inattention, as is done in the DR(eye)VE dataset, may be justified. A viable option is to use expert drivers (\eg driving instructors) to provide eye-tracking data. Alternatively, in-lab gaze data or human annotations as in \cite{2017_PR_Ohn-Bar} may be used although it would be difficult to establish whether the resulting gaze patterns are optimal or even safe as the human subjects/annotators are not in control of the vehicle.

Currently, evaluation of all driver attention models follows the protocol established in the bottom-up saliency research and uses metrics that assess how closely generated gaze maps resemble those of human subjects \cite{2018_PAMI_Bylinskii}. Comparisons are typically made with static bottom-up approaches that compute saliency maps per frame. Video saliency models \cite{2019_PAMI_Wang} that can take into account dynamic aspects of the scene are not considered for evaluation. In addition to standard bottom-up metrics, Pal \etal \cite{2020_CVPR_Pal} measure how well the model captures the semantic context of the scene represented by detected road users and infrastructure using F1-score and MAE metrics. However, it is unclear whether their semantic ground truth adequately captures what objects should be attended by the driver. Alternatively, Xia \etal \cite{2018_ACCV_Xia} propose using human subjects to evaluate their model by showing them videos of traffic scenes attenuated by human and predicted saliency maps. The authors report that human subjects preferred the output of their model in $41\%$ of trials and maps produced by the competing DR(eye)VE model \cite{2017_IV_Palazzi} in $29\%$ of trials. The significance of these results is difficult to interpret as they are not tied to safety or driving performance.

\subsubsection{Predicting gaze inside the vehicle}
\label{sec:in_vehicle_gaze}
Identifying where the driver is looking inside the vehicle is an important source of information for many driver monitoring and assistance applications which will be described later in Section \ref{sec:driver_monitoring}. Here we consider approaches that determine approximate gaze AOI given a view of the driver's face from the driver-facing camera installed inside the vehicle, although some methods derive this information from 3D head poses without visual input \cite{2018_ITSC_Jha, 2016_TITS_Lundgren}. Table \ref{tab:driver_in_vehicle_gaze} lists proposed models with their properties.

\begin{table}[t!]
  \centering
\resizebox{1\textwidth}{!}{

    \begin{tabular}{lccllcl}
    Reference & Input & \specialcell{Observation\\length} & \specialcell{Processing\\pipeline} &  \# AOIs & Dataset & Metrics \\
    \midrule
    Vora \etal 2018 \cite{2018_TIV_Vora} & RGB   &  1 frame & DL & 6+EC   & custom: on-road  & accuracy, confusion matrix \\
    Vora \etal 2017 \cite{2017_IV_Vora} & RGB   & 1 frame & DL &  6+EC  & custom: on-road & accuracy \\
    Vasli \etal 2016 \cite{2016_ITSC_Vasli} & RGB &  1 frame & hybrid & 6   & custom: on-road & accuracy, confusion matrix \\
    Fridman \etal 2016 \cite{2016_IS_Fridman} & RGB & 1 frame & hybrid & 6   & custom: on-road & accuracy \\
    Fridman \etal 2016 \cite{2016_IET_Fridman} & RGB & 1 frame & hybrid & 11   & custom: on-road & accuracy \\
    Choi \etal 2016 \cite{2016_BigComp_Choi} & RGB & 1 frame & DL & 8+EC   & custom: on-road & accuracy \\
    Tawari \etal 2014 \cite{2014_IV_Tawari} & RGB & 6s    & hybrid &  8   & custom: on-road & accuracy, confusion matrix \\
    Tawari \etal 2014 \cite{2014_ITSC_Tawari} & RGB & 1 frame & hybrid &  6   & custom: on-road & accuracy, confusion matrix\\
    Lee \etal 2011 \cite{2011_TITS_Lee} & RGB & 1 frame & hybrid & 18   & custom: on-road & SCER, LCER \\
    \end{tabular}%
    }
  \caption[Algorithms for estimating driver gaze AOI inside the vehicle]{List of algorithms for estimating driver gaze AOI inside the vehicle. Some models treat ``eyes closed'' (EC) or blink as a separate AOI. DL denotes deep-learning processing pipeline. Metrics SCER and LCER are defined as ratio of number of strictly correct frames to total frames and ratio of loosely correct frames to total frames respectively.}

  \label{tab:driver_in_vehicle_gaze}%
\end{table}%

\noindent
\textbf{Processing pipeline.} Current approaches to estimating drivers' gaze combine geometric modeling of head and eye in 3D and machine learning methods for classification of geometric and other features into specified AOI classes. A full geometric pipeline starts with face detection using the off-the-shelf algorithms (\eg DLIB implementation \cite{2009_JMLR_King} or Viola-Jones \cite{2001_IJCV_Viola}), followed by detection of facial landmarks and optionally iris or pupil detection. Facial landmarks can be used to recover head pose (roll, yaw, and pitch) using a generic 3D face model (\eg POS \cite{1992_ECCV_DeMenthon}). From detected eye locations and head pose a 3D gaze vector and its intersection with the area inside the vehicle can be determined analytically (see also a review of gaze estimation algorithms in \cite{2017_IEEEAccess_Kar}).

Many works replace some of the explicit 3D computations with machine learning approaches. For example, in \cite{2016_ITSC_Vasli} head pose, gaze angle, and the intersection point of gaze vector are fed into an SVM to determine the AOI, Fridman \etal \cite{2016_IS_Fridman, 2016_IET_Fridman} apply a random forest to classify detected facial landmarks, and Lee \etal \cite{2011_TITS_Lee} instead of landmarks and eye corners use yaw, pitch, face pose and face size as features for the SVM classifier. 

Deep learning approaches reduce the explicit computations even further and generalize better than analytical methods to occlusions (\eg by eyewear) and head or eye rotations. CNN-based models reach high accuracy by directly classifying cropped images of drivers' faces \cite{2016_BigComp_Choi, 2017_IV_Vora, 2018_TIV_Vora}. In a large study, Vora \etal \cite{2018_TIV_Vora} experimented with multiple face cropping techniques and several common CNN architectures and determined that the upper half of the face provided optimal information although even heuristic-based crops containing drivers face were sufficient.  

\noindent
\textbf{Evaluation.} As shown in Table \ref{tab:driver_in_vehicle_gaze}, accuracy and confusion matrix are the most commonly used evaluation metrics. Other classification metrics, such as precision or recall, are not provided. Accuracy is used as a global performance assessment while the confusion matrix provides accuracy per AOI and shows which areas are often confused by the algorithms. For example, the speedometer zone is difficult to detect because it is directly below the road where the majority of gaze is allocated and a subtle eye movement is sufficient to switch the gaze between these two areas \cite{2014_ITSC_Tawari, 2014_IV_Tawari}. Some works propose to train and evaluate models on data from individual users \cite{2016_IS_Fridman} to account for significant individual differences between drivers \cite{2016_IET_Fridman}.

Due to the lack of publicly available data for gaze zone estimation, each work uses a custom dataset with varying properties. Some are recorded at different times of the day and involve multiple drivers, with and without eyewear \cite{2011_TITS_Lee, 2014_IV_Tawari, 2016_IS_Fridman, 2017_IV_Vora}. The annotations of AOIs also vary across studies. Typically, 6 zones are defined: road, instrument cluster, center stack, rear-view mirror, left, and right. Studies using 8 gaze zones consider side windows and mirrors separately. More fine-grained zones are rarely used. Given the lack of code and public datasets and inconsistent definitions of gaze zones, comparisons between different approaches are virtually impossible.


\subsection{Driver monitoring}
\label{sec:driver_monitoring}

\subsubsection{Inattention detection}

Detecting driver inattention, which includes various types of distractions as well as drowsiness, is a major part of driver monitoring systems that are becoming commonly implemented in vehicles. Since it is a well-surveyed area, here we will present a summary of the existing approaches. Table \ref{tab:inattention_algorithms} lists algorithms for detecting inattention divided into three groups: distraction detection, drowsiness detection, and algorithms that can detect both. The underlying assumption behind all these works is that gaze is tightly coupled with attention, therefore inattention can be detected by observing changes in gaze patterns due to secondary task involvement or drowsiness.

\begin{table}[t!]
  \centering
\resizebox{1\textwidth}{!}{
    \begin{tabular}{llllllll}
    Reference & \specialcell{Inattention\\type} & \specialcell{Observation\\Length} & Input & Output & \specialcell{Algorithm\\pipeline} & Dataset & Metrics \\
    \midrule
    Fan \etal 2019 \cite{2019_TMC_Fan} & VD, VMD, CD & 1-5s  & RGB   & distraction type & FE $\rightarrow$
    LSTM & custom: on-road, sim & P, R, Acc, F1 \\
    Wang \etal 2018 \cite{2018_IROS_Wang} & CD    & 5, 10, 15s & GC    & driver status & LSTM  & custom: sim & P, R, F1 \\
    Liao \etal 2016 \cite{2016_TITS_Liao} & CD    & -     & EV, GC & driver status & SVM   & custom: sim & DR, F1 \\
    Li \etal 2016 \cite{2016_TITS_Li} & VD, VMD, CD & 2, 5s & RGB$^\ast$ , EV & distraction type & FE $\rightarrow$ RB & custom: on-road & F1 \\
    Liao \etal 2016 \cite{2016_IV_Liao} & CD    & 2, 5s & EV, GC, HP & driver status & SVM   & custom: sim & DR, F1 \\
    Liu \etal 2015 \cite{2015_TITS_Liu}  & CD  & 10s   & HP, GC & driver status & SVM   & custom: on-road & P, R, Acc, F1 \\
    Li \etal 2015 \cite{2015_TITS_Li}  & VD, VMD, CD & 10s   & RGB, EV & distraction type & FE $\rightarrow$ BC & custom: on-road & P, R, F1 \\
    Hirayama \etal 2012 \cite{2012_ITSC_Hirayama}  & CD    & 10s   & GL    & driver status & BC    & custom: on-road & Acc \\
    Liang \etal 2012 \cite{2012_HumanFactors_Liang} & D     & 3, 6, 12, 24s & GL    & driver status & LR    & 100-Car NDS & PC \\
    Wollmer \etal 2011 \cite{2011_TITS_Wollmer} & VMD   & 3s    & EV, HR & driver status & LSTM  & custom: on-road & P, R, Acc, F1 \\ \hdashline
    Zhang \etal 2019 \cite{2019_IEEEAccess_Zhang} & DR    & 10s   & RGB, PS & driver status & FE $\rightarrow$ TD & custom: on-road & P, R, Acc, F1\\
    Deng \etal 2019 \cite{2019_IEEEAccess_Deng} & DR    & 60s   & RGB   & driver status & FE $\rightarrow$ CNN $\rightarrow$ TD & custom: on-road & P, Acc \\
    Yu \etal 2018 \cite{2018_TITS_Yu}  & DR    & 5 frames & NIR   & driver status & CNN   & DDD   & ROC, P, R, F1 \\
    Zhao \etal 2017 \cite{2017_IET_Zhao} & DR    & 1s    & RGB   & driver status & FE $\rightarrow$ DBN & custom: on-road & Acc \\
    Weng \etal 2017 \cite{2017_ACCV_Weng}  & DR    & 300 frames & NIR   & driver status & FE $\rightarrow$ DBN & DDD   & Acc, F1 \\
    Shih \etal 2017 \cite{2017_ACCV_Shih}  & DR    & 50 frames & NIR   & driver status & FE $\rightarrow$ CNN $\rightarrow$ LSTM & DDD   & Acc, F1 \\
    Choi \etal 2016 \cite{2016_ApplSci_Choi} & DR    & 15 frames & RGB   & driver status & FE $\rightarrow$ HMM & custom: on-road & Q \\
    Wang \etal 2016 \cite{2016_AccidentAnalysis_Wang}  & DR    & 60s   & EV, BL, EC & driver status & LR $\rightarrow$ MLP & custom: sim & Acc \\
    Jin \etal 2013 \cite{2013_AdvMechEng_Jin} & DR    & 10s   & EC, GC & driver status & SVM   & custom: sim & Acc \\
    Zhang \etal 2010 \cite{2010_JCTA_Zhang} & DR    & 6m    & NIR   & driver status & FE $\rightarrow$ TD & custom: on-road & DR \\
    Friedrichs \etal 2010 \cite{2010_IV_Friedrichs} & DR    & 5 timesteps & EC, HP, EV, BL & driver status & MLP   & custom: on-road & DR \\ \hdashline
    Chiou \etal 2019 \cite{2019_TITS_Chiou}  & \specialcell{VD, VMD,\\CD, DR} & 1, 2, 4, 8, 16 frames & RGB   &\specialcell{driver status,\\distraction type} & FE $\rightarrow$ TD & \specialcell{custom: on-road,\\YawDD, DDD} & P, R, Acc, F1 \\
    Mbouna \etal 2013 \cite{2013_TITS_Mbouna} & D, DR & 120 frames & RGB   & driver status & FE $\rightarrow$ SVM & custom: on-road & TP, FP, FN, TN \\
    Jo \etal 2011 \cite{2011_OptEng_Jo} & D, DR & 10s   & NIR   & driver status & FE $\rightarrow$ TD & custom: on-road & FP, FN \\
    Flores \etal 2011 \cite{2011_IET_Flores} & D, DR & 1 frame & NIR   & driver status & FE $\rightarrow$ TD & custom: on-road & DR \\
    \end{tabular}%
 }%

	\caption[Algorithms for detecting distraction, drowsiness and both]{A table listing algorithms for detecting distraction, drowsiness and both (separated by dashed line). The following abbreviations are used in the table. Inattention type: VD - visual distraction, VMD - visual-manual distraction, CD - cognitive distraction, D - any distraction, DR - drowsiness. Input: RGB and NIR - color and near-infrared images from driver-facing camera, EV - ego-vehicle data, GC - gaze coordinates, GL - glance locations, EC - eye closure, BL - blink features, HP - head pose, PS - physiological signals. Algorithm pipeline: FE - feature extraction, BC - Bayesian classifier, LR - logistic regression, SVM - support vector machine, TD - threshold, CNN - convolutional neural network, LSTM - long-short term memory, MLP - multi-layer perceptron, DBN - deep belief network, HMM - Hidden Markov Model, RB - RUSBoost. Metrics: P - precision, R - recall, Acc - accuracy, F1 - F-score, ROC - receiver operating characteristic curve, DR - detection rate (ratio of detected samples to the total number of samples), Q - qualitative. Note that common units for observation length cannot be established since not all studies provide sampling rate used in the recording of data.\\ $^\ast$ - both driver-facing and road-facing cameras are used.}
  \label{tab:inattention_algorithms}%
\end{table}%

\noindent
\textbf{Distraction detection} algorithms either focus on the specific distraction type (\eg cognitive distraction) or learn to distinguish between visual, visual-manual, and cognitive distractions. As discussed in Section \ref{sec:secondary_tasks_effects}, different secondary tasks lead to different gaze patterns and driving behaviors, therefore driver monitoring systems often rely on features such as gaze coordinates or coarsely labeled glance locations together with ego-vehicle information. Since not only gaze locations but also their temporal distribution change depending on the secondary activity, all algorithms use temporal data.

Algorithms that receive visual input usually follow the steps described for in-vehicle gaze detection (see Section \ref{sec:in_vehicle_gaze}) \cite{2019_TMC_Fan}. Using gaze data provided by the remote eye-tracking systems is more reliable and leads to faster processing times since heavy computations necessary for processing visual information are eliminated. Often, in addition to the raw gaze data, its various statistical functionals are computed (\eg mean, standard deviation, variance, percentile, \etc) \cite{2016_IV_Liao}. Then various methods for feature elimination may be applied to find the optimal set \cite{2011_TITS_Wollmer, 2016_TITS_Liao}. For instance, Wollmer \etal \cite{2011_TITS_Wollmer} found that raw head rotation angle and its derivatives and steering wheel angle are sensitive to visual-manual tasks tested, whereas Liao \etal \cite{2016_IV_Liao} determined that gaze coordinates can help detect cognitive distractions. Liang \etal \cite{2012_HumanFactors_Liang} used crash data from 100-Car NDS to show that glance durations due to secondary task involvement were associated with increased accident risk, however, glance history and glance locations did not further increase the sensitivity of the algorithm. Moreover, their study indicated that aggregating glances across longer time intervals dilutes distraction signal. 

Once the necessary features are extracted, various classifiers may be applied, ranging from simple linear regression \cite{2012_HumanFactors_Liang} to boosting \cite{2016_TITS_Li}, SVMs \cite{2016_TITS_Liao, 2016_IV_Liao, 2015_TITS_Liu}, and deep learning models \cite{2019_TMC_Fan, 2018_IROS_Wang, 2011_TITS_Wollmer}. Data is labeled as distracted or normal according to the timed intervals during which the drivers were asked to perform secondary tasks \cite{2016_TITS_Liao, 2016_IV_Liao, 2015_TITS_Liu, 2012_ITSC_Hirayama}.

\noindent
\textbf{Drowsiness detection} methods rely more on the driver's appearance to detect signs of fatigue such as frequent blinking, eyes closed for long periods of time, yawning, and nodding. Near-infrared (NIR) cameras are used more often for this purpose since they can be used in night conditions and are generally more robust to illumination changes. Eye, mouth, and head features necessary for estimating drowsiness state \cite{2017_IET_Zhao, 2010_IV_Friedrichs} can be detected using methods similar to those described in Section \ref{sec:in_vehicle_gaze}. These include face detection \cite{2017_ACCV_Shih, 2017_IET_Zhao} and tracking\cite{2019_IEEEAccess_Deng}, locating facial landmarks \cite{2017_IET_Zhao, 2017_ACCV_Weng}, detecting the state of eyes to measure blinks \cite{2016_ApplSci_Choi} and eye closure \cite{2017_ACCV_Weng}, recognizing the state of mouth to identify signs of drooping and yawning \cite{2017_ACCV_Weng}, and measuring head pose and movement \cite{2016_ApplSci_Choi}.

Detected features can be converted into drowsiness measures established in the behavioral literature, such as the percentage of eyes closed (PERCLOS) \cite{2013_AdvMechEng_Jin, 2010_JCTA_Zhang, 2010_IV_Friedrichs} and blink frequency \cite{2019_IEEEAccess_Zhang, 2016_AccidentAnalysis_Wang, 2010_IV_Friedrichs}. Then, simple thresholding may be applied to detect drowsiness \cite{2019_IEEEAccess_Zhang, 2010_JCTA_Zhang}. Other methods, such as SVM  \cite{2013_AdvMechEng_Jin} or MLP \cite{2010_IV_Friedrichs}, may be more robust to individual differences between drivers \cite{2019_IEEEAccess_Zhang} 

Machine learning and recent deep learning approaches can replace some of the explicit computations. For instance, Zhao \etal \cite{2017_IET_Zhao} feed facial landmarks and raw images of the eyes and mouths of the drivers into a deep belief network (DBN) to classify drowsiness expressions. Weng \etal \cite{2017_ACCV_Weng} use three separate DBNs for mouth, head, and eye to encode features, on top of which two HMMs are adopted to continuously model temporal relationships between features for alert and drowsy states. Drowsiness is determined via inverse logit transform applied to the difference of likelihoods computed by each HMM. Shih \etal \cite{2017_ACCV_Shih} use a pretrained CNN to encode cropped images of drivers' faces that are then passed to an LSTM. 3D CNNs have also been applied to extract spatio-temporal representations that can be used for classification \cite{2019_IEEEAccess_Zhang, 2018_TITS_Yu}.

Despite high accuracies above 90\% reported by many algorithms, dealing with extreme head angles \cite{2016_ApplSci_Choi, 2017_IET_Zhao}, changes in illumination, and glare from eyewear \cite{2019_IEEEAccess_Deng, 2010_IV_Friedrichs} must be resolved before deploying these models. Furthermore, temporal window should be selected to balance the sensitivity and latency of the detection. For example, when using blink frequency or PERCLOS, longer time intervals typically work best but also increase the risk of missing microsleeps or issuing alerts  too late.

\noindent
\textbf{Distraction and drowsiness detection} algorithms combine approaches described above in a single system. For example, Chiou \etal \cite{2019_TITS_Chiou} compute temporal face descriptors for normal, distracted, and drowsy states. At test time, incoming data is compared to each of the models and is classified accordingly. Mbouna \etal \cite{2013_TITS_Mbouna} do not explicitly distinguish between distraction and drowsiness and instead compute an overall alertness score based on the temporal eye activity and head pose. Heuristic approaches apply drowsiness detection when the driver is looking straight ahead and start identifying distraction when drivers' eyes are averted away from the road \cite{2011_OptEng_Jo, 2011_IET_Flores}.

\noindent
\textbf{Evaluation.} Standard metrics for classification problems, such as precision, recall, accuracy, and F-score, are used for evaluating distraction and drowsiness detection. High recall minimizes the risk of missing the driver getting distracted or falling asleep. At the same time, high precision avoids alerting the driver unnecessarily. In almost all studies that report recall and precision, precision is somewhat higher \cite{2011_TITS_Wollmer, 2015_TITS_Li, 2015_TITS_Liu, 2018_IROS_Wang, 2019_TMC_Fan}. 

In general, there is a lack of well-defined and publicly available benchmarks for inattention detection that include a comprehensive set of tasks recorded under systematically varied and realistic conditions with a broad pool of participants. Unpublished custom datasets used to evaluate nearly all algorithms have different (and often under-specified) properties, leading to two important implications. First, given that source code is often not provided, models cannot be compared. As a result, most approaches are evaluated against versions of themselves on different sets of features and temporal windows, making it difficult to establish the relative effectiveness of different methods and overall progress in the field. Second, the practical applicability of the models is uncertain. As mentioned earlier, it is common practice to ask drivers to act drowsy and to induce distraction by making them perform artificial tasks at timed intervals. Even though detectable changes in gaze allocation and driving performance can be achieved, nothing guarantees that they reflect inattention that naturally occurs during driving. According to behavioral literature, secondary task involvement is voluntary and highly depends on other conditions (see Section \ref{sec:secondary_task}), therefore forcing the driver to engage in often meaningless tasks on-demand and incentivizing high performance may not produce the same effect. 

Potential practical validity and user acceptance of the proposed inattention detection systems is not addressed. Although some models that we reviewed report high precision and recall well above $90\%$ it is hard to estimate how these metrics translate to practice since datasets may not represent the conditions well or capture user feedback. Out of all models listed in Table \ref{tab:inattention_algorithms} only one was deployed in a small field study  which revealed visual processing issues caused by vibration of the vehicle on the curves and on rough surfaces \cite{2019_IEEEAccess_Zhang}. As an earlier and larger field study of the AttenD algorithm demonstrated, a number of additional issues may be revealed by user studies. For instance, warnings issued once per 10 km were considered excessive by the users and had to be suppressed in the urban areas with speeds below 50 km/h \cite{2013_TITS_Ahlstrom}. 

\subsubsection{Driver maneuver recognition and prediction}
A driver monitoring system should be able to recognize and anticipate drivers' intended maneuvers. As was discussed in Sections \ref{sec:bu_td_attention} and \ref{sec:vehicle_control}, drivers' gaze is strongly associated with the goal and actions being performed. Here we consider models that exploit this property.  

\begin{table}[t!]
  \centering
\resizebox{1\textwidth}{!}{  
    \begin{tabular}{llclllll}
    Reference & Input & \specialcell{Observation\\length} & Maneuvers & TTE   & Model & Dataset & Metrics \\
    \midrule
    Akai \etal 2019 \cite{2019_IV_Akai} & GD, EV & 0.5s  & LT, RT, S & 0 & HMM & custom: on-road & Q \\
    Martin \etal 2018 \cite{2018_TIV_Martin} & RGB$_D$   & 5s    & LLC, RLC, S & 0-5s  & MVN &custom: on-road & R, Q \\
    Martin \etal 2017 \cite{2017_IV_Martin} & RGB$_D$   & 5s    & LLC, RLC, S & 0-2s  & MVN & custom: on-road & P, R \\
    \href{https://github.com/asheshjain399/RNNexp}{Fusion-RNN} \cite{2016_ICRA_Jain}  & RGB$_{D,S}$   & 0.8s  & LLC, RLC, LT, RT, S & 3-4s  & RNN & Brain4Cars & P, R \\
    AIO-HMM \cite{2015_ICCV_Jain} & RGB$_{D,S}$   & 0.8s  & LLC, RLC, LT, RT, S & 3-4s & HMM & Brain4Cars & P, R, CM, F1, FP \\
    Li \etal 2016 \cite{2016_TITS_Li} & RGB$_{D,S}$, EV & 2, 5s & LT, RT, S & 0     & RB & custom: on-road & P, R, F1 \\
    \end{tabular}%
 }
 \caption[Algorithms for detecting and predicting driver actions]{A list of algorithms for detecting and predicting driver actions. Column TTM (time-to-maneuver) specifies how far into future the action is predicted, '0' denotes action detection. The following abbreviations are used. Input: RGB - 3-channel images from driver- ($D$) and scene-facing ($S$) camera, EV - ego-vehicle data, GD - gaze data. Actions: LT/RT - left and right turns, LLC/RLC - left and right lane change, S - straight. Models: HMM - Hidden Markov Model, RF - random forest, MVN - multivariate normal distribution, RNN - recurrent neural network, RB - RUSBoost.  Metrics: Q - qualitative, P - precision, R - recall, CM - confusion matrix, F1 - F-score, FP - false positives.}
  \label{tab:action_prediction}%
\end{table}%

Table \ref{tab:action_prediction} lists algorithms that recognize or predict drivers' maneuvers given observations extracted from driver- and optionally scene-facing cameras. Typical maneuvers considered are turns and lane changes. Each of these requires several actions from the driver such as checking mirrors or looking over the shoulder to check the blind spots, therefore all algorithms operate on observations gathered over some time. 

As with other driver monitoring algorithms, the first step is the estimation of drivers' gaze direction (unless data provided by eye-tracker is used as in \cite{2019_IV_Akai}). The processing pipeline usually includes face detection and tracking and facial landmark detection. Then explicit gaze information can be extracted, such as gaze zones \cite{2018_TIV_Martin, 2017_IV_Martin, 2016_TITS_Li}, gaze duration, frequency, and blinks \cite{2016_TITS_Li,2018_TIV_Martin}. Alternatively, action prediction can be made using an implicit representation of gaze such as tracked facial landmarks aggregated over time as proposed in \cite{2015_ICCV_Jain, 2016_ICRA_Jain} or mirror-checking actions \cite{2016_TITS_Li}.

Many methods have been proposed for capturing temporal changes in the representations. Some authors aggregate detected features across observation time and apply classifiers to predict the upcoming maneuver. For instance, Li \etal \cite{2016_TITS_Li} apply a random undersampling boosting algorithm (RUSBoost) \cite{2009_IEEE_Seiffert} to a combination of mirror-checking actions and vehicle dynamics derived from the CAN bus. Martin \etal \cite{2017_IV_Martin, 2018_TIV_Martin} propose to model maneuvers using an unnormalized multivariate normal distribution (MVN) of spatio-temporal descriptors that capture the duration of gaze in relevant AOIs. At test time, descriptors computed from incoming data are compared to MVNs for each maneuver based on Mahalanobis distance. Besides discriminative models, temporal modeling that fits the data more naturally has also been applied. Jain \etal \cite{2015_ICCV_Jain} and Akai \etal \cite{2019_IV_Akai} propose auto-regressive input-output HMMs to model driver's actions given driver gaze and vehicle dynamics. RNNs have also been shown effective for combining multiple sensory streams to estimate future maneuvers \cite{2016_ICRA_Jain} but lack explainability of HMM models.

\noindent
\textbf{Evaluation.} Since action prediction is framed as classification, typical metrics such as precision, recall, F1 are used to evaluate the results. As expected, it is more difficult to predict maneuver several seconds in advance, but precision and recall improve as time-to-maneuver (TTM) decreases \cite{2017_IV_Martin}. Besides issues with face detection and tracking due to illumination changes \cite{2015_ICCV_Jain}, some maneuvers are more difficult to predict because of the overlap in behaviors (\eg mirror checking is not always a precursor to lane changing \cite{2017_IV_Martin}) and lack of visual cues from the driver (when they are familiar with the road or turn from a dedicated lane \cite{2015_ICCV_Jain}).

\subsubsection{Driver awareness estimation}
\label{sec:driver_awareness}
Driver monitoring systems discussed so far focus on drivers' behaviors but do not connect them to the environment outside the car. In addition to understanding what the driver and other road users are doing it is helpful to know what the driver is aware of. For example, at the intersection the system may check whether the driver looked at the traffic lights and signs and noticed approaching pedestrians and vehicles, and issue warnings if necessary.

\begin{table}[t!]
  \centering
\resizebox{1\textwidth}{!}{  
    \begin{tabular}{lllllll}
    Reference & Input & \multicolumn{1}{l}{\specialcell{Observation\\length}} & Output & \specialcell{Objects\\of interest} & \multicolumn{1}{l}{Dataset} & Metrics \\
    \midrule
    Ma \etal 2019 \cite{2019_ACM_Ma} & RGB$_{D,S}$ &  \multicolumn{1}{r}{\multirow{11}[1]{*}{\rotatebox{90}{\specialcell{continuous\\\\}}}}     & warning & \specialcell{vehicles,\\pedestrians} &    \multicolumn{1}{r}{\multirow{11}[1]{*}{\rotatebox{90}{\specialcell{custom: on-road\\\\}}}} & TP, FP \\
    Schwehr \etal 2018 \cite{2018_ITSC_Schwehr} & GD, RD  &       & PoG  & road users &        & Q \\
    Zabihi \etal 2017 \cite{2017_IV_Zabihi} & RGB$_{D,S}$ &       & BB, PoG &   signs &    & Q \\
    Schwehr \etal 2017 \cite{2017_ITSC_Schwehr} & GD, RD &       & PoG  & road users &       & Q \\
    Langner \etal 2016 \cite{2016_ICRA_Langner} & GD, RGB$_S$ &       & warning & traffic lights &       & Q \\
    Roth \etal 2016, \cite{2016_IV_Roth} & HP$_{D,P}$, BB, EV &       & warning & pedestrians &       & Q \\
    Kowsari \etal 2014 \cite{2014_IV_Kowsari} & GD, RGB$_S$ &       & PoG   & - &       & E \\
    Tawari \etal 2014 \cite{2014_ITSC_Tawari_1} & RGB$_{D,S}$ &       & PoG, BB & pedestrians &      & DR \\
    Rezaei \etal 2014 \cite{2014_CVPR_Rezaei} & RGB$_{D,S}$ &  & risk level, warning & vehicles &  & Q \\
    Bar \etal 2013 \cite{2013_IV_Bar} & RGB$_{D,S}$, LS, EV &       & BB + IS, warning & vehicles &       & Q \\
    Mori \etal 2012 \cite{2012_ITSC_Mori} & RGB$_{D,S}$, LS &       & awareness & vehicles &       & Q \\
    \end{tabular}%
  }
  \caption[Algorithms for driver situation awareness estimation]{A list of algorithms for driver assistance that use information about both driver gaze and objects outside the vehicle. The following abbreviations are used. Input: RGB - 3-channel images from driver- ($D$) or scene-facing ($S$) camera, RD - range data, LS - laser scanner data, EV - ego-vehicle data, GD - driver gaze data, HP - head pose (driver's ($D$) or pedestrian's ($P$)). Output: PoG - point of driver's gaze in the environment, BB - bounding box, IS - importance score. Metrics: Q - qualitative, DR - detection rate,  E - error in PoG estimation, TP/FP - true/false positive.}
  \label{tab:awareness_detection}%
\end{table}%

Table \ref{tab:awareness_detection} lists algorithms for driver assistance systems that can infer whether the driver is aware of the vehicles, pedestrians, and other road users around the ego-vehicle as well as signs and traffic signals and provide appropriate warning signals. Most follow the same procedure: 1) detect drivers' direction of gaze in 3D from driver-facing camera images via geometric or machine learning methods, 2) convert gaze to vehicle's frame of reference using geometric transformations, 3) detect objects in the scene-facing camera images, 4) match gaze with objects to determine whether the driver has noticed them and 5) issue a warning when appropriate. 

Drivers' gaze direction estimation in most models follows procedures described earlier based on either explicit geometric calculations or machine learning (see Section \ref{sec:in_vehicle_gaze}). Object detection relies on off-the-shelf algorithms \cite{2019_ACM_Ma}, classical vision pipelines \cite{2014_CVPR_Rezaei, 2016_ICRA_Langner, 2017_IV_Zabihi}, or is manually annotated \cite{2014_ITSC_Tawari_1}. Distances to the detected objects and their relative velocities may be inferred from a stereo camera \cite{2016_ICRA_Langner}, provided by sensors \cite{2018_ITSC_Schwehr, 2016_ICRA_Langner, 2014_IV_Kowsari, 2013_IV_Bar, 2012_ITSC_Mori}, or determined using simple heuristics \cite{2019_ACM_Ma}. 

Different strategies have been proposed for detecting what objects the driver is aware of. The simplest solution is to detect whether the driver's gaze is within the object's bounding box \cite{2019_ACM_Ma, 2014_ITSC_Tawari_1, 2013_IV_Bar}. However, to account for errors in gaze estimation, instead of a 3D vector, gaze can be represented by a cone of attention projected from the drivers' eye towards the windshield. Consequently, all objects within the intersection area are considered as detected \cite{2016_ICRA_Langner, 2017_IV_Zabihi}. Some algorithms also take into account that drivers retain information about objects for some time after looking at them \cite{2012_ITSC_Mori, 2019_ACM_Ma}. Schwehr \etal \cite{2017_ITSC_Schwehr, 2018_ITSC_Schwehr} propose a Dynamic Bayesian Network to model the joint probability distribution of the object states in the 2D vehicle coordinate system and the gaze state of the driver given the static and mobile object coordinates as well as gaze direction measurements in 2D. The system can estimate which static and mobile objects have been fixated on or tracked.

\noindent
\textbf{Evaluation.} Although performance is reported for individual components of the models such as driver gaze estimation and object detection, often the entire system is qualitatively evaluated on several illustrative scenarios. Since such assistive systems are in the realm of HCI, user studies assessing their reliability in a variety of real-life scenarios, ease of use, and acceptance are desirable in addition to quantitative performance metrics. Out of the reviewed systems, only one was evaluated in a small field study involving 3 routes around the university campus and 4 drivers \cite{2019_ACM_Ma}. The proposed gaze-directed forward collision warning (FCW) system had improved rates of valid and false alarms compared to the standard FCW. The authors also reported that their system did not issue warnings in approximately 9\% of cases on some routes. No information on user experience was provided.

\noindent
\textbf{Challenges.} Multiple challenges are yet to be resolved to make such driver assistance systems viable. Some of these have already been discussed in Sections \ref{ch:what_driver_sees} and \ref{ch:attention_measures} but will be reiterated in short here. Technically, associating human gaze with objects outside of the vehicle and using this as a basis for awareness estimation is prone to error for several reasons. First, gaze does not always land exactly on the objects and most of the objects in the scene are detected peripherally. Second, both the depth and size of the objects matter. For example, to compensate for the small size of pedestrians and traffic infrastructure, detection area around the object may be enlarged, leading to issues in the crowded scenes. On the other hand, if the driver is focusing on one object, they may miss other objects farther away or closer even if the objects are spatially close in the image plane. Third, even if the gaze landed on the object, this does not guarantee processing. Hence, other measurements such as ego-vehicle information should be used in addition to gaze data. For example, if the driver glanced at the stop sign but is not slowing down, there is a chance that they might have missed the sign.

Another issue debated in the behavioral literature is how to warn the driver (see Section \ref{sec:automation_guidance_warning}). For simpler applications, such as distraction detection, an audio signal, seat or wheel vibrations and visual indicators shown on the instrument panel may be sufficient and have been successfully deployed in commercial DMS. When it comes to warning the driver about making an unsafe maneuver or potentially missed hazards, message contents is highly dependent on the situation, thus, the appropriate mode of delivery may be different depending on the context. For example, if a driver is about to miss a stop sign, a blinking icon on the in-vehicle display may suffice. However, it is not clear what approach is better for more complex situations, such as failing to see an approaching motorcyclist at an intersection or when a hidden pedestrian is about to appear from behind a parked vehicle.

\subsection{Attention for self-driving vehicles}
The works discussed in this section are at the intersection of several major trends in robotics, intelligent transportation, and computer vision, namely the development of automated driving technology, use of deep learning approaches for perception and reasoning, and widespread adoption of attention mechanisms (loosely analogous to those in human visual and cognitive systems) in said deep learning approaches.

Self-driving technology is a long, ongoing effort towards improving safety by automating control of the vehicle. Creating robust AI-driven systems capable of matching human driver performance relies on solving multiple problems in virtually every area of computer vision and robotics. Perception alone involves overcoming significant challenges in object detection, tracking, scene segmentation, depth estimation, and optical flow (see \cite{2020_Now_Janai} for a recent detailed review of vision for autonomous driving). Decision-making for motion planning, behavior selection, and vehicle control requires precise mapping and localization \cite{2020_ESA_Badue} as well as understanding the behaviors of vulnerable road users \cite{2019_TITS_Rasouli}. Current self-driving systems that tackle these issues can be broadly subdivided into modular and end-to-end \cite{2020_arXiv_Tampuu}. The former use dedicated modules for various elements and stages of processing whereas the latter uses a single complex neural network that receives input from sensors and outputs control commands. Lastly, one of the recent advances in deep learning is the inclusion of attention mechanisms for perception and reasoning that serve to both improve the performance and explainability of the models. Due to the overwhelming number of approaches relevant to various aspects of self-driving, here we focus on several latest examples of using attention in the end-to-end driving models for vehicle control listed in Table \ref{tab:self_driving_attention}.

\begin{table}[t!]
  \centering
\resizebox{1\textwidth}{!}{  
    \begin{tabular}{llllll}
    Reference & \specialcell{Observation\\length} & Input & Output & Dataset & Metrics \\
    \midrule
    Cultrera \etal 2020 \cite{2020_CVPRW_Cultrera} & 1 frame & \multicolumn{1}{r}{\multirow{9}[1]{*}{\rotatebox{90}{\specialcell{image from scene-facing\\camera}}}}   & steering angle & CARLA \cite{2017_CRL_Dosovitskiy} & SR\\
    Wang \etal 2019 \cite{2019_ICRA_Wang} & 1 frame & & steering angle, speed & BDDV \cite{2017_CVPR_Xu} & I/C, DBI \\
    Kim \etal 2019 \cite{2019_CVPR_Kim} & 20s &   & steering angle, speed & HAD   & Corr, MAE \\
    Yuying \etal 2019 \cite{2019_IROSW_Yuying} & 1 frame &   & steering angle & TORCS \cite{2000_TR_Wymann} & DBI \\
    Liu \etal 2019 \cite{2019_ACM_Liu} & 1 frame &   & steering angle & TORCS & MAE\\
    He \etal \cite{2018_ICPR_He} & 1 frame &   & steering angle & DIPLECS, \texttt{comma.ai}  \cite{2016_arXiv_Santana} & MAE \\
    \href{https://github.com/JinkyuKimUCB/explainable-deep-driving}{Kim \etal 2018} \cite{2018_ECCV_Kim} & 4 frames (0.4s) & & steering angle, speed & BDD-X & MAE, Corr \\
    Kim \etal 2017 \cite{2017_ICCV_Kim} & 20 frames (1s) &   & steering angle & \specialcell{\texttt{comma.ai}, Udacity \cite{2000_Udacity},\\custom: on-road} & MAE \\
    Bojarski \etal 2017 \cite{2017_arXiv_Bojarski} & 1 frame &   & steering angle & custom: on-road & Q\\
    \end{tabular}%
}
  \caption[End-to-end self-driving algorithms with attention]{End-to-end self-driving approaches incorporating attention mechanisms for improving performance and/or explainability. The following abbreviations are used. Metrics: SR - success rate, I/C - number of infractions, interventions or collisions, DBI - distance travelled between infractions, MAE - mean absolute error, Corr - distance correlation.}
  \label{tab:self_driving_attention}%
\end{table}%

\begin{figure}[t!]
\centering
\begin{subfigure}{0.70\linewidth}
\centering
 \includegraphics[height=1in]{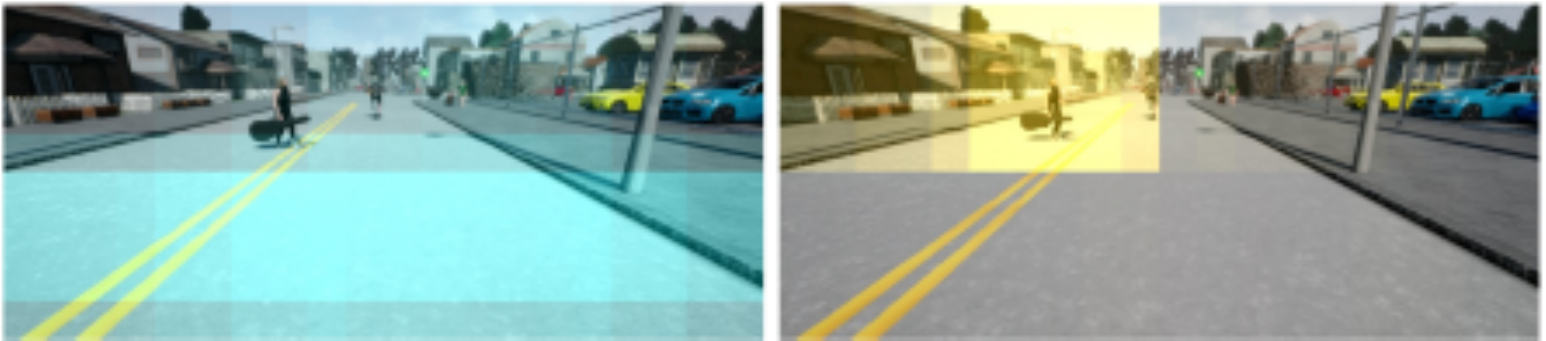}
  \caption{ROI attention}
  \label{fig:ROI_attention}
\end{subfigure}
\begin{subfigure}{0.29\linewidth}
\centering
  \includegraphics[height=1in]{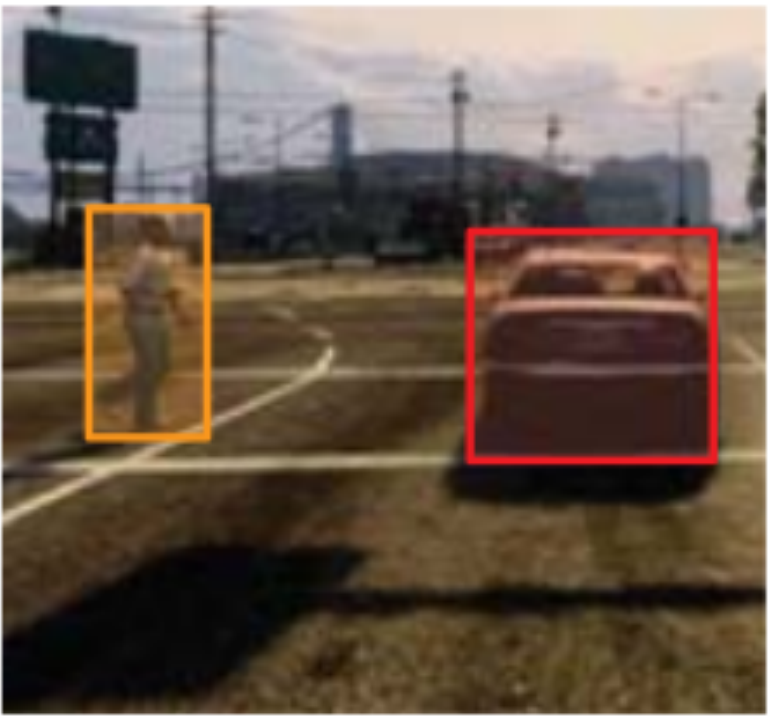}
\caption{Object attention}
\label{fig:object_attention}
\end{subfigure}
\begin{subfigure}{1\linewidth}
\centering
  \includegraphics[height=1in]{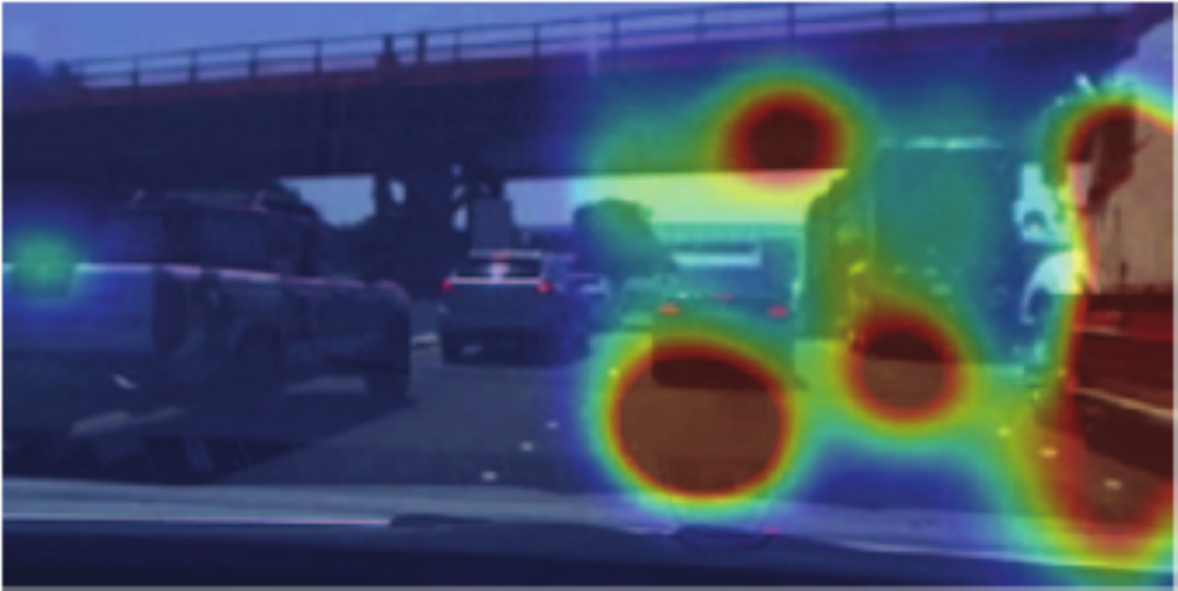}
   ~
  \includegraphics[height=1in]{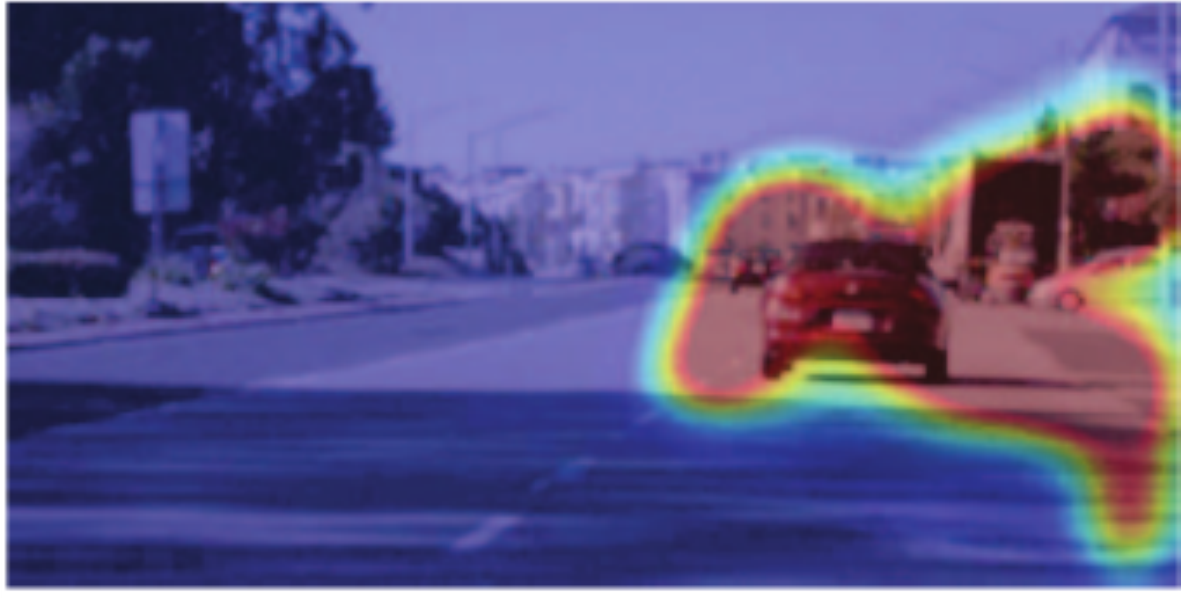}
\caption{Pixel attention}
\label{fig:pixel_attention}
\end{subfigure}
\caption[Examples of spatial attention visualizations]{Examples of spatial attention visualizations. Sources: a) \cite{2020_CVPRW_Cultrera}, b) \cite{2019_ICRA_Wang} and c) \cite{2017_ICCV_Kim}.}
\end{figure}

One of the early end-to-end approaches, PilotNet, proposed by Bojarski \etal \cite{2016_arXiv_Bojarski} consisted of a single CNN that directly mapped images from the monocular front-facing camera to steering controls. While the authors did not implement attention mechanisms in the model, they later proposed a method to highlight regions in the input that were correlated with the output \cite{2017_arXiv_Bojarski}. Starting with the last convolutional layer, each feature map is averaged and scaled up to the size of the preceding layer via deconvolution operation. The result is element-wise multiplied with the averaged map from the preceding layer. This continues until the input layer is reached. Inspecting the salient regions in the input and intermediate convolutional layers helps elucidate the internal processing of the model. Additionally, the authors show that modifying input outside of salient areas does not affect the output, but shifts in pixels inside the salient regions do change the output.

PilotNet demonstrated that end-to-end models learn to focus on certain areas such as edges of the road and ignore other visually salient areas in the input, somewhat analogous to task-based attention. Works that followed experimented with incorporating explicit spatial attention within the architecture as opposed to using it for visualization. A simple approach is to highlight important areas in the input by modulating it with the saliency map produced by an existing gaze prediction algorithm \cite{2019_ACM_Liu, 2019_IROSW_Yuying}. Both the original and modulated images are then passed to a CNN that generates steering output helping it to ignore irrelevant parts of the image and focus on the areas important for the task, improving its performance. 

Cultrera \etal \cite{2020_CVPRW_Cultrera} instead of relying on existing gaze prediction approaches, use a simple spatial attention block that is trained as part of the network. Their model starts by extracting features from the input image using a pretrained CNN and passes them through ROI pooling layer \cite{2015_ICCV_Girshick} (using a coarse grid of rectangular ROIs). An attention block consisting of fully-connected layer followed by softmax produces weights that are element-wise multiplied with the output of ROI pooling layer. The resulting attention map is shown in Figure \ref{fig:ROI_attention}. He \etal \cite{2018_ICPR_He} propose to use sparsemax for computing attention weights for input features. This results in a more selective and compact focus of attention since values with low probabilities are truncated to zero as opposed to softmax which generates probability distributions with full support. Instead of learning weights for every pixel in the image, Wang \etal \cite{2019_ICRA_Wang} propose to focus only on the important objects. Their approach relies on an object detector to find relevant objects in the scene and use a trained selector model to rank objects based on their importance for the given maneuver (Figure \ref{fig:object_attention}). Top-k scoring objects are then used to learn actions. 

Kim \etal \cite{2017_ICCV_Kim} implement an encoder-decoder architecture where CNN-based encoder computes visual features over input frames and an LSTM decoder produces steering angle output. A spatial attention block consisting of a dense layer and softmax reweights encoded features corresponding to different spatial locations conditioned on the previous hidden state of the control network and current feature vectors. This weighted input is passed to an LSTM. Another attention block implements a causality test for visualizing the most important areas in the input similar in spirit to \cite{2016_arXiv_Bojarski}. Instead of deconvolution, the authors sample salient particles from the attention map and cluster them. Then each cluster is masked (by setting the corresponding attention weight to 0) and if the output is not affected, the cluster is removed from the visualization (see the visualization of refined attention maps in Figure \ref{fig:pixel_attention}).

A different approach leverages video captioning techniques to direct the vehicle control model towards important elements of the scene \cite{2019_CVPR_Kim}. Here, the outputs of two encoders, a CNN for visual features and an LSTM for text description (\eg ''There are construction cones on the road``), are combined into a single feature vector which is fed into the vehicle controller. A multi-layer perceptron generates weights for different image areas conditioned on the previous hidden state of the controller LSTM and current feature vector.

\noindent
\textbf{Evaluation} of self-driving algorithms naturally assesses their driving performance in dedicated simulators, such as CARLA \cite{2017_CRL_Dosovitskiy} or TORCS \cite{2000_TR_Wymann}, or using datasets that provide ego-vehicle information for videos, \eg \texttt{comma.ai} \cite{2016_arXiv_Santana}, Berkeley DeepDrive Video dataset (BDDV) \cite{2017_CVPR_Xu}, and Udacity \cite{2000_Udacity}. In simulators, models are evaluated by measuring the number of collisions or infractions, and the distance traveled between infractions. For example, Wang \etal \cite{2019_ICRA_Wang} show a reduction in interventions when various attention mechanisms are added to the vehicle control network. On prerecorded datasets, the algorithms are evaluated based on how well they can match the recorded actions, steering wheel, and speed, measured using mean absolute error (MAE) or distance correlation metrics.

It is more challenging to assess the performance of attention modules separately, therefore, qualitative evaluation is common. The example shown in Figure \ref{fig:ROI_attention} shows changes in attention distribution depending on the maneuver. When turning right, the intensity of the saliency map around the crossing pedestrian is low, however, when the car is going straight, the area is highlighted, presumably reflecting its importance. Likewise, object-based attention shown in Figure \ref{fig:object_attention} is easy to interpret by looking at the bounding boxes around detected objects and their importance score visualized as color. Pixel-based attention (Figure \ref{fig:pixel_attention}) is more difficult to understand as blobs are amorphous and do not match specific objects in the scene. Besides, as Kim \etal \cite{2017_ICCV_Kim} report in their paper, more than half of the salient areas identified by their model are spurious, \ie do not affect the controls.

\subsection{Summary}
Practical solutions focus on several interconnected areas: driver monitoring, predicting drivers' actions for assistance and warning, and mimicking attention for autonomous driving. Detecting where the driver is looking now and predicting where they will look next is a basic component for all of these applications. Drivers' gaze can be used for monitoring drivers' current condition (drowsy or distracted) and anticipating their maneuvers. The most advanced applications detect driver awareness of the objects in the scene and issue warnings when appropriate. Self-driving applications aim to replace the driver altogether and also build upon research in human visual attention, either by simulating the driver's gaze distribution directly or by adopting neural attention mechanisms that focus on the important objects and can be used for explaining the actions of the autonomous system. 

Despite significant progress in all these areas, there is an acute data availability issue, particularly, lack of high-quality eye-tracking data recorded in diverse and challenging conditions. As a result, most practical research is conducted using custom unpublished datasets with unknown properties, making it difficult to replicate the results and compare the effectiveness of different approaches. Due to difficulties with large-scale naturalistic data collection, most of the data used in research is recorded in low-fidelity driving simulators, and behaviors such as distraction or drowsiness are scripted. Thus, it is unknown whether models derived from such data would generalize well to more realistic conditions. Besides data, multiple other challenges must be addressed to produce viable driver assistance systems. These include addressing the limitations of gaze as a proxy for attention, developing explicit task representation, important for modeling drivers' awareness of the objects in the scene and decision-making, and improving evaluation by taking into account driving performance and safety. Another open problem is warning the driver, \ie what modality, visual, auditory, tactile, or a combination thereof, to use to effectively direct drivers' gaze to the appropriate regions of the scene without compromising their attention to other driving tasks.

\section{Discussion and conclusions}
\label{ch:discussion_conclusions}
This report provides a comprehensive analysis of the past decade of research on the topic of attention allocation during driving and links theoretical and behavioral findings to practical implementations. Here we will summarize some of the remaining issues that should be addressed moving forward.

\subsection{Remaining challenges in behavioral research}

\vspace{0.5em}
\noindent
\textbf{The limitations of using gaze as a proxy for attention.} Although gaze has been extensively used as a proxy for attention in driving, it has significant limitations that must be taken into account in the design of the experiments and analysis of the results. The utility of gaze is limited for visual processing beyond foveal vision and for covert changes of attention, thus the bulk of the literature is dedicated to central vision. Only a few experimental and theoretical works consider the role of periphery in driving which remains an open research problem. Likewise, decision-making processes are difficult to infer from gaze since it is a product of interaction between data- and task-driven factors whose relative contributions and interaction is an active area of research. 

\vspace{0.5em}
\noindent
\textbf{Lack of multi-modal sensory perception.} In addition to the visual signal, the driver receives sensory information from other modalities, including sound (\eg noise, honks) and tactile feedback (wheel traction, speed bumps). While it is universally agreed that vision is the predominant source of information for driving, the contribution of the other sensory modalities and their effect on attention is understudied.

\vspace{0.5em}
\noindent
\textbf{Inconsistent definitions and procedures.} The definitions of the basic attention measures, data recording, and analysis procedures in the literature are often inconsistent or under-specified. Besides, some of the commonly used terms, such as \textit{forward roadway}, \textit{novice} or \textit{older} driver, have no formal definitions. As a result, gathered evidence cannot be aggregated across studies, and conclusions regarding some of the factors are contradictory. Meta-reviews conclude the same and show that only a fraction of papers uses comparable experimental setups and approaches to statistical analysis of data. To resolve this issue, effects of various definitions and thresholds on the outcomes of the studies should be provided. Note, however, that even the most precise gaze recording in ecologically valid conditions does not reduce the fundamental limitations of gaze as a tool for studying human attention.

\vspace{0.5em}
\noindent
\textbf{Ecological validity and lack of validation studies.} The majority of the behavioral works are conducted in low- to mid-fidelity driving simulators rarely validated against on-road data. Several studies indicate significant changes in attention allocation caused by simulators, especially when subjects are not actively controlling the vehicle, thus conclusions and models based on such data may not apply in real-world conditions.

\vspace{0.5em}
\noindent
\textbf{Results are rarely tied to safety.} Even though most of the research on driving and attention is motivated by safety, the results are rarely tied directly to crash risk. Currently, only the naturalistic driving studies provide accident statistics with relevant attention data but for limited scenarios, \eg rear-end collisions. Accident data from simulated and closed track studies is difficult to aggregate due to the specificity of the scenarios and validation issues.

\vspace{0.5em}
\noindent
\textbf{Focus on correlations.} The bulk of behavioral literature establishes correlations between measures of attention and various factors. Conclusions that indicate safety concerns may be used to inform policy-making. Relatively few works attempt to infer internal decision-making processes or attentional modulation that lead to the observed gaze distribution and subsequent motor actions. Existing analytical models either predict aggregate statistics of drivers' gaze or consider dynamic gaze changes in very limited driving scenarios. 

\vspace{0.5em}
\noindent
\textbf{Uneven coverage of factors that affect attention.} Factors related to environment, demographics, driving experience, and certain types of distractions are understudied. First, conditions during most studies are optimal for driving. Environmental factors such as bad weather, unfamiliar environment, or heavy traffic are avoided in on-road studies and rarely modeled in simulation. Some common in-vehicle distractions such as passengers requiring attention (small children or pets) or driver's emotional state (distress or excitement) are not well-studied. Some cognitive distractions such as mind-wondering or anxiety are difficult to study directly unlike visual-manual distractions that attract a lot of research attention. It is unclear whether artificial tasks that induce such states (\eg n-back or sound counting) are adequate substitutes. Finally, the diversity of the subject pool in terms of age, gender, driving experience, and socio-economic factors is lacking in many studies.

\vspace{0.5em}
\noindent
\textbf{Fragmented nature of behavioral research.} A large number of disciplines are involved in studying attention and driving, \eg psychology, transportation, human factors, and ergonomics, each with their own goals, approaches, and terminology. Within a single discipline, the majority of works are narrowly focused on one or two factors considered in isolation. Including a wider array of factors and conditions will enable direct comparisons between them and lead to more robust conclusions. Furthermore, increasing interdisciplinary connections is necessary to achieve a unified understanding of drivers' attention allocation.

\vspace{0.5em}
\noindent
\textbf{Research may become obsolete} as driver education, infrastructure design, vehicle, and consumer electronics technology continuously improve and cause shifts in driver behavior. For example, touchscreens have recently become the primary input method and should be reflected in the existing benchmarks for secondary tasks. These changes should be taken into account, particularly when dealing with data from naturalistic studies that take a significant amount of time to record and process. For example, the data from 100-Car NDS recorded in 2004 may not reflect drivers' behaviors a decade and a half later.

\subsection{Remaining challenges in applied research}

\vspace{0.5em}
\noindent
\textbf{Lack of publicly available data and models.} There are few sufficiently large and varied public datasets for model evaluation and benchmarking. As a result, nearly two-thirds of the models are based on data with unknown properties. Furthermore, code for the statistical analysis and implemented models is not available for more than 90\% of the works. As a result, it is impossible to ensure the reproducibility of the results and directly compare the effectiveness of different approaches.

\vspace{0.5em}
\noindent
\textbf{Driving as an active problem.} All available datasets consist of pre-recorded driving footage accompanied by gaze information (driver's gaze and gaze of passive observers) or manual annotations. Such data provides limited ability to estimate the changes in drivers' gaze depending on the task, \eg where one would look when turning left or right in the same conditions. Counterfactual studies allow testing how the presence or absence of some objects affects attention allocation. However, it is virtually impossible to simulate the effect of the drivers' actions on other road users using pre-recorded data.

\vspace{0.5em}
\noindent
\textbf{A disconnect between behavioral and practical works.} Practical works focus mainly on driver monitoring and assistive applications as well as automating vehicle control. Despite encouraging results, most algorithms consider only a fraction of the factors identified in behavioral studies; for instance, driving experience and the effects of the environment are rarely taken into account. Data collection procedures for datasets do not always follow the best practices established in the behavioral literature regarding conditions and equipment used for data recording, statistical analysis, and validation of the results.

\vspace{0.5em}
\noindent
\textbf{Challenges in implementing task-based attention.} It is generally acknowledged that gaze allocation during driving is affected more by the top-down factors rather than the saliency of the objects in the scene. However, the majority of current approaches focus on capturing implicit dependencies between visual data and drivers' gaze and vehicle control signals without explicit task representation.

\vspace{0.5em}
\noindent
\textbf{Limited functions of attention.} Existing implementations focus on the selective and explanatory functions of attention, \ie highlighting and ranking objects or areas in the scenes given the current task and tying them to the output of the algorithm. Given the safety-critical nature of driving, the speed and transparency of the models are highly desirable. Presently, interpretations of attentional weights remain qualitative and somewhat speculative in nature. Thus, the development of more reliable quantitative methods for assessment is needed. Other aspects of attention, such as its sequential nature, relation to memory, decision-making, and allocation of cognitive resources, are not considered. Part of the reason is the limited scope of the proposed models. As the model complexity increases towards incorporating analysis of the relationships between road users, infrastructure and drivers' actions, more sophisticated attention mechanisms will be necessary. 

\vspace{0.5em}
\noindent
\textbf{Lack of evaluation with focus on safety.} Similar to behavioral research, most assistive and autonomous driving applications are motivated by safety concerns, however, quantitative evaluations of generated attention distributions can only assess how well they align with human gaze or manual annotations. Currently, two approaches are taken towards solving this issue in the literature. One is modeling eye-tracking data for accident videos. Another is integrating attention into vehicle control models and testing them in simulation. Accident datasets provide information about various types of collisions and their timelines, but without the gaze data of the original driver and active control it is impossible to predict whether visual strategies of the passive observers could have prevented or reduced the severity of the crash. Simulated experiments provide both the active control and the ability to replicate the same scenarios but are not validated in on-road conditions.

\vspace{0.5em}
\noindent
\textbf{Practical viability of the proposed solutions is unknown}, partly due to the lack of real-world data for validation and partly because it is unclear how they could be integrated into existing ADAS. For instance, both the mode and contents of the issued warnings are highly context dependent. Currently available ADAS are specialized for certain functions (\eg lane departure, forward collision) and rely on a combination of audio-tactile stimulation to warn the drivers, which may not be appropriate for complex situations. Extending the ADAS functionality from warning the driver to a more cooperative role would require new solutions for adapting to drivers' attention and directing it to relevant elements of the scene without interfering with the driving task.

\vspace{0.5em}

In conclusion, studying human attention in a safety-critical activity such as driving is an important and challenging problem. This report brings together theoretical findings and practical results from various related fields, such as human factors, transportation, psychology, human-computer interaction, and computer vision. This is particularly important given the fractured nature of research on drivers’ attention and may provide helpful cross-disciplinary links to researchers with different backgrounds. An extensive overview of the limitations of gaze as a proxy of attention, inconsistencies in data collection, processing, analysis, and modeling, as well as a summary of open research problems, may inform future research.

\section*{Acknowledgements}
This research was supported by grants to the senior author (JKT) from the following sources: Air Force Office of Scientific Research USA (FA9550-18-1-0054), The Canada Research Chairs Program (950- 231659)  and Natural Sciences and Engineering Research Council of Canada  (RGPIN-2016-05352). The funders had no role in study design, data collection and analysis, decision to publish, or preparation of the manuscript.

\FloatBarrier
{\small
\bibliographystyle{ieee}
\bibliography{behavioral,practical,surveys,other}

\begin{thebibliography}{100}\itemsep=-1pt

\bibitem{2018_Traffic_Safety}
{2018 Traffic Safety Culture Index}.
\newblock
  \url{https://aaafoundation.org/wp-content/uploads/2019/06/2018-TSCI-FINAL-061819_updated.pdf}.
\newblock Accessed: 2020-11-28.

\bibitem{2014_Research_Brief}
{AAA Foundation for Traffic Safety. Rates of Motor Vehicle Crashes, Injuries,
  and Deaths in Relation to Driver Age, United States, 2014 – 2015}.
\newblock
  \url{http://aaafoundation.org/wp-content/uploads/2017/11/CrashesInjuriesDeathsInRelationToAge2014-2015Brief.pdf}.
\newblock Accessed: 2020-11-28.

\bibitem{2017_Household_Travel_Survey}
{Bureau of Transportation Statistics. National Household Travel Survey Daily
  Travel Quick Facts}.
\newblock
  \url{https://www.bts.gov/statistical-products/surveys/national-household-travel-survey-daily-travel-quick-facts}.
\newblock Accessed: 2020-11-28.

\bibitem{2019_Road_Safety}
{European Comission. Road safety: Commission welcomes agreement on new EU rules
  to help save lives}.
\newblock
  \url{https://ec.europa.eu/commission/presscorner/detail/en/IP_19_1793}.
\newblock Accessed: 2020-11-28.

\bibitem{2020_EU_Road_Safety}
{European Commission. Road safety. Distraction.}
\newblock
  \url{https://ec.europa.eu/transport/road_safety/topics/behaviour/distraction_en}.
\newblock Accessed: 2020-11-28.

\bibitem{2015_Distraction_Study}
{European Commission. sStudy on good practices for reducing road safety risks
  caused by road user distractions}.
\newblock
  \url{https://ec.europa.eu/transport/road_safety/sites/roadsafety/files/pdf/behavior/distraction_study.pdf}.
\newblock Accessed: 2020-11-28.

\bibitem{2017_Highway_Statistics}
{Federal Highway Administration. Highway Statistics 2017. Distribution of
  Licensed Drivers by Sex and Percentage in each Age Group and Relation to
  Population}.
\newblock
  \url{https://www.fhwa.dot.gov/policyinformation/statistics/2017/dl20.cfm}.
\newblock Accessed: 2020-11-28.

\bibitem{2020_FHWA_Intersection_Safety}
{Federal Highway Administration. Intersection Safety}.
\newblock
  \url{https://ec.europa.eu/transport/road_safety/topics/behaviour/distraction_en}.
\newblock Accessed: 2020-11-28.

\bibitem{2018_Fatality_Facts}
{Insurance Institute for Highway Safety. Fatality Facts 2018. Gender}.
\newblock \url{https://www.iihs.org/topics/fatality-statistics/detail/gender}.
\newblock Accessed: 2020-11-28.

\bibitem{ISO_14198}
{ISO/TS 14198 (11.2012). Road vehicles-Ergonomic aspects of transport
  information and control systems-Calibration tasks for methods which assess
  driver demand due to the use of in-vehicle systems: ISO International
  Organization for Standardization}.

\bibitem{2017_Japan_Driver_Statistics}
{National Police Agency. Statistics 5-14 Number of driver's license holders by
  age group and gender (2017)}.
\newblock \url{https://www.npa.go.jp/hakusyo/h30/toukei/05/14.xlsx}.
\newblock Accessed: 2020-11-28.

\bibitem{2017_Eye_Link_Specs}
{SR Research. EyeLink 1000 Plus Technical Specifications }.
\newblock
  \url{https://www.sr-research.com/wp-content/uploads/2017/11/eyelink-1000-plus-specifications.pdf}.
\newblock Accessed: 2020-11-28.

\bibitem{2016_Canadian_Collision_Stats}
{Transport Canada. Canadian Motor Vehicle Traffic Collision Statistics: 2016}.
\newblock
  \url{https://tc.canada.ca/en/canadian-motor-vehicle-traffic-collision-statistics-2016}.
\newblock Accessed: 2020-11-28.

\bibitem{2000_Traffic_Collision_Statistics}
{Transport Canada. Canadian Motor VehicleTraffic Collision Statistics (2000)}.
\newblock
  \url{https://ccmta.ca/images/publications/pdf//collision_stats00_e.pdf}.
\newblock Accessed: 2020-11-28.

\bibitem{2000_Udacity}
{Udacity. Public Driving Dataset}.
\newblock \url{https://www.udacity.com/self-driving-car}.
\newblock Accessed: 2020-12-01.

\bibitem{2011_Statistics_Canada}
{Statistics Canada. Commuting to work}.
\newblock
  \url{https://www12.statcan.gc.ca/nhs-enm/2011/as-sa/99-012-x/99-012-x2011003_1-eng.cfm},
  2011.
\newblock Accessed: 2020-11-28.

\bibitem{2012_NHTSA_Driver_Distraction}
{National Highway Traffic Safety Administration. Visual-manual NHTSA Driver
  Distraction Guidelines for In-vehicle Devices. Docket No NHTSA-2010-0053}.
\newblock
  \url{https://www.nhtsa.gov/sites/nhtsa.dot.gov/files/distraction_npfg-02162012.pdf},
  2012.
\newblock Accessed: 2020-10-19.

\bibitem{2013_TR_NHTSA}
{National Highway Traffic Safety Administration. Visual-manual NHTSA driver
  distraction guidelines for in-vehicle electronic devices}.
\newblock 2012.
\newblock Accessed: 28-11-2020.

\bibitem{2015_EU_Traffic_Safety}
{European Commission. Traffic Safety Basic Facts}.
\newblock
  \url{https://ec.europa.eu/transport/road_safety/sites/roadsafety/files/pdf/statistics/dacota/bfs2015_motomoped.pdf},
  2015.
\newblock Accessed: 2020-11-28.

\bibitem{2020_ISO}
{ISO 15007. Road vehicles -- Measurement of driver visual behaviour with
  respect to transport information and control systems}, 2020.

\bibitem{2014_ACM_Abtahi}
S.~Abtahi, M.~Omidyeganeh, S.~Shirmohammadi, and B.~Hariri.
\newblock {YawDD: A yawning detection dataset}.
\newblock In {\em Proceedings of the ACM Multimedia Systems Conference}, pages
  24--28, 2014.

\bibitem{2016_SigProcMag_Aghaei}
A.~S. Aghaei, B.~Donmez, C.~C. Liu, D.~He, G.~Liu, K.~N. Plataniotis, H.-Y.~W.
  Chen, and Z.~Sojoudi.
\newblock {Smart driver monitoring: when signal processing meets human factors:
  In the driver's seat}.
\newblock {\em IEEE Signal Processing Magazine}, 33(6):35--48, 2016.

\bibitem{2017_AppliedErgonomics_Ahlstrom}
C.~Ahlstrom and K.~Kircher.
\newblock {Changes in glance behaviour when using a visual eco-driving
  system--A field study}.
\newblock {\em {Applied Ergonomics}}, 58:414--423, 2017.

\bibitem{2013_TITS_Ahlstrom}
C.~Ahlstrom, K.~Kircher, and A.~Kircher.
\newblock A gaze-based driver distraction warning system and its effect on
  visual behavior.
\newblock {\em IEEE Transactions on Intelligent Transportation Systems},
  14(2):965--973, 2013.

\bibitem{2019_IV_Akai}
N.~Akai, T.~Hirayama, L.~Y. Morales, Y.~Akagi, H.~Liu, and H.~Murase.
\newblock {Driving behavior modeling based on hidden Markov models with
  driver's eye-gaze measurement and ego-vehicle localization}.
\newblock In {\em Proceedings of the IEEE Intelligent Vehicles Symposium (IV)},
  pages 949--956. IEEE, 2019.

\bibitem{1999_IJN_Akerstedt}
T.~{\AA}kerstedt and M.~Gillberg.
\newblock Subjective and objective sleepiness in the active individual.
\newblock {\em International Journal of Neuroscience}, 52(1-2):29--37, 1990.

\bibitem{2014_TR_Alberti}
C.~F. Alberti, A.~Shahar, and D.~Crundall.
\newblock Are experienced drivers more likely than novice drivers to benefit
  from driving simulations with a wide field of view?
\newblock {\em {Transportation Research Part F: Traffic Psychology and
  Behaviour}}, 27:124--132, 2014.

\bibitem{2016_CVPRW_Alletto}
S.~Alletto, A.~Palazzi, F.~Solera, S.~Calderara, and R.~Cucchiara.
\newblock Dr (eye) ve: a dataset for attention-based tasks with applications to
  autonomous and assisted driving.
\newblock In {\em Proceedings of the Conference on Computer Vision and Pattern
  Recognition Workshops (CVPR)}, pages 54--60, 2016.

\bibitem{1990_Perception_Psychophysics_Andersen}
G.~J. Andersen.
\newblock Focused attention in three-dimensional space.
\newblock {\em {Perception \& Psychophysics}}, 47(2):112--120, 1990.

\bibitem{2011_AccidentAnalysis_Andersen}
G.~J. Andersen, R.~Ni, Z.~Bian, and J.~Kang.
\newblock Limits of spatial attention in three-dimensional space and dual-task
  driving performance.
\newblock {\em {Accident Analysis \& Prevention}}, 43(1):381--390, 2011.

\bibitem{2004_PsychReview_Anderson}
J.~R. Anderson, D.~Bothell, M.~D. Byrne, S.~Douglass, C.~Lebiere, and Y.~Qin.
\newblock {An integrated theory of the mind}.
\newblock {\em Psychological review}, 111(4):1036, 2004.

\bibitem{2013_TR_Angell}
L.~Angell, M.~Perez, and W.~R. Garrott.
\newblock {Explanatory Material About the Definition of a Task Used in NHTSA's
  Driver Distraction Guidelines, and Task Examples}.
\newblock Technical report, {National Highway Traffic Safety Administration},
  2013.

\bibitem{2013_TrendsCognSci_Awh}
E.~Awh, A.~V. Belopolsky, and J.~Theeuwes.
\newblock {Top-down versus bottom-up attentional control: A failed theoretical
  dichotomy}.
\newblock {\em Trends in Cognitive Sciences}, 16(8):437--443, 2012.

\bibitem{2020_ESA_Badue}
C.~Badue, R.~Guidolini, R.~V. Carneiro, P.~Azevedo, V.~B. Cardoso, A.~Forechi,
  L.~Jesus, R.~Berriel, T.~M. Paix{\~a}o, F.~Mutz, et~al.
\newblock {Self-driving cars: A survey}.
\newblock {\em {Expert Systems with Applications}}, page 113816, 2020.

\bibitem{2019_arXiv_Baee}
S.~Baee, E.~Pakdamanian, V.~O. Roman, I.~Kim, L.~Feng, and L.~Barnes.
\newblock Eyecar: Modeling the visual attention allocation of drivers in
  semi-autonomous vehicles.
\newblock {\em arXiv preprint arXiv:1912.07773}, 2019.

\bibitem{2016_iPerception_Baldwin}
J.~Baldwin, A.~Burleigh, R.~Pepperell, and N.~Ruta.
\newblock The perceived size and shape of objects in peripheral vision.
\newblock {\em i-Perception}, 7(4):2041669516661900, 2016.

\bibitem{1991_HumanFactors_Ball}
K.~Ball and C.~Owsley.
\newblock Identifying correlates of accident involvement for the older driver.
\newblock {\em Human factors}, 33(5):583--595, 1991.

\bibitem{1988_JOpnSocAm_Ball}
K.~K. Ball, B.~L. Beard, D.~L. Roenker, R.~L. Miller, and D.~S. Griggs.
\newblock {Age and visual search: Expanding the useful field of view}.
\newblock {\em {Journal of Optical Society of America}}, 5(12):2210--2219,
  1988.

\bibitem{2006_JAGS_Ball}
K.~K. Ball, D.~L. Roenker, V.~G. Wadley, J.~D. Edwards, D.~L. Roth,
  G.~McGwin~Jr, R.~Raleigh, J.~J. Joyce, G.~M. Cissell, and T.~Dube.
\newblock Can high-risk older drivers be identified through performance-based
  measures in a department of motor vehicles setting?
\newblock {\em {Journal of the American Geriatrics Society}}, 54(1):77--84,
  2006.

\bibitem{2009_VisualCognition_Ballard}
D.~H. Ballard and M.~M. Hayhoe.
\newblock Modelling the role of task in the control of gaze.
\newblock {\em Visual cognition}, 17(6-7):1185--1204, 2009.

\bibitem{1991_JOSA_Banks}
M.~S. Banks, A.~B. Sekuler, and S.~J. Anderson.
\newblock Peripheral spatial vision: Limits imposed by optics, photoreceptors,
  and receptor pooling.
\newblock {\em Journal of the Optical Society of America A}, 8(11):1775--1787,
  1991.

\bibitem{2013_IV_Bar}
T.~B{\"a}r, D.~Linke, D.~Nienh{\"u}ser, and J.~M. Z{\"o}llner.
\newblock Seen and missed traffic objects: A traffic object-specific awareness
  estimation.
\newblock In {\em Proceedings of the IEEE Intelligent Vehicles Symposium (IV)},
  pages 31--36. IEEE, 2013.

\bibitem{2015_TransRes_Bargman}
J.~B{\"a}rgman, V.~Lisovskaja, T.~Victor, C.~Flannagan, and M.~Dozza.
\newblock {How does glance behavior influence crash and injury risk? A
  ‘what-if’counterfactual simulation using crashes and near-crashes from
  SHRP2}.
\newblock {\em {Transportation Research Part F: Traffic Psychology and
  Behaviour}}, 35:152--169, 2015.

\bibitem{2017_AccidentAnalysis_Beanland}
V.~Beanland, A.~J. Filtness, and R.~Jeans.
\newblock {Change detection in urban and rural driving scenes: Effects of
  target type and safety relevance on change blindness}.
\newblock {\em {Accident Analysis \& Prevention}}, 100:111--122, 2017.

\bibitem{2019_HumanFactorsErgonomics_Beanland}
V.~Beanland and R.~A. Wynne.
\newblock {Does familiarity breed competence or contempt? Effects of driver
  experience, road type and familiarity on hazard perception}.
\newblock In {\em {Proceedings of the Human Factors and Ergonomics Society
  Annual Meeting}}, volume~63, pages 2006--2010, 2019.

\bibitem{2010_AccidentAnalysis_Belanger}
A.~B{\'e}langer, S.~Gagnon, and S.~Yamin.
\newblock {Capturing the serial nature of older drivers’ responses towards
  challenging events: A simulator study}.
\newblock {\em {Accident Analysis \& Prevention}}, 42(3):809--817, 2010.

\bibitem{2016_AccidentAnalysis_Belyusar}
D.~Belyusar, B.~Reimer, B.~Mehler, and J.~F. Coughlin.
\newblock A field study on the effects of digital billboards on glance behavior
  during highway driving.
\newblock {\em {Accident Analysis \& Prevention}}, 88:88--96, 2016.

\bibitem{2013_TransRes_Benedetto}
S.~Benedetto, M.~Pedrotti, R.~Bremond, and T.~Baccino.
\newblock Leftward attentional bias in a simulated driving task.
\newblock {\em {Transportation Research Part F: Traffic Psychology and
  Behaviour}}, 20:147--153, 2013.

\bibitem{2005_SSJ_Bergdahl}
J.~Bergdahl.
\newblock Sex differences in attitudes toward driving: A survey.
\newblock {\em Social Science Journal}, 42(4):595--601, 2005.

\bibitem{2010_TRR_Bian}
Z.~Bian, J.~J. Kang, and G.~J. Andersen.
\newblock Changes in extent of spatial attention with increased workload in
  dual-task driving.
\newblock {\em {Transportation Research Record}}, 2185(1):8--14, 2010.

\bibitem{2014_TransRes_Birrell}
S.~A. Birrell and M.~Fowkes.
\newblock {Glance behaviours when using an in-vehicle smart driving aid: A
  real-world, on-road driving study}.
\newblock {\em {Transportation Research Part F: Traffic Psychology and
  Behaviour}}, 22:113--125, 2014.

\bibitem{2016_arXiv_Bojarski}
M.~Bojarski, D.~Del~Testa, D.~Dworakowski, B.~Firner, B.~Flepp, P.~Goyal, L.~D.
  Jackel, M.~Monfort, U.~Muller, J.~Zhang, et~al.
\newblock End to end learning for self-driving cars.
\newblock {\em {arXiv preprint arXiv:1604.07316}}, 2016.

\bibitem{2017_arXiv_Bojarski}
M.~Bojarski, P.~Yeres, A.~Choromanska, K.~Choromanski, B.~Firner, L.~Jackel,
  and U.~Muller.
\newblock Explaining how a deep neural network trained with end-to-end learning
  steers a car.
\newblock {\em arXiv preprint arXiv:1704.07911}, 2017.

\bibitem{2011_BMVC_Borji}
A.~Borji, D.~N. Sihite, and L.~Itti.
\newblock Computational modeling of top-down visual attention in interactive
  environments.
\newblock In {\em Proceedings of the British Machine Vision Conference (BMVC)},
  volume~85, pages 1--12, 2011.

\bibitem{2012_CVPR_Borji}
A.~Borji, D.~N. Sihite, and L.~Itti.
\newblock Probabilistic learning of task-specific visual attention.
\newblock In {\em Proceedings of the IEEE Conference on Computer Vision and
  Pattern Recognition (CVPR)}, pages 470--477. IEEE, 2012.

\bibitem{2014_TransSysManCybernetics_Borji}
A.~Borji, D.~N. Sihite, and L.~Itti.
\newblock {What/where to look next? Modeling top-down visual attention in
  complex interactive environments}.
\newblock {\em IEEE Transactions on Systems, Man, and Cybernetics: Systems},
  44(5):523--538, 2013.

\bibitem{2016_AutoUI_Borojeni}
S.~S. Borojeni, L.~Chuang, W.~Heuten, and S.~Boll.
\newblock Assisting drivers with ambient take-over requests in highly automated
  driving.
\newblock In {\em Proceedings of the 8th International Conference on Automotive
  User Interfaces and Interactive Vehicular Applications}, pages 237--244,
  2016.

\bibitem{2015_TrafficInjuryPrevention_Borowsky}
A.~Borowsky, W.~J. Horrey, Y.~Liang, A.~Garabet, L.~Simmons, and D.~L. Fisher.
\newblock The effects of momentary visual disruption on hazard anticipation and
  awareness in driving.
\newblock {\em Traffic injury prevention}, 16(2):133--139, 2015.

\bibitem{2013_AccidentAnalysis_Borowsky}
A.~Borowsky and T.~Oron-Gilad.
\newblock Exploring the effects of driving experience on hazard awareness and
  risk perception via real-time hazard identification, hazard classification,
  and rating tasks.
\newblock {\em {Accident Analysis \& Prevention}}, 59:548--565, 2013.

\bibitem{2012_AccidentAnalysis_Borowsky}
A.~Borowsky, T.~Oron-Gilad, A.~Meir, and Y.~Parmet.
\newblock {Drivers’ perception of vulnerable road users: A hazard perception
  approach}.
\newblock {\em {Accident Analysis \& Prevention}}, 44(1):160--166, 2012.

\bibitem{2010_AccidentAnalysis_Borowsky}
A.~Borowsky, D.~Shinar, and T.~Oron-Gilad.
\newblock Age, skill, and hazard perception in driving.
\newblock {\em {Accident Analysis \& Prevention}}, 42(4):1240--1249, 2010.

\bibitem{2008_HumanFactors_Borowsky}
A.~Borowsky, D.~Shinar, and Y.~Parmet.
\newblock The relation between driving experience and recognition of road signs
  relative to their locations.
\newblock {\em Human factors}, 50(2):173--182, 2008.

\bibitem{2016_JoV_Boucart}
M.~Boucart, Q.~Lenoble, J.~Quettelart, S.~Szaffarczyk, P.~Despretz, and S.~J.
  Thorpe.
\newblock Finding faces, animals, and vehicles in far peripheral vision.
\newblock {\em Journal of vision}, 16(2):10--10, 2016.

\bibitem{2019_SAP_Bozkir}
E.~Bozkir, D.~Geisler, and E.~Kasneci.
\newblock {Assessment of driver attention during a safety critical situation in
  VR to generate VR-based training}.
\newblock In {\em {Proceedings of the ACM Symposium on Applied Perception}},
  pages 1--5, 2019.

\bibitem{2011_TR_Briggs}
G.~F. Briggs, G.~J. Hole, and M.~F. Land.
\newblock Emotionally involving telephone conversations lead to driver error
  and visual tunnelling.
\newblock {\em {Transportation Research Part F: Traffic Psychology and
  Behaviour}}, 14(4):313--323, 2011.

\bibitem{2016_AccidentAnalysis_Burdett}
B.~R. Burdett, S.~G. Charlton, and N.~J. Starkey.
\newblock {Not all minds wander equally: The influence of traits, states and
  road environment factors on self-reported mind wandering during everyday
  driving}.
\newblock {\em {Accident Analysis \& Prevention}}, 95:1--7, 2016.

\bibitem{1958_QJEP_Bursill}
A.~Bursill.
\newblock The restriction of peripheral vision during exposure to hot and humid
  conditions.
\newblock {\em Quarterly Journal of Experimental Psychology}, 10(3):113--129,
  1958.

\bibitem{2018_PAMI_Bylinskii}
Z.~Bylinskii, T.~Judd, A.~Oliva, A.~Torralba, and F.~Durand.
\newblock What do different evaluation metrics tell us about saliency models?
\newblock {\em {IEEE Transactions on Pattern Analysis and Machine
  Intelligence}}, 41(3):740--757, 2018.

\bibitem{2011_Handbook_Caird}
J.~K. Caird and W.~J. Horrey.
\newblock Twelve practical and useful questions about driving simulation.
\newblock {\em Handbook of driving simulation for engineering, medicine, and
  psychology}, pages 5--1, 2011.

\bibitem{2018_HumanFactors_Caird}
J.~K. Caird, S.~M. Simmons, K.~Wiley, K.~A. Johnston, and W.~J. Horrey.
\newblock {Does talking on a cell phone, with a passenger, or dialing affect
  driving performance? An updated systematic review and meta-analysis of
  experimental studies}.
\newblock {\em Human Factors}, 60(1):101--133, 2018.

\bibitem{2016_ACCV_Chan}
F.-H. Chan, Y.-T. Chen, Y.~Xiang, and M.~Sun.
\newblock Anticipating accidents in dashcam videos.
\newblock In {\em Proceedings of the Asian Conference on Computer Vision
  (ACCV)}, pages 136--153. Springer, 2016.

\bibitem{1998_Perception_Chapman}
P.~R. Chapman and G.~Underwood.
\newblock {Visual search of driving situations: Danger and experience}.
\newblock {\em Perception}, 27(8):951--964, 1998.

\bibitem{2013_TrafficPsychology_Charlton}
S.~G. Charlton and N.~J. Starkey.
\newblock {Driving on familiar roads: Automaticity and inattention blindness}.
\newblock {\em {Transportation Research Part F: Traffic Psychology and
  Behaviour}}, 19:121--133, 2013.

\bibitem{2019_TransRes_Chen}
W.~Chen, X.~Zhuang, Z.~Cui, and G.~Ma.
\newblock {Drivers’ recognition of pedestrian road-crossing intentions:
  Performance and process}.
\newblock {\em {Transportation Research Part F: Traffic Psychology and
  Behaviour}}, 64:552--564, 2019.

\bibitem{2019_IROSW_Yuying}
Y.~Chen, C.~Liu, L.~Tai, M.~Liu, and B.~E. Shi.
\newblock Gaze training by modulated dropout improves imitation learning.
\newblock In {\em Proceedings of the International Conference on Intelligent
  Robots and Systems Workshop (IROS)}, 2019.

\bibitem{2016_PLOS_Cheng}
Y.~Cheng, L.~Gao, Y.~Zhao, and F.~Du.
\newblock Drivers’ visual characteristics when merging onto or exiting an
  urban expressway.
\newblock {\em PloS one}, 11(9):e0162298, 2016.

\bibitem{2019_TITS_Chiou}
C.-Y. Chiou, W.-C. Wang, S.-C. Lu, C.-R. Huang, P.-C. Chung, and Y.-Y. Lai.
\newblock Driver monitoring using sparse representation with part-based
  temporal face descriptors.
\newblock {\em IEEE Transactions on Intelligent Transportation Systems},
  21(1):346--361, 2019.

\bibitem{2016_BigComp_Choi}
I.-H. Choi, S.~K. Hong, and Y.-G. Kim.
\newblock Real-time categorization of driver's gaze zone using the deep
  learning techniques.
\newblock In {\em 2016 International Conference on Big Data and Smart Computing
  (BigComp)}, pages 143--148. IEEE, 2016.

\bibitem{2016_ApplSci_Choi}
I.-H. Choi, C.-H. Jeong, and Y.-G. Kim.
\newblock Tracking a driver’s face against extreme head poses and inference
  of drowsiness using a hidden markov model.
\newblock {\em Applied Sciences}, 6(5):137, 2016.

\bibitem{2014_TransRes_Ciceri}
M.~R. Ciceri and D.~Ruscio.
\newblock {Does driving experience in video games count? Hazard anticipation
  and visual exploration of male gamers as function of driving experience}.
\newblock {\em {Transportation Research Part F: Traffic Psychology and
  Behaviour}}, 22:76--85, 2014.

\bibitem{2017_AccidentAnalysis_Clark}
H.~Clark and J.~Feng.
\newblock Age differences in the takeover of vehicle control and engagement in
  non-driving-related activities in simulated driving with conditional
  automation.
\newblock {\em {Accident Analysis \& Prevention}}, 106:468--479, 2017.

\bibitem{2019_TransRes_Clark}
J.~R. Clark, N.~A. Stanton, and K.~M. Revell.
\newblock Directability, eye-gaze, and the usage of visual displays during an
  automated vehicle handover task.
\newblock {\em {Transportation Research Part F: Traffic Psychology and
  Behaviour}}, 67:29--42, 2019.

\bibitem{2005_OVS_Clay}
O.~J. Clay, V.~G. Wadley, J.~D. Edwards, D.~L. Roth, D.~L. Roenker, and K.~K.
  Ball.
\newblock Cumulative meta-analysis of the relationship between useful field of
  view and driving performance in older adults: Current and future
  implications.
\newblock {\em Optometry and vision science}, 82(8):724--731, 2005.

\bibitem{2011_Encyclopedia_Cohen}
R.~A. Cohen.
\newblock {\em Cortical Magnification}, pages 718--719.
\newblock Springer New York, 2011.

\bibitem{2010_Ergonomics_Collet1}
C.~Collet, A.~Guillot, and C.~Petit.
\newblock {Phoning while driving I: A review of epidemiological, psychological,
  behavioural and physiological studies}.
\newblock {\em Ergonomics}, 53(5):589--601, 2010.

\bibitem{2010_Ergonomics_Collet}
C.~Collet, A.~Guillot, and C.~Petit.
\newblock {Phoning while driving II: A review of driving conditions influence}.
\newblock {\em Ergonomics}, 53(5):602--616, 2010.

\bibitem{2010_HumanFactors_Cooper}
J.~M. Cooper, N.~Medeiros-Ward, and D.~L. Strayer.
\newblock The impact of eye movements and cognitive workload on lateral
  position variability in driving.
\newblock {\em {Human Factors}}, 55(5):1001--1014, 2013.

\bibitem{2019_AppliedErgonomics_Costa}
M.~Costa, L.~Bonetti, V.~Vignali, A.~Bichicchi, C.~Lantieri, and A.~Simone.
\newblock Driver's visual attention to different categories of roadside
  advertising signs.
\newblock {\em {Applied Ergonomics}}, 78:127--136, 2019.

\bibitem{2018_Ergonomics_Costa}
M.~Costa, L.~Bonetti, V.~Vignali, C.~Lantieri, and A.~Simone.
\newblock The role of peripheral vision in vertical road sign identification
  and discrimination.
\newblock {\em Ergonomics}, 61(12):1619--1634, 2018.

\bibitem{2014_TransRes_Costa}
M.~Costa, A.~Simone, V.~Vignali, C.~Lantieri, A.~Bucchi, and G.~Dondi.
\newblock Looking behavior for vertical road signs.
\newblock {\em {Transportation Research Part F: Traffic Psychology and
  Behaviour}}, 23:147--155, 2014.

\bibitem{2018_AppliedErgonomics_Costa}
M.~Costa, A.~Simone, V.~Vignali, C.~Lantieri, and N.~Palena.
\newblock Fixation distance and fixation duration to vertical road signs.
\newblock {\em {Applied Ergonomics}}, 69:48--57, 2018.

\bibitem{2012_AccidentAnalysis_Crundall_1}
D.~Crundall, P.~Chapman, S.~Trawley, L.~Collins, E.~Van~Loon, B.~Andrews, and
  G.~Underwood.
\newblock {Some hazards are more attractive than others: Drivers of varying
  experience respond differently to different types of hazard}.
\newblock {\em {Accident Analysis \& Prevention}}, 45:600--609, 2012.

\bibitem{2012_AccidentAnalysis_Crundall}
D.~Crundall, E.~Crundall, D.~Clarke, and A.~Shahar.
\newblock Why do car drivers fail to give way to motorcycles at t-junctions?
\newblock {\em {Accident Analysis \& Prevention}}, 44(1):88--96, 2012.

\bibitem{2008_TransRes_Crundall}
D.~Crundall, K.~Humphrey, and D.~Clarke.
\newblock Perception and appraisal of approaching motorcycles at junctions.
\newblock {\em {Transportation Research Part F: Traffic Psychology and
  Behaviour}}, 11(3):159--167, 2008.

\bibitem{1999_Perception_Crundall}
D.~Crundall, G.~Underwood, and P.~Chapman.
\newblock Driving experience and the functional field of view.
\newblock {\em Perception}, 28(9):1075--1087, 1999.

\bibitem{2002_ACP_Crundall}
D.~Crundall, G.~Underwood, and P.~Chapman.
\newblock Attending to the peripheral world while driving.
\newblock {\em {Applied Cognitive Psychology}}, 16(4):459--475, 2002.

\bibitem{2015_AccidentAnalysis_Cuenen}
A.~Cuenen, E.~M. Jongen, T.~Brijs, K.~Brijs, M.~Lutin, K.~Van~Vlierden, and
  G.~Wets.
\newblock Does attention capacity moderate the effect of driver distraction in
  older drivers?
\newblock {\em {Accident Analysis \& Prevention}}, 77:12--20, 2015.

\bibitem{2020_CVPRW_Cultrera}
L.~Cultrera, L.~Seidenari, F.~Becattini, P.~Pala, and A.~Del~Bimbo.
\newblock {Explaining Autonomous Driving by Learning End-to-End Visual
  Attention}.
\newblock In {\em Proceedings of the IEEE/CVF Conference on Computer Vision and
  Pattern Recognition Workshops (CVPRW)}, pages 340--341, 2020.

\bibitem{2017_IET_Cunningham}
M.~L. Cunningham and M.~A. Regan.
\newblock Driver distraction and inattention in the realm of automated driving.
\newblock {\em IET Intelligent Transport Systems}, 12(6):407--413, 2017.

\bibitem{1990_JCN_Curcio}
C.~A. Curcio, K.~R. Sloan, R.~E. Kalina, and A.~E. Hendrickson.
\newblock Human photoreceptor topography.
\newblock {\em Journal of Comparative Neurology}, 292(4):497--523, 1990.

\bibitem{2014_TransRes_Winter}
J.~C. De~Winter, R.~Happee, M.~H. Martens, and N.~A. Stanton.
\newblock {Effects of adaptive cruise control and highly automated driving on
  workload and situation awareness: A review of the empirical evidence}.
\newblock {\em Transportation Research Part F: Traffic Psychology and
  Behaviour}, 27:196--217, 2014.

\bibitem{2015_TrafficInjuryPrevention_Decker}
J.~S. Decker, S.~J. Stannard, B.~McManus, S.~M. Wittig, V.~P. Sisiopiku, and
  D.~Stavrinos.
\newblock {The impact of billboards on driver visual behavior: A systematic
  literature review}.
\newblock {\em Traffic Injury Prevention}, 16(3):234--239, 2015.

\bibitem{1992_ECCV_DeMenthon}
D.~F. DeMenthon and L.~S. Davis.
\newblock Model-based object pose in 25 lines of code.
\newblock In {\em {Proceedings of the European Conference on Computer Cision}},
  pages 335--343. Springer, 1992.

\bibitem{2018_TITS_Deng}
T.~Deng, H.~Yan, and Y.-J. Li.
\newblock Learning to boost bottom-up fixation prediction in driving
  environments via random forest.
\newblock {\em IEEE Transactions on Intelligent Transportation Systems},
  19(9):3059--3067, 2017.

\bibitem{2020_TITS_Deng}
T.~Deng, H.~Yan, L.~Qin, T.~Ngo, and B.~Manjunath.
\newblock {How do drivers allocate their potential attention? Driving fixation
  prediction via convolutional neural networks}.
\newblock {\em IEEE Transactions on Intelligent Transportation Systems},
  21(5):2146--2154, 2019.

\bibitem{2016_TITS_Deng}
T.~Deng, K.~Yang, Y.~Li, and H.~Yan.
\newblock Where does the driver look? top-down-based saliency detection in a
  traffic driving environment.
\newblock {\em IEEE Transactions on Intelligent Transportation Systems},
  17(7):2051--2062, 2016.

\bibitem{2019_IEEEAccess_Deng}
W.~Deng and R.~Wu.
\newblock Real-time driver-drowsiness detection system using facial features.
\newblock {\em IEEE Access}, 7:118727--118738, 2019.

\bibitem{1994_TheHeartsEye_DerryBerry}
D.~Derryberry and D.~M. Tucker.
\newblock Motivating the focus of attention.
\newblock In P.~M. Niedenthal and S.~Kitayama, editors, {\em {The hearts eye:
  Emotional influence in perception and attention}}, pages 167--196. Academic
  Press, 1994.

\bibitem{1998_FHWA_Dinges}
D.~F. Dinges and R.~Grace.
\newblock {PERCLOS: A valid psychophysiological measure of alertness as
  assessed by psychomotor vigilance. FHWA-MCRT-98-006}.
\newblock 1998.

\bibitem{2006_TechRep_Dingus}
T.~A. Dingus, S.~G. Klauer, V.~L. Neale, A.~Petersen, S.~E. Lee, J.~Sudweeks,
  M.~A. Perez, J.~Hankey, D.~Ramsey, S.~Gupta, et~al.
\newblock {The 100-car naturalistic driving study, Phase II-results of the
  100-car field experiment}.
\newblock Technical report, United States. Department of Transportation.
  National Highway Traffic Safety Administration, 2006.

\bibitem{2012_TRR_Divekar}
G.~Divekar, A.~K. Pradhan, A.~Pollatsek, and D.~L. Fisher.
\newblock {Effect of external distractions: Behavior and vehicle control of
  novice and experienced drivers evaluated}.
\newblock {\em {Transportation Research Record}}, 2321(1):15--22, 2012.

\bibitem{2010_TransEng_Bonmez}
B.~Donmez, L.~N. Boyle, and J.~D. Lee.
\newblock {Differences in off-road glances: Effects on young drivers’
  performance}.
\newblock {\em Journal of Transportation Engineering}, 136(5):403--409, 2010.

\bibitem{2017_CRL_Dosovitskiy}
A.~Dosovitskiy, G.~Ros, F.~Codevilla, A.~Lopez, and V.~Koltun.
\newblock {CARLA: An Open Urban Driving Simulator}.
\newblock In {\em {Conference on Robot Learning}}, pages 1--16, 2017.

\bibitem{2020_IJITSR_Doudou}
M.~Doudou, A.~Bouabdallah, and V.~Berge-Cherfaoui.
\newblock {Driver Drowsiness Measurement Technologies: Current Research, Market
  Solutions, and Challenges}.
\newblock {\em International Journal of Intelligent Transportation Systems
  Research}, pages 1--23, 2019.

\bibitem{2013_AccidentAnalysis_Dozza}
M.~Dozza.
\newblock What factors influence drivers’ response time for evasive maneuvers
  in real traffic?
\newblock {\em {Accident Analysis \& Prevention}}, 58:299--308, 2013.

\bibitem{2013_TrafficInjuryPrevention_Dukic}
T.~Dukic, C.~Ahlstrom, C.~Patten, C.~Kettwich, and K.~Kircher.
\newblock Effects of electronic billboards on driver distraction.
\newblock {\em Traffic injury prevention}, 14(5):469--476, 2013.

\bibitem{2011_AppliedErgonomics_Edquist}
J.~Edquist, T.~Horberry, S.~Hosking, and I.~Johnston.
\newblock Effects of advertising billboards during simulated driving.
\newblock {\em {Applied Ergonomics}}, 42(4):619--626, 2011.

\bibitem{2019_TITS_Khatib}
A.~El~Khatib, C.~Ou, and F.~Karray.
\newblock {Driver Inattention Detection in the Context of Next-Generation
  Autonomous Vehicles Design: A Survey}.
\newblock {\em IEEE Transactions on Intelligent Transportation Systems}, 2019.

\bibitem{2013_TechReport_Engstrom}
J.~Engstr{\"o}m, C.~A. Monk, R.~Hanowski, W.~Horrey, J.~Lee, D.~McGehee,
  M.~Regan, A.~Stevens, E.~Traube, M.~Tuukkanen, et~al.
\newblock A conceptual framework and taxonomy for understanding and
  categorizing driver inattention.
\newblock Technical report, European Commission, 2013.

\bibitem{2015_TR_Eyraud}
R.~Eyraud, E.~Zibetti, and T.~Baccino.
\newblock Allocation of visual attention while driving with simulated augmented
  reality.
\newblock {\em {Transportation Research Part F: Traffic Psychology and
  Behaviour}}, 32:46--55, 2015.

\bibitem{2019_TMC_Fan}
X.~Fan, F.~Wang, D.~Song, Y.~Lu, and J.~Liu.
\newblock {GazMon: Eye Gazing Enabled Driving Behavior Monitoring and
  Prediction}.
\newblock {\em IEEE Transactions on Mobile Computing}, 2019.

\bibitem{2019_ITSC_Fang}
J.~Fang, D.~Yan, J.~Qiao, J.~Xue, H.~Wang, and S.~Li.
\newblock {DADA-2000: Can Driving Accident be Predicted by Driver Attentionƒ
  Analyzed by A Benchmark}.
\newblock In {\em Proceedings of the IEEE Intelligent Transportation Systems
  Conference (ITSC)}, pages 4303--4309. IEEE, 2019.

\bibitem{2020_Information_Feierle}
A.~Feierle, S.~Danner, S.~Steininger, and K.~Bengler.
\newblock {Information Needs and Visual Attention during Urban, Highly
  Automated Driving—An Investigation of Potential Influencing Factors}.
\newblock {\em Information}, 11(2):62, 2020.

\bibitem{2017_AdvErgonomics_Feldhutter}
A.~Feldh{\"u}tter, C.~Gold, S.~Schneider, and K.~Bengler.
\newblock How the duration of automated driving influences take-over
  performance and gaze behavior.
\newblock In {\em Advances in ergonomic design of systems, products and
  processes}, pages 309--318. Springer, 2017.

\bibitem{2018_AHAT_Feldhutter}
A.~Feldh{\"u}tter, N.~H{\"a}rtwig, C.~Kurpiers, J.~M. Hernandez, and
  K.~Bengler.
\newblock Effect on mode awareness when changing from conditionally to
  partially automated driving.
\newblock In {\em Congress of the International Ergonomics Association}, pages
  314--324. Springer, 2018.

\bibitem{2013_AJPH_Ferdinand}
A.~O. Ferdinand and N.~Menachemi.
\newblock {Associations between driving performance and engaging in secondary
  tasks: A systematic review}.
\newblock {\em American Journal of Public Health}, 104(3):e39--e48, 2014.

\bibitem{2016_TransRes_Fitzpatrick}
C.~D. Fitzpatrick, S.~Samuel, and M.~A. Knodler~Jr.
\newblock Evaluating the effect of vegetation and clear zone width on driver
  behavior using a driving simulator.
\newblock {\em {Transportation Research Part F: Traffic Psychology and
  Behaviour}}, 42:80--89, 2016.

\bibitem{2011_IET_Flores}
M.~J. Flores, J.~M. Armingol, and A.~de~la Escalera.
\newblock Driver drowsiness detection system under infrared illumination for an
  intelligent vehicle.
\newblock {\em IET Intelligent Transport Systems}, 5(4):241--251, 2011.

\bibitem{2016_IS_Fridman}
L.~Fridman, P.~Langhans, J.~Lee, and B.~Reimer.
\newblock Driver gaze region estimation without use of eye movement.
\newblock {\em IEEE Intelligent Systems}, 31(3):49--56, 2016.

\bibitem{2016_IET_Fridman}
L.~Fridman, J.~Lee, B.~Reimer, and T.~Victor.
\newblock ‘owl’and ‘lizard’: Patterns of head pose and eye pose in
  driver gaze classification.
\newblock {\em IET Computer Vision}, 10(4):308--314, 2016.

\bibitem{2017_CHI_Fridman}
L.~Fridman, H.~Toyoda, S.~Seaman, B.~Seppelt, L.~Angell, J.~Lee, B.~Mehler, and
  B.~Reimer.
\newblock What can be predicted from six seconds of driver glances?
\newblock In {\em Proceedings of the Conference on Human Factors in Computing
  Systems}, pages 2805--2813, 2017.

\bibitem{2010_IV_Friedrichs}
F.~Friedrichs and B.~Yang.
\newblock Camera-based drowsiness reference for driver state classification
  under real driving conditions.
\newblock In {\em Proceedings of the IEEE Intelligent Vehicles Symposium (IV)},
  pages 101--106. IEEE, 2010.

\bibitem{2016_HumanFactorsErgonomics_Funke}
G.~Funke, E.~Greenlee, M.~Carter, A.~Dukes, R.~Brown, and L.~Menke.
\newblock {Which eye tracker is right for your research? Performance evaluation
  of several cost variant eye trackers}.
\newblock In {\em Proceedings of the Human Factors and Ergonomics Society
  Annual Meeting}, volume~60, pages 1240--1244. SAGE Publications Sage CA: Los
  Angeles, CA, 2016.

\bibitem{2018_FrontiersPsychology_Gajewski}
P.~D. Gajewski, E.~Hanisch, M.~Falkenstein, S.~Th{\"o}nes, and E.~Wascher.
\newblock What does the n-back task measure as we get older? relations between
  working-memory measures and other cognitive functions across the lifespan.
\newblock {\em {Frontiers in Psychology}}, 9:2208, 2018.

\bibitem{2013_ACP_Garrison}
T.~M. Garrison and C.~C. Williams.
\newblock Impact of relevance and distraction on driving performance and visual
  attention in a simulated driving environment.
\newblock {\em Applied cognitive psychology}, 27(3):396--405, 2013.

\bibitem{2019_HumanFactors_Gaspar}
J.~Gaspar and C.~Carney.
\newblock The effect of partial automation on driver attention: a naturalistic
  driving study.
\newblock {\em {Human Factors}}, 61(8):1261--1276, 2019.

\bibitem{2016_HumanFactors_Gaspar}
J.~G. Gaspar, N.~Ward, M.~B. Neider, J.~Crowell, R.~Carbonari, H.~Kaczmarski,
  R.~V. Ringer, A.~P. Johnson, A.~F. Kramer, and L.~C. Loschky.
\newblock Measuring the useful field of view during simulated driving with
  gaze-contingent displays.
\newblock {\em {Human Factors}}, 58(4):630--641, 2016.

\bibitem{2013_CVPR_Geiger}
A.~Geiger, P.~Lenz, and R.~Urtasun.
\newblock Are we ready for autonomous driving? the kitti vision benchmark
  suite.
\newblock In {\em Proceedings of the IEEE Conference on Computer Vision and
  Pattern Recognition (CVPR)}, pages 3354--3361. IEEE, 2012.

\bibitem{Geiger2012CVPR}
A.~Geiger, P.~Lenz, and R.~Urtasun.
\newblock {Are we ready for Autonomous Driving? The KITTI Vision Benchmark
  Suite}.
\newblock In {\em Proceedings of the Conference on Computer Vision and Pattern
  Recognition}, 2012.

\bibitem{1998_HV_Geisler}
W.~S. Geisler and J.~S. Perry.
\newblock Real-time foveated multiresolution system for low-bandwidth video
  communication.
\newblock In {\em Human vision and electronic imaging III}, volume 3299, pages
  294--305, 1998.

\bibitem{1938_AJP_Gibson}
J.~J. Gibson and L.~E. Crooks.
\newblock A theoretical field-analysis of automobile-driving.
\newblock {\em {The American journal of psychology}}, 51(3):453--471, 1938.

\bibitem{2015_ICCV_Girshick}
R.~Girshick.
\newblock {Fast R-CNN}.
\newblock In {\em Proceedings of the IEEE International Conference on Computer
  Vision (ICCV)}, pages 1440--1448, 2015.

\bibitem{1966_HumanFactors_Gordon}
D.~A. Gordon.
\newblock Experimental isolation of the driver's visual input.
\newblock {\em Human factors}, 8(2):129--138, 1966.

\bibitem{2016_JTSS_Grippenkoven}
J.~Grippenkoven and S.~Dietsch.
\newblock Gaze direction and driving behavior of drivers at level crossings.
\newblock {\em {Journal of Transportation Safety \& Security}}, 8(sup1):4--18,
  2016.

\bibitem{2018_TransRes_Hammit}
B.~E. Hammit, A.~Ghasemzadeh, R.~M. James, M.~M. Ahmed, and R.~K. Young.
\newblock Evaluation of weather-related freeway car-following behavior using
  the shrp2 naturalistic driving study database.
\newblock {\em Transportation Research Part F: Traffic Psychology and
  Behaviour}, 59:244--259, 2018.

\bibitem{2016_TR_Hankey}
J.~M. Hankey, M.~A. Perez, and J.~A. McClafferty.
\newblock {Description of the SHRP 2 naturalistic database and the crash,
  near-crash, and baseline data sets}.
\newblock Technical report, Virginia Tech Transportation Institute, 2016.

\bibitem{2009_JoV_Hansen}
T.~Hansen, L.~Pracejus, and K.~R. Gegenfurtner.
\newblock Color perception in the intermediate periphery of the visual field.
\newblock {\em Journal of vision}, 9(4):26--26, 2009.

\bibitem{2016_TransRes_Harms}
I.~M. Harms and K.~A. Brookhuis.
\newblock Dynamic traffic management on a familiar road: Failing to detect
  changes in variable speed limits.
\newblock {\em {Transportation Research Part F: Traffic Psychology and
  Behaviour}}, 38:37--46, 2016.

\bibitem{2019_TransRes_Hashash}
M.~Hashash, M.~Abou~Zeid, and N.~M. Moacdieh.
\newblock Social media browsing while driving: effects on driver performance
  and attention allocation.
\newblock {\em {Transportation Research Part F: Traffic Psychology and
  Behaviour}}, 63:67--82, 2019.

\bibitem{2005_TrendsCogSci_Ballard}
M.~Hayhoe and D.~Ballard.
\newblock Eye movements in natural behavior.
\newblock {\em {Trends in Cognitive Sciences}}, 9(4):188--194, 2005.

\bibitem{2014_CB_Hayhoe}
M.~Hayhoe and D.~Ballard.
\newblock Modeling task control of eye movements.
\newblock {\em Current Biology}, 24(13):R622--R628, 2014.

\bibitem{2003_JoV_Hayhoe}
M.~M. Hayhoe, A.~Shrivastava, R.~Mruczek, and J.~B. Pelz.
\newblock Visual memory and motor planning in a natural task.
\newblock {\em Journal of vision}, 3(1):6--6, 2003.

\bibitem{2014_HumanFactors_He}
J.~He, J.~S. McCarley, and A.~F. Kramer.
\newblock Lane keeping under cognitive load: performance changes and
  mechanisms.
\newblock {\em {Human Factors}}, 56(2):414--426, 2014.

\bibitem{2017_CVPR_He}
K.~He, G.~Gkioxari, P.~Doll{\'a}r, and R.~Girshick.
\newblock Mask r-cnn.
\newblock In {\em Proceedings of the IEEE International Conference on Computer
  Vision (CVPR)}, pages 2961--2969, 2017.

\bibitem{2018_ICPR_He}
S.~He, D.~Kangin, Y.~Mi, and N.~Pugeault.
\newblock {Aggregated Sparse Attention for Steering Angle Prediction}.
\newblock In {\em Proceedings of the International Conference on Pattern
  Recognition (ICPR)}, pages 2398--2403. IEEE, 2018.

\bibitem{2016_HumanFactors_Hergeth}
S.~Hergeth, L.~Lorenz, R.~Vilimek, and J.~F. Krems.
\newblock {Keep your scanners peeled: Gaze behavior as a measure of automation
  trust during highly automated driving}.
\newblock {\em {Human Factors}}, 58(3):509--519, 2016.

\bibitem{1980_Perception_Hills}
B.~L. Hills.
\newblock Vision, visibility, and perception in driving.
\newblock {\em Perception}, 9(2):183--216, 1980.

\bibitem{2012_ITSC_Hirayama}
T.~Hirayama, K.~Mase, and K.~Takeda.
\newblock Detection of driver distraction based on temporal relationship
  between eye-gaze and peripheral vehicle behavior.
\newblock In {\em Proceedings of the International IEEE Conference on
  Intelligent Transportation Systems (ITSC)}, pages 870--875. IEEE, 2012.

\bibitem{2014_THMS_Ho}
C.~Ho, R.~Gray, and C.~Spence.
\newblock To what extent do the findings of laboratory-based spatial attention
  research apply to the real-world setting of driving?
\newblock {\em IEEE Transactions on Human-Machine Systems}, 44(4):524--530,
  2014.

\bibitem{1973_Psychophysiology_Hoddes}
E.~Hoddes, V.~Zarcone, H.~Smythe, R.~Phillips, and W.~C. Dement.
\newblock Quantification of sleepiness: a new approach.
\newblock {\em Psychophysiology}, 10(4):431--436, 1973.

\bibitem{2011_OUP_Holmqvist}
K.~Holmqvist, M.~Nystr{\"o}m, R.~Andersson, R.~Dewhurst, H.~Jarodzka, and
  J.~Van~de Weijer.
\newblock {\em Eye tracking: A comprehensive guide to methods and measures}.
\newblock Oxford University Press, 2011.

\bibitem{2019_VisionResearch_Hooge}
I.~T. Hooge, R.~S. Hessels, and M.~Nystr{\"o}m.
\newblock Do pupil-based binocular video eye trackers reliably measure
  vergence?
\newblock {\em Vision Research}, 156:1--9, 2019.

\bibitem{2006_JEP_Horrey}
W.~J. Horrey, C.~D. Wickens, and K.~P. Consalus.
\newblock Modeling drivers' visual attention allocation while interacting with
  in-vehicle technologies.
\newblock {\em {Journal of Experimental Psychology: Applied}}, 12(2):67, 2006.

\bibitem{2016_JoV_Huestegge}
L.~Huestegge and A.~B{\"o}ckler.
\newblock {Out of the corner of the driver's eye: Peripheral processing of
  hazards in static traffic scenes}.
\newblock {\em Journal of vision}, 16(2):11--11, 2016.

\bibitem{2010_TR_Huestegge}
L.~Huestegge, E.-M. Skottke, S.~Anders, J.~M{\"u}sseler, and G.~Debus.
\newblock {The development of hazard perception: Dissociation of visual
  orientation and hazard processing}.
\newblock {\em {Transportation Research Part F: Traffic Psychology and
  Behaviour}}, 13(1):1--8, 2010.

\bibitem{2016_AutoUI_Hurtado}
S.~Hurtado and S.~Chiasson.
\newblock An eye-tracking evaluation of driver distraction and unfamiliar road
  signs.
\newblock In {\em Proceedings of the 8th International Conference on Automotive
  User Interfaces and Interactive Vehicular Applications}, pages 153--160,
  2016.

\bibitem{2012_Cell_Hutchinson}
J.~B. Hutchinson and N.~B. Turk-Browne.
\newblock {Memory-guided attention: Control from multiple memory systems}.
\newblock {\em Trends in Cognitive Sciences}, 16(12):576--579, 2012.

\bibitem{2015_PONE_Itkonen}
T.~Itkonen, J.~Pekkanen, and O.~Lappi.
\newblock Driver gaze behavior is different in normal curve driving and when
  looking at the tangent point.
\newblock {\em PloS one}, 10(8):e0135505, 2015.

\bibitem{2009_Gerontechnology_Itoh}
N.~Itoh, K.~Sagawa, and Y.~Fukunaga.
\newblock Useful visual field at a homogeneous background for old and young
  subjects.
\newblock {\em Gerontechnology}, 8(1):42--51, 2009.

\bibitem{2001_NatRevNeuroscience_Itti}
L.~Itti and C.~Koch.
\newblock Computational modelling of visual attention.
\newblock {\em {Nature Reviews Neuroscience}}, 2(3):194--203, 2001.

\bibitem{1998_PAMI_Itti}
L.~Itti, C.~Koch, and E.~Niebur.
\newblock A model of saliency-based visual attention for rapid scene analysis.
\newblock {\em {IEEE Transactions on Pattern Analysis and Machine
  Intelligence}}, 20(11):1254--1259, 1998.

\bibitem{2016_TraffInjuryPrevention_Jackson}
M.~L. Jackson, S.~Raj, R.~J. Croft, A.~C. Hayley, L.~A. Downey, G.~A. Kennedy,
  and M.~E. Howard.
\newblock Slow eyelid closure as a measure of driver drowsiness and its
  relationship to performance.
\newblock {\em Traffic injury prevention}, 17(3):251--257, 2016.

\bibitem{2015_ICCV_Jain}
A.~Jain, H.~S. Koppula, B.~Raghavan, S.~Soh, and A.~Saxena.
\newblock Car that knows before you do: Anticipating maneuvers via learning
  temporal driving models.
\newblock In {\em Proceedings of the IEEE International Conference on Computer
  Vision (ICCV)}, pages 3182--3190, 2015.

\bibitem{2016_ICRA_Jain}
A.~Jain, A.~Singh, H.~S. Koppula, S.~Soh, and A.~Saxena.
\newblock Recurrent neural networks for driver activity anticipation via
  sensory-fusion architecture.
\newblock In {\em Proceedings of the International Conference on Robotics and
  Automation (ICRA)}, pages 3118--3125. IEEE, 2016.

\bibitem{2013_TR_Jamson}
A.~H. Jamson, N.~Merat, O.~M. Carsten, and F.~C. Lai.
\newblock Behavioural changes in drivers experiencing highly-automated vehicle
  control in varying traffic conditions.
\newblock {\em {Transportation Research Part C: Emerging Technologies}},
  30:116--125, 2013.

\bibitem{2020_Now_Janai}
J.~Janai, F.~G{\"u}ney, A.~Behl, A.~Geiger, et~al.
\newblock {Computer vision for autonomous vehicles: Problems, datasets and
  state of the art}.
\newblock {\em {Foundations and Trends{\textregistered} in Computer Graphics
  and Vision}}, 12(1--3):1--308, 2020.

\bibitem{2010_CHI_Jensen}
B.~S. Jensen, M.~B. Skov, and N.~Thiruravichandran.
\newblock Studying driver attention and behaviour for three configurations of
  gps navigation in real traffic driving.
\newblock In {\em Proceedings of the SIGCHI Conference on Human Factors in
  Computing Systems}, pages 1271--1280, 2010.

\bibitem{2011_WIR_Jensen}
M.~S. Jensen, R.~Yao, W.~N. Street, and D.~J. Simons.
\newblock Change blindness and inattentional blindness.
\newblock {\em Wiley Interdisciplinary Reviews: Cognitive Science},
  2(5):529--546, 2011.

\bibitem{2018_ITSC_Jha}
S.~Jha and C.~Busso.
\newblock Probabilistic estimation of the gaze region of the driver using dense
  classification.
\newblock In {\em Proceedings of the International Conference on Intelligent
  Transportation Systems (ITSC)}, pages 697--702. IEEE, 2018.

\bibitem{2012_TITS_Jimenez}
P.~Jim{\'e}nez, L.~M. Bergasa, J.~Nuevo, N.~Hern{\'a}ndez, and I.~G. Daza.
\newblock Gaze fixation system for the evaluation of driver distractions
  induced by ivis.
\newblock {\em IEEE Transactions on Intelligent Transportation Systems},
  13(3):1167--1178, 2012.

\bibitem{2013_AdvMechEng_Jin}
L.~Jin, Q.~Niu, Y.~Jiang, H.~Xian, Y.~Qin, and M.~Xu.
\newblock Driver sleepiness detection system based on eye movements variables.
\newblock {\em Advances in Mechanical Engineering}, 5:648431, 2013.

\bibitem{2011_OptEng_Jo}
J.~Jo, S.~J. Lee, J.~Kim, H.~G. Jung, and K.~R. Park.
\newblock Vision-based method for detecting driver drowsiness and distraction
  in driver monitoring system.
\newblock {\em Optical Engineering}, 50(12):127202, 2011.

\bibitem{2013_RSTB_Johnson}
L.~Johnson, B.~Sullivan, M.~Hayhoe, and D.~Ballard.
\newblock Predicting human visuomotor behaviour in a driving task.
\newblock {\em Philosophical Transactions of the Royal Society B: Biological
  Sciences}, 369(1636):20130044, 2014.

\bibitem{2014_AccidentPrevention_Jones}
M.~Jones, P.~Chapman, and K.~Bailey.
\newblock The influence of image valence on visual attention and perception of
  risk in drivers.
\newblock {\em {Accident Analysis \& Prevention}}, 73:296--304, 2014.

\bibitem{2015_AppliedErgonomics_Kaber}
D.~Kaber, C.~Pankok~Jr, B.~Corbett, W.~Ma, J.~Hummer, and W.~Rasdorf.
\newblock Driver behavior in use of guide and logo signs under distraction and
  complex roadway conditions.
\newblock {\em {Applied Ergonomics}}, 47:99--106, 2015.

\bibitem{2012_TR_Kaber}
D.~B. Kaber, Y.~Liang, Y.~Zhang, M.~L. Rogers, and S.~Gangakhedkar.
\newblock Driver performance effects of simultaneous visual and cognitive
  distraction and adaptation behavior.
\newblock {\em {Transportation Research Part F: Traffic Psychology and
  Behaviour}}, 15(5):491--501, 2012.

\bibitem{1966_Science_Kahneman}
D.~Kahneman and J.~Beatty.
\newblock Pupil diameter and load on memory.
\newblock {\em Science}, 154(3756):1583--1585, 1966.

\bibitem{1967_PerceptionPsychopysica_Kahneman}
D.~Kahnemann and J.~Beatty.
\newblock Pupillary responses in a pitch-discrimination task.
\newblock {\em Perception \& Psychophysics}, 2(3):101--105, 1967.

\bibitem{1969_TechRep_Kaluger}
N.~A. Kaluger and G.~Smith~Jr.
\newblock Driver eye-movement patterns under conditions of prolonged driving
  and sleep deprivation.
\newblock Technical report, Ohio State University, 1969.

\bibitem{2017_IEEEAccess_Kar}
A.~Kar and P.~Corcoran.
\newblock A review and analysis of eye-gaze estimation systems, algorithms and
  performance evaluation methods in consumer platforms.
\newblock {\em IEEE Access}, 5:16495--16519, 2017.

\bibitem{2013_TechRep_Karlsson}
J.~Karlsson, C.~Apoy, H.~Lind, S.~Dombrovskis, M.~Axest{\aa}l, and
  M.~Johansson.
\newblock {EyesOnRoad -- an anti-distraction field operational test}.
\newblock Technical report, FFI-Vehicle and Traffic Safety Program, 2016.

\bibitem{2012_TechRep_Kessler}
C.~Kessler, A.~Etemad, G.~Alessendretti, K.~Heinig, B.~R. Selpi,
  A.~Cserpinszky, W.~Hagleitner, and M.~Benmimoun.
\newblock {European large-scale field operational tests on in-vehicle systems:
  Deliverable D11. 3}.
\newblock Technical report, {euroFOT Consortium}, 2012.

\bibitem{2016_AccidentAnalysis_Kidd}
D.~G. Kidd and A.~T. McCartt.
\newblock Differences in glance behavior between drivers using a rearview
  camera, parking sensor system, both technologies, or no technology during
  low-speed parking maneuvers.
\newblock {\em {Accident Analysis \& Prevention}}, 87:92--101, 2016.

\bibitem{2018_TransRes_Kidd}
D.~G. Kidd, B.~Reimer, J.~Dobres, and B.~Mehler.
\newblock Changes in driver glance behavior when using a system that automates
  steering to perform a low-speed parallel parking maneuver.
\newblock {\em {Transportation Research Part F: Traffic Psychology and
  Behaviour}}, 58:629--639, 2018.

\bibitem{2019_HumanFactors_Kim}
H.~Kim and J.~L. Gabbard.
\newblock Assessing distraction potential of augmented reality head-up displays
  for vehicle drivers.
\newblock {\em {Human Factors}}, page 0018720819844845, 2019.

\bibitem{2019_HumanFactorsErgonomics_Kim}
H.~Kim, J.~L. Gabbard, S.~Martin, A.~Tawari, and T.~Misu.
\newblock {Toward Prediction of Driver Awareness of Automotive Hazards:
  Driving-Video-Based Simulation Approach}.
\newblock In {\em {Proceedings of the Human Factors and Ergonomics Society
  Annual Meeting}}, volume~63, pages 2099--2103. SAGE Publications Sage CA: Los
  Angeles, CA, 2019.

\bibitem{2017_ICCV_Kim}
J.~Kim and J.~Canny.
\newblock Interpretable learning for self-driving cars by visualizing causal
  attention.
\newblock In {\em Proceedings of the IEEE International Conference on Computer
  Vision (ICCV)}, pages 2942--2950, 2017.

\bibitem{2019_CVPR_Kim}
J.~Kim, T.~Misu, Y.-T. Chen, A.~Tawari, and J.~Canny.
\newblock Grounding human-to-vehicle advice for self-driving vehicles.
\newblock In {\em Proceedings of the IEEE Conference on Computer Vision and
  Pattern Recognition (CVPR)}, pages 10591--10599, 2019.

\bibitem{2018_ECCV_Kim}
J.~Kim, A.~Rohrbach, T.~Darrell, J.~Canny, and Z.~Akata.
\newblock Textual explanations for self-driving vehicles.
\newblock In {\em Proceedings of the European Conference on Computer Vision
  (ECCV)}, pages 563--578, 2018.

\bibitem{2012_SAE_Kim}
R.~Kim, R.~Rauschenberger, G.~Heckman, D.~Young, and R.~Lange.
\newblock Efficacy and usage patterns for three types of rearview camera
  displays during backing up.
\newblock Technical report, SAE Technical Paper, 2012.

\bibitem{2009_JMLR_King}
D.~E. King.
\newblock {Dlib-ml: A machine learning toolkit}.
\newblock {\em Journal of Machine Learning Research}, 10:1755--1758, 2009.

\bibitem{2017_HumanFactors_Kircher}
K.~Kircher and C.~Ahlstrom.
\newblock Minimum required attention: a human-centered approach to driver
  inattention.
\newblock {\em {Human Factors}}, 59(3):471--484, 2017.

\bibitem{2018_AccidentAnalysis_Kircher}
K.~Kircher and C.~Ahlstrom.
\newblock Evaluation of methods for the assessment of attention while driving.
\newblock {\em {Accident Analysis \& Prevention}}, 114:40--47, 2018.

\bibitem{1958_JEP_Kirchner}
W.~K. Kirchner.
\newblock Age differences in short-term retention of rapidly changing
  information.
\newblock {\em {Journal of Experimental Psychology}}, 55(4):352, 1958.

\bibitem{2006_TR_Klauer}
S.~G. Klauer, T.~A. Dingus, V.~L. Neale, J.~D. Sudweeks, D.~J. Ramsey, et~al.
\newblock The impact of driver inattention on near-crash/crash risk: An
  analysis using the 100-car naturalistic driving study data.
\newblock Technical report, National Highway Traffic Safety Administration,
  2006.

\bibitem{2010_TR_Klauer}
S.~G. Klauer, F.~Guo, J.~Sudweeks, and T.~A. Dingus.
\newblock {An analysis of driver inattention using a case-crossover approach on
  100-Car data}.
\newblock Technical report, {US Department of Transportation. National Highway
  Traffic Safety Administration}, 2010.

\bibitem{2010_AccidentAnalysis_Konstantopoulos}
P.~Konstantopoulos, P.~Chapman, and D.~Crundall.
\newblock Driver's visual attention as a function of driving experience and
  visibility. using a driving simulator to explore drivers’ eye movements in
  day, night and rain driving.
\newblock {\em {Accident Analysis \& Prevention}}, 42(3):827--834, 2010.

\bibitem{2016_AccidentAnalysis_Kountouriotis}
G.~K. Kountouriotis and N.~Merat.
\newblock {Leading to distraction: Driver distraction, lead car, and road
  environment}.
\newblock {\em {Accident Analysis \& Prevention}}, 89:22--30, 2016.

\bibitem{2014_IV_Kowsari}
T.~Kowsari, S.~S. Beauchemin, M.~A. Bauer, D.~Laurendeau, and N.~Teasdale.
\newblock Multi-depth cross-calibration of remote eye gaze trackers and
  stereoscopic scene systems.
\newblock In {\em Proceedings of the IEEE Intelligent Vehicles Symposium (IV)},
  pages 1245--1250. IEEE, 2014.

\bibitem{2018_TransRes_Kraft}
A.-K. Kraft, F.~Naujoks, J.~W{\"o}rle, and A.~Neukum.
\newblock The impact of an in-vehicle display on glance distribution in
  partially automated driving in an on-road experiment.
\newblock {\em {Transportation Research Part F: Traffic Psychology and
  Behaviour}}, 52:40--50, 2018.

\bibitem{2017_ACM_Krizhevsky}
A.~Krizhevsky, I.~Sutskever, and G.~E. Hinton.
\newblock Imagenet classification with deep convolutional neural networks.
\newblock {\em Communications of the ACM}, 60(6):84--90, 2017.

\bibitem{2013_PUC_Kujala}
T.~Kujala.
\newblock Browsing the information highway while driving: three in-vehicle
  touch screen scrolling methods and driver distraction.
\newblock {\em Personal and Ubiquitous Computing}, 17(5):815--823, 2013.

\bibitem{2019_SafetyScience_Kuo}
J.~Kuo, M.~G. Lenn{\'e}, M.~Mulhall, T.~Sletten, C.~Anderson, M.~Howard,
  S.~Rajaratnam, M.~Magee, and A.~Collins.
\newblock Continuous monitoring of visual distraction and drowsiness in
  shift-workers during naturalistic driving.
\newblock {\em Safety Science}, 119:112--116, 2019.

\bibitem{1994_Nature_Land}
M.~F. Land and D.~N. Lee.
\newblock Where we look when we steer.
\newblock {\em Nature}, 369(6483):742--744, 1994.

\bibitem{2016_ICRA_Langner}
T.~Langner, D.~Seifert, B.~Fischer, D.~Goehring, T.~Ganjineh, and R.~Rojas.
\newblock Traffic awareness driver assistance based on stereovision,
  eye-tracking, and head-up display.
\newblock In {\em Proceedings of the International Conference on Robotics and
  Automation (ICRA)}, pages 3167--3173. IEEE, 2016.

\bibitem{2014_JoV_Lappi}
O.~Lappi.
\newblock Future path and tangent point models in the visual control of
  locomotion in curve driving.
\newblock {\em Journal of Vision}, 14(12):21--21, 2014.

\bibitem{2016_NBR_Lappi}
O.~Lappi.
\newblock {Eye movements in the wild: Oculomotor control, gaze behavior \&
  frames of reference}.
\newblock {\em Neuroscience \& Biobehavioral Reviews}, 69:49--68, 2016.

\bibitem{2013_JoV_Lappi}
O.~Lappi, E.~Lehtonen, J.~Pekkanen, and T.~Itkonen.
\newblock Beyond the tangent point: gaze targets in naturalistic driving.
\newblock {\em Journal of Vision}, 13(13):11--11, 2013.

\bibitem{2017_FPsych_Lappi}
O.~Lappi, P.~Rinkkala, and J.~Pekkanen.
\newblock Systematic observation of an expert driver's gaze strategy -- an
  on-road case study.
\newblock {\em Frontiers in psychology}, 8:620, 2017.

\bibitem{2006_TechRep_LeBlanc}
D.~LeBlanc.
\newblock Road departure crash warning system field operational test:
  methodology and results.
\newblock Technical report, Transportation Research Institute, University of
  Michigan, Ann Arbor, 2006.

\bibitem{2012_HumanFactors_Lee}
J.~D. Lee, S.~C. Roberts, J.~D. Hoffman, and L.~S. Angell.
\newblock Scrolling and driving: How an mp3 player and its aftermarket
  controller affect driving performance and visual behavior.
\newblock {\em {Human Factors}}, 54(2):250--263, 2012.

\bibitem{2016_CHI_Lee}
J.~Y. Lee, M.~C. Gibson, and J.~D. Lee.
\newblock Error recovery in multitasking while driving.
\newblock In {\em Proceedings of the 2016 CHI Conference on Human Factors in
  Computing Systems}, pages 5104--5113, 2016.

\bibitem{2018_AccidentAnalysis_Lee}
J.~Y. Lee, J.~D. Lee, J.~B{\"a}rgman, J.~Lee, and B.~Reimer.
\newblock How safe is tuning a radio?: using the radio tuning task as a
  benchmark for distracted driving.
\newblock {\em {Accident Analysis \& Prevention}}, 110:29--37, 2018.

\bibitem{2011_TITS_Lee}
S.~J. Lee, J.~Jo, H.~G. Jung, K.~R. Park, and J.~Kim.
\newblock Real-time gaze estimator based on driver's head orientation for
  forward collision warning system.
\newblock {\em IEEE Transactions on Intelligent Transportation Systems},
  12(1):254--267, 2011.

\bibitem{2016_OptometryVisionScience_Lee}
S.~S.-Y. Lee, A.~A. Black, P.~Lacherez, and J.~M. Wood.
\newblock Eye movements and road hazard detection: effects of blur and
  distractors.
\newblock {\em Optometry and Vision Science}, 93(9):1137--1146, 2016.

\bibitem{2015_OPO_Lee}
S.~S.-Y. Lee, J.~M. Wood, and A.~A. Black.
\newblock Blur, eye movements and performance on a driving visual recognition
  slide test.
\newblock {\em Ophthalmic and physiological optics}, 35(5):522--529, 2015.

\bibitem{1987_JOSA_Legge}
G.~E. Legge and D.~Kersten.
\newblock Contrast discrimination in peripheral vision.
\newblock {\em Journal of the Optical Society of America A}, 4(8):1594--1598,
  1987.

\bibitem{2014_AccidentAnalysis_Lehtonen}
E.~Lehtonen, O.~Lappi, I.~Koirikivi, and H.~Summala.
\newblock Effect of driving experience on anticipatory look-ahead fixations in
  real curve driving.
\newblock {\em {Accident Analysis \& Prevention}}, 70:195--208, 2014.

\bibitem{2013_Ergonomics_Lehtonen}
E.~Lehtonen, O.~Lappi, H.~Kotkanen, and H.~Summala.
\newblock Look-ahead fixations in curve driving.
\newblock {\em Ergonomics}, 56(1):34--44, 2013.

\bibitem{2014_SafetyScience_Lemercier}
C.~Lemercier, C.~Pecher, G.~Berthi{\'e}, B.~Valery, V.~Vidal, P.-V. Paubel,
  M.~Cour, A.~Fort, C.~Gal{\'e}ra, C.~Gabaude, et~al.
\newblock {Inattention behind the wheel: How factual internal thoughts impact
  attentional control while driving}.
\newblock {\em Safety Science}, 62:279--285, 2014.

\bibitem{2014_JEMR_Lemonnier}
S.~Lemonnier, R.~Br{\'e}mond, and T.~Baccino.
\newblock Discriminating cognitive processes with eye movements in a
  decision-making driving task.
\newblock {\em Journal of Eye Movement Research}, 7(4):1--14, 2014.

\bibitem{2015_TR_Lemonnier}
S.~Lemonnier, R.~Br{\'e}mond, and T.~Baccino.
\newblock {Gaze behavior when approaching an intersection: Dwell time
  distribution and comparison with a quantitative prediction}.
\newblock {\em {Transportation Research Part F: Traffic Psychology and
  Behaviour}}, 35:60--74, 2015.

\bibitem{2016_TheorIssErgonSci_Lenne}
M.~G. Lenn{\'e} and E.~E. Jacobs.
\newblock {Predicting drowsiness-related driving events: A review of recent
  research methods and future opportunities}.
\newblock {\em Theoretical Issues in Ergonomics Science}, 17(5-6):533--553,
  2016.

\bibitem{2019_JSR_Li}
G.~Li, Y.~Wang, F.~Zhu, X.~Sui, N.~Wang, X.~Qu, and P.~Green.
\newblock {Drivers’ visual scanning behavior at signalized and unsignalized
  intersections: A naturalistic driving study in China}.
\newblock {\em {Journal of Safety Research}}, 71:219--229, 2019.

\bibitem{2015_TITS_Li}
N.~Li and C.~Busso.
\newblock Predicting perceived visual and cognitive distractions of drivers
  with multimodal features.
\newblock {\em IEEE Transactions on Intelligent Transportation Systems},
  16(1):51--65, 2014.

\bibitem{2016_TITS_Li}
N.~Li and C.~Busso.
\newblock Detecting drivers' mirror-checking actions and its application to
  maneuver and secondary task recognition.
\newblock {\em IEEE Transactions on Intelligent Transportation Systems},
  17(4):980--992, 2015.

\bibitem{2018_TRR_Li}
X.~Li, A.~Rakotonirainy, X.~Yan, and Y.~Zhang.
\newblock Driver’s visual performance in rear-end collision avoidance process
  under the influence of cell phone use.
\newblock {\em {Transportation Research Record}}, 2672(37):55--63, 2018.

\bibitem{2020_TransRes_Li}
X.~Li, R.~Schroeter, A.~Rakotonirainy, J.~Kuo, and M.~G. Lenn{\'e}.
\newblock Effects of different non-driving-related-task display modes on
  drivers’ eye-movement patterns during take-over in an automated vehicle.
\newblock {\em {Transportation Research Part F: Traffic Psychology and
  Behaviour}}, 70:135--148, 2020.

\bibitem{2015_HumanFactors_Liang}
Y.~Liang, W.~J. Horrey, and J.~D. Hoffman.
\newblock Reading text while driving: Understanding drivers’ strategic and
  tactical adaptation to distraction.
\newblock {\em {Human Factors}}, 57(2):347--359, 2015.

\bibitem{2014_HFES_Liang}
Y.~Liang, J.~D. Lee, and W.~J. Horrey.
\newblock A looming crisis: the distribution of off-road glance duration in
  moments leading up to crashes/near-crashes in naturalistic driving.
\newblock In {\em Proceedings of the Human Factors and Ergonomics Society
  Annual Meeting}, volume~58, pages 2102--2106. SAGE Publications Sage CA: Los
  Angeles, CA, 2014.

\bibitem{2012_HumanFactors_Liang}
Y.~Liang, J.~D. Lee, and L.~Yekhshatyan.
\newblock How dangerous is looking away from the road? algorithms predict crash
  risk from glance patterns in naturalistic driving.
\newblock {\em Human Factors}, 54(6):1104--1116, 2012.

\bibitem{2016_IV_Liao}
Y.~Liao, S.~E. Li, G.~Li, W.~Wang, B.~Cheng, and F.~Chen.
\newblock Detection of driver cognitive distraction: An svm based real-time
  algorithm and its comparison study in typical driving scenarios.
\newblock In {\em Proceedings of the IEEE Intelligent Vehicles Symposium (IV)},
  pages 394--399. IEEE, 2016.

\bibitem{2016_TITS_Liao}
Y.~Liao, S.~E. Li, W.~Wang, Y.~Wang, G.~Li, and B.~Cheng.
\newblock Detection of driver cognitive distraction: A comparison study of
  stop-controlled intersection and speed-limited highway.
\newblock {\em IEEE Transactions on Intelligent Transportation Systems},
  17(6):1628--1637, 2016.

\bibitem{2013_TR_Lim}
P.~C. Lim, E.~Sheppard, and D.~Crundall.
\newblock Cross-cultural effects on drivers’ hazard perception.
\newblock {\em {Transportation Research Part F: Traffic Psychology and
  Behaviour}}, 21:194--206, 2013.

\bibitem{2019_ACM_Liu}
C.~Liu, Y.~Chen, L.~Tai, H.~Ye, M.~Liu, and B.~E. Shi.
\newblock A gaze model improves autonomous driving.
\newblock In {\em Proceedings of the ACM Symposium on Eye Tracking Research \&
  Applications}, pages 1--5, 2019.

\bibitem{2015_TITS_Liu}
T.~Liu, Y.~Yang, G.-B. Huang, Y.~K. Yeo, and Z.~Lin.
\newblock Driver distraction detection using semi-supervised machine learning.
\newblock {\em IEEE transactions on Intelligent Transportation Systems},
  17(4):1108--1120, 2015.

\bibitem{2019_AIR_Liu}
X.~Liu, Z.~Deng, and Y.~Yang.
\newblock Recent progress in semantic image segmentation.
\newblock {\em {Artificial Intelligence Review}}, 52(2):1089--1106, 2019.

\bibitem{2011_OUP_Liversedge}
S.~Liversedge, I.~Gilchrist, and S.~Everling.
\newblock {\em The Oxford handbook of eye movements}.
\newblock Oxford University Press, 2011.

\bibitem{2014_HumanFactorsErognomics_Lorenz}
L.~Lorenz, P.~Kerschbaum, and J.~Schumann.
\newblock Designing take over scenarios for automated driving: How does
  augmented reality support the driver to get back into the loop?
\newblock In {\em Proceedings of the Human Factors and Ergonomics Society
  Annual Meeting}, volume~58, pages 1681--1685. SAGE Publications Sage CA: Los
  Angeles, CA, 2014.

\bibitem{2017_JoV_Loschky}
L.~C. Loschky, A.~Nuthmann, F.~C. Fortenbaugh, and D.~M. Levi.
\newblock Scene perception from central to peripheral vision.
\newblock {\em Journal of vision}, 17(1):6--6, 2017.

\bibitem{2019_TransRes_Louw}
T.~Louw, J.~Kuo, R.~Romano, V.~Radhakrishnan, M.~G. Lenn{\'e}, and N.~Merat.
\newblock {Engaging in NDRTs affects drivers’ responses and glance patterns
  after silent automation failures}.
\newblock {\em {Transportation Research Part F: Traffic Psychology and
  Behaviour}}, 62:870--882, 2019.

\bibitem{2017_TransRes_Louw}
T.~Louw and N.~Merat.
\newblock {Are you in the loop? Using gaze dispersion to understand driver
  visual attention during vehicle automation}.
\newblock {\em {Transportation Research Part C: Emerging Technologies}},
  76:35--50, 2017.

\bibitem{1958_ArchOpht_Lowenstein}
O.~Lowenstein and I.~E. Loewenfeld.
\newblock Electronic pupillography: a new instrument and some clinical
  applications.
\newblock {\em {AMA Archives of Ophthalmology}}, 59(3):352--363, 1958.

\bibitem{2017_AppErgonomics_Lu}
Z.~Lu, X.~Coster, and J.~De~Winter.
\newblock {How much time do drivers need to obtain situation awareness? A
  laboratory-based study of automated driving}.
\newblock {\em {Applied Ergonomics}}, 60:293--304, 2017.

\bibitem{2020_TR_Lu}
Z.~Lu, R.~Happee, and J.~C. de~Winter.
\newblock {Take over! A video-clip study measuring attention, situation
  awareness, and decision-making in the face of an impending hazard}.
\newblock {\em {Transportation Research Part F: Traffic Psychology and
  Behaviour}}, 72:211--225, 2020.

\bibitem{2019_TransRes_Lu}
Z.~Lu, B.~Zhang, A.~Feldh{\"u}tter, R.~Happee, M.~Martens, and J.~De~Winter.
\newblock {Beyond mere take-over requests: The effects of monitoring requests
  on driver attention, take-over performance, and acceptance}.
\newblock {\em {Transportation Research Part F: Traffic Psychology and
  Behaviour}}, 63:22--37, 2019.

\bibitem{2016_TITS_Lundgren}
M.~Lundgren, L.~Hammarstrand, and T.~McKelvey.
\newblock Driver-gaze zone estimation using bayesian filtering and gaussian
  processes.
\newblock {\em IEEE Transactions on Intelligent Transportation Systems},
  17(10):2739--2750, 2016.

\bibitem{2019_ACM_Ma}
Y.~Ma, J.~Wu, and C.~Long.
\newblock Gazefcw: Filter collision warning triggers by detecting driver's gaze
  area.
\newblock In {\em Proceedings of the Workshop on Machine Learning on Edge in
  Sensor Systems}, pages 13--18, 2019.

\bibitem{2015_VC_Mackenzie}
A.~K. Mackenzie and J.~M. Harris.
\newblock Eye movements and hazard perception in active and passive driving.
\newblock {\em Visual cognition}, 23(6):736--757, 2015.

\bibitem{2012_PLOS_Mars}
F.~Mars and J.~Navarro.
\newblock Where we look when we drive with or without active steering wheel
  control.
\newblock {\em PloS one}, 7(8):e43858, 2012.

\bibitem{1999_TechRep_Martens}
M.~Martens and W.~Van~Winsum.
\newblock Measuring distraction: the peripheral detection task.
\newblock {\em TNO Human Factors}, 2000.

\bibitem{2007_TransRes_Martens}
M.~H. Martens and M.~R. Fox.
\newblock Do familiarity and expectations change perception? drivers’ glances
  and response to changes.
\newblock {\em {Transportation Research Part F: Traffic Psychology and
  Behaviour}}, 10(6):476--492, 2007.

\bibitem{2017_IV_Martin}
S.~Martin and M.~M. Trivedi.
\newblock Gaze fixations and dynamics for behavior modeling and prediction of
  on-road driving maneuvers.
\newblock In {\em Proceedings of the IEEE Intelligent Vehicles Symposium (IV)},
  pages 1541--1545. IEEE, 2017.

\bibitem{2018_TIV_Martin}
S.~Martin, S.~Vora, K.~Yuen, and M.~M. Trivedi.
\newblock Dynamics of driver's gaze: Explorations in behavior modeling and
  maneuver prediction.
\newblock {\em IEEE Transactions on Intelligent Vehicles}, 3(2):141--150, 2018.

\bibitem{2013_TITS_Mbouna}
R.~O. Mbouna, S.~G. Kong, and M.-G. Chun.
\newblock Visual analysis of eye state and head pose for driver alertness
  monitoring.
\newblock {\em IEEE Transactions on Intelligent Transportation Systems},
  14(3):1462--1469, 2013.

\bibitem{2014_ACP_Mccarley}
J.~S. McCarley, K.~S. Steelman, and W.~J. Horrey.
\newblock {The View from the Driver's Seat: What Good Is Salience?}
\newblock {\em Applied Cognitive Psychology}, 28(1):47--54, 2014.

\bibitem{2014_TR_Merat}
N.~Merat, A.~H. Jamson, F.~C. Lai, M.~Daly, and O.~M. Carsten.
\newblock {Transition to manual: Driver behaviour when resuming control from a
  highly automated vehicle}.
\newblock {\em {Transportation Research Part F: Traffic Psychology and
  Behaviour}}, 27:274--282, 2014.

\bibitem{2014_TransRes_Metz}
B.~Metz and H.-P. Kr{\"u}ger.
\newblock Do supplementary signs distract the driver?
\newblock {\em {Transportation Research Part F: Traffic Psychology and
  Behaviour}}, 23:1--14, 2014.

\bibitem{2011_TR_Metz}
B.~Metz, N.~Sch{\"o}mig, and H.-P. Kr{\"u}ger.
\newblock {Attention during visual secondary tasks in driving: Adaptation to
  the demands of the driving task}.
\newblock {\em {Transportation Research Part F: Traffic Psychology and
  Behaviour}}, 14(5):369--380, 2011.

\bibitem{2019_TransRes_Miller}
E.~E. Miller and L.~N. Boyle.
\newblock Adaptations in attention allocation: implications for takeover in an
  automated vehicle.
\newblock {\em {Transportation Research Part F: Traffic Psychology and
  Behaviour}}, 66:101--110, 2019.

\bibitem{2016_AccidentAnalysis_Morando}
A.~Morando, T.~Victor, and M.~Dozza.
\newblock {Drivers anticipate lead-vehicle conflicts during automated
  longitudinal control: Sensory cues capture driver attention and promote
  appropriate and timely responses}.
\newblock {\em {Accident Analysis \& Prevention}}, 97:206--219, 2016.

\bibitem{2018_TITS_Morando}
A.~Morando, T.~Victor, and M.~Dozza.
\newblock {A reference model for driver attention in automation: Glance
  behavior changes during lateral and longitudinal assistance}.
\newblock {\em {IEEE Transactions on Intelligent Transportation Systems}},
  20(8):2999--3009, 2018.

\bibitem{2012_ITSC_Mori}
M.~Mori, C.~Miyajima, P.~Angkititrakul, T.~Hirayama, Y.~Li, N.~Kitaoka, and
  K.~Takeda.
\newblock Measuring driver awareness based on correlation between gaze behavior
  and risks of surrounding vehicles.
\newblock In {\em Proceedings of the International IEEE Conference on
  Intelligent Transportation Systems (ITSC)}, pages 644--647. IEEE, 2012.

\bibitem{1970_SAE_Mourant}
R.~R. Mourant and T.~H. Rockwell.
\newblock Visual information seeking of novice drivers.
\newblock Technical report, SAE International, 1970.

\bibitem{1972_HumanFactors_Mourant}
R.~R. Mourant and T.~H. Rockwell.
\newblock Strategies of visual search by novice and experienced drivers.
\newblock {\em {Human Factors}}, 14(4):325--335, 1972.

\bibitem{2016_TransRes_Munoz}
M.~Mu{\~n}oz, B.~Reimer, J.~Lee, B.~Mehler, and L.~Fridman.
\newblock Distinguishing patterns in drivers’ visual attention allocation
  using hidden markov models.
\newblock {\em Transportation Research Part F: Traffic Psychology and
  Behaviour}, 43:90--103, 2016.

\bibitem{2011_CTW_Nabatilan}
L.~B. Nabatilan, F.~Aghazadeh, A.~D. Nimbarte, C.~C. Harvey, and S.~K.
  Chowdhury.
\newblock Effect of driving experience on visual behavior and driving
  performance under different driving conditions.
\newblock {\em Cognition, technology \& work}, 14(4):355--363, 2012.

\bibitem{2019_IJHCI_Navarro}
J.~Navarro, F.~Osiurak, M.~Ovigue, L.~Charrier, and E.~Reynaud.
\newblock {Highly Automated Driving Impact on Drivers’ Gaze Behaviors during
  a Car-Following Task}.
\newblock {\em {International Journal of Human-Computer Interaction}},
  35(11):1008--1017, 2019.

\bibitem{2005_TechRep_Neale}
V.~L. Neale, T.~A. Dingus, S.~G. Klauer, J.~Sudweeks, and M.~Goodman.
\newblock {An overview of the 100-Car naturalistic study and findings}.
\newblock Technical report, {National Highway Traffic Safety Administration},
  2005.

\bibitem{1940_JAMA_Newman}
H.~Newman and E.~Fletcher.
\newblock The effect of alcohol on driving skill.
\newblock {\em Journal of the American Medical Association},
  115(19):1600--1602, 1940.

\bibitem{2015_TransRes_Niezgoda}
M.~Niezgoda, A.~Tarnowski, M.~Kruszewski, and T.~Kami{\'n}ski.
\newblock {Towards testing auditory--vocal interfaces and detecting distraction
  while driving: A comparison of eye-movement measures in the assessment of
  cognitive workload}.
\newblock {\em {Transportation Research Part F: Traffic Psychology and
  Behaviour}}, 32:23--34, 2015.

\bibitem{2019_ITSC_Ning}
M.~Ning, C.~Lu, and J.~Gong.
\newblock {An Efficient Model for Driving Focus of Attention Prediction using
  Deep Learning}.
\newblock In {\em Proceedings of the IEEE Intelligent Transportation Systems
  Conference (ITSC)}, pages 1192--1197. IEEE, 2019.

\bibitem{2017_PR_Ohn-Bar}
E.~Ohn-Bar and M.~M. Trivedi.
\newblock Are all objects equal? deep spatio-temporal importance prediction in
  driving videos.
\newblock {\em Pattern Recognition}, 64:425--436, 2017.

\bibitem{2019_JEMR_Ojstersek}
T.~C. Ojster{\v{s}}ek and D.~Topol{\v{s}}ek.
\newblock {Eye tracking use in researching driver distraction: A scientometric
  and qualitative literature review approach}.
\newblock {\em Journal of Eye Movement Research}, 12(3), 2019.

\bibitem{2007_TransRes_Olsen}
E.~C. Olsen, S.~E. Lee, and B.~G. Simons-Morton.
\newblock Eye movement patterns for novice teen drivers: does 6 months of
  driving experience make a difference?
\newblock {\em {Transportation Research Record}}, 2009(1):8--14, 2007.

\bibitem{1935_AO_Osterberg}
G.~Osterberg.
\newblock Topography of the layer of the rods and cones in the human retima.
\newblock {\em Acta ophthalmol}, 13(6):1--102, 1935.

\bibitem{2008_JoV_Otero-Millan}
J.~Otero-Millan, X.~G. Troncoso, S.~L. Macknik, I.~Serrano-Pedraza, and
  S.~Martinez-Conde.
\newblock Saccades and microsaccades during visual fixation, exploration, and
  search: foundations for a common saccadic generator.
\newblock {\em Journal of vision}, 8(14):21--21, 2008.

\bibitem{2016_TR_OviedoTrespalacios}
O.~Oviedo-Trespalacios, M.~M. Haque, M.~King, and S.~Washington.
\newblock {Understanding the impacts of mobile phone distraction on driving
  performance: A systematic review}.
\newblock {\em Transportation Research Part C: Emerging Technologies},
  72:360--380, 2016.

\bibitem{2019_TR_OviedoTrespalacios}
O.~Oviedo-Trespalacios, V.~Truelove, B.~Watson, and J.~A. Hinton.
\newblock {The impact of road advertising signs on driver behaviour and
  implications for road safety: A critical systematic review}.
\newblock {\em Transportation Research Part A: Policy and Practice},
  122:85--98, 2019.

\bibitem{2011_AccidentAnalysis_Owens}
J.~M. Owens, S.~B. McLaughlin, and J.~Sudweeks.
\newblock Driver performance while text messaging using handheld and in-vehicle
  systems.
\newblock {\em {Accident Analysis \& Prevention}}, 43(3):939--947, 2011.

\bibitem{1998_JAMA_Owsley}
C.~Owsley, K.~Ball, G.~McGwin~Jr, M.~E. Sloane, D.~L. Roenker, M.~F. White, and
  E.~T. Overley.
\newblock Visual processing impairment and risk of motor vehicle crash among
  older adults.
\newblock {\em {Journal of the American Medical Association}},
  279(14):1083--1088, 1998.

\bibitem{2010_VisionResearch_Owsley}
C.~Owsley and G.~McGwin~Jr.
\newblock Vision and driving.
\newblock {\em Vision Research}, 50(23):2348--2361, 2010.

\bibitem{2019_TRR_Mangalore}
G.~Pai~Mangalore, Y.~Ebadi, S.~Samuel, M.~A. Knodler, and D.~L. Fisher.
\newblock The promise of virtual reality headsets: Can they be used to measure
  accurately drivers’ hazard anticipation performance?
\newblock {\em {Transportation Research Record}}, 2673(10):455--464, 2019.

\bibitem{2020_CVPR_Pal}
A.~Pal, S.~Mondal, and H.~I. Christensen.
\newblock {" Looking at the Right Stuff"-Guided Semantic-Gaze for Autonomous
  Driving}.
\newblock In {\em Proceedings of the IEEE/CVF Conference on Computer Vision and
  Pattern Recognition (CVPR)}, pages 11883--11892, 2020.

\bibitem{2018_PAMI_Palazzi}
A.~Palazzi, D.~Abati, F.~Solera, R.~Cucchiara, et~al.
\newblock {Predicting the Driver's Focus of Attention: the DR (eye) VE
  Project}.
\newblock {\em IEEE transactions on pattern analysis and machine intelligence},
  41(7):1720--1733, 2018.

\bibitem{2017_IV_Palazzi}
A.~Palazzi, F.~Solera, S.~Calderara, S.~Alletto, and R.~Cucchiara.
\newblock Learning where to attend like a human driver.
\newblock In {\em Proceedings of the IEEE Intelligent Vehicles Symposium (IV)},
  pages 920--925. IEEE, 2017.

\bibitem{2019_TRR_Pankok}
C.~Pankok~Jr, D.~Kaber, W.~Rasdorf, and J.~Hummer.
\newblock Effects of guide and logo signs on freeway driving behavior.
\newblock {\em {Transportation Research Record}}, 2518(1):73--78, 2015.

\bibitem{2018_TIES_Parnell}
K.~J. Parnell, N.~A. Stanton, and K.~Plant.
\newblock {Where are we on driver distraction? Methods, approaches and
  recommendations}.
\newblock {\em Theoretical Issues in Ergonomics Science}, 19(5):578--605, 2018.

\bibitem{2014_ITSC_Pech}
T.~Pech, P.~Lindner, and G.~Wanielik.
\newblock Head tracking based glance area estimation for driver behaviour
  modelling during lane change execution.
\newblock In {\em Proceedings of the International IEEE Conference on
  Intelligent Transportation Systems (ITSC)}, pages 655--660. IEEE, 2014.

\bibitem{2015_AccidentAnalysis_Peng}
Y.~Peng and L.~N. Boyle.
\newblock Driver's adaptive glance behavior to in-vehicle information systems.
\newblock {\em {Accident Analysis \& Prevention}}, 85:93--101, 2015.

\bibitem{2013_AccidentAnalysis_Peng}
Y.~Peng, L.~N. Boyle, and S.~L. Hallmark.
\newblock Driver's lane keeping ability with eyes off road: Insights from a
  naturalistic study.
\newblock {\em {Accident Analysis \& Prevention}}, 50:628--634, 2013.

\bibitem{2005_AAP_Philip}
P.~Philip, P.~Sagaspe, N.~Moore, J.~Taillard, A.~Charles, C.~Guilleminault, and
  B.~Bioulac.
\newblock Fatigue, sleep restriction and driving performance.
\newblock {\em Accident Analysis \& Prevention}, 37(3):473--478, 2005.

\bibitem{2016_VisionResearch_Poletti}
M.~Poletti and M.~Rucci.
\newblock A compact field guide to the study of microsaccades: Challenges and
  functions.
\newblock {\em Vision research}, 118:83--97, 2016.

\bibitem{2012_ACM_Pomarjanschi}
L.~Pomarjanschi, M.~Dorr, and E.~Barth.
\newblock Gaze guidance reduces the number of collisions with pedestrians in a
  driving simulator.
\newblock {\em ACM Transactions on Interactive Intelligent Systems (TiiS)},
  1(2):1--14, 2012.

\bibitem{2010_ACCV_Pugeault}
N.~Pugeault and R.~Bowden.
\newblock Learning pre-attentive driving behaviour from holistic visual
  features.
\newblock In {\em Proceedings of the European Conference on Computer Vision
  (ECCV)}, pages 154--167. Springer, 2010.

\bibitem{2015_TranVehTech_Pugeault}
N.~Pugeault and R.~Bowden.
\newblock How much of driving is preattentive?
\newblock {\em IEEE Transactions on Vehicular Technology}, 64(12):5424--5438,
  2015.

\bibitem{2018_CVPR_Ramanishka}
V.~Ramanishka, Y.-T. Chen, T.~Misu, and K.~Saenko.
\newblock Toward driving scene understanding: A dataset for learning driver
  behavior and causal reasoning.
\newblock In {\em Proceedings of the IEEE Conference on Computer Vision and
  Pattern Recognition (CVPR)}, pages 7699--7707, 2018.

\bibitem{2017_ICCVW_Rasouli}
A.~Rasouli, I.~Kotseruba, and J.~K. Tsotsos.
\newblock Are they going to cross? a benchmark dataset and baseline for
  pedestrian crosswalk behavior.
\newblock In {\em Proceedings of the IEEE International Conference on Computer
  Vision Workshops (ICCVW)}, pages 206--213, 2017.

\bibitem{2019_TITS_Rasouli}
A.~Rasouli and J.~K. Tsotsos.
\newblock {Autonomous vehicles that interact with pedestrians: A survey of
  theory and practice}.
\newblock {\em {IEEE Transactions on Intelligent Transportation Systems}},
  21(3):900--918, 2019.

\bibitem{2020_JAppGeront_Ratnapradipa}
K.~L. Ratnapradipa, C.~N. Pope, A.~Nwosu, and M.~Zhu.
\newblock {Older Driver Crash Involvement and Fatalities, by Age and Sex,
  2000--2017}.
\newblock {\em {Journal of Applied Gerontology}}, page 0733464820956507, 2020.

\bibitem{2000_JEP_Recarte}
M.~A. Recarte and L.~M. Nunes.
\newblock Effects of verbal and spatial-imagery tasks on eye fixations while
  driving.
\newblock {\em {Journal of Experimental Psychology: Applied}}, 6(1):31, 2000.

\bibitem{2003_JEP_Recarte}
M.~A. Recarte and L.~M. Nunes.
\newblock Mental workload while driving: effects on visual search,
  discrimination, and decision making.
\newblock {\em {Journal of Experimental Psychology: Applied}}, 9(2):119, 2003.

\bibitem{2011_AccidentAnalysis_Regan}
M.~A. Regan, C.~Hallett, and C.~P. Gordon.
\newblock {Driver distraction and driver inattention: Definition, relationship
  and taxonomy}.
\newblock {\em Accident Analysis \& Prevention}, 43(5):1771--1781, 2011.

\bibitem{2014_AAAM_Regan}
M.~A. Regan and D.~L. Strayer.
\newblock {Towards an understanding of driver inattention: Taxonomy and
  theory}.
\newblock {\em Annals of advances in automotive medicine}, 58:5, 2014.

\bibitem{2014_ACM_Reimer}
B.~Reimer, B.~Mehler, J.~Dobres, H.~McAnulty, A.~Mehler, D.~Munger, and
  A.~Rumpold.
\newblock Effects of an 'expert mode' voice command system on task performance,
  glance behavior \& driver physiology.
\newblock In {\em Proceedings of the International Conference on Automotive
  User Interfaces and Interactive Vehicular Applications}, pages 1--9, 2014.

\bibitem{2014_TR_Reimer}
B.~Reimer, B.~Mehler, and B.~Donmez.
\newblock A study of young adults examining phone dialing while driving using a
  touchscreen vs. a button style flip-phone.
\newblock {\em {Transportation Research Part F: Traffic Psychology and
  Behaviour}}, 23:57--68, 2014.

\bibitem{2010_HumanFactorsErgonomics_Reimer}
B.~Reimer, B.~Mehler, Y.~Wang, and J.~F. Coughlin.
\newblock The impact of systematic variation of cognitive demand on drivers'
  visual attention across multiple age groups.
\newblock In {\em Proceedings of the Human Factors and Ergonomics Society
  Annual Meeting}, volume~54, pages 2052--2055. SAGE Publications Sage CA: Los
  Angeles, CA, 2010.

\bibitem{2012_HumanFactors_Reimer}
B.~Reimer, B.~Mehler, Y.~Wang, and J.~F. Coughlin.
\newblock A field study on the impact of variations in short-term memory
  demands on drivers’ visual attention and driving performance across three
  age groups.
\newblock {\em {Human Factors}}, 54(3):454--468, 2012.

\bibitem{2005_Elsevier_Rensink}
R.~A. Rensink.
\newblock Change blindness.
\newblock In {\em Neurobiology of attention}, pages 76--81. Elsevier, 2005.

\bibitem{1993_GeriatricMedicine_Retchin}
S.~M. Retchin and J.~Anapolle.
\newblock An overview of the older driver.
\newblock {\em {Clinics in Geriatric Medicine}}, 9(2):279--296, 1993.

\bibitem{1988_JAGS_Reuben}
D.~B. Reuben, R.~A. Silliman, and M.~Traines.
\newblock The aging driver medicine, policy, and ethics.
\newblock {\em {Journal of the American Geriatrics Society}},
  36(12):1135--1142, 1988.

\bibitem{2014_CVPR_Rezaei}
M.~Rezaei and R.~Klette.
\newblock {Look at the driver, look at the road: No distraction! No accident!}
\newblock In {\em Proceedings of the IEEE Conference on Computer Vision and
  Pattern Recognition (CVPR)}, pages 129--136, 2014.

\bibitem{2011_AccidentAnalysis_Rhodes}
N.~Rhodes and K.~Pivik.
\newblock Age and gender differences in risky driving: The roles of positive
  affect and risk perception.
\newblock {\em {Accident Analysis \& Prevention}}, 43(3):923--931, 2011.

\bibitem{1960_AJO_Richards}
O.~W. Richards.
\newblock Seeing for night driving.
\newblock {\em The Australian Journal of Optometry}, 43(12):614--617, 1960.

\bibitem{2019_AccidentAnalysis_Robbins}
C.~Robbins and P.~Chapman.
\newblock {How does drivers’ visual search change as a function of
  experience? A systematic review and meta-analysis}.
\newblock {\em Accident Analysis \& Prevention}, 132:105266, 2019.

\bibitem{2019_AppliedErgonomics_Robbins}
C.~J. Robbins, H.~A. Allen, and P.~Chapman.
\newblock Comparing drivers’ visual attention at junctions in real and
  simulated environments.
\newblock {\em {Applied Ergonomics}}, 80:89--101, 2019.

\bibitem{2018_HumanFactors_Robbins}
C.~J. Robbins and P.~Chapman.
\newblock Drivers’ visual search behavior toward vulnerable road users at
  junctions as a function of cycling experience.
\newblock {\em {Human Factors}}, 60(7):889--901, 2018.

\bibitem{1970_TechRep_Rockwell}
T.~H. Rockwell, R.~L. Ernst, and M.~J. Rulon.
\newblock Visual requirements in night driving.
\newblock Technical report, National Cooperative Highway Research Program,
  1970.

\bibitem{2004_VisionResearch_Roge}
J.~Rog{\'e}, T.~P{\'e}bayle, E.~Lambilliotte, F.~Spitzenstetter,
  D.~Giselbrecht, and A.~Muzet.
\newblock Influence of age, speed and duration of monotonous driving task in
  traffic on the driver’s useful visual field.
\newblock {\em {Vision Research}}, 44(23):2737--2744, 2004.

\bibitem{2013_TR_Romoser}
M.~R. Romoser, A.~Pollatsek, D.~L. Fisher, and C.~C. Williams.
\newblock {Comparing the glance patterns of older versus younger experienced
  drivers: Scanning for hazards while approaching and entering the
  intersection}.
\newblock {\em {Transportation Research Part F: Traffic Psychology and
  Behaviour}}, 16:104--116, 2013.

\bibitem{2015_ICMIC_Ronneberger}
O.~Ronneberger, P.~Fischer, and T.~Brox.
\newblock {U-Net: Convolutional networks for biomedical image segmentation}.
\newblock In {\em Proceedings of the International Conference on Medical Image
  Computing and Computer-Assisted Intervention}, pages 234--241. Springer,
  2015.

\bibitem{2016_AnnRevVisScience_Rosenholtz}
R.~Rosenholtz.
\newblock Capabilities and limitations of peripheral vision.
\newblock 2:437--457, 2016.

\bibitem{2012_FPsych_Rosenholtz}
R.~Rosenholtz, J.~Huang, and K.~A. Ehinger.
\newblock {Rethinking the role of top-down attention in vision: Effects
  attributable to a lossy representation in peripheral vision}.
\newblock {\em Frontiers in psychology}, 3:13, 2012.

\bibitem{2016_IV_Roth}
M.~Roth, F.~Flohr, and D.~M. Gavrila.
\newblock Driver and pedestrian awareness-based collision risk analysis.
\newblock In {\em Proceedings of the IEEE Intelligent Vehicles Symposium (IV)},
  pages 454--459. IEEE, 2016.

\bibitem{1936_AJP_Ryan}
A.~Ryan and M.~Warner.
\newblock The effect of automobile driving on the reactions of the driver.
\newblock {\em The American Journal of Psychology}, 48(3):403--421, 1936.

\bibitem{1998_ICCV_Salgian}
G.~Salgian and D.~H. Ballard.
\newblock Visual routines for autonomous driving.
\newblock In {\em Proceedings of the International Conference on Computer
  Vision}, pages 876--882. IEEE, 1998.

\bibitem{1998_Springer_Salgian}
G.~Salgian and D.~H. Ballard.
\newblock Visual routines for vehicle control.
\newblock In {\em The confluence of vision and control}, pages 244--256.
  Springer, 1998.

\bibitem{2008_PsychReview_Salvucci}
D.~D. Salvucci and N.~A. Taatgen.
\newblock {Threaded cognition: An integrated theory of concurrent
  multitasking}.
\newblock {\em Psychological review}, 115(1):101, 2008.

\bibitem{2015_TRR_Samuel}
S.~Samuel and D.~L. Fisher.
\newblock Evaluation of the minimum forward roadway glance duration.
\newblock {\em {Transportation Research Record}}, 2518(1):9--17, 2015.

\bibitem{1970_Ergonomics_Sanders}
A.~F. Sanders.
\newblock Some aspects of the selective process in the functional visual field.
\newblock {\em Ergonomics}, 13(1):101--117, 1970.

\bibitem{2016_arXiv_Santana}
E.~Santana and G.~Hotz.
\newblock Learning a driving simulator.
\newblock {\em {arXiv preprint arXiv:1608.01230}}, 2016.

\bibitem{2013_TransportRes_Savage}
S.~W. Savage, D.~D. Potter, and B.~W. Tatler.
\newblock {Does preoccupation impair hazard perception? A simultaneous EEG and
  eye tracking study}.
\newblock {\em {Transportation Research Part F: Traffic Psychology and
  Behaviour}}, 17:52--62, 2013.

\bibitem{1956_PRSM_Davenport}
L.~Savin and H.~Weston.
\newblock Discussion on the visual problems of night driving.
\newblock In {\em Proceedings of the Royal Society of Medicine}, volume~50,
  pages 173--179, 1957.

\bibitem{2014_HumanFactorsErgonomics_Schieber}
F.~Schieber, K.~Limrick, R.~McCall, and A.~Beck.
\newblock Evaluation of the visual demands of digital billboards using a hybrid
  driving simulator.
\newblock In {\em Proceedings of the Human Factors and Ergonomics Society
  Annual Meeting}, volume~58, pages 2214--2218. SAGE Publications Sage CA: Los
  Angeles, CA, 2014.

\bibitem{2016_IV_Schmidt}
J.~Schmidt, C.~Braunagel, W.~Stolzmann, and K.~Karrer-Gau{\ss}.
\newblock Driver drowsiness and behavior detection in prolonged conditionally
  automated drives.
\newblock In {\em 2016 IEEE intelligent vehicles symposium (IV)}, pages
  400--405. IEEE, 2016.

\bibitem{2017_ITSC_Schwehr}
J.~Schwehr and V.~Willert.
\newblock Driver's gaze prediction in dynamic automotive scenes.
\newblock In {\em Proceedings of the IEEE 20th International Conference on
  Intelligent Transportation Systems (ITSC)}, pages 1--8. IEEE, 2017.

\bibitem{2018_ITSC_Schwehr}
J.~Schwehr and V.~Willert.
\newblock Multi-hypothesis multi-model driver's gaze target tracking.
\newblock In {\em Proceedings of the International Conference on Intelligent
  Transportation Systems (ITSC)}, pages 1427--1434. IEEE, 2018.

\bibitem{2013_SafetyResearch_Scott}
H.~Scott, L.~Hall, D.~Litchfield, and D.~Westwood.
\newblock {Visual information search in simulated junction negotiation: Gaze
  transitions of young novice, young experienced and older experienced
  drivers}.
\newblock {\em {Journal of Safety Research}}, 45:111--116, 2013.

\bibitem{2009_IEEE_Seiffert}
C.~Seiffert, T.~M. Khoshgoftaar, J.~Van~Hulse, and A.~Napolitano.
\newblock {RUSBoost: A hybrid approach to alleviating class imbalance}.
\newblock {\em {IEEE Transactions on Systems, Man, and Cybernetics-Part A:
  Systems and Humans}}, 40(1):185--197, 2009.

\bibitem{1967_TR_Senders}
J.~W. Senders, A.~Kristofferson, W.~Levison, C.~Dietrich, J.~Ward, et~al.
\newblock The attentional demand of automobile driving.
\newblock 1967.

\bibitem{2017_AccidentAnalysis_Seppelt}
B.~D. Seppelt, S.~Seaman, J.~Lee, L.~S. Angell, B.~Mehler, and B.~Reimer.
\newblock {Glass half-full: On-road glance metrics differentiate crashes from
  near-crashes in the 100-Car data}.
\newblock {\em {Accident Analysis \& Prevention}}, 107:48--62, 2017.

\bibitem{2010_AccidentAnalysis_Shahar}
A.~Shahar, C.~F. Alberti, D.~Clarke, and D.~Crundall.
\newblock Hazard perception as a function of target location and the field of
  view.
\newblock {\em {Accident Analysis \& Prevention}}, 42(6):1577--1584, 2010.

\bibitem{2014_CurrDrivAbuseRev_Shiferaw}
B.~Shiferaw, C.~Stough, and L.~Downey.
\newblock {Drivers’ visual scanning impairment under the influences of
  alcohol and distraction: A literature review}.
\newblock {\em Current drug abuse reviews}, 7(3):174--182, 2014.

\bibitem{2019_DrugAlcoDependence_Shiferaw}
B.~A. Shiferaw, D.~P. Crewther, and L.~A. Downey.
\newblock Gaze entropy measures detect alcohol-induced driver impairment.
\newblock {\em Drug and alcohol dependence}, 204:107519, 2019.

\bibitem{2018_NatSciReports_Shiferaw}
B.~A. Shiferaw, L.~A. Downey, J.~Westlake, B.~Stevens, S.~M. Rajaratnam, D.~J.
  Berlowitz, P.~Swann, and M.~E. Howard.
\newblock Stationary gaze entropy predicts lane departure events in
  sleep-deprived drivers.
\newblock {\em Scientific reports}, 8(1):1--10, 2018.

\bibitem{2017_ACCV_Shih}
T.-H. Shih and C.-T. Hsu.
\newblock Mstn: Multistage spatial-temporal network for driver drowsiness
  detection.
\newblock In {\em Proceedings of the Asian Conference on Computer Vision
  (ACCV)}, pages 146--153. Springer, 2016.

\bibitem{2017_AutoUI_Shinohara}
Y.~Shinohara, R.~Currano, W.~Ju, and Y.~Nishizaki.
\newblock {Visual attention during simulated autonomous driving in the US and
  Japan}.
\newblock In {\em Proceedings of the 9th International Conference on Automotive
  User Interfaces and Interactive Vehicular Applications}, pages 144--153,
  2017.

\bibitem{2017_SNPD_Shinohara}
Y.~Shinohara and Y.~Nishizaki.
\newblock Where do drivers look when driving in a foreign country?
\newblock In {\em International Conference on Software Engineering, Artificial
  Intelligence, Networking and Parallel/Distributed Computing}, pages 151--163.
  Springer, 2017.

\bibitem{2013_IJVT_Sigari}
M.-H. Sigari, M.~Fathy, and M.~Soryani.
\newblock A driver face monitoring system for fatigue and distraction
  detection.
\newblock {\em International Journal of Vehicular Technology}, 2013, 2013.

\bibitem{2018_TITS_Sikander}
G.~Sikander and S.~Anwar.
\newblock {Driver fatigue detection systems: A review}.
\newblock {\em IEEE Transactions on Intelligent Transportation Systems},
  20(6):2339--2352, 2018.

\bibitem{2014_JAH_SimonsMorton}
B.~G. Simons-Morton, F.~Guo, S.~G. Klauer, J.~P. Ehsani, and A.~K. Pradhan.
\newblock Keep your eyes on the road: Young driver crash risk increases
  according to duration of distraction.
\newblock {\em Journal of Adolescent Health}, 54(5):S61--S67, 2014.

\bibitem{2014_arXiv_Simonyan}
K.~Simonyan and A.~Zisserman.
\newblock Very deep convolutional networks for large-scale image recognition.
\newblock {\em arXiv preprint arXiv:1409.1556}, 2014.

\bibitem{1996_Perception_Sivak}
M.~Sivak.
\newblock The information that drivers use: is it indeed 90\% visual?
\newblock {\em Perception}, 25(9):1081--1089, 1996.

\bibitem{2016_AutoUI_Smith}
M.~Smith, J.~L. Gabbard, and C.~Conley.
\newblock Head-up vs. head-down displays: examining traditional methods of
  display assessment while driving.
\newblock In {\em Proceedings of the 8th international conference on Automotive
  User Interfaces and Interactive Vehicular Applications}, pages 185--192,
  2016.

\bibitem{2011_IET_Spiessl}
W.~Spiessl and H.~Hussmann.
\newblock Assessing error recognition in automated driving.
\newblock {\em {IET Intelligent Transport Systems}}, 5(2):103--111, 2011.

\bibitem{2003_NIPS_Sprague}
N.~Sprague and D.~Ballard.
\newblock Eye movements for reward maximization.
\newblock {\em Advances in Neural Information Processing Systems},
  16:1467--1474, 2003.

\bibitem{2007_ACM_Sprague}
N.~Sprague, D.~Ballard, and A.~Robinson.
\newblock Modeling embodied visual behaviors.
\newblock {\em {ACM Transactions on Applied Perception}}, 4(2):11--es, 2007.

\bibitem{2019_AccidentAnalysis_Stahl}
P.~Stahl, B.~Donmez, and G.~A. Jamieson.
\newblock Eye glances towards conflict-relevant cues: the roles of anticipatory
  competence and driver experience.
\newblock {\em {Accident Analysis \& Prevention}}, 132:105255, 2019.

\bibitem{2016_TR_Stavrinos}
D.~Stavrinos, P.~R. Mosley, S.~M. Wittig, H.~D. Johnson, J.~S. Decker, V.~P.
  Sisiopiku, and S.~C. Welburn.
\newblock {Visual behavior differences in drivers across the lifespan: A
  digital billboard simulator study}.
\newblock {\em {Transportation Research Part F: Traffic Psychology and
  Behaviour}}, 41:19--28, 2016.

\bibitem{2017_CHB_Stenberger}
F.~Steinberger, R.~Schroeter, and C.~N. Watling.
\newblock {From road distraction to safe driving: Evaluating the effects of
  boredom and gamification on driving behaviour, physiological arousal, and
  subjective experience}.
\newblock {\em {Computers in Human Behavior}}, 75:714--726, 2017.

\bibitem{2020_iPerception_Strasburger}
H.~Strasburger.
\newblock Seven myths on crowding and peripheral vision.
\newblock {\em i-Perception}, 11(3):2041669520913052, 2020.

\bibitem{2011_JoV_Strasburger}
H.~Strasburger, I.~Rentschler, and M.~J{\"u}ttner.
\newblock Peripheral vision and pattern recognition: A review.
\newblock {\em Journal of vision}, 11(5):13--13, 2011.

\bibitem{2012_JoV_Sullivan}
B.~T. Sullivan, L.~Johnson, C.~A. Rothkopf, D.~Ballard, and M.~Hayhoe.
\newblock The role of uncertainty and reward on eye movements in a virtual
  driving task.
\newblock {\em Journal of Vision}, 12(13):19--19, 2012.

\bibitem{1998_AccidentAnalysis_Summala}
H.~Summala, D.~Lamble, and M.~Laakso.
\newblock Driving experience and perception of the lead car's braking when
  looking at in-car targets.
\newblock {\em {Accident Analysis \& Prevention}}, 30(4):401--407, 1998.

\bibitem{1996_HumanFactors_Summala}
H.~Summala, T.~Nieminen, and M.~Punto.
\newblock Maintaining lane position with peripheral vision during in-vehicle
  tasks.
\newblock {\em Human factors}, 38(3):442--451, 1996.

\bibitem{2016_JEMR_Sun}
Q.~Sun, j.~Xia, T.~Falkmer, and H.~Lee.
\newblock {Investigating the Spatial Pattern of Older Drivers Eye Fixation
  Behaviour and Associations with Their Visual Capacity}.
\newblock {\em Journal of Eye Movement Research}, 9(6), 2016.

\bibitem{2018_AccidentAnalysis_Sun}
Q.~C. Sun, J.~C. Xia, J.~He, J.~Foster, T.~Falkmer, and H.~Lee.
\newblock {Towards unpacking older drivers’ visual-motor coordination: A
  gaze-based integrated driving assessment}.
\newblock {\em {Accident Analysis \& Prevention}}, 113:85--96, 2018.

\bibitem{2017_NatSciData_Taamneh}
S.~Taamneh, P.~Tsiamyrtzis, M.~Dcosta, P.~Buddharaju, A.~Khatri, M.~Manser,
  T.~Ferris, R.~Wunderlich, and I.~Pavlidis.
\newblock A multimodal dataset for various forms of distracted driving.
\newblock {\em Scientific Data}, 4:170110, 2017.

\bibitem{2020_arXiv_Tampuu}
A.~Tampuu, M.~Semikin, N.~Muhammad, D.~Fishman, and T.~Matiisen.
\newblock {A Survey of End-to-End Driving: Architectures and Training Methods}.
\newblock {\em {arXiv preprint arXiv:2003.06404}}, 2020.

\bibitem{2011_JoV_Tatler}
B.~W. Tatler, M.~M. Hayhoe, M.~F. Land, and D.~H. Ballard.
\newblock {Eye guidance in natural vision: Reinterpreting salience}.
\newblock {\em Journal of Vision}, 11(5):5--5, 2011.

\bibitem{2019_WACV_Tavakoli}
H.~R. Tavakoli, E.~Rahtu, J.~Kannala, and A.~Borji.
\newblock Digging deeper into egocentric gaze prediction.
\newblock In {\em Proceedings of the IEEE Winter Conference on Applications of
  Computer Vision (WACV)}, pages 273--282. IEEE, 2019.

\bibitem{2014_ITSC_Tawari}
A.~Tawari, K.~H. Chen, and M.~M. Trivedi.
\newblock Where is the driver looking: Analysis of head, eye and iris for
  robust gaze zone estimation.
\newblock In {\em Proceedings of the International IEEE Conference on
  Intelligent Transportation Systems (ITSC)}, pages 988--994. IEEE, 2014.

\bibitem{2017_IV_Tawari}
A.~Tawari and B.~Kang.
\newblock A computational framework for driver's visual attention using a fully
  convolutional architecture.
\newblock In {\em Proceedings of the IEEE Intelligent Vehicles Symposium (IV)},
  pages 887--894. IEEE, 2017.

\bibitem{2018_ITSC_Tawari}
A.~Tawari, P.~Mallela, and S.~Martin.
\newblock Learning to attend to salient targets in driving videos using fully
  convolutional rnn.
\newblock In {\em Proceedings of the International Conference on Intelligent
  Transportation Systems (ITSC)}, pages 3225--3232. IEEE, 2018.

\bibitem{2014_ITSC_Tawari_1}
A.~Tawari, A.~M{\o}gelmose, S.~Martin, T.~B. Moeslund, and M.~M. Trivedi.
\newblock Attention estimation by simultaneous analysis of viewer and view.
\newblock In {\em Proceedings fothe International IEEE Conference on
  Intelligent Transportation Systems (ITSC)}, pages 1381--1387. IEEE, 2014.

\bibitem{2014_IV_Tawari}
A.~Tawari and M.~M. Trivedi.
\newblock Robust and continuous estimation of driver gaze zone by dynamic
  analysis of multiple face videos.
\newblock In {\em Proceedings of the IEEE Intelligent Vehicles Symposium (IV)},
  pages 344--349. IEEE, 2014.

\bibitem{2011_DrivSymposium_Taylor}
T.~G. Taylor, K.~M. Masserang, A.~K. Pradhan, G.~Divekar, S.~Samuel, J.~W.
  Muttart, A.~Pollatsek, and D.~L. Fisher.
\newblock Long term effects of hazard anticipation training on novice drivers
  measured on the open road.
\newblock In {\em Proceedings of the... International Driving Symposium on
  Human Factors in Driver Assessment, Training, and Vehicle Design}, volume
  2011, page 187. NIH Public Access, 2011.

\bibitem{2016_SJOVS_Thorslund}
B.~Thorslund and N.~Strand.
\newblock {Vision measurability and its impact on safe driving: A literature
  review}.
\newblock {\em Scandinavian Journal of Optometry and Visual Science},
  9(1):1--9, 2016.

\bibitem{2014_TransRes_Tivesten}
E.~Tivesten and M.~Dozza.
\newblock Driving context and visual-manual phone tasks influence glance
  behavior in naturalistic driving.
\newblock {\em {Transportation Research Part F: Traffic Psychology and
  Behaviour}}, 26:258--272, 2014.

\bibitem{2015_DDI_Tivesten}
E.~Tivesten, A.~Morando, and T.~Victor.
\newblock The timecourse of driver visual attention in naturalistic driving
  with adaptive cruise control and forward collision warning.
\newblock In {\em Proceedings of the International Conference on Driver
  Distraction and Inattention}, number 15349, 2015.

\bibitem{2016_TR_Topolsek}
D.~Topol{\v{s}}ek, I.~Areh, and T.~Cvahte.
\newblock Examination of driver detection of roadside traffic signs and
  advertisements using eye tracking.
\newblock {\em {Transportation Research Part F: Traffic Psychology and
  Behaviour}}, 43:212--224, 2016.

\bibitem{2015_ICCV_Tran}
D.~Tran, L.~Bourdev, R.~Fergus, L.~Torresani, and M.~Paluri.
\newblock Learning spatiotemporal features with 3d convolutional networks.
\newblock In {\em Proceedings of the International Conference on Computer
  Vision}, pages 4489--4497, 2015.

\bibitem{2012_VR_Traschulz}
A.~Trasch{\"u}tz, W.~Zinke, and D.~Wegener.
\newblock Speed change detection in foveal and peripheral vision.
\newblock {\em Vision Research}, 72:1--13, 2012.

\bibitem{1980_CognitivePsychology_Treisman}
A.~M. Treisman and G.~Gelade.
\newblock A feature-integration theory of attention.
\newblock {\em Cognitive psychology}, 12(1):97--136, 1980.

\bibitem{2011_DrivSymposium_Trutschel}
U.~Trutschel, B.~Sirois, D.~Sommer, M.~Golz, and D.~Edwards.
\newblock {PERCLOS: An alertness measure of the past}.
\newblock In {\em {International Driving Symposium on Human Factors in Driver
  Assessment}}, pages 172--179, 2011.

\bibitem{2016_JEMR_Tsotsos}
J.~Tsotsos, I.~Kotseruba, and C.~Wloka.
\newblock A focus on selection for fixation.
\newblock {\em Journal of Eye Movement Research}, 9(5), 2016.

\bibitem{2011_MIT_Tsotsos}
J.~K. Tsotsos.
\newblock {\em A computational perspective on visual attention}.
\newblock MIT Press, 2011.

\bibitem{1984_Cognition_Ullman}
S.~Ullman.
\newblock Visual routines.
\newblock {\em Cognition}, 18:97--159, 1984.

\bibitem{2003_TransRes_Underwood}
G.~Underwood, P.~Chapman, Z.~Berger, and D.~Crundall.
\newblock Driving experience, attentional focusing, and the recall of recently
  inspected events.
\newblock {\em {Transportation Research Part F: Traffic Psychology and
  Behaviour}}, 6(4):289--304, 2003.

\bibitem{2011_VR_Underwood}
G.~Underwood, K.~Humphrey, and E.~Van~Loon.
\newblock Decisions about objects in real-world scenes are influenced by visual
  saliency before and during their inspection.
\newblock {\em Vision research}, 51(18):2031--2038, 2011.

\bibitem{2015_BMCGeriatrics_Urwyler}
P.~Urwyler, N.~Gruber, R.~M. M{\"u}ri, M.~J{\"a}ger, R.~Bieri, T.~Nyffeler,
  U.~P. Mosimann, and T.~Nef.
\newblock {Age-dependent visual exploration during simulated day-and night
  driving on a motorway: A cross-sectional study}.
\newblock {\em BMC Geriatrics}, 15(1):18, 2015.

\bibitem{2014_TransRes_vanLeeuwen}
P.~Van~Leeuwen, R.~Happee, and J.~De~Winter.
\newblock {Vertical field of view restriction in driver training: A
  simulator-based evaluation}.
\newblock {\em {Transportation Research Part F: Traffic Psychology and
  Behaviour}}, 24:169--182, 2014.

\bibitem{2017_PONE_VanLeeuwen}
P.~M. Van~Leeuwen, S.~de~Groot, R.~Happee, and J.~C. de~Winter.
\newblock {Differences between racing and non-racing drivers: A simulator study
  using eye-tracking}.
\newblock {\em PLoS one}, 12(11):e0186871, 2017.

\bibitem{2015_Ergonomics_vanLeeuwen}
P.~M. van Leeuwen, C.~G{\'o}mez~i Subils, A.~Ramon~Jimenez, R.~Happee, and
  J.~C. de~Winter.
\newblock Effects of visual fidelity on curve negotiation, gaze behaviour and
  simulator discomfort.
\newblock {\em Ergonomics}, 58(8):1347--1364, 2015.

\bibitem{2010_Perception_vanLoon}
E.~M. Van~Loon, F.~Khashawi, and G.~Underwood.
\newblock Visual strategies used for time-to-arrival judgments in driving.
\newblock {\em Perception}, 39(9):1216--1229, 2010.

\bibitem{2016_ITSC_Vasli}
B.~Vasli, S.~Martin, and M.~M. Trivedi.
\newblock On driver gaze estimation: Explorations and fusion of geometric and
  data driven approaches.
\newblock In {\em Proceedings of the International Conference on Intelligent
  Transportation Systems (ITSC)}, pages 655--660. IEEE, 2016.

\bibitem{2002_TransRes_Velichkovsky}
B.~M. Velichkovsky, A.~Rothert, M.~Kopf, S.~M. Dornh{\"o}fer, and M.~Joos.
\newblock Towards an express-diagnostics for level of processing and hazard
  perception.
\newblock {\em {Transportation Research Part F: Traffic Psychology and
  Behaviour}}, 5(2):145--156, 2002.

\bibitem{2015_TITS_Vicente}
F.~Vicente, Z.~Huang, X.~Xiong, F.~De~la Torre, W.~Zhang, and D.~Levi.
\newblock Driver gaze tracking and eyes off the road detection system.
\newblock {\em IEEE Transactions on Intelligent Transportation Systems},
  16(4):2014--2027, 2015.

\bibitem{2015_TR_Victor}
T.~Victor, M.~Dozza, J.~B{\"a}rgman, C.-N. Boda, J.~Engstr{\"o}m, C.~Flannagan,
  J.~D. Lee, and G.~Markkula.
\newblock {Analysis of naturalistic driving study data: Safer glances, driver
  inattention, and crash risk}.
\newblock Technical report, Transportation Research Board, 2015.

\bibitem{2005_TransRes_Victor}
T.~W. Victor, J.~L. Harbluk, and J.~A. Engstr{\"o}m.
\newblock Sensitivity of eye-movement measures to in-vehicle task difficulty.
\newblock {\em {Transportation Research Part F: Traffic Psychology and
  Behaviour}}, 8(2):167--190, 2005.

\bibitem{2018_HumanFactors_Victor}
T.~W. Victor, E.~Tivesten, P.~Gustavsson, J.~Johansson, F.~Sangberg, and
  M.~Ljung~Aust.
\newblock Automation expectation mismatch: Incorrect prediction despite eyes on
  threat and hands on wheel.
\newblock {\em {Human Factors}}, 60(8):1095--1116, 2018.

\bibitem{2001_IJCV_Viola}
P.~Viola, M.~Jones, et~al.
\newblock Robust real-time object detection.
\newblock {\em {International Journal of Computer Vision}}, 4(34-47):4, 2001.

\bibitem{2011_TransRes_Vkalveld}
W.~Vlakveld, M.~R. Romoser, H.~Mehranian, F.~Diete, A.~Pollatsek, and D.~L.
  Fisher.
\newblock Do crashes and near crashes in simulator-based training enhance
  novice drivers’ visual search for latent hazards?
\newblock {\em {Transportation Research Record}}, 2265(1):153--160, 2011.

\bibitem{2018_TR_Vlakveld}
W.~Vlakveld, N.~van Nes, J.~de~Bruin, L.~Vissers, and M.~van~der Kroft.
\newblock {Situation awareness increases when drivers have more time to take
  over the wheel in a Level 3 automated car: A simulator study}.
\newblock {\em {Transportation Research Part F: Traffic Psychology and
  Behaviour}}, 58:917--929, 2018.

\bibitem{2019_AccidentAnalysis_Vogelpohl}
T.~Vogelpohl, M.~K{\"u}hn, T.~Hummel, and M.~Vollrath.
\newblock {Asleep at the automated wheel -- Sleepiness and fatigue during
  highly automated driving}.
\newblock {\em {Accident Analysis \& Prevention}}, 126:70--84, 2019.

\bibitem{2017_IV_Vora}
S.~Vora, A.~Rangesh, and M.~M. Trivedi.
\newblock On generalizing driver gaze zone estimation using convolutional
  neural networks.
\newblock In {\em Proceedings of the IEEE Intelligent Vehicles Symposium (IV)},
  pages 849--854. IEEE, 2017.

\bibitem{2018_TIV_Vora}
S.~Vora, A.~Rangesh, and M.~M. Trivedi.
\newblock Driver gaze zone estimation using convolutional neural networks: A
  general framework and ablative analysis.
\newblock {\em IEEE Transactions on Intelligent Vehicles}, 3(3):254--265, 2018.

\bibitem{2019_AutomotiveUI_Walch}
M.~Walch, D.~Lehr, M.~Colley, and M.~Weber.
\newblock {Don't you see them? Towards gaze-based interaction adaptation for
  driver-vehicle cooperation}.
\newblock In {\em {Proceedings of the International Conference on Automotive
  User Interfaces and Interactive Vehicular Applications (AutomotiveUI)}},
  pages 232--237, 2019.

\bibitem{1991_HumanFactors_Waller}
P.~F. Waller.
\newblock The older driver.
\newblock {\em {Human Factors}}, 33(5):499--505, 1991.

\bibitem{2019_ICRA_Wang}
D.~Wang, C.~Devin, Q.-Z. Cai, F.~Yu, and T.~Darrell.
\newblock Deep object-centric policies for autonomous driving.
\newblock In {\em Proceedings of the International Conference on Robotics and
  Automation (ICRA)}, pages 8853--8859. IEEE, 2019.

\bibitem{2018_IROS_Wang}
R.~Wang, P.~V. Amadori, and Y.~Demiris.
\newblock Real-time workload classification during driving using hypernetworks.
\newblock In {\em Proceedings of the International Conference on Intelligent
  Robots and Systems (IROS)}, pages 3060--3065. IEEE, 2018.

\bibitem{2019_arXiv_Wang}
W.~Wang, Q.~Lai, H.~Fu, J.~Shen, H.~Ling, and R.~Yang.
\newblock Salient object detection in the deep learning era: An in-depth
  survey.
\newblock {\em arXiv preprint arXiv:1904.09146}, 2019.

\bibitem{2019_PAMI_Wang}
W.~Wang, J.~Shen, J.~Xie, M.-M. Cheng, H.~Ling, and A.~Borji.
\newblock Revisiting video saliency prediction in the deep learning era.
\newblock {\em IEEE Transactions on Pattern Analysis and Machine Intelligence},
  43(1):220--237, 2019.

\bibitem{2016_AccidentAnalysis_Wang}
X.~Wang and C.~Xu.
\newblock Driver drowsiness detection based on non-intrusive metrics
  considering individual specifics.
\newblock {\em Accident Analysis \& Prevention}, 95:350--357, 2016.

\bibitem{2017_JSR_Wang}
Y.~Wang, S.~Bao, W.~Du, Z.~Ye, and J.~R. Sayer.
\newblock Examining drivers' eye glance patterns during distracted driving:
  Insights from scanning randomness and glance transition matrix.
\newblock {\em {Journal of Safety Research}}, 63:149--155, 2017.

\bibitem{2017_TrafficInjuryPrevention_Wang}
Y.~Wang, M.~Xin, H.~Bai, and Y.~Zhao.
\newblock {Can variations in visual behavior measures be good predictors of
  driver sleepiness? A real driving test study}.
\newblock {\em Traffic injury prevention}, 18(2):132--138, 2017.

\bibitem{2018_TIV_Wang}
Z.~Wang, R.~Zheng, T.~Kaizuka, and K.~Nakano.
\newblock Relationship between gaze behavior and steering performance for
  driver--automation shared control: a driving simulator study.
\newblock {\em IEEE Transactions on Intelligent Vehicles}, 4(1):154--166, 2018.

\bibitem{2000_Nature_Wann}
J.~P. Wann and D.~K. Swapp.
\newblock Why you should look where you are going.
\newblock {\em {Nature Neuroscience}}, 3(7):647--648, 2000.

\bibitem{1992_PP_Warren}
W.~H. Warren and K.~J. Kurtz.
\newblock The role of central and peripheral vision in perceiving the direction
  of self-motion.
\newblock {\em Perception \& psychophysics}, 51(5):443--454, 1992.

\bibitem{2017_ACCV_Weng}
C.-H. Weng, Y.-H. Lai, and S.-H. Lai.
\newblock Driver drowsiness detection via a hierarchical temporal deep belief
  network.
\newblock In {\em Proceedings of the Asian Conference on Computer Vision
  (ACCV)}, pages 117--133. Springer, 2016.

\bibitem{2012_AccidentAnalysis_Werneke}
J.~Werneke and M.~Vollrath.
\newblock What does the driver look at? the influence of intersection
  characteristics on attention allocation and driving behavior.
\newblock {\em {Accident Analysis \& Prevention}}, 45:610--619, 2012.

\bibitem{2014_CogTechWork_Werneke}
J.~Werneke and M.~Vollrath.
\newblock How do environmental characteristics at intersections change in their
  relevance for drivers before entering an intersection: analysis of drivers’
  gaze and driving behavior in a driving simulator study.
\newblock {\em Cognition, technology \& work}, 16(2):157--169, 2014.

\bibitem{2010_AccidentAnalysis_White}
C.~B. White and J.~K. Caird.
\newblock {The blind date: The effects of change blindness, passenger
  conversation and gender on looked-but-failed-to-see (LBFTS) errors}.
\newblock {\em {Accident Analysis \& Prevention}}, 42(6):1822--1830, 2010.

\bibitem{2011_Cell_Whitney}
D.~Whitney and D.~M. Levi.
\newblock {Visual crowding: A fundamental limit on conscious perception and
  object recognition}.
\newblock {\em {Trends in Cognitive Sciences}}, 15(4):160--168, 2011.

\bibitem{2001_TechReport_Wickens}
C.~D. Wickens, J.~Helleberg, J.~Goh, X.~Xu, and W.~J. Horrey.
\newblock Pilot task management: Testing an attentional expected value model of
  visual scanning.
\newblock Technical report, {Aviation Research Lab Institute of Aviation,
  University of Illinois at Urbana-Champaign}, 2001.

\bibitem{1994_TechRep_Wierwille}
W.~W. Wierwille, S.~Wreggit, C.~Kirn, L.~Ellsworth, and R.~Fairbanks.
\newblock {Research on vehicle-based driver status/performance monitoring;
  development, validation, and refinement of algorithms for detection of driver
  drowsiness. Final report}.
\newblock Technical report, 1994.

\bibitem{2016_JAR_Wilson}
R.~T. Wilson and J.~Casper.
\newblock {The role of location and visual saliency in capturing attention to
  outdoor advertising: How location attributes increase the likelihood for a
  driver to notice a billboard ad}.
\newblock {\em Journal of Advertising Research}, 56(3):259--273, 2016.

\bibitem{2017_AppliedErgonomics_Wolfe}
B.~Wolfe, J.~Dobres, R.~Rosenholtz, and B.~Reimer.
\newblock {More than the Useful Field: Considering peripheral vision in
  driving}.
\newblock {\em Applied Ergonomics}, 65:316--325, 2017.

\bibitem{2019_APP_Wolfe}
B.~Wolfe, B.~D. Sawyer, A.~Kosovicheva, B.~Reimer, and R.~Rosenholtz.
\newblock Detection of brake lights while distracted: separating peripheral
  vision from cognitive load.
\newblock {\em {Attention, Perception, \& Psychophysics}}, 81(8):2798--2813,
  2019.

\bibitem{2020_HumanFactors_Wolfe}
B.~Wolfe, B.~D. Sawyer, and R.~Rosenholtz.
\newblock Toward a theory of visual information acquisition in driving.
\newblock {\em {Human Factors}}, page 0018720820939693, 2020.

\bibitem{2020_JEP_Wolfe}
B.~Wolfe, B.~Seppelt, B.~Mehler, B.~Reimer, and R.~Rosenholtz.
\newblock Rapid holistic perception and evasion of road hazards.
\newblock {\em {Journal of Experimental Psychology: General}}, 149(3):490,
  2020.

\bibitem{2005_Nature_Wolfe}
J.~M. Wolfe, T.~S. Horowitz, and N.~M. Kenner.
\newblock Rare items often missed in visual searches.
\newblock {\em Nature}, 435(7041):439--440, 2005.

\bibitem{2011_TITS_Wollmer}
M.~Wollmer, C.~Blaschke, T.~Schindl, B.~Schuller, B.~Farber, S.~Mayer, and
  B.~Trefflich.
\newblock Online driver distraction detection using long short-term memory.
\newblock {\em IEEE Transactions on Intelligent Transportation Systems},
  12(2):574--582, 2011.

\bibitem{2013_AccidentPrevention_Wong}
J.-T. Wong and S.-H. Huang.
\newblock Attention allocation patterns in naturalistic driving.
\newblock {\em {Accident Analysis \& Prevention}}, 58:140--147, 2013.

\bibitem{2016_JARMAC_Wood}
G.~Wood, G.~Hartley, P.~Furley, and M.~Wilson.
\newblock Working memory capacity, visual attention and hazard perception in
  driving.
\newblock {\em Journal of Applied Research in Memory and Cognition},
  5(4):454--462, 2016.

\bibitem{2013_TR_Wortelen}
B.~Wortelen, M.~Baumann, and A.~L{\"u}dtke.
\newblock Dynamic simulation and prediction of drivers’ attention
  distribution.
\newblock {\em Transportation Research Part F: Traffic Psychology and
  Behaviour}, 21:278--294, 2013.

\bibitem{2019_TrafficInjuryPrevention_Wundersitz}
L.~Wundersitz.
\newblock {Driver distraction and inattention in fatal and injury crashes:
  Findings from in-depth road crash data}.
\newblock {\em Traffic Injury Prevention}, 20(7):696--701, 2019.

\bibitem{2000_TR_Wymann}
B.~Wymann, E.~Espi{\'e}, C.~Guionneau, C.~Dimitrakakis, R.~Coulom, and
  A.~Sumner.
\newblock {TORCS, The open racing car simulator}.
\newblock Technical report, 2000.

\bibitem{2019_SafetyScience_Wynne}
R.~A. Wynne, V.~Beanland, and P.~M. Salmon.
\newblock Systematic review of driving simulator validation studies.
\newblock {\em Safety Science}, 117:138--151, 2019.

\bibitem{2018_ACCV_Xia}
Y.~Xia, D.~Zhang, J.~Kim, K.~Nakayama, K.~Zipser, and D.~Whitney.
\newblock Predicting driver attention in critical situations.
\newblock In {\em {Proceedings of the Asian Conference on Computer Vision}},
  pages 658--674. Springer, 2018.

\bibitem{2017_CVPR_Xu}
H.~Xu, Y.~Gao, F.~Yu, and T.~Darrell.
\newblock End-to-end learning of driving models from large-scale video
  datasets.
\newblock In {\em Proceedings of the IEEE Conference on Computer Vision and
  Pattern Recognition}, pages 2174--2182, 2017.

\bibitem{2017_DrivingAssessmentConference_Yamani}
Y.~Yamani, P.~B{\i}{\c{c}}aks{\i}z, D.~B. Palmer, J.~M. Cronauer, and
  S.~Samuel.
\newblock {Following expert’s eyes: Evaluation of the effectiveness of a
  gaze-based training intervention on young drivers’ latent hazard
  anticipation skills}.
\newblock In {\em {Driving Assessment Conference}}. University of Iowa, 2017.

\bibitem{2016_PLOS_Yamani}
Y.~Yamani, W.~J. Horrey, Y.~Liang, and D.~L. Fisher.
\newblock Age-related differences in vehicle control and eye movement patterns
  at intersections: older and middle-aged drivers.
\newblock {\em PLoS one}, 11(10):e0164124, 2016.

\bibitem{2016_PLOS_Yan}
X.~Yan, X.~Zhang, Y.~Zhang, X.~Li, and Z.~Yang.
\newblock Changes in drivers’ visual performance during the collision
  avoidance process as a function of different field of views at intersections.
\newblock {\em PLoS one}, 11(10):e0164101, 2016.

\bibitem{2015_IET_Yang}
B.~Yang, R.~Zheng, Y.~Yin, S.~Yamabe, and K.~Nakano.
\newblock Analysis of influence on driver behaviour while using in-vehicle
  traffic lights with application of head-up display.
\newblock {\em IET Intelligent Transport Systems}, 10(5):347--353, 2016.

\bibitem{2018_ITSC_Yang}
Y.~Yang, B.~Karakaya, G.~C. Dominioni, K.~Kawabe, and K.~Bengler.
\newblock {An HMI Concept to Improve Driver's Visual Behavior and Situation
  Awareness in Automated Vehicle}.
\newblock In {\em Proceedings of the IEEE International Conference on
  Intelligent Transportation Systems}, pages 650--655. IEEE, 2018.

\bibitem{1998_Attention_Yantis}
S.~Yantis.
\newblock Control of visual attention.
\newblock {\em Attention}, 1(1):223--256, 1998.

\bibitem{1967_EyeMovements_Yarbus}
A.~L. Yarbus.
\newblock Eye movements during perception of complex objects.
\newblock In {\em Eye movements and vision}, pages 171--211. Springer, 1967.

\bibitem{2013_TITS_Yekhshatyan}
L.~Yekhshatyan and J.~D. Lee.
\newblock Changes in the correlation between eye and steering movements
  indicate driver distraction.
\newblock {\em IEEE Transactions on Intelligent Transportation Systems},
  14(1):136--145, 2012.

\bibitem{2018_TransRes_Young}
A.~H. Young, A.~K. Mackenzie, R.~L. Davies, and D.~Crundall.
\newblock {Familiarity breeds contempt for the road ahead: The real-world
  effects of route repetition on visual attention in an expert driver}.
\newblock {\em {Transportation Research Part F: Traffic Psychology and
  Behaviour}}, 57:4--9, 2018.

\bibitem{2007_DistractedDriving_Young}
K.~Young, M.~Regan, and M.~Hammer.
\newblock {Driver distraction: A review of the literature}.
\newblock In I.~Faulks, M.~Regan, M.~Stevenson, J.~Brown, A.~Porter, and
  J.~Irwi, editors, {\em Distracted driving}, volume 2007, pages 379--405.
  Sydney, NSW: Australasian College of Road Safety, 2007.

\bibitem{2014_SafetyScience_Young}
K.~L. Young, C.~M. Rudin-Brown, C.~Patten, R.~Ceci, and M.~G. Lenn{\'e}.
\newblock Effects of phone type on driving and eye glance behaviour while
  text-messaging.
\newblock {\em Safety science}, 68:47--54, 2014.

\bibitem{2012_SafetyScience_Young}
K.~L. Young and P.~M. Salmon.
\newblock {Examining the relationship between driver distraction and driving
  errors: A discussion of theory, studies and methods}.
\newblock {\em Safety Science}, 50(2):165--174, 2012.

\bibitem{2012_SAE_Young}
R.~Young.
\newblock {Cognitive distraction while driving: A critical review of
  definitions and prevalence in crashes}.
\newblock {\em SAE International Journal of Passenger Cars-Electronic and
  Electrical Systems}, 5(2012-01-0967):326--342, 2012.

\bibitem{2018_TITS_Yu}
J.~Yu, S.~Park, S.~Lee, and M.~Jeon.
\newblock Driver drowsiness detection using condition-adaptive representation
  learning framework.
\newblock {\em IEEE Transactions on Intelligent Transportation Systems},
  20(11):4206--4218, 2018.

\bibitem{2017_IV_Zabihi}
S.~J. Zabihi, S.~Zabihi, S.~S. Beauchemin, and M.~A. Bauer.
\newblock Detection and recognition of traffic signs inside the attentional
  visual field of drivers.
\newblock In {\em Proceedings of the IEEE Intelligent Vehicles Symposium (IV)},
  pages 583--588. IEEE, 2017.

\bibitem{2018_AppliedErgonomics_Zahabi}
M.~Zahabi and D.~Kaber.
\newblock Effect of police mobile computer terminal interface design on officer
  driving distraction.
\newblock {\em {Applied Ergonomics}}, 67:26--38, 2018.

\bibitem{2018_HumanFactorsErgonomics_Zahabi}
M.~Zahabi, P.~Machado, M.~Lau, Y.~Deng, C.~Pankok, J.~Hummer, W.~Rasdorf, and
  D.~Kaber.
\newblock {Effect of Driver Age and Distance Guide Sign Format on Driver
  Attention Allocation and Performance}.
\newblock In {\em {Proceedings of the Human Factors and Ergonomics Society
  Annual Meeting}}, volume~62, pages 1903--1907. SAGE Publications Sage CA: Los
  Angeles, CA, 2018.

\bibitem{2017_AppliedErgonomics_Zahabi}
M.~Zahabi, P.~Machado, M.~Y. Lau, Y.~Deng, C.~Pankok~Jr, J.~Hummer, W.~Rasdorf,
  and D.~B. Kaber.
\newblock Driver performance and attention allocation in use of logo signs on
  freeway exit ramps.
\newblock {\em {Applied Ergonomics}}, 65:70--80, 2017.

\bibitem{2016_AccidentAnalysis_Zeeb}
K.~Zeeb, A.~Buchner, and M.~Schrauf.
\newblock {Is take-over time all that matters? The impact of visual-cognitive
  load on driver take-over quality after conditionally automated driving}.
\newblock {\em {Accident Analysis \& Prevention}}, 92:230--239, 2016.

\bibitem{1969_TechRep_Zell}
J.~K. Zell.
\newblock {\em {Driver's Eye Movements as a Function of Driving Experience}}.
\newblock Ohio State University, 1969.

\bibitem{2019_IEEEAccess_Zhang}
C.~Zhang, X.~Wu, X.~Zheng, and S.~Yu.
\newblock Driver drowsiness detection using multi-channel second order blind
  identifications.
\newblock {\em IEEE Access}, 7:11829--11843, 2019.

\bibitem{2019_JTEPBS_Zhang}
L.~Zhang, J.~Kong, B.~Cui, and T.~Fu.
\newblock {Safety Effects of Freeway Roadside Electronic Billboards on Visual
  Properties of Drivers: Insights from Field Experiments}.
\newblock {\em {Journal of Transportation Engineering, Part A: Systems}},
  146(2):04019071, 2020.

\bibitem{2013_AppliedErgonomics_Zhang}
Y.~Zhang, E.~Harris, M.~Rogers, D.~Kaber, J.~Hummer, W.~Rasdorf, and J.~Hu.
\newblock Driver distraction and performance effects of highway logo sign
  design.
\newblock {\em {Applied Ergonomics}}, 44(3):472--479, 2013.

\bibitem{2010_JCTA_Zhang}
Z.~Zhang and J.~Zhang.
\newblock A new real-time eye tracking based on nonlinear unscented kalman
  filter for monitoring driver fatigue.
\newblock {\em Journal of Control Theory and Applications}, 8(2):181--188,
  2010.

\bibitem{2017_IET_Zhao}
L.~Zhao, Z.~Wang, X.~Wang, and Q.~Liu.
\newblock Driver drowsiness detection using facial dynamic fusion information
  and a dbn.
\newblock {\em IET Intelligent Transport Systems}, 12(2):127--133, 2017.

\bibitem{2020_TR_Zheng}
X.~Zheng, Y.~Yang, S.~Easa, W.~Lin, and E.~Cherchi.
\newblock The effect of leftward bias on visual attention for driving tasks.
\newblock {\em {Transportation Research Part F: Traffic Psychology and
  Behaviour}}, 70:199--207, 2020.

\end{thebibliography}
}
\end{document}